\documentclass[preprint,5p,twocolumn]{elsarticle}

\usepackage{amssymb}
\usepackage{amsmath}
\usepackage{algorithmic}
\usepackage{algorithm}
\usepackage{url}
\usepackage[colorlinks,linkcolor=blue]{hyperref}
\usepackage{threeparttable}
\usepackage{multirow}
\usepackage{array}
\usepackage{makecell}
\usepackage{xcolor}         
\usepackage{graphicx}       
\graphicspath{{figures/}}
\usepackage{colortbl}
\usepackage{subcaption}     
\usepackage{enumitem}
\usepackage{cancel}

\begin{document}

\begin{frontmatter}



\title{Attack Anything: Blind DNNs via Universal Background Adversarial Attack}


\author[organization1,organization2]{Jiawei Lian} 
\author[organization1]{Shaohui Mei\corref{cor1}}
\cortext[cor1]{Corresponding author.}
\ead{meish@nwpu.edu.cn}
\author[organization1]{Xiaofei Wang}
\author[organization2]{Yi Wang}
\author[organization1]{Lefan Wang}
\author[organization1]{Yingjie Lu}
\author[organization1]{Mingyang Ma}
\author[organization2]{Lap-Pui Chau}

\affiliation[organization1]{
            organization={School of Electronics and Information, Northwestern Polytechnical University},
            city={Xi'an},
            postcode={710129},
            country={China}}
\affiliation[organization2]{
            organization={Department of Electrical and Electronic Engineering, The Hong Kong Polytechnic University},
            city={Hong Kong SAR}}

\begin{abstract}
  It has been widely substantiated that deep neural networks (DNNs) are susceptible and vulnerable to adversarial perturbations.
  Existing studies mainly focus on performing attacks by corrupting targeted objects (physical attack) or images (digital attack), which is intuitively acceptable and understandable in terms of the attack's effectiveness.
  In contrast, our focus lies in conducting background adversarial attacks in both digital and physical domains, without causing any disruptions to the targeted objects themselves.
  Specifically, an effective background adversarial attack framework is proposed to attack anything, by which the attack efficacy generalizes well between diverse objects, models, and tasks.
  Technically, we approach the background adversarial attack as an iterative optimization problem, analogous to the process of DNN learning.
  Besides, we offer a theoretical demonstration of its convergence under a set of mild but sufficient conditions.
  To strengthen the attack efficacy and transferability, we propose a new ensemble strategy tailored for adversarial perturbations and introduce an improved smooth constraint for the seamless connection of integrated perturbations.
  We conduct comprehensive and rigorous experiments in both digital and physical domains across various objects, models, and tasks, demonstrating the effectiveness of attacking anything of the proposed method.
  The findings of this research substantiate the significant discrepancy between human and machine vision on the value of background variations, which play a far more critical role than previously recognized, necessitating a reevaluation of the robustness and reliability of DNNs. 
  The code will be publicly available at \url{https://github.com/JiaweiLian/Attack_Anything}.
\end{abstract}



\begin{keyword}
DNNs \sep Universal \sep Background \sep Adversarial Attack



\end{keyword}

\end{frontmatter}



\section{Introduction}
The remarkable advancements of deep learning have revolutionized various domains of artificial intelligence (AI), enabling significant achievements in computer vision, natural language processing, and other complex tasks.
However, these achievements have also unveiled a critical vulnerability of deep neural networks (DNNs) to adversarial perturbations \cite{szegedy2014intriguing,chen2024content,scheurer2024detection,tang2025sfa,pi2025efficient,cao2025subspectrum,du2024efficient}. 
Numerous studies \cite{huang2024towards,li2024survey,ma2024adversarial,bai2024wasserstein,brown2017adversarial,wei2024physical} have demonstrated the alarming ease with which state-of-the-art (SOTA) models can be manipulated through carefully crafted perturbations, raising great concerns about DNNs' reliability and security.

\begin{figure}[!t]
  \centering
  \begin{subfigure}{0.325\linewidth}
    \includegraphics[width=1\linewidth]{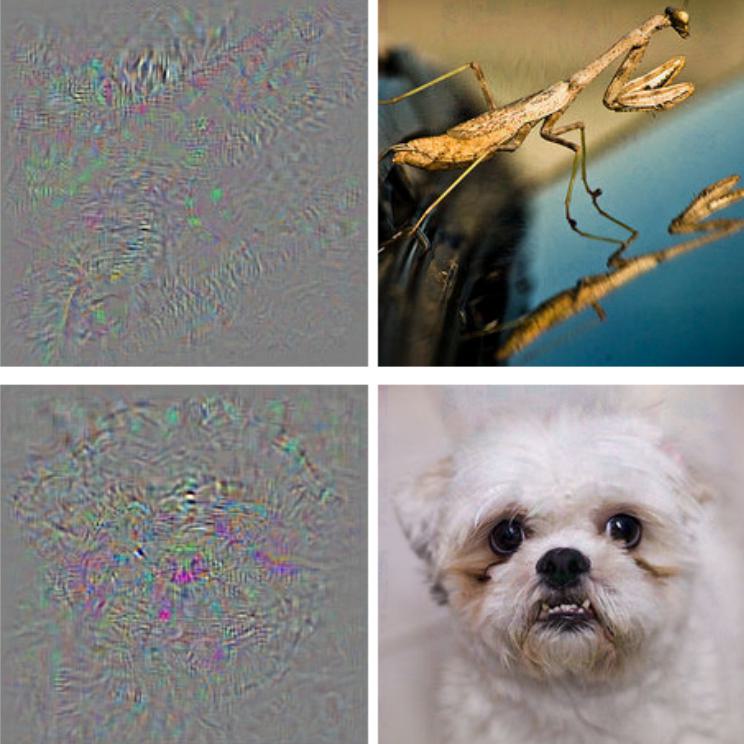}
    \caption{Digital noise\cite{szegedy2014intriguing}}
  \end{subfigure}
  \begin{subfigure}{0.325\linewidth}
    \includegraphics[width=1\linewidth]{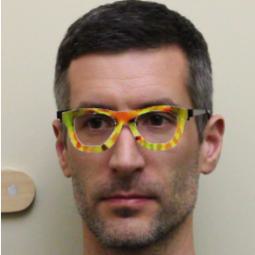}
    \caption{Eyeglass\cite{sharif2016accessorize}}
  \end{subfigure}
  \begin{subfigure}{0.325\linewidth}
    \includegraphics[width=1\linewidth]{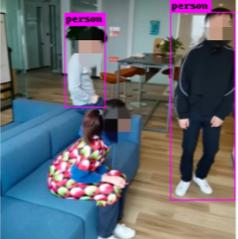}
    \caption{Clothes\cite{hu2022adversarial}}
  \end{subfigure}
  \begin{subfigure}{0.325\linewidth}
    \includegraphics[width=1\linewidth]{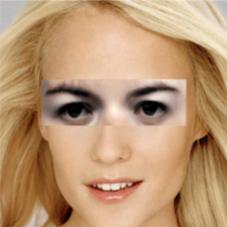}
    \caption{Mask\cite{xiao2021improving}}
  \end{subfigure}
  \begin{subfigure}{0.325\linewidth}
    \includegraphics[width=1\linewidth]{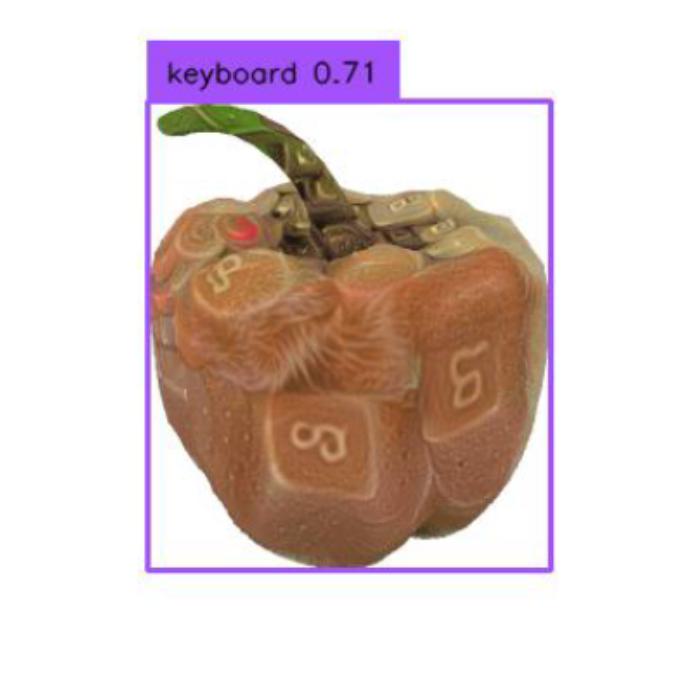}
    \caption{3D Mesh\cite{huang2024towards}}
  \end{subfigure}
  \begin{subfigure}{0.325\linewidth}
    \includegraphics[width=1\linewidth]{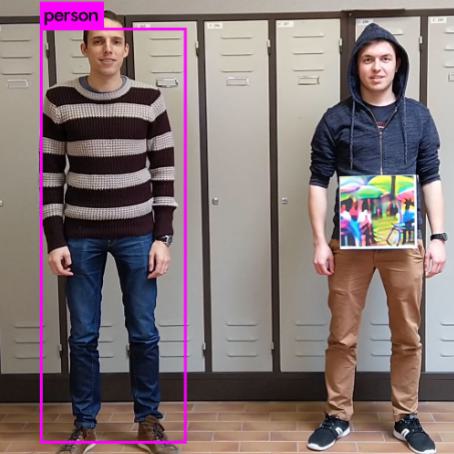}
    \caption{Patch\cite{thys2019fooling}}
  \end{subfigure}
    \begin{subfigure}{0.325\linewidth}
    \includegraphics[width=1\linewidth]{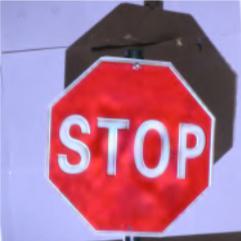}
    \caption{Light\cite{gnanasambandam2021optical}}
  \end{subfigure}
  \begin{subfigure}{0.325\linewidth}
    \includegraphics[width=1\linewidth]{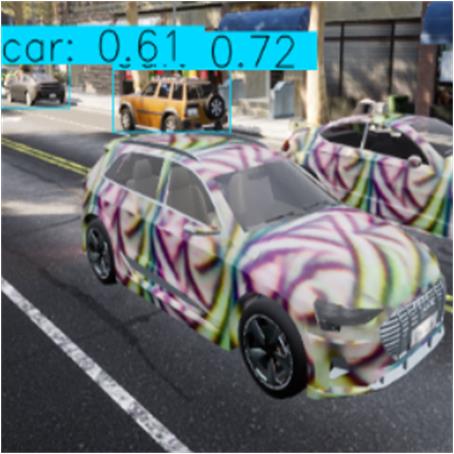}
    \caption{Texture\cite{suryanto2023active}}
  \end{subfigure}
  \begin{subfigure}{0.325\linewidth}
    \includegraphics[width=1\linewidth]{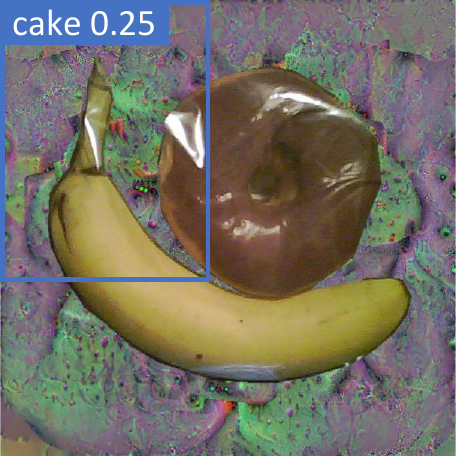}
    \caption{Background (Ours)}
    \label{subfig_background_attack}
  \end{subfigure}
  \caption{
  Comparison of adversarial perturbations in diverse forms.
  (a) conducts digital attacks with imperceptible perturbations entirely covering the images \cite{szegedy2014intriguing}.
  (b)-(h) perform physical attacks by corrupting targeted objects with physical perturbations in various forms \cite{sharif2016accessorize,hu2022adversarial,xiao2021improving,thys2019fooling,gnanasambandam2021optical,wei2022adversarial,huang2024towards}. 
  (i) is our adversarial attack with background perturbation preserving the integrity of the targeted objects.
  }
  \label{fig_1}
\end{figure}

Existing studies \cite{wang2020hamiltonian,bai2020adversarial,wei2023efficient,wei2023adaptive,zhao2023variational} have primarily centered on adversarial attacks that corrupt targeted objects (physical attack) or images (digital attack) as shown in Fig. \ref{fig_1} (a)-(h).
These attacks are designed to be ``visually" camouflaged for DNNs, a strategy that is intuitively plausible and comprehensible given that humans can also be deceived by visually camouflaged objects.
However, an interesting divergence arises when considering the impact of background variations on the targeted objects. 
While such variations do not significantly affect human recognition, DNNs exhibit a high degree of sensitivity to these changes, as exemplified by the banana and donut in Fig. \ref{fig_1} (i). 
This discrepancy underscores a fundamental difference in the role of background features in human and machine vision.
Historically, adversarial attacks have overlooked the potential of exploiting background features, resulting in an incomplete understanding of their role in adversarial contexts. 
Moreover, the prevailing focus on a specific object (physical attack) or whole image (digital attack) manipulation may not sufficiently address the need for generalizing adversarial attacks.
These limitations impede progress in exploring the adversarial robustness of DNNs.

\begin{figure*}[!t]
  \centering
  \begin{subfigure}{0.329\linewidth}
    \includegraphics[width=1\linewidth]{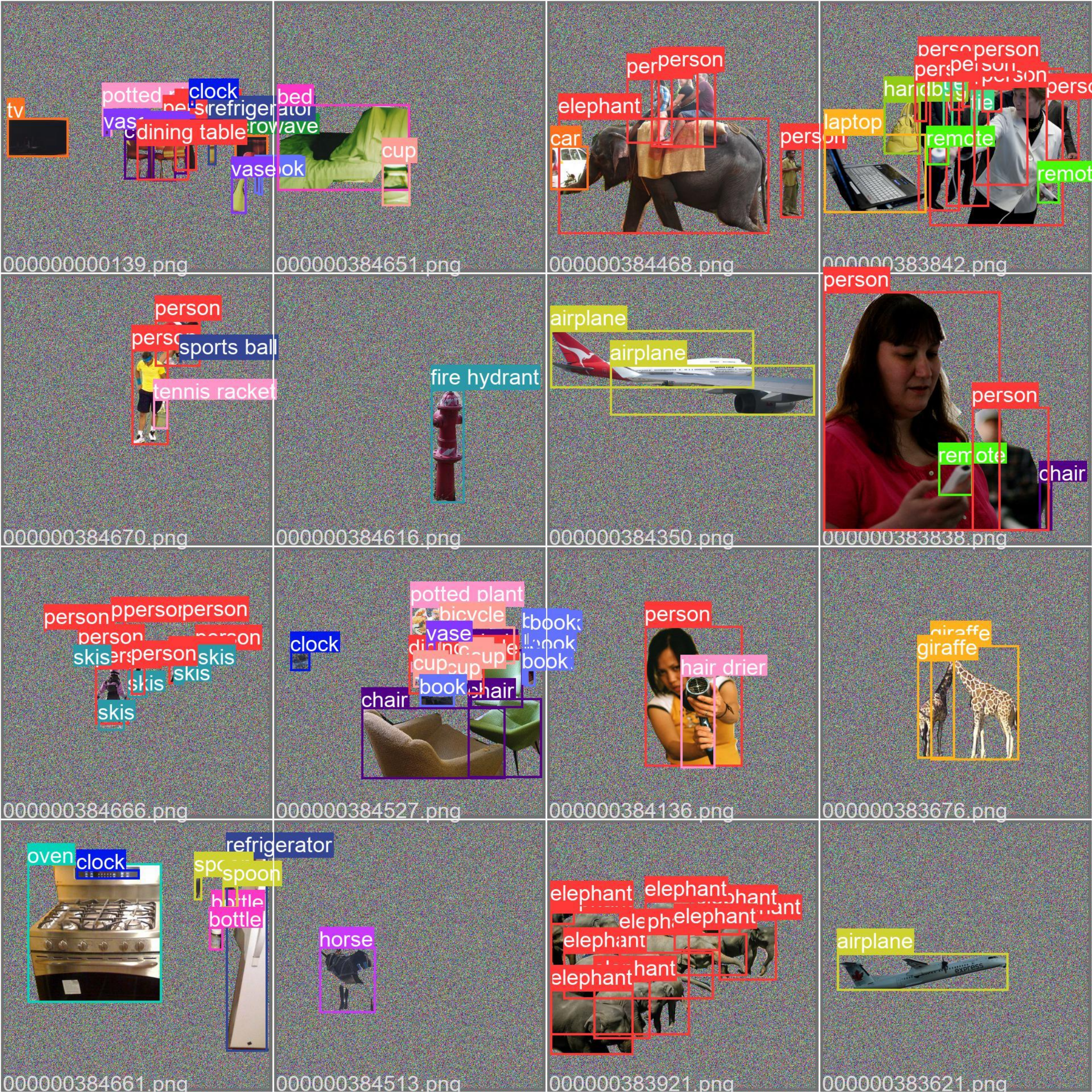}
    \caption{Ground truth}
  \end{subfigure}
  \begin{subfigure}{0.329\linewidth}
    \includegraphics[width=1\linewidth]{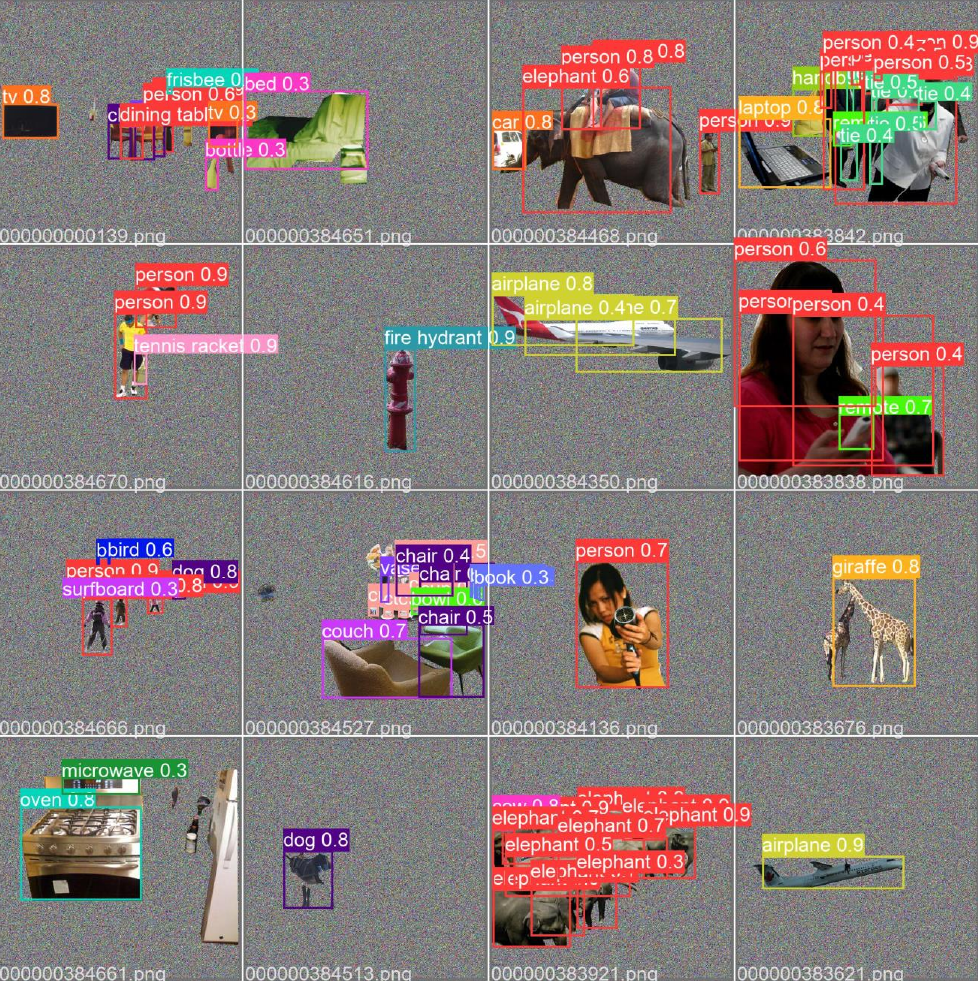}
    \caption{Random background perturbation}
  \end{subfigure}
  \begin{subfigure}{0.329\linewidth}
    \includegraphics[width=1\linewidth]{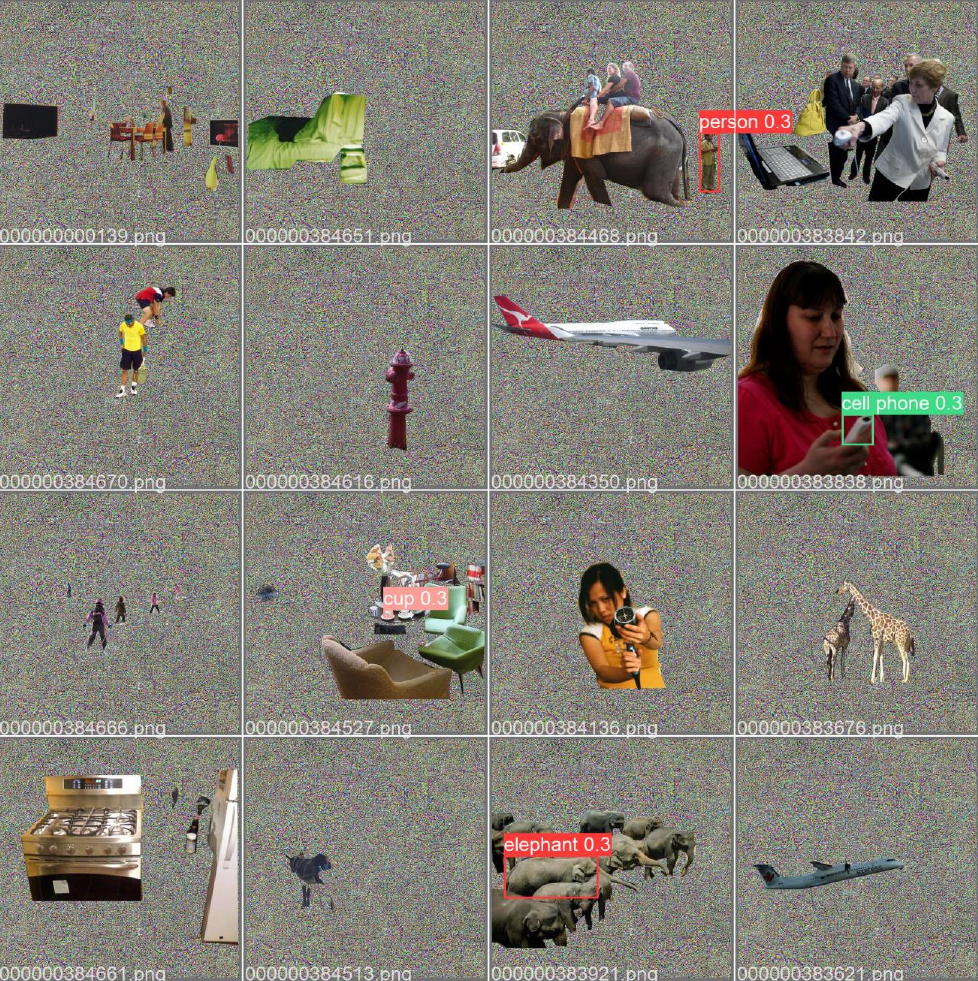}
    \caption{Digital background perturbation}
    \label{fig_digital_background_attack_exhibition}
  \end{subfigure}
  \begin{subfigure}{0.9998\linewidth}
    \includegraphics[width=1\linewidth]{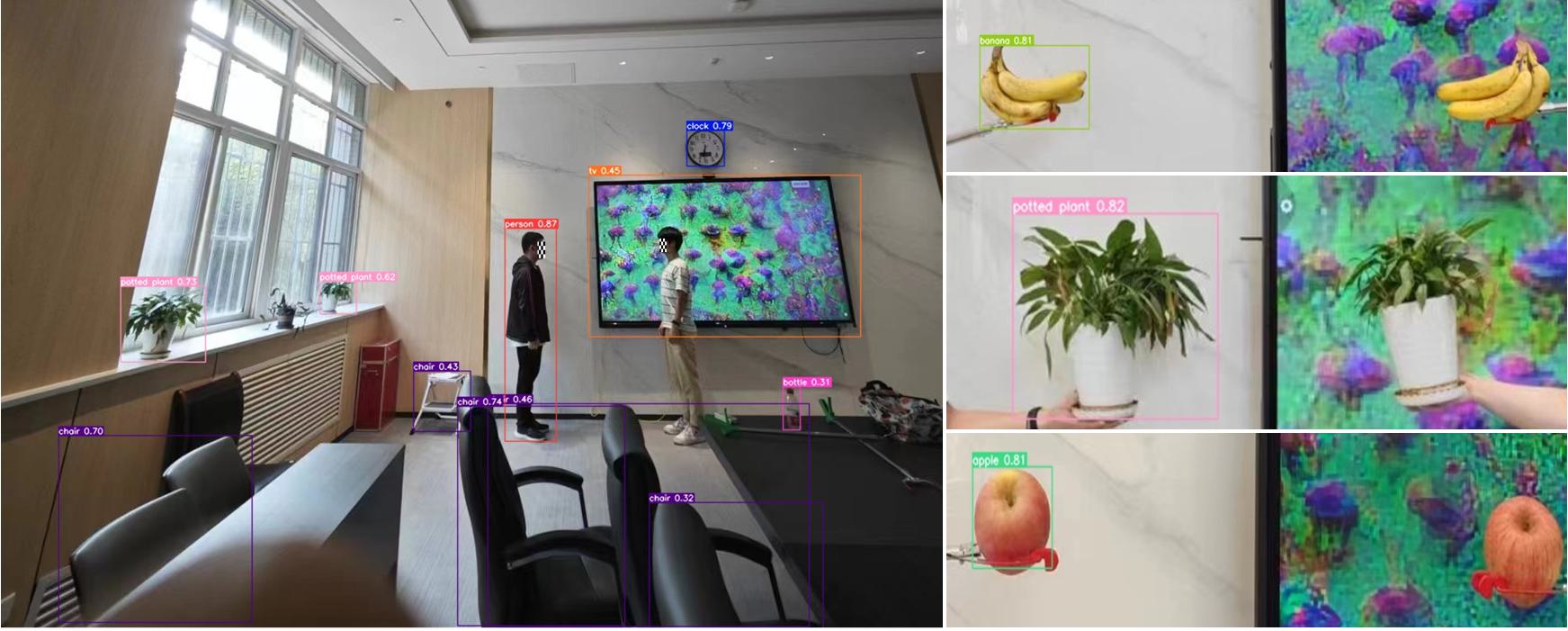}
    \caption{Physical background perturbation}
    \label{fig_physical_background_attack_exhibition}
  \end{subfigure}
  \caption{Proposed background adversarial attack against YOLOv5 \cite{glenn_jocher_2020_4154370} in both digital and physical realms. (a) shows the ground truth. (b) is the detection results of images with random background noise. (c) and (d) are detection results of images under digital and physical background attacks where physical perturbations are displayed by an LED screen. The objectiveness confidence threshold is set as 0.25. Please zoom in for the details.}
  \label{fig_background_attack_exhibition}
\end{figure*}


In this paper, we redirect the attention toward background adversarial attacks that are executed smoothly across digital and physical domains, transferring well across various objects, models, and tasks. 
By manipulating the background environment without directly interfering with objects, we introduce a novel approach to adversarial attacks, i.e., we propose an innovative framework, capitalizing on the untapped potential of background features to deceive DNNs, as shown in Fig. \ref{fig_background_attack_exhibition}. 
Methodologically, we formulate the background adversarial attack as an iterative optimization problem, analogous to the process of DNN learning, and provide a theoretical demonstration of its convergence under certain moderate but sufficient conditions.
To enhance the attack transferability and efficacy, we introduce a novel ensemble strategy tailored to the unique attributes of adversarial perturbations, effectively strengthening their capability in various scenarios. 
Additionally, we propose a sophisticated smooth constraint that ensures the harmonious integration of perturbations.
To validate the efficacy and robustness of the proposed method, we undertake an extensive series of experiments. 
These experiments span across both the digital and physical realms, white-box and black-box conditions, involving diverse objects, models, and tasks. 
The experimental results underscore the formidable effectiveness of the introduced background adversarial attack framework, revealing its potential to disrupt a wide range of AI applications in real-world scenarios.
The implications of our findings extend beyond the realm of adversarial attacks, prompting a profound reevaluation of the principles that underpin DNNs. 
In summary, our contributions are as follows:

\begin{itemize}[left=0pt]
    \item We propose an innovative attack anything paradigm, i.e., blinding DNNs via background adversarial attack, which achieves robust and generalizable attack efficacy across a wide range of objects, models, and tasks.
    \item We conceptualize the background adversarial attack as an iterative optimization problem similar to learning a DNN and theoretically demonstrate its convergence under certain mild but sufficient conditions.
    \item To enhance the attack effectiveness and transferability, we introduce a new ensemble strategy tailored for adversarial perturbations and devise a novel smooth loss to integrate adversarial perturbations seamlessly.
    \item Comprehensive and rigorous experiments are conducted in both digital and physical domains across various objects, models, and tasks, demonstrating the effectiveness of attacking anything of the proposed method.
    \item This work provides substantial evidence that the background feature's significance surpasses our initial expectations, highlighting the need to reassess and further explore the robustness and reliability of DNNs.
\end{itemize}

The remainder of this paper is organized as follows.
Section \ref{sec_backgrounds} briefly reviews adversarial attacks and convergence analysis concerning DNNs.
Section \ref{sec_methodology} details the proposed universal background adversarial attack.
Section \ref{sec_experiments} presents the experimental results and analyses.
Section \ref{sec_discussion} discusses the implications of the findings.
Section \ref{sec_conclusion} concludes the paper. 

\section{Backgrounds}
\label{sec_backgrounds}

In this section, we give the backgrounds of adversarial attacks according to different attack domains (\ref{bg_digital_attack} and \ref{bg_physical_attack}) and convergence analysis concerning DNNs (\ref{bg_convergence_analysis}).

\subsection{Digital Attack}
\label{bg_digital_attack}

The adversarial phenomenon was originally identified from image classification in digital space, which has driven concentrated research on adversarial attacks within this domain.
Adversarial attack methods are presently categorized as gradient-based and optimization-based, depending on the adopted strategy for generating adversarial examples.
Gradient-based adversarial attack techniques, exemplified by the fast gradient sign method (FGSM) \cite{goodfellow2015explaining}, iterative FGSM (I-FGSM) \cite{kurakin2018adversarial}, momentum iterative FGSM (MI-FGSM) \cite{dong2018boosting}, AutoAttack \cite{croce2020reliable}, etc., are designed to generate adversarial perturbations that reside at a significant distance from the decision boundary within predefined perturbation bounds. 
Conversely, optimization-based approaches such as L-BFGS \cite{szegedy2014intriguing}, Deepfool \cite{moosavi2016deepfool}, C\&W \cite{carlini2017towards}, etc., focus on minimizing the magnitude of the adversarial perturbations while adhering to the separation between adversarial and clean examples within a specified perturbation scope. 
Consequently, gradient-based adversarial attack strategies tend to yield more effective misclassifications, whereas the perturbations introduced by optimization-based methods exhibit greater visual imperceptibility.
Additionally, some studies commit to conducting attacks under black-box conditions \cite{dong2021query,shi2022query,williams2023black,yin2023generalizable}, i.e., without prior information about the victim models.
However, prevailing digital attack methods frequently tailor adversarial perturbations individually for each image, and encompass the entirety of the image.

\subsection{Physical Attack}
\label{bg_physical_attack}

Physical adversarial attacks, in contrast, extend the concept of adversarial attacks into the physical realm.  The primary motivation behind physical attacks is to craft physical modifications, causing the deep learning models to be misinterpreted.
Numerous AI systems have fallen under physical attacks, such as face recognition \cite{wei2022simultaneously,li2023physical}, autonomous driving \cite{hu2023physically,hu2022adversarial}, remote sensing \cite{mei2023comprehensive,lian2023contextual}, and so on.
Researchers have demonstrated that by applying adversarially designed stickers \cite{wei2022adversarial,li2019adversarial}, patterns \cite{wang2022fca,yang2023towards}, makeup \cite{zhu2019generating,lin2022real}, light \cite{duan2021adversarial,jin2023pla}, 3D mesh \cite{huang2024towards}, etc., to an object, DNNs-based AI systems can misidentify the object as something entirely different. 
However, the aforementioned physical attacks share a commonality in that they all need to corrupt the targets of interest in varying forms.
Some studies \cite{lian2022benchmarking,xu2022universal,du2022physical} have endeavored to manipulate the backgrounds of targeted objects for adversarial purposes, causing slight sway in the model's predictions, yet often devoid of comprehensive empirical substantiation. 
Additionally, a fraction of these effects might stem from data augmentations beyond the model's training regimen.
Research \cite{lian2023cba} and \cite{wang2024fooling} propose to perform background attack on aerial detection, which achieves comparable performance while lacking theoretical demonstration and comprehensive reflection of the potential of background attack.

\subsection{Convergence Analysis}
\label{bg_convergence_analysis}

Convergence analysis is a critical aspect of studying DNNs. It involves understanding how the iterative learning process of a DNN progresses and whether it will eventually reach a point where the model's parameters no longer change significantly, indicating that the model has learned the underlying patterns in the training data.
Yang et al. \cite{yang2016unified} first explore the convergence of training DNNs with stochastic momentum methods, in particular for non-convex optimization, which fills the gap between practice and theory by developing a basic convergence analysis of two stochastic momentum methods.
Work \cite{zhou2018convergence} provides a fine-grained convergence analysis for a general class of adaptive gradient methods including AMSGrad \cite{reddi2018convergence}, RMSProp \cite{hinton2012rmsprop} and AdaGrad \cite{duchi2011adaptive}.
The authors of \cite{reddi2018convergence} fix the convergence issue of Adam-type algorithms by endowing them with long-term memory of past gradients.
In paper \cite{chen2019convergence}, the researchers develop an analysis framework with sufficient conditions, which guarantee the convergence of the Adam-type methods for non-convex stochastic optimization.

In the context of adversarial attacks, convergence analysis can help understand how the iterative process of crafting adversarial examples progresses and whether it will eventually produce an example that can successfully fool the model. 
This can provide valuable insights for developing more effective and efficient adversarial attack methods.
In work \cite{wang2019convergence}, the researchers propose the First-Order Stationary Condition for constrained optimization (FOSC), which quantitatively evaluates the convergence quality of adversarial examples.
Study \cite{gao2019convergence} partially explains the success of adversarial training by showing its convergence to a network.
Liu et al. \cite{liu2019min} introduce ZO-Min-Max by integrating a zeroth-order (ZO) gradient estimator with an alternating projected stochastic gradient descent-ascent method, which is subject to a sublinear convergence rate under mild conditions and scales gracefully with problem size.
To obtain a smooth loss convergence process, Zhao et al. \cite{zhao2023fast} propose a novel oscillatory constraint to limit the loss difference between adjacent epochs.
Long et al. \cite{long2024convergence} derive a regret upper bound for general convex functions of adversarial attacks.
However, the convergence analysis of adversarial attacks in the context of non-convex functions remains relatively unexplored.
This paper fills the gap between practice and theory by developing a basic convergence analysis of background adversarial attacks, which provides a theoretical illustration of its convergence under certain mild yet adequate conditions.

\section{Methodology}
\label{sec_methodology}

In this section, we first formulate the problem of background adversarial attack in \ref{subsection_problem_formulation} and give a detailed illustration of the proposed paradigm of attack anything in \ref{subsection_attack_anything}.
Then we describe the ensemble strategy in \ref{subsection_ensemble_strategy} and objective loss in \ref{subsection_objective_loss} for attacking anything, respectively.
Finally, we conduct a convergence analysis of the devised background attack in \ref{subsection_convergence_analysis}.

\begin{figure*}[t]
\centering
\includegraphics[width=0.99\linewidth]{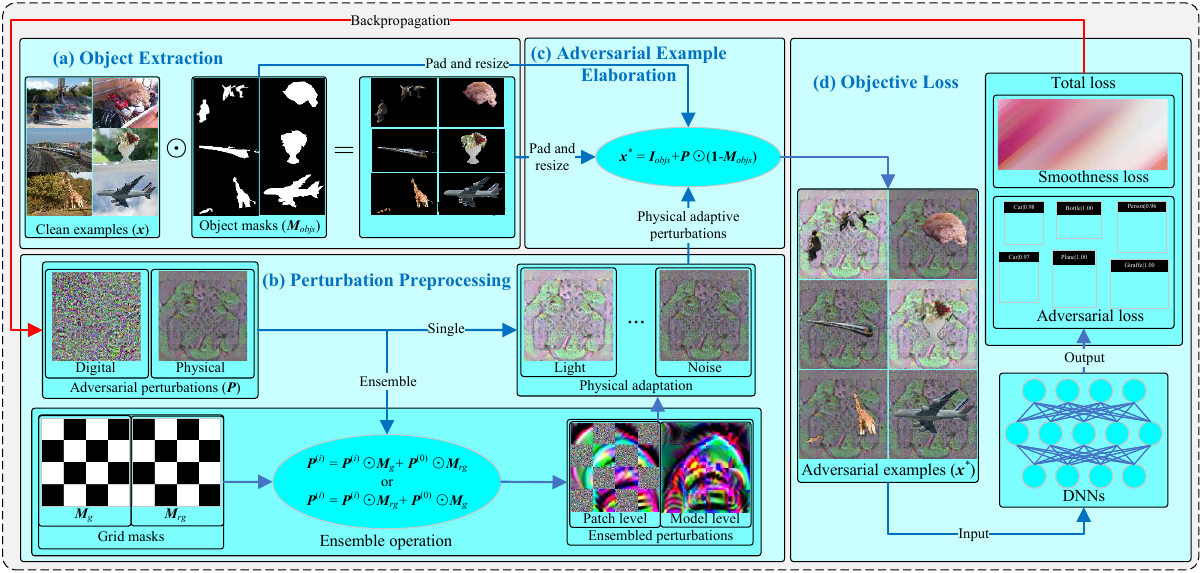} 
\caption{Overall background adversarial attack paradigm.
(a) Object Extraction: we adopt the object's mask to separate the foreground and background regions. 
(b) Perturbation Preprocessing: the adversarial perturbations are preprocessed before elaborating adversarial examples, including physical adaptation and ensemble fortification. 
(c) Adversarial Example Elaboration: adversarial examples are crafted by replacing the background area of the objects with the preprocessed adversarial background perturbation, which is optionally trained in the single or ensemble mode.
(d) Objective Loss: The adversarial examples are fed into the DNNs, and the adversarial loss is extracted from the prediction results. The total loss consists of the adversarial loss and the smoothness loss. The adversarial perturbation is then optimized through backpropagation.
}
\label{fig_overview}
\end{figure*}

\subsection{Problem Formulation}
\label{subsection_problem_formulation}

Previous studies have predominantly focused on carrying out adversarial attacks by directly corrupting targeted objects or images. 
These attacks aim to ``visually" blind DNNs, which is intuitively feasible and understandable since humans can also be deceived by visually camouflaged objects. 
However, an interesting divergence arises when considering the impact of background variations. 
While such variations hardly affect human recognition, DNNs exhibit a high degree of sensitivity to these changes. 
This discrepancy underscores a fundamental difference in the role of background features in the visual perception of humans and machines. 
Historically, adversarial attacks have overlooked the potential of exploiting background features, resulting in an incomplete understanding of their role in adversarial contexts. 
In contrast, this paper redirects the focus toward background adversarial attacks that can easily blind DNNs even without causing any disruptions to the targeted objects themselves.

Technically, we choose object detection as the targeted task as it is a basic computer vision problem and is widely applied in autonomous driving, security surveillance, embodied AI, and other safety-critical applications.
Our background adversarial attack aims to hide the targeted objects from being detected, i.e., the targeted objects are misrecognized as no-objects or backgrounds.
We denote by $D:\mathbb R^m \longrightarrow \left\{[\textbf{\textit{l}}_1, s^{conf}_1, \textbf{\textit{p}}_1^{cls}], \cdots, [\textbf{\textit{l}}_k, s^{conf}_k, \textbf{\textit{p}}_k^{cls}]\right\}$ an object detector $D$ mapping image tensors belong to $\mathbb R^m$, where $m$ represents the dimensionality of the input image, to a discrete detected object set, including object's \textbf{l}ocation $\textbf{\textit{l}}$, objectiveness \textbf{s}core $s$, and category \textbf{p}robabilities $\textbf{\textit{p}}$.
For a given adversarial example $\boldsymbol{x}^* \in \mathbb R^m$, the attack purpose is mathematically defined as:
\begin{equation}
    \label{eq_recognize_nothing}
    D(\boldsymbol{x}^*, \boldsymbol{\theta}) = \left\{[\textbf{\textit{l}}_1, s^{conf}_1, \textbf{\textit{p}}_1^{cls}], \cdots, [\textbf{\textit{l}}_k, s^{conf}_k, \textbf{\textit{p}}_k^{cls}]\right\} \longrightarrow \varnothing,
\end{equation}
where $D(\cdot)$ is parameterized with $\boldsymbol{\theta}$, $\varnothing$ means recognition results are no-objects or background.
To achieve the aforementioned attack purpose, we construct the objective loss of the background adversarial attack as $L(D(\boldsymbol{x}^*, \boldsymbol{\theta}), \boldsymbol{x}^*)$, which is concretely explained in Sec. \ref{subsection_objective_loss}.
We mathematically formulate this attack as an optimization problem similar to training a DNN as follows:
\begin{equation}
    \label{eq_optimization_problem}
    \mathop{\arg\min}\limits_{\boldsymbol{x}^*} L(D(\boldsymbol{x}^*, \boldsymbol{\theta}), \boldsymbol{x}^*) \quad s.t. \quad \boldsymbol{x}^* \in [0, 1]^m.
\end{equation}
Then comes the problem of designing adversarial example $\boldsymbol{x}^*$.
Given a benign example $\boldsymbol{x}$, we aim to blind detectors from detecting anything via background adversarial attack.
Technically, we craft adversarial example $\boldsymbol{x}^*$ by adding elaborated background perturbations $\boldsymbol{P}$ to the benign example $\boldsymbol{x}$, which is formulated as:
\begin{equation}
    \label{eq_problem_formulation}
    \boldsymbol{x}^* = \boldsymbol{x} \odot \boldsymbol{M}_{objs} +  \boldsymbol{P} \odot \boldsymbol{M}_{bg},
\end{equation}
where $\boldsymbol{M}_{objs}$ and $\boldsymbol{M}_{bg}$ are the masks of objects and background respectively, and $\boldsymbol{M}_{objs}+\boldsymbol{M}_{bg}=\boldsymbol{1}$.
$\odot$ means Hadamard product.
Then we aim to generate background adversarial perturbations $\boldsymbol{P}$.
The optimization problem can be expressed as:
\begin{equation}
    \label{eq_optimization_problem2}
    \begin{split}
    \mathop{\arg\min}\limits_{\boldsymbol{P}} L(D(\boldsymbol{x}, \boldsymbol{M}_{objs}, \boldsymbol{M}_{bg}, \boldsymbol{P}, \boldsymbol{\theta}), \boldsymbol{P})\\ \quad s.t. \quad \boldsymbol{P} \in [0, 1]^m.
    \end{split}
\end{equation}
Considering the background perturbations will be iteratively trained in batch form with a large dataset, where iteration number and batch size are $T$ and $B$, the optimization problem of background perturbation can be revised to 
\begin{equation}
    \label{eq_optimization_problem_iterative}
    \begin{split}
    \mathop{\arg\min}\limits_{\boldsymbol{P}} \sum\limits_{t=1}^T \sum\limits_{b=1}^{B_t} L(D(\boldsymbol{x}^{*}_{tb}, \boldsymbol{\theta}), \boldsymbol{P})\\ \quad s.t. \quad \boldsymbol{P} \in [0, 1]^m,
    \end{split}
\end{equation}
which is shorted as:
\begin{equation}
    \label{eq_optimization_problem_shorted}
    \begin{split}
    \mathop{\arg\min}\limits_{\boldsymbol{P}} \sum\limits_{t=1}^T f_t(\boldsymbol{P}) = \mathop{\arg\min}\limits_{\boldsymbol{P}} f(\boldsymbol{P}) \\ \quad s.t. \quad \boldsymbol{P} \in [0, 1]^m.
    \end{split}
\end{equation}

\begin{algorithm}[tb]
\caption{Attack Anything (AA)}
\label{alg_algorithm}
\textbf{Input}: DNNs-based detector $D(\cdot)$, clean example $\boldsymbol{x}$, initial perturbation $\boldsymbol{P}^{(0)}$, loss function $L$, grid mask $\boldsymbol{M}_g$, reversed grid mask $\boldsymbol{M}_{rg}$, and objects mask $\boldsymbol{M}_{objs}$. \\
\textbf{Parameter}: Iteration number $T$, hyperparameter $\alpha, \lambda, \eta$.\\
\textbf{Output}: Background perturbation $\boldsymbol{P}$.

\begin{algorithmic}[1] 
    \FOR{$i=0$ to $T$}
        \IF{Ensemble}
        \STATE $\boldsymbol{P}^{(i)} = \boldsymbol{P}^{(i)} \odot \boldsymbol{M}_{g} + \boldsymbol{P}^{(0)} \odot \boldsymbol{M}_{rg}$ or \\
        $\boldsymbol{P}^{(i)} = \boldsymbol{P}^{(i)} \odot \boldsymbol{M}_{rg} + \boldsymbol{P}^{(0)} \odot \boldsymbol{M}_{g}$;
        \ENDIF
        \STATE $\boldsymbol{P}^{(i)} = PA({\boldsymbol{P}^{(i)}})$;
        \STATE $\boldsymbol{x}^*_{i} = \boldsymbol{x}_{i} \odot \boldsymbol{M}_{objs} + \boldsymbol{P}^{(i)} \odot (\boldsymbol{1} - \boldsymbol{M}_{objs})$;
        \STATE $[\boldsymbol{x}^1_{i},\boldsymbol{y}^1_{i},\boldsymbol{x}^2_{i},\boldsymbol{y}^2_{i},\boldsymbol{s}^{conf}_{i},\boldsymbol{p}^{cls}_{i}] \leftarrow D(\boldsymbol{x}^*_{i})$;
        \STATE $L_{obj}, L_{box} \leftarrow [\boldsymbol{x}^1_{i},\boldsymbol{y}^1_{i},\boldsymbol{x}^2_{i},\boldsymbol{y}^2_{i},\boldsymbol{s}^{conf}_{i},\boldsymbol{p}^{cls}_{i}]$;
        \STATE $L = L_{obj} + \eta \cdot L_{abtv} + \lambda \cdot L_{box}$;
        \STATE $\mathbf{g}_i = \nabla \sum\limits_{b=1}^{B_i} L$;
        \STATE $\boldsymbol{m}^{(i)} = \beta_1 \cdot \boldsymbol{m}^{(i-1)} + (1-\beta_1) \cdot \mathbf{g}_i$;
        \STATE $\boldsymbol{v}^{(i)} = \beta_2 \cdot \boldsymbol{v}^{(i-1)} + (1-\beta_2) \cdot \mathbf{g}_i^2$;
        \STATE $\hat{\boldsymbol{m}}^{(i)} = \frac{\boldsymbol{m}^{(i)}}{1-\beta_1^i}$;
        \STATE $\hat{\boldsymbol{v}}^{(i)} = \max(\hat{\boldsymbol{v}}^{(i-1)}, \frac{\boldsymbol{v}^{(i)}}{1-\beta_2^i})$;
        \STATE $\boldsymbol{P}^{(i+1)} = \boldsymbol{P}^{(i)}-\boldsymbol{\alpha}_i \cdot \frac{\hat{\boldsymbol{m}}^{(i)}}{\sqrt{\hat{\boldsymbol{v}}^{(i)}}+\epsilon}$;
    \ENDFOR
    \STATE $\boldsymbol{P} = \boldsymbol{P}^{(T)}$;
    \STATE \textbf{return} $\boldsymbol{P}$.
\end{algorithmic}
\end{algorithm}

\subsection{Attack Anything}
\label{subsection_attack_anything}

To blind DNNs, we design an attack anything paradigm via manipulating contextual background features.
The overview of the devised paradigm is displayed in Fig. \ref{fig_overview}.
Firstly, we randomly initiate the background perturbation $\boldsymbol{P}$.
To overcome the loss of attack efficacy caused by cross-domain transformation, we conduct physical adaptation $PA(\cdot)$ to simulate dynamic conditions in real-world scenarios, such as varying lighting conditions, various physical noises, etc., similar to \cite{thys2019fooling}.
Next, the adversarial examples $\boldsymbol{x}^*$ are fed into the object detector $D(\cdot)$. 
We then decompose the detection results and further process them as the adversarial losses $L_{obj}$ and $L_{box}$.
Additionally, an adaptive bi-directional smooth loss $L_{abtv}$ is introduced to bridge the gap between adjacent pixels in perturbations, which cannot be properly captured by imaging devices.
Consequently, the total loss $L$ consists of adversarial loss ($L_{obj}$ and $L_{box}$) and smoothness loss ($L_{abtv}$).
Finally, the background perturbation $\boldsymbol{P}$ is iteratively optimized using the gradient descent algorithm. 

\begin{figure}[!htbp]
  \centering
  \begin{subfigure}{0.325\linewidth}
    \includegraphics[width=1\linewidth]{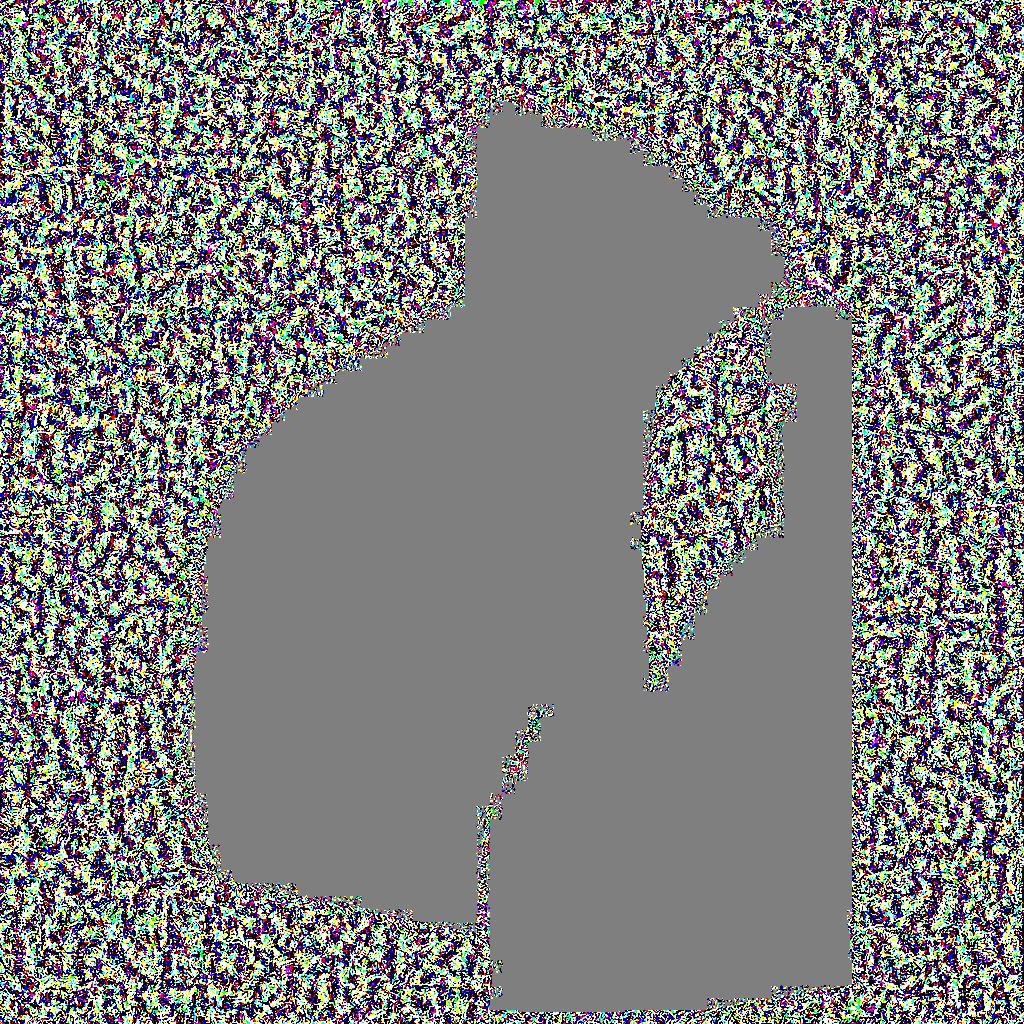}
    \caption{}
  \end{subfigure}
  \begin{subfigure}{0.325\linewidth}
    \includegraphics[width=1\linewidth]{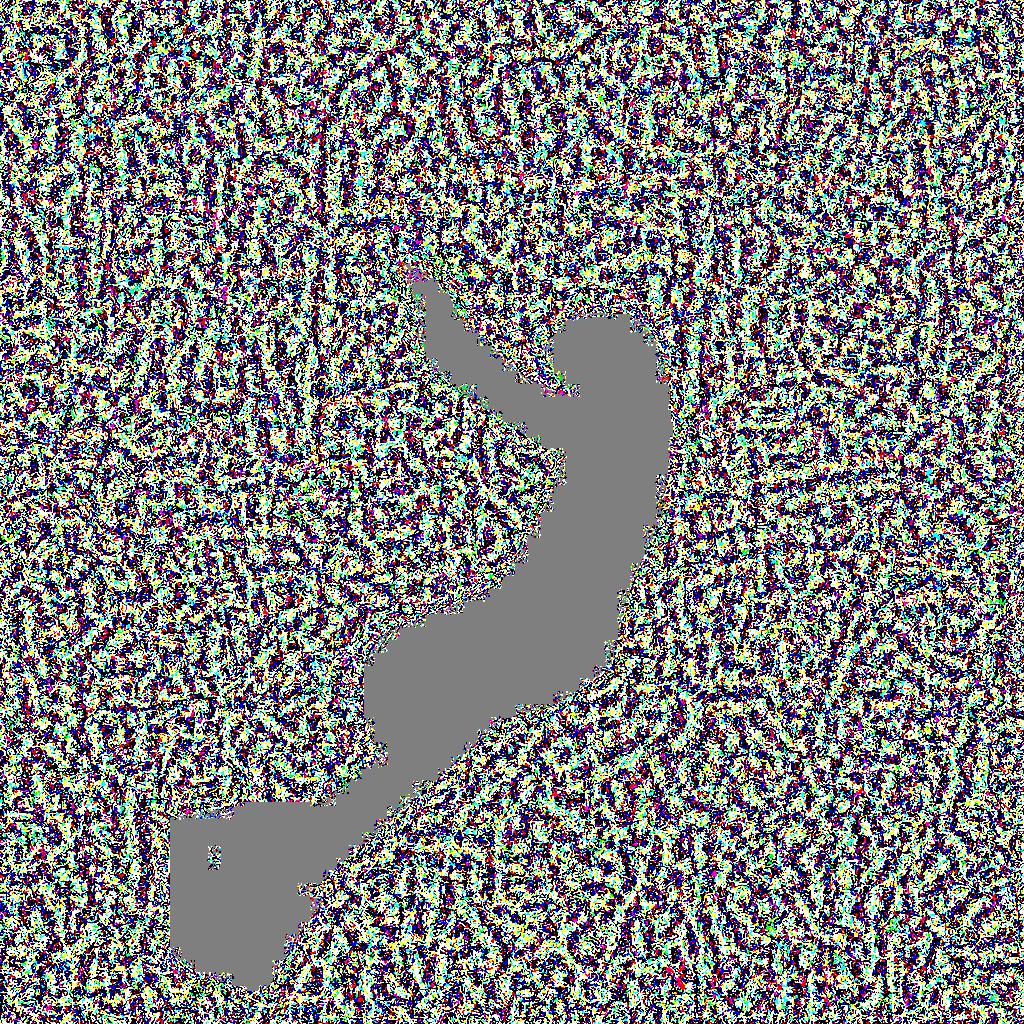}
    \caption{}
  \end{subfigure}
  \begin{subfigure}{0.325\linewidth}
    \includegraphics[width=1\linewidth]{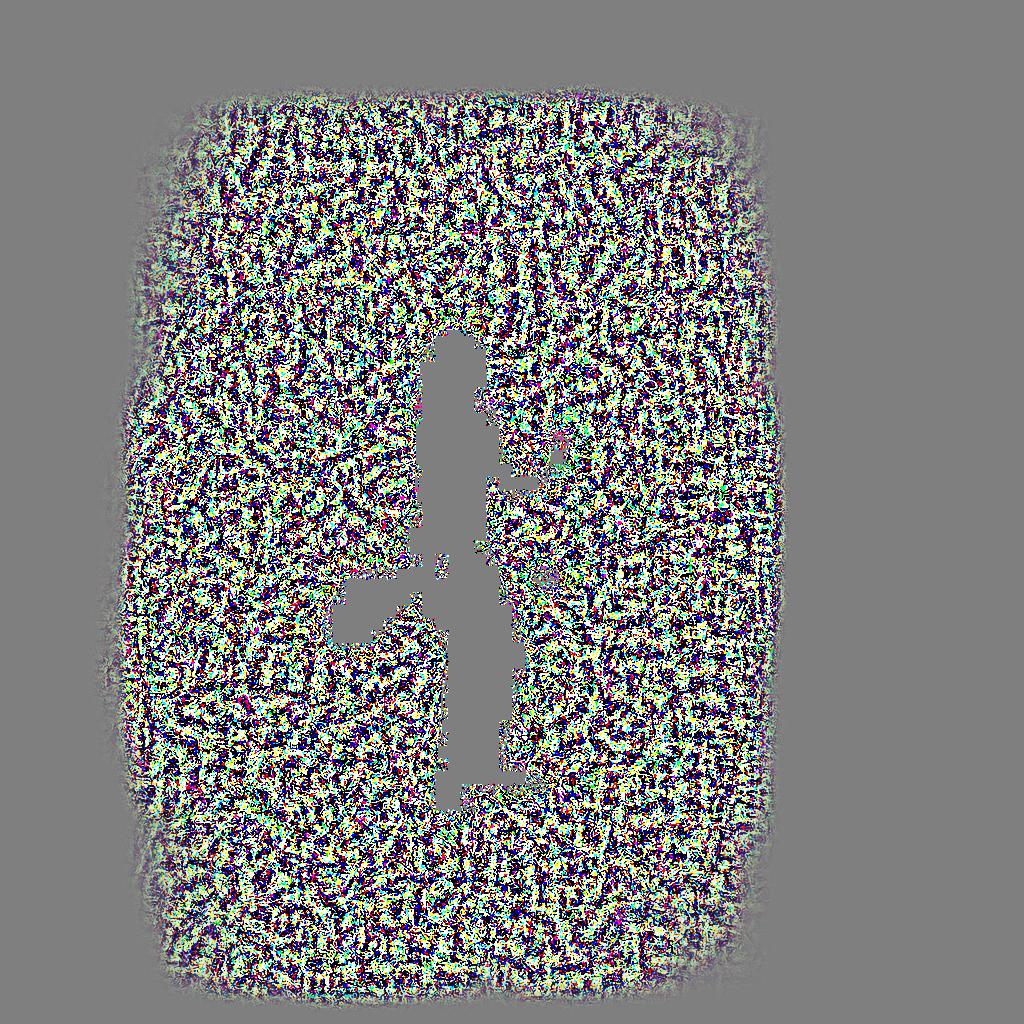}
    \caption{}
  \end{subfigure}
  \caption{Dropout operation for perturbation optimization, in which the pixels of object area are suppressed in an iteration.}
  \label{fig_dropout}
\end{figure}

AMSGrad \cite{reddi2018convergence} is adopted as the optimizer, which is an improved version of Adam \cite{kingma2014adam} by retaining the original performance of Adam to the greatest extent while overcoming its convergence analysis issues even in the non-convex setting.
The optimization process is detailed as follows.
Firstly, the gradient $\mathbf{g}_t$ is computed by Eq. \ref{eq_g}.
\begin{equation}
    \label{eq_g}
    \begin{split}
    \begin{aligned}
        \mathbf{g}_t & = \nabla \sum\limits_{b=1}^{B_t} L(D(\boldsymbol{x}, \boldsymbol{M}_{objs}, \boldsymbol{M}_{bg}, \boldsymbol{P}^{(t)}, \boldsymbol{\theta}), \boldsymbol{P}^{(t)})
        \\ & = \nabla f_t({\boldsymbol{P}^{(t)}}), 
    \end{aligned}
    \end{split}
\end{equation}
where $\boldsymbol{P}^{(0)}$ is randomly initialized.
Secondly, the first and second moments $\boldsymbol{m}^{(t)}$ and $\boldsymbol{v}^{(t)}$ are updated by Eqs. \ref{eq_m} and \ref{eq_v}.
\begin{equation}
    \label{eq_m}
    \boldsymbol{m}^{(t)} = \beta_1 \cdot \boldsymbol{m}^{(t-1)} + (1-\beta_1) \cdot \mathbf{g}_t, \ \boldsymbol{m}^{(0)} = \boldsymbol{0},
\end{equation}
\begin{equation}
    \label{eq_v}
    \boldsymbol{v}^{(t)} = \beta_2 \cdot \boldsymbol{v}^{(t-1)} + (1-\beta_2) \cdot \mathbf{g}_t^2, \ \boldsymbol{v}^{(0)} = \boldsymbol{0},
\end{equation}
where the hyperparameters $\beta_1$ and $\beta_2$ are the exponential decay rates of the first and second moments, respectively.
Thirdly, the bias-corrected moments $\hat{\boldsymbol{m}}^{(t)}$ and $\hat{\boldsymbol{v}}^{(t)}$ are calculated by Eqs. \ref{eq_m_hat} and \ref{eq_v_hat}.
\begin{equation}
    \label{eq_m_hat}
    \hat{\boldsymbol{m}}^{(t)} = \frac{\boldsymbol{m}^{(t)}}{1-\beta_1^t},
\end{equation}
\begin{equation}
    \label{eq_v_hat}
    \hat{\boldsymbol{v}}^{(t)} = \max(\hat{\boldsymbol{v}}^{(t-1)}, \frac{\boldsymbol{v}^{(t)}}{1-\beta_2^t}).
\end{equation}
Finally, the perturbation $\boldsymbol{P}^{(t+1)}$ is optimized by Eq. \ref{eq_theta}.
\begin{equation}
    \label{eq_theta}
    \boldsymbol{P}^{(t+1)} = \boldsymbol{P}^{(t)}-\boldsymbol{\alpha}_t \cdot \frac{\hat{\boldsymbol{m}}^{(t)}}{\sqrt{\hat{\boldsymbol{v}}^{(t)}}+\epsilon},
\end{equation}
where $\epsilon$ is a small constant added for numerical stability.
Please refer to AMSGrad \cite{reddi2018convergence} for more details.
The optimization process is iteratively conducted until the perturbation converges or the maximum iteration number is reached.
The previous attack methods mainly optimize $\boldsymbol{P}$ by placing it on the targets of interest or covering the entire image, while we put targeted objects on the background perturbations.
Through this approach, certain regions of the perturbations become selectively suppressed in each training iteration as shown in Fig. \ref{fig_dropout}, bearing a resemblance to the underlying principles of dropout \cite{srivastava2014dropout} employed in DNNs' training.

Algorithm \ref{alg_algorithm} summarizes the overall optimization scheme of the devised attack anything framework, where the ensemble operation detailed in the following section is optional for fortifying attack efficacy and transferability.

\subsection{Ensemble Strategy}
\label{subsection_ensemble_strategy}

\begin{figure}[t]
\centering
\includegraphics[width=0.8\linewidth]{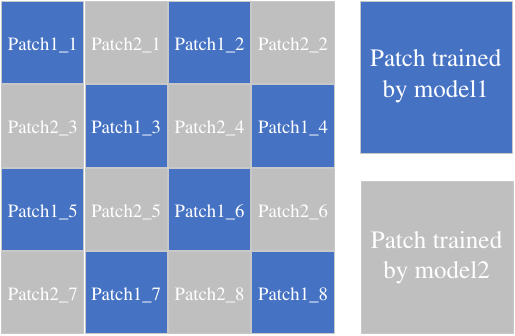} 
\caption{Illustration of the two-level ensemble strategy.}
\label{fig_ensemble}
\end{figure}

To strengthen attack efficacy and transferability, we design a novel ensemble strategy customized for adversarial perturbations, as shown in Fig \ref{fig_ensemble}.
Specifically, we use a pair of opposite grid masks to separate the background perturbations into $n \times n$ small patches.
We take $n=4$ as an example.
Then, We first optimize the non-adjacent 8 among the 16 patches, as the ensemble operation shown in Fig. \ref{fig_overview} (b), which can be deemed as an ensemble at the patch level and mathematically written as:
\begin{equation}
    \label{eq_ensemble_patch_level}
    \boldsymbol{P}^{(i)} = \boldsymbol{P}^{(i)} \odot \boldsymbol{M}_{g} + \boldsymbol{P}^{(0)} \odot \boldsymbol{M}_{rg}.
\end{equation}
Next, the rest 8 of the 16 patches are trained with a different model, which is viewed as another ensemble at the model level and mathematically written as:
\begin{equation}
    \label{eq_ensemble_model_level}
    \boldsymbol{P}^{(i)} = \boldsymbol{P}^{(i)} \odot \boldsymbol{M}_{rg} + \boldsymbol{P}^{(0)} \odot \boldsymbol{M}_{g}.
\end{equation}
After the ensemble operation, the perturbation will be sent to the next procedure.

\subsection{Objective Loss}
\label{subsection_objective_loss}

\subsubsection{Adversarial Loss}

In this work, the adversarial loss consists of the objectiveness loss $L_{obj}$ and the bounding box loss $L_{box}$. The objective is to deceive DNNs into not detecting any objects. 
If there are any objects detected, the goal is to minimize their confidence scores and bounding boxes.
Specifically, we use all objectiveness scores of detected objects, including every object of all classes, to calculate the \textbf{objectiveness loss}, which is defined as:
\begin{equation}
    \label{eq_box_loss}
    L_{obj} = \frac{1}{N_c} \sum\limits_{j=1}^{N_c} \frac{1}{N_{c_j}} \sum\limits_{i=1}^{N_{c_j}} s^{conf}_{j,i},
\end{equation}
where $N_c$ represents the number of detected classes and $N_{c_j}$ is the number of objects in detected class $j$.

For \textbf{bounding box loss}, we adopt the width and height of the bounding box weighted by their corresponding confidence score as box loss, i.e., the higher the corresponding confidence score of the bounding box, the bigger the corresponding box loss, which is calculated as:
\begin{equation}
    \label{eq_objectiveness_loss}
    L_{box} = \frac{1}{N_c} \sum\limits_{j=1}^{N_c} \frac{1}{N_{c_j}} \sum\limits_{i=1}^{N_{c_j}} s^{conf}_{j,i} \cdot (|x^2_{j,i} - x^1_{j,i}| + |y^2_{j,i} - y^1_{j,i}|).
\end{equation}
The adversarial loss of the proposed paradigm can be flexibly customized according to attackers' desire.

\subsubsection{Smoothness Loss}

To ensure the smoothness of the generated perturbations, we utilize the total variation (TV) \cite{mahendran2015understanding} to fill the gap between adjacent pixels.
The $L_{tv}$ of background perturbation is defined as:
\begin{equation}
    \label{eq_tv_loss}
    L_{tv} = \sum\limits_{j,i} (p_{j+1,i}-p_{j,i})^2 + (p_{j,i+1}-p_{j,i})^2,
\end{equation}
where $p_{j,i}$ is the pixel value of $\boldsymbol{P}$ at position $(j,i)$.

\begin{figure}[!htbp]
  \centering
  \begin{subfigure}{0.24\linewidth}
    \includegraphics[width=1\linewidth]{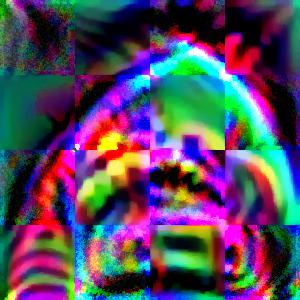}
    \caption{Not smooth}
  \end{subfigure}
  \begin{subfigure}{0.24\linewidth}
    \includegraphics[width=1\linewidth]{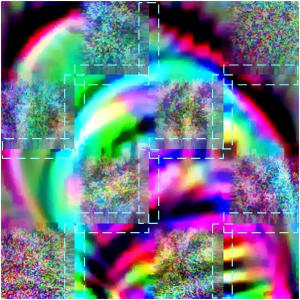}
    \caption{Half-smooth}
  \end{subfigure}
  \begin{subfigure}{0.24\linewidth}
    \includegraphics[width=1\linewidth]{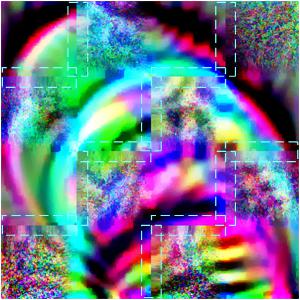}
    \caption{Half-smooth}
  \end{subfigure}
  \begin{subfigure}{0.24\linewidth}
    \includegraphics[width=1\linewidth]{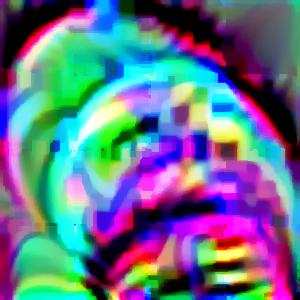}
    \caption{Smooth}
  \end{subfigure}
  \caption{Comparison of ensembled perturbations with different smoothness loss. Please zoom in for a better view.}
  \label{fig_different_tv}
\end{figure}

However, grid artifacts are observed in the perturbations generated through ensemble operations, as depicted in Figure \ref{fig_different_tv} (a). 
This indicates that the previously applied Total Variation (TV) loss is insufficient for effectively smoothing the concatenated perturbations. 
To address this issue, we propose a distance-adaptive smoothness loss tailored for the ensembled perturbations. 
This approach involves assigning a higher smoothness weight, denoted as $w$, to pixels proximal to the integration boundaries, indexed by $k \in \left\{k_1, k_2, \ldots, k_{n-1}\right\}$, where $n$ signifies the count of ensemble patches per row or column, and the proximity is defined within a distance $\delta$. 
The formulation of the \textbf{a}daptive total variation is as follows:
\begin{equation}
    \label{eq_distance_adaptive_smooth_loss}
    L_{atv} = \sum\limits_{j,i} (p_{j+1,i}-p_{j,i})^2 \cdot w_j + (p_{j,i+1}-p_{j,i})^2 \cdot w_i,
\end{equation}
where the adaptive weight $w_j$ is calculated as:
\begin{equation}
    w_{i}=\begin{cases}
      1, & |i-k|\geq\delta\\
      \frac{\delta}{|i-k|+\epsilon}, & 0<|i-k|<\delta\\
      \delta, & |i-k|=0
    \end{cases},
\end{equation}
where $\epsilon$ is a small constant added for numerical stability.
$w_j$ is calculated similarly.

Additionally, we discover the directionality of the smoothness loss from the generated half-smooth perturbations by Eq. \ref{eq_distance_adaptive_smooth_loss}, as shown in Fig. \ref{fig_different_tv} (b) and (c).
We accommodate this problem by introducing an \textbf{a}daptive \textbf{b}i-directional total variation as:
\begin{equation}
    \label{eq_distance_adaptive_full_smooth_loss}
    \begin{split}
    L_{abtv} = \sum\limits_{j,i} ((p_{j+1,i}-p_{j,i})^2+(p_{j,i}-p_{j+1,i})^2) \cdot w_j + \\((p_{j,i+1}-p_{j,i})^2+(p_{j,i}-p_{j,i+1})^2) \cdot w_i,        
    \end{split}
\end{equation}
by which the generated full-smooth perturbation is exhibited as Fig. \ref{fig_different_tv} (d).

\subsubsection{Total Loss}
Overall, the total loss is formulated as:
\begin{equation}
    \label{eq_total_loss}
    L = L_{obj} + \eta \cdot L_{abtv} + \lambda \cdot L_{box},
\end{equation}
where $\eta$ and $\lambda$ are adopted to balance different parts of the total loss.

\subsection{Convergence Analysis}
\label{subsection_convergence_analysis}

This section treats the proposed background adversarial attack as a non-convex optimization problem and theoretically demonstrates its convergence.
We formalize the \textbf{assumptions} required in the convergence analysis based on the commonality between DNN training \cite{chen2019convergence} and perturbations generation as follows:

\textbf{A1}: The objective function $f(\boldsymbol{P})$ is the global loss function, defined as:
\begin{equation}
    f(\boldsymbol{P})=\lim\limits_{T\longrightarrow\infty}\frac{1}{T}\sum\limits_{t=1}^{T}f_t(\boldsymbol{P}),
\end{equation}
where $f_t(\boldsymbol{P})$ denotes the loss function updated at the $t$th iteration for $t=1,2,\cdots,T$.
$f(\boldsymbol{P})$ is a non-convex but L-smooth function, i.e., it satisfies \textit{1)} $f(\boldsymbol{P})$ is differentiable, namely $\nabla f$ exists everywhere within the defined domain, and \textit{2)} exists $L > 0$, for any $\boldsymbol{P}_1$ and $\boldsymbol{P}_2$ within the defined domain satisfy:
\begin{equation}
\label{lpc_ctn_1}
    f\left(\boldsymbol{P}_{2}\right)\leq f\left(\boldsymbol{P}_{1}\right)+\left\langle \nabla f\left(\boldsymbol{P}_{1}\right),\boldsymbol{P}_{2}-\boldsymbol{P}_{1}\right\rangle +\frac{L}{2}\left\Vert \boldsymbol{P}_{2}-\boldsymbol{P}_{1}\right\Vert _{2}^{2}
\end{equation}
and
\begin{equation}
\label{lpc_ctn_2}
    \left\Vert \nabla f\left(\boldsymbol{P}_{1}\right)-\nabla f\left(\boldsymbol{P}_{2}\right)\right\Vert _{2}\leq L\left\Vert \boldsymbol{P}_{1}-\boldsymbol{P}_{2}\right\Vert _{2},
\end{equation}
which is also known as Lipschitz continuous.

\textbf{A2}: The background perturbations are bounded:
\begin{equation}
    \left\Vert \boldsymbol{P}-\boldsymbol{P}^{\prime}\right\Vert _{2}\leq D,\,\,\forall\boldsymbol{P},\boldsymbol{P}^{\prime}
\end{equation}
or for each dimension $i$ is subject to
\begin{equation}
    \left\Vert P_{i}-P_{i}^{\prime}\right\Vert_{2}\leq D_{i},\,\,\forall P_{i},P_{i}^{\prime}.
\end{equation}

\textbf{A3}: The gradients are bounded:
\begin{equation}
    \left\Vert \nabla f\left(\boldsymbol{P}^{(t)}\right)\right\Vert _{2}\leq G,\,\forall t,
\end{equation}
\begin{equation}
    \left\Vert \mathbf{g}_t\right\Vert _{2}\leq G,\,\forall t,
\end{equation}
\begin{equation}
     \left\Vert \mathbf{g}_{1}\right\Vert _{2}\geq c,
\end{equation}
or for each dimension $i$ is subject to
\begin{equation}
    \left\Vert \left[\nabla f\left(\boldsymbol{P}^{(t)}\right)\right]_{i}\right\Vert _{2}\leq G_{i},\,\forall t,
\end{equation}
\begin{equation}
    \left\Vert g_{t,i}\right\Vert _{2}\leq G_{i},\,\forall t,
\end{equation}
\begin{equation}
    \left\Vert g_{1,i}\right\Vert _{2}\geq c,
\end{equation}
where $c$ is the lower bound of the gradients.

\textbf{A4}: The index that determines convergence is a statistic $E\left(T\right)$:
\begin{equation}
    E\left(T\right)=\min_{t=1,2,\ldots,T}\mathbb{E}_{t-1}\left[\left\Vert \nabla f\left(\boldsymbol{P}^{(t)}\right)\right\Vert _{2}^{2}\right].
\end{equation}
When $T\rightarrow\infty$, if $E\left(T\right)/T\rightarrow0$, we believe that such an algorithm is convergent, and it is generally believed that the slower $E(T)$ grows with $T$, the faster the algorithm converges.

\textbf{A5}: For $\forall t$, random variable $\mathbf{n}_{t}$ is defined as:
\begin{equation}
    \mathbf{n}_{t}=\mathbf{g}_{t}-\nabla f\left(\boldsymbol{P}^{(t)}\right),
\end{equation}
which satisfies:
\begin{equation}
    \mathbb{E}\left[\mathbf{n}_{t}\right]=\mathbf{0} \quad \& \quad \mathbb{E}\left[\left\Vert \mathbf{n}_{t}\right\Vert _{2}^{2}\right]\leq\sigma^{2}.
\end{equation}
In addition, $\mathbf{n}_{t_{1}}$ and $\mathbf{n}_{t_{2}}$ are statistically independent when $t_{1}\neq t_{2}$.

\textbf{Theorem 1}: \textit{Assume that assumptions A1-A5 are satisfied, which yields}
\begin{equation}
    \begin{aligned}
        & E\left(T\right) 
        \\ = & \min_{t=1,2,\ldots,T}\mathbb{E}_{t-1}\left[\left\Vert \nabla f\left(\boldsymbol{P}^{(t)}\right)\right\Vert_{2}^{2}\right]
        \\ \leq & \frac{\max_{i}\left(G_{i}\right)}{\sum_{t=1}^{T}\alpha_{t}} \cdot \left(\left( \frac{L}{2}\frac{\beta_{1}^{2}}{\left(1-\beta_{1}\right)^{2}}\sum_{i=1}^{d}G_{i}^{2}/c^{2} \right.\right.
        \\& \left.\left. + L\cdot2\frac{1}{\left(1-\beta_{1}\right)^{2}}\sum_{i=1}^{d}G_{i}^{2}/c^{2} \right )\sum_{t=1}^{T}\alpha_{t}^{2} + f\left(\boldsymbol{P}^{(1)}\right) \right.
        \\& \left. -f\left(\boldsymbol{P}^{\star}\right)+\frac{\alpha_{t}}{1-\beta_{1}^{t}}\left(\max_{i}G_{i}\right)\left(2\max_{i}G_{i}\right)d/c \right.
        \\& \left. +\left(L\cdot2\frac{\beta_{1}^{2}}{\left(1-\beta_{1}\right)^{2}}\left(\max_{i}G_{i}\right)^{2}\frac{\alpha_{1}}{\left(1-\beta_{1}\right)c} \right.\right.
        \\& \left.\left. +\frac{\beta_{1}}{1-\beta_{1}}\left(\max_{i}G_{i}\right)\left(\max_{i}G_{i}\right) \right.\right.
        \\& \left.\left. +\left(\max_{i}G_{i}\right)\left(2\max_{i}G_{i}\right)\right)\frac{\alpha_{1}d}{\left(1-\beta_{1}\right)c} \right)
        \\ \triangleq & \frac{C^{\prime\prime}\sum_{t=1}^{T}\alpha_{t}^{2}+C^{\prime\prime\prime}}{C^{\prime}\sum_{t=1}^{T}\alpha_{t}},
    \end{aligned}
\end{equation}
\textit{where $\boldsymbol{P}^{\star}=\min_{\boldsymbol{P}}f\left(\boldsymbol{P}\right)$, $d$ is the element number of $\boldsymbol{m}$ and $\boldsymbol{v}$ in AMSGrad algorithms \cite{reddi2018convergence}, and $C^{\prime},C^{\prime\prime},C^{\prime\prime\prime}$ are constants independent of $T$}. 

Please refer to the Appendix for the detailed proof.

Then, we set the learning rate $\alpha_{t}=\alpha/t^{e}$ and appears polynomially decayed, we have
\begin{equation}
    E\left(T\right)\leq\frac{C^{\prime\prime}\sum_{t=1}^{T}\alpha_{t}^{2}+C^{\prime\prime\prime}}{C^{\prime}\sum_{t=1}^{T}\alpha_{t}}=\frac{C^{\prime\prime}\alpha^{2}\sum_{t=1}^{T}1/t^{2e}+C^{\prime\prime\prime}}{C^{\prime}\alpha\sum_{t=1}^{T}1/t^{e}}.
\end{equation}
In general, $C^{\prime\prime}\alpha^{2}\sum_{t=1}^{T}1/t^{2e}=\mathcal{O}\left(T^{1-2e}\right)$, $C^{\prime\prime\prime}=\mathcal{O}\left(1\right)$, $C^{\prime}\alpha\sum_{t=1}^{T}1/t^{e}=\mathcal{O}\left(T^{1-e}\right)$, $E\left(T\right)=\mathcal{O}\left(T^{\max\left(-e,e-1\right)}\right)$, when $e=1/2$, $E\left(T\right)$ has the lowest upper bounds.
Let's take a closer look at when $e=1/2$:
\begin{equation}
    \begin{aligned}
        E\left(T\right)&\leq\frac{C^{\prime\prime}\alpha^{2}\sum_{t=1}^{T}1/t+C^{\prime\prime\prime}}{C^{\prime}\alpha\sum_{t=1}^{T}1/t^{1/2}}
        \\&\leq\frac{C^{\prime\prime}\alpha^{2}\left(1+\log T\right)+C^{\prime\prime\prime}}{C^{\prime}\alpha\left(2\left(T+1\right)^{1/2}-2\right)},
    \end{aligned}
\end{equation}
when $T\longrightarrow\infty$,
\begin{equation}
    E\left(T\right)=\mathcal{O}\left(\frac{\log T}{T^{1/2}}\right),
\end{equation}
\begin{equation}
    \frac{E\left(T\right)}{T}=\mathcal{O}\left(\frac{\log T}{T^{3/2}}\right)\longrightarrow0.
\end{equation}

\begin{table*}[!t]
  \centering
  \tiny
  \setlength{\tabcolsep}{1.224mm}
  \begin{tabular*}{\hsize}{cccccccccccccccccccc}
  \hline
  & \cellcolor[HTML]{EBF1DE}\rotatebox{75}{SSD}   & \cellcolor[HTML]{EBF1DE}\rotatebox{75}{Faster R-CNN} & \cellcolor[HTML]{EBF1DE}\rotatebox{75}{Swin Transformer} & \cellcolor[HTML]{EBF1DE}\rotatebox{75}{YOLOv3} & \cellcolor[HTML]{EBF1DE}\rotatebox{75}{YOLOv5n} & \cellcolor[HTML]{EBF1DE}\rotatebox{75}{YOLOv5s} & \cellcolor[HTML]{EBF1DE}\rotatebox{75}{YOLOv5m} & \cellcolor[HTML]{EBF1DE}\rotatebox{75}{YOLOv5l} & \cellcolor[HTML]{EBF1DE}\rotatebox{75}{YOLOv5x} & \cellcolor[HTML]{EBF1DE}\rotatebox{75}{Cascade R-CNN} & \cellcolor[HTML]{EBF1DE}\rotatebox{75}{RetinaNet} & \cellcolor[HTML]{EBF1DE}\rotatebox{75}{Mask R-CNN}& \cellcolor[HTML]{EBF1DE}\rotatebox{75}{FreeAnchor} & \cellcolor[HTML]{EBF1DE}\rotatebox{75}{FSAF}  & \cellcolor[HTML]{EBF1DE}\rotatebox{75}{RepPoints} & \cellcolor[HTML]{EBF1DE}\rotatebox{75}{TOOD}  & \cellcolor[HTML]{EBF1DE}\rotatebox{75}{ATSS}  & \cellcolor[HTML]{EBF1DE}\rotatebox{75}{FoveaBox} & \cellcolor[HTML]{EBF1DE}\rotatebox{75}{VarifocalNet} \\ \hline
  \cellcolor[HTML]{FDE9D9}Clean            & \cellcolor[HTML]{FEC8D2}0.354 & \cellcolor[HTML]{D2F1FD}0.590        & \cellcolor[HTML]{A3E3FA}0.681            & \cellcolor[HTML]{ADE6FB}0.661  & \cellcolor[HTML]{FEEEF1}0.457   & \cellcolor[HTML]{DDF5FD}0.568   & \cellcolor[HTML]{B8E9FB}0.641   & \cellcolor[HTML]{A7E4FA}0.673   & \cellcolor[HTML]{9FE2FA}0.689   & \cellcolor[HTML]{D0F1FD}0.594         & \cellcolor[HTML]{E3F7FE}0.556     & \cellcolor[HTML]{D3F2FD}0.587      & \cellcolor[HTML]{DAF4FD}0.573      & \cellcolor[HTML]{DDF5FD}0.568 & \cellcolor[HTML]{DDF5FD}0.567     & \cellcolor[HTML]{C3EDFC}0.619 & \cellcolor[HTML]{D9F3FD}0.576 & \cellcolor[HTML]{DEF5FE}0.565    & \cellcolor[HTML]{CFF0FD}0.595        \\
  \cellcolor[HTML]{FDE9D9}Random Noise     & \cellcolor[HTML]{FEA6B6}0.265 & \cellcolor[HTML]{FEF5F6}0.474        & \cellcolor[HTML]{D0F1FD}0.594            & \cellcolor[HTML]{DAF4FD}0.574  & \cellcolor[HTML]{FEEAEE}0.446   & \cellcolor[HTML]{FEF3F5}0.470   & \cellcolor[HTML]{E8F8FE}0.546   & \cellcolor[HTML]{DBF4FD}0.571   & \cellcolor[HTML]{D0F1FD}0.593   & \cellcolor[HTML]{FEF7F8}0.480         & \cellcolor[HTML]{FEEDF0}0.454     & \cellcolor[HTML]{FEFAFA}0.487      & \cellcolor[HTML]{FEF4F6}0.473      & \cellcolor[HTML]{FEF5F7}0.476 & \cellcolor[HTML]{FEF3F5}0.470     & \cellcolor[HTML]{F0FBFF}0.530 & \cellcolor[HTML]{FEF9FA}0.486 & \cellcolor[HTML]{FEF2F4}0.467    & \cellcolor[HTML]{FEFFFF}0.503        \\
  \cellcolor[HTML]{FDE9D9}SSD              & \cellcolor[HTML]{FEA1B2}\textbf{0.252} & \cellcolor[HTML]{FEE7EB}0.437        & \cellcolor[HTML]{E0F6FE}0.562            & \cellcolor[HTML]{E7F8FE}0.548  & \cellcolor[HTML]{FEDCE2}0.407   & \cellcolor[HTML]{FEE4E9}0.430   & \cellcolor[HTML]{FEF9FA}0.485   & \cellcolor[HTML]{ECF9FE}0.538   & \cellcolor[HTML]{EBF9FE}0.540   & \cellcolor[HTML]{FEE8EC}0.439         & \cellcolor[HTML]{FEE0E5}0.418     & \cellcolor[HTML]{FEECEF}0.450      & \cellcolor[HTML]{FEE5EA}0.433      & \cellcolor[HTML]{FEE7EB}0.437 & \cellcolor[HTML]{FEE4E9}0.430     & \cellcolor[HTML]{FEF9FA}0.485 & \cellcolor[HTML]{FEE5EA}0.433 & \cellcolor[HTML]{FEE3E8}0.427    & \cellcolor[HTML]{FEEEF1}0.456        \\
  \cellcolor[HTML]{FDE9D9}Faster R-CNN     & \cellcolor[HTML]{FE9FB0}0.245 & \cellcolor[HTML]{FED3DB}\textbf{0.385}        & \cellcolor[HTML]{E9F9FE}0.544            & \cellcolor[HTML]{EBF9FE}0.540  & \cellcolor[HTML]{FEDCE2}0.407   & \cellcolor[HTML]{FEE2E7}0.423   & \cellcolor[HTML]{FEEEF1}0.457   & \cellcolor[HTML]{EFFAFF}0.532   & \cellcolor[HTML]{F1FBFF}0.528   & \cellcolor[HTML]{FEDAE0}0.402         & \cellcolor[HTML]{FECDD6}0.369     & \cellcolor[HTML]{FEDAE1}0.403      & \cellcolor[HTML]{FED0D9}0.377      & \cellcolor[HTML]{FED2DA}0.382 & \cellcolor[HTML]{FED1D9}0.379     & \cellcolor[HTML]{FEE6EB}0.436 & \cellcolor[HTML]{FED1D9}0.379 & \cellcolor[HTML]{FED0D9}0.377    & \cellcolor[HTML]{FEDBE2}0.406        \\
  \cellcolor[HTML]{FDE9D9}Swin Transformer & \cellcolor[HTML]{FEA3B3}0.256 & \cellcolor[HTML]{FEEBEF}0.449        & \cellcolor[HTML]{DDF5FD}\textbf{0.567}            & \cellcolor[HTML]{E6F8FE}0.550  & \cellcolor[HTML]{FEE2E7}0.423   & \cellcolor[HTML]{FEE7EB}0.437   & \cellcolor[HTML]{FDFFFF}0.505   & \cellcolor[HTML]{EEFAFE}0.534   & \cellcolor[HTML]{EEFAFE}0.535   & \cellcolor[HTML]{FEEDF0}0.454         & \cellcolor[HTML]{FEE4E9}0.429     & \cellcolor[HTML]{FEF0F2}0.461      & \cellcolor[HTML]{FEEAEE}0.446      & \cellcolor[HTML]{FEEBEE}0.448 & \cellcolor[HTML]{FEE9ED}0.442     & \cellcolor[HTML]{FEFEFE}0.499 & \cellcolor[HTML]{FEEBEE}0.448 & \cellcolor[HTML]{FEE9ED}0.443    & \cellcolor[HTML]{FEF4F6}0.471        \\
  \cellcolor[HTML]{FDE9D9}YOLOv3           & \cellcolor[HTML]{FEA3B4}0.257 & \cellcolor[HTML]{FEECF0}0.452        & \cellcolor[HTML]{DCF4FD}0.570            & \cellcolor[HTML]{FECAD3}\textbf{0.360}  & \cellcolor[HTML]{FEE3E8}0.426   & \cellcolor[HTML]{FEE5E9}0.431   & \cellcolor[HTML]{FEF8FA}0.484   & \cellcolor[HTML]{FFFFFF}0.500   & \cellcolor[HTML]{F9FDFF}0.513   & \cellcolor[HTML]{FEEEF1}0.456         & \cellcolor[HTML]{FEE6EA}0.435     & \cellcolor[HTML]{FEF0F2}0.461      & \cellcolor[HTML]{FEEBEE}0.448      & \cellcolor[HTML]{FEECF0}0.452 & \cellcolor[HTML]{FEEBEE}0.448     & \cellcolor[HTML]{FEFFFF}0.502 & \cellcolor[HTML]{FEECF0}0.452 & \cellcolor[HTML]{FEE9ED}0.443    & \cellcolor[HTML]{FEF5F7}0.475        \\
  \cellcolor[HTML]{FDE9D9}YOLOv5n          & \cellcolor[HTML]{FEA1B2}0.250 & \cellcolor[HTML]{FEE3E8}0.427        & \cellcolor[HTML]{DEF5FE}0.566            & \cellcolor[HTML]{E2F6FE}0.558  & \cellcolor[HTML]{FE91A5}\textbf{0.209}   & \cellcolor[HTML]{FEE3E8}0.428   & \cellcolor[HTML]{FEFCFD}0.494   & \cellcolor[HTML]{E1F6FE}0.560   & \cellcolor[HTML]{DEF5FE}0.565   & \cellcolor[HTML]{FEE4E9}0.429         & \cellcolor[HTML]{FEDCE2}0.408     & \cellcolor[HTML]{FEE7EB}0.438      & \cellcolor[HTML]{FEE2E7}0.424      & \cellcolor[HTML]{FEE3E8}0.428 & \cellcolor[HTML]{FEE1E6}0.422     & \cellcolor[HTML]{FEF4F6}0.471 & \cellcolor[HTML]{FEE1E6}0.422 & \cellcolor[HTML]{FEDEE4}0.414    & \cellcolor[HTML]{FEE9ED}0.443        \\
  \cellcolor[HTML]{FDE9D9}YOLOv5s          & \cellcolor[HTML]{FEA1B2}0.251 & \cellcolor[HTML]{FEE9ED}0.442        & \cellcolor[HTML]{D9F4FD}0.575            & \cellcolor[HTML]{E5F7FE}0.551  & \cellcolor[HTML]{FEDFE5}0.417   & \cellcolor[HTML]{FE9FB0}\textbf{0.246}   & \cellcolor[HTML]{FEF3F5}0.469   & \cellcolor[HTML]{F8FDFF}0.514   & \cellcolor[HTML]{F4FCFF}0.522   & \cellcolor[HTML]{FEE9ED}0.444         & \cellcolor[HTML]{FEE1E6}0.422     & \cellcolor[HTML]{FEEDF0}0.453      & \cellcolor[HTML]{FEE7EB}0.437      & \cellcolor[HTML]{FEE8EC}0.441 & \cellcolor[HTML]{FEE7EB}0.437     & \cellcolor[HTML]{FEFAFA}0.487 & \cellcolor[HTML]{FEE8EC}0.439 & \cellcolor[HTML]{FEE5EA}0.432    & \cellcolor[HTML]{FEF0F2}0.461        \\
  \cellcolor[HTML]{FDE9D9}YOLOv5m          & \cellcolor[HTML]{FEA3B4}0.257 & \cellcolor[HTML]{FEECEF}0.450        & \cellcolor[HTML]{D8F3FD}0.577            & \cellcolor[HTML]{E6F8FE}0.550  & \cellcolor[HTML]{FEE3E8}0.427   & \cellcolor[HTML]{FEE2E7}0.423   & \cellcolor[HTML]{FEB4C1}\textbf{0.301}   & \cellcolor[HTML]{FEF8FA}0.484   & \cellcolor[HTML]{FEFBFB}0.490   & \cellcolor[HTML]{FEECF0}0.452         & \cellcolor[HTML]{FEE4E9}0.430     & \cellcolor[HTML]{FEF0F2}0.461      & \cellcolor[HTML]{FEE9ED}0.444      & \cellcolor[HTML]{FEEBEE}0.448 & \cellcolor[HTML]{FEEAEE}0.446     & \cellcolor[HTML]{FEFDFD}0.496 & \cellcolor[HTML]{FEEBEE}0.448 & \cellcolor[HTML]{FEE8EC}0.439    & \cellcolor[HTML]{FEF3F5}0.470        \\
  \cellcolor[HTML]{FDE9D9}YOLOv5l          & \cellcolor[HTML]{FEA4B4}0.259 & \cellcolor[HTML]{FEEBEE}0.448        & \cellcolor[HTML]{D8F3FD}0.578            & \cellcolor[HTML]{E6F8FE}0.550  & \cellcolor[HTML]{FEE2E7}0.424   & \cellcolor[HTML]{FEDEE4}0.413   & \cellcolor[HTML]{FEF3F5}0.469   & \cellcolor[HTML]{FEAEBC}\textbf{0.285}   & \cellcolor[HTML]{FEF7F8}0.479   & \cellcolor[HTML]{FEECF0}0.452         & \cellcolor[HTML]{FEE3E8}0.426     & \cellcolor[HTML]{FEEFF2}0.460      & \cellcolor[HTML]{FEE9ED}0.443      & \cellcolor[HTML]{FEEBEE}0.447 & \cellcolor[HTML]{FEE9ED}0.444     & \cellcolor[HTML]{FEFDFD}0.496 & \cellcolor[HTML]{FEEBEF}0.449 & \cellcolor[HTML]{FEE8EC}0.439    & \cellcolor[HTML]{FEF3F5}0.469        \\
  \cellcolor[HTML]{FDE9D9}YOLOv5x          & \cellcolor[HTML]{FEA3B4}0.257 & \cellcolor[HTML]{FEECEF}0.450        & \cellcolor[HTML]{D7F3FD}0.579            & \cellcolor[HTML]{E8F8FE}0.547  & \cellcolor[HTML]{FEE3E8}0.426   & \cellcolor[HTML]{FEE3E8}0.426   & \cellcolor[HTML]{FEF5F7}0.476   & \cellcolor[HTML]{FEF4F6}0.472   & \cellcolor[HTML]{FEA5B5}\textbf{0.261}   & \cellcolor[HTML]{FEEDF0}0.454         & \cellcolor[HTML]{FEE5E9}0.431     & \cellcolor[HTML]{FEF1F3}0.463      & \cellcolor[HTML]{FEEBEE}0.448      & \cellcolor[HTML]{FEECEF}0.451 & \cellcolor[HTML]{FEEBEE}0.448     & \cellcolor[HTML]{FEFFFF}0.502 & \cellcolor[HTML]{FEEDF0}0.454 & \cellcolor[HTML]{FEE9ED}0.443    & \cellcolor[HTML]{FEF5F7}0.475        \\
  \cellcolor[HTML]{FDE9D9}Cascade R-CNN    & \cellcolor[HTML]{FE9FB1}0.247 & \cellcolor[HTML]{FED1D9}0.379        & \cellcolor[HTML]{EBF9FE}0.541            & \cellcolor[HTML]{EDFAFE}0.537  & \cellcolor[HTML]{FEDDE3}0.412   & \cellcolor[HTML]{FEE0E6}0.419   & \cellcolor[HTML]{FEF1F4}0.465   & \cellcolor[HTML]{F3FCFF}0.525   & \cellcolor[HTML]{FBFEFF}0.508   & \cellcolor[HTML]{FED3DB}\textbf{0.385}         & \cellcolor[HTML]{FECBD4}0.362     & \cellcolor[HTML]{FED7DE}0.394      & \cellcolor[HTML]{FECBD4}0.362      & \cellcolor[HTML]{FECFD8}0.374 & \cellcolor[HTML]{FECDD6}0.369     & \cellcolor[HTML]{FEE3E8}0.427 & \cellcolor[HTML]{FECCD5}0.366 & \cellcolor[HTML]{FECCD6}0.367    & \cellcolor[HTML]{FED3DB}0.384        \\
  \cellcolor[HTML]{FDE9D9}RetinaNet        & \cellcolor[HTML]{FE9FB1}0.247 & \cellcolor[HTML]{FED3DB}0.385        & \cellcolor[HTML]{E9F8FE}0.545            & \cellcolor[HTML]{EAF9FE}0.543  & \cellcolor[HTML]{FEDAE1}0.403   & \cellcolor[HTML]{FEE2E7}0.424   & \cellcolor[HTML]{FEF0F2}0.461   & \cellcolor[HTML]{EDFAFE}0.537   & \cellcolor[HTML]{F3FCFF}0.524   & \cellcolor[HTML]{FEDAE1}0.404         & \cellcolor[HTML]{FECDD6}\textbf{0.368}     & \cellcolor[HTML]{FEDAE0}0.402      & \cellcolor[HTML]{FED0D9}0.377      & \cellcolor[HTML]{FED3DB}0.384 & \cellcolor[HTML]{FED0D8}0.376     & \cellcolor[HTML]{FEE5EA}0.433 & \cellcolor[HTML]{FECFD8}0.374 & \cellcolor[HTML]{FED1D9}0.379    & \cellcolor[HTML]{FED9E0}0.401        \\
  \cellcolor[HTML]{FDE9D9}Mask R-CNN       & \cellcolor[HTML]{FE9FB0}0.245 & \cellcolor[HTML]{FED5DD}0.390        & \cellcolor[HTML]{EAF9FE}0.542            & \cellcolor[HTML]{ECF9FE}0.539  & \cellcolor[HTML]{FEDAE1}0.404   & \cellcolor[HTML]{FEE0E6}0.420   & \cellcolor[HTML]{FEEFF2}0.460   & \cellcolor[HTML]{EFFAFF}0.532   & \cellcolor[HTML]{F5FCFF}0.521   & \cellcolor[HTML]{FEDAE1}0.404         & \cellcolor[HTML]{FECFD8}0.374     & \cellcolor[HTML]{FED7DE}\textbf{0.396}      & \cellcolor[HTML]{FED2DA}0.381      & \cellcolor[HTML]{FED4DB}0.386 & \cellcolor[HTML]{FED2DA}0.381     & \cellcolor[HTML]{FEE8EC}0.440 & \cellcolor[HTML]{FED3DA}0.383 & \cellcolor[HTML]{FED0D9}0.377    & \cellcolor[HTML]{FEDAE0}0.402        \\
  \cellcolor[HTML]{FDE9D9}FreeAnchor       & \cellcolor[HTML]{FEA0B1}0.248 & \cellcolor[HTML]{FED8DF}0.398        & \cellcolor[HTML]{EAF9FE}0.543            & \cellcolor[HTML]{EBF9FE}0.540  & \cellcolor[HTML]{FEDCE2}0.409   & \cellcolor[HTML]{FEDFE5}0.417   & \cellcolor[HTML]{FEF0F3}0.462   & \cellcolor[HTML]{F0FBFF}0.530   & \cellcolor[HTML]{F5FCFF}0.521   & \cellcolor[HTML]{FEDEE4}0.413         & \cellcolor[HTML]{FED5DD}0.390     & \cellcolor[HTML]{FEDFE4}0.415      & \cellcolor[HTML]{FED3DB}\textbf{0.385}      & \cellcolor[HTML]{FED7DE}0.396 & \cellcolor[HTML]{FED5DD}0.390     & \cellcolor[HTML]{FEE9ED}0.444 & \cellcolor[HTML]{FED4DC}0.388 & \cellcolor[HTML]{FED6DD}0.392    & \cellcolor[HTML]{FEE1E6}0.421        \\
  \cellcolor[HTML]{FDE9D9}FSAF             & \cellcolor[HTML]{FE9FB0}0.246 & \cellcolor[HTML]{FED4DB}0.386        & \cellcolor[HTML]{EBF9FE}0.540            & \cellcolor[HTML]{EAF9FE}0.542  & \cellcolor[HTML]{FEDCE2}0.408   & \cellcolor[HTML]{FEE0E6}0.420   & \cellcolor[HTML]{FEEEF1}0.457   & \cellcolor[HTML]{F3FCFF}0.524   & \cellcolor[HTML]{F6FDFF}0.518   & \cellcolor[HTML]{FED9DF}0.399         & \cellcolor[HTML]{FECED6}0.370     & \cellcolor[HTML]{FED9E0}0.401      & \cellcolor[HTML]{FED1D9}0.378      & \cellcolor[HTML]{FED0D9}\textbf{0.377} & \cellcolor[HTML]{FED1D9}0.379     & \cellcolor[HTML]{FEE6EA}0.434 & \cellcolor[HTML]{FED0D9}0.377 & \cellcolor[HTML]{FECFD8}0.374    & \cellcolor[HTML]{FEDAE1}0.404        \\
  \cellcolor[HTML]{FDE9D9}RepPoints        & \cellcolor[HTML]{FE9FB1}0.247 & \cellcolor[HTML]{FEDAE1}0.404        & \cellcolor[HTML]{E7F8FE}0.549            & \cellcolor[HTML]{E9F9FE}0.544  & \cellcolor[HTML]{FEDDE3}0.410   & \cellcolor[HTML]{FEE2E7}0.424   & \cellcolor[HTML]{FEF2F4}0.467   & \cellcolor[HTML]{EBF9FE}0.541   & \cellcolor[HTML]{ECF9FE}0.539   & \cellcolor[HTML]{FEDEE4}0.414         & \cellcolor[HTML]{FED4DC}0.388     & \cellcolor[HTML]{FEE0E5}0.418      & \cellcolor[HTML]{FED6DD}0.392      & \cellcolor[HTML]{FED9DF}0.399 & \cellcolor[HTML]{FED3DB}\textbf{0.385}     & \cellcolor[HTML]{FEEAEE}0.446 & \cellcolor[HTML]{FED4DB}0.386 & \cellcolor[HTML]{FED7DE}0.396    & \cellcolor[HTML]{FEE1E6}0.421        \\
  \cellcolor[HTML]{FDE9D9}TOOD             & \cellcolor[HTML]{FE9FB0}0.246 & \cellcolor[HTML]{FEE1E6}0.422        & \cellcolor[HTML]{E2F6FE}0.558            & \cellcolor[HTML]{E7F8FE}0.548  & \cellcolor[HTML]{FEE0E5}0.418   & \cellcolor[HTML]{FEE3E8}0.428   & \cellcolor[HTML]{FEF7F9}0.481   & \cellcolor[HTML]{E7F8FE}0.549   & \cellcolor[HTML]{E5F7FE}0.552   & \cellcolor[HTML]{FEE3E8}0.428         & \cellcolor[HTML]{FEDAE0}0.402     & \cellcolor[HTML]{FEE6EA}0.435      & \cellcolor[HTML]{FEDEE4}0.413      & \cellcolor[HTML]{FEDFE5}0.417 & \cellcolor[HTML]{FEDDE3}0.410     & \cellcolor[HTML]{FEF1F3}\textbf{0.463} & \cellcolor[HTML]{FEDDE3}0.411 & \cellcolor[HTML]{FEDDE3}0.411    & \cellcolor[HTML]{FEE7EB}0.438        \\
  \cellcolor[HTML]{FDE9D9}ATSS             & \cellcolor[HTML]{FE9FB0}0.246 & \cellcolor[HTML]{FEDDE3}0.412        & \cellcolor[HTML]{E5F7FE}0.552            & \cellcolor[HTML]{EBF9FE}0.541  & \cellcolor[HTML]{FEDEE4}0.413   & \cellcolor[HTML]{FEE1E6}0.421   & \cellcolor[HTML]{FEF4F6}0.473   & \cellcolor[HTML]{ECF9FE}0.538   & \cellcolor[HTML]{EBF9FE}0.540   & \cellcolor[HTML]{FEE0E6}0.419         & \cellcolor[HTML]{FED7DE}0.395     & \cellcolor[HTML]{FEE1E6}0.421      & \cellcolor[HTML]{FED9E0}0.400      & \cellcolor[HTML]{FEDAE0}0.402 & \cellcolor[HTML]{FED8DF}0.398     & \cellcolor[HTML]{FEECEF}0.450 & \cellcolor[HTML]{FED6DE}\textbf{0.393} & \cellcolor[HTML]{FED8DF}0.397    & \cellcolor[HTML]{FEE2E7}0.423        \\
  \cellcolor[HTML]{FDE9D9}FoveaBox         & \cellcolor[HTML]{FEA0B1}0.248 & \cellcolor[HTML]{FEDCE2}0.408        & \cellcolor[HTML]{E6F8FE}0.550            & \cellcolor[HTML]{EAF9FE}0.542  & \cellcolor[HTML]{FEDEE4}0.414   & \cellcolor[HTML]{FEDFE5}0.417   & \cellcolor[HTML]{FEEEF1}0.457   & \cellcolor[HTML]{EFFAFF}0.532   & \cellcolor[HTML]{F5FCFF}0.521   & \cellcolor[HTML]{FEE0E5}0.418         & \cellcolor[HTML]{FED6DD}0.391     & \cellcolor[HTML]{FEE0E6}0.419      & \cellcolor[HTML]{FED6DD}0.392      & \cellcolor[HTML]{FED9DF}0.399 & \cellcolor[HTML]{FED7DE}0.394     & \cellcolor[HTML]{FEEAEE}0.445 & \cellcolor[HTML]{FED4DC}0.388 & \cellcolor[HTML]{FED4DB}\textbf{0.386}    & \cellcolor[HTML]{FEE0E6}0.420        \\
  \cellcolor[HTML]{FDE9D9}VarifocalNet     & \cellcolor[HTML]{FE9EAF}0.243 & \cellcolor[HTML]{FED6DE}0.393        & \cellcolor[HTML]{E8F8FE}0.546            & \cellcolor[HTML]{E8F8FE}0.547  & \cellcolor[HTML]{FEDDE3}0.411   & \cellcolor[HTML]{FEE3E8}0.426   & \cellcolor[HTML]{FEEDF0}0.454   & \cellcolor[HTML]{EEFAFE}0.535   & \cellcolor[HTML]{EEFAFE}0.534   & \cellcolor[HTML]{FEDBE1}0.405         & \cellcolor[HTML]{FECFD8}0.374     & \cellcolor[HTML]{FEDBE1}0.405      & \cellcolor[HTML]{FED2DA}0.382      & \cellcolor[HTML]{FED4DB}0.386 & \cellcolor[HTML]{FED1D9}0.379     & \cellcolor[HTML]{FEE5E9}0.431 & \cellcolor[HTML]{FECFD8}0.374 & \cellcolor[HTML]{FED3DA}0.383    & \cellcolor[HTML]{FED6DD}\textbf{0.392}
  \\  
  \hline
  \end{tabular*}
  \caption{
    Experimental results of digital background attack on the validation set of COCO in the metric of mAP, where white-box attacks are highlighted in bold and the rest are black-box attacks. 
  The \textbf{redder} the cell, the \textbf{worse} the \textbf{detection performance}.
  The \textbf{bluer} the cell, the \textbf{better} the \textbf{detection performance}.
  Clean and Random Noise mean experiments on clean images and images with random noise, respectively.
The 19 detectors of the first row and the first column are for detection and perturbation optimization, respectively.
  }
  \label{table_digital_attack}
\end{table*}

\begin{figure}[!htbp]
  \centering
  \begin{subfigure}{0.49\linewidth}
    \includegraphics[width=1\linewidth]{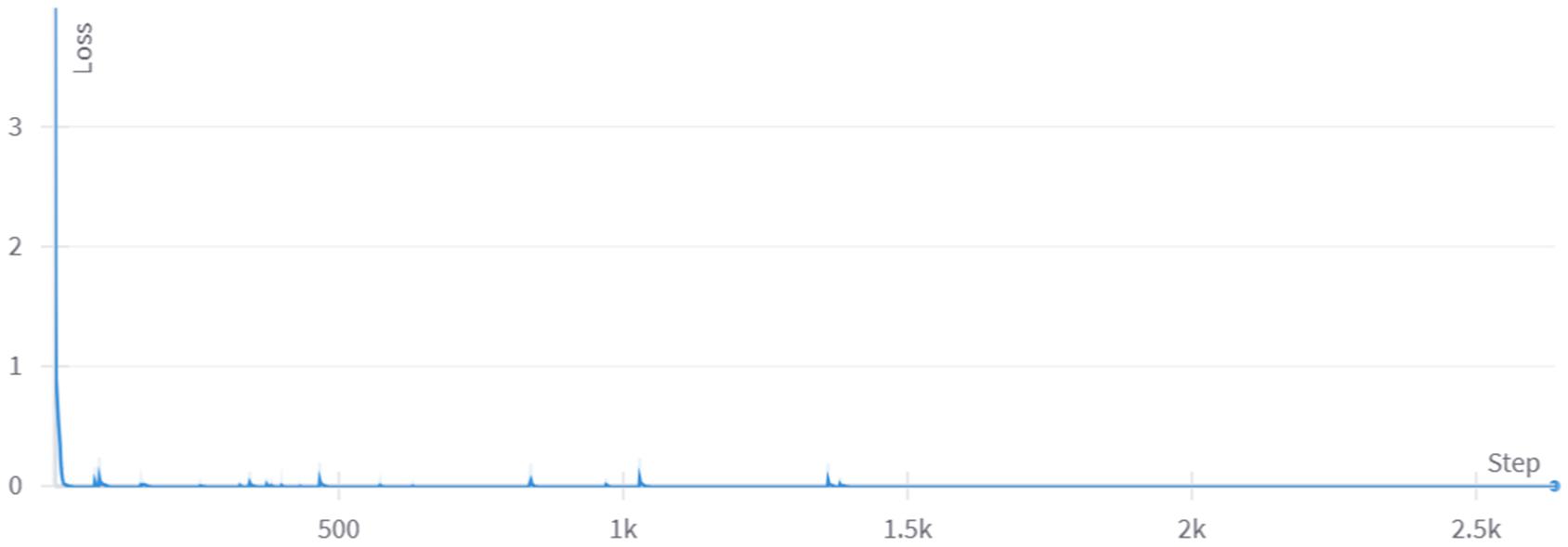}
    \caption{Bounding box loss ($L_{box}$)}
  \end{subfigure}
  \begin{subfigure}{0.49\linewidth}
    \includegraphics[width=1\linewidth]{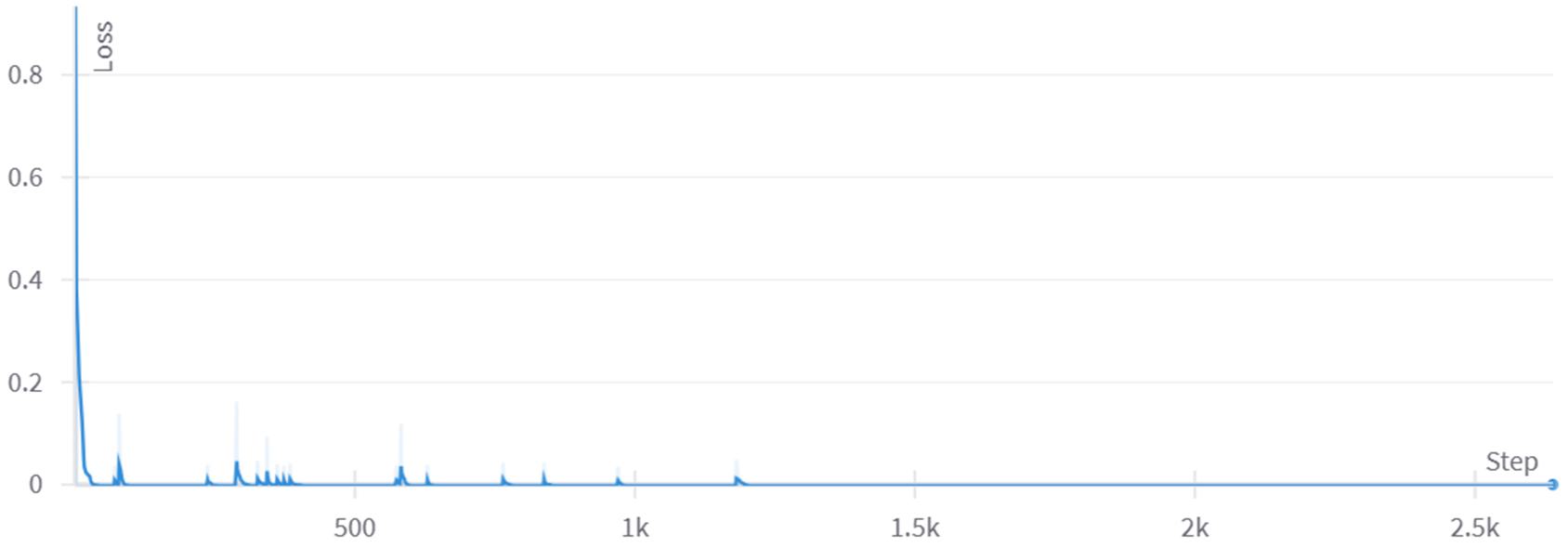}
    \caption{Ojectiveness loss ($L_{obj}$)}
  \end{subfigure}
  \begin{subfigure}{0.49\linewidth}
    \includegraphics[width=1\linewidth]{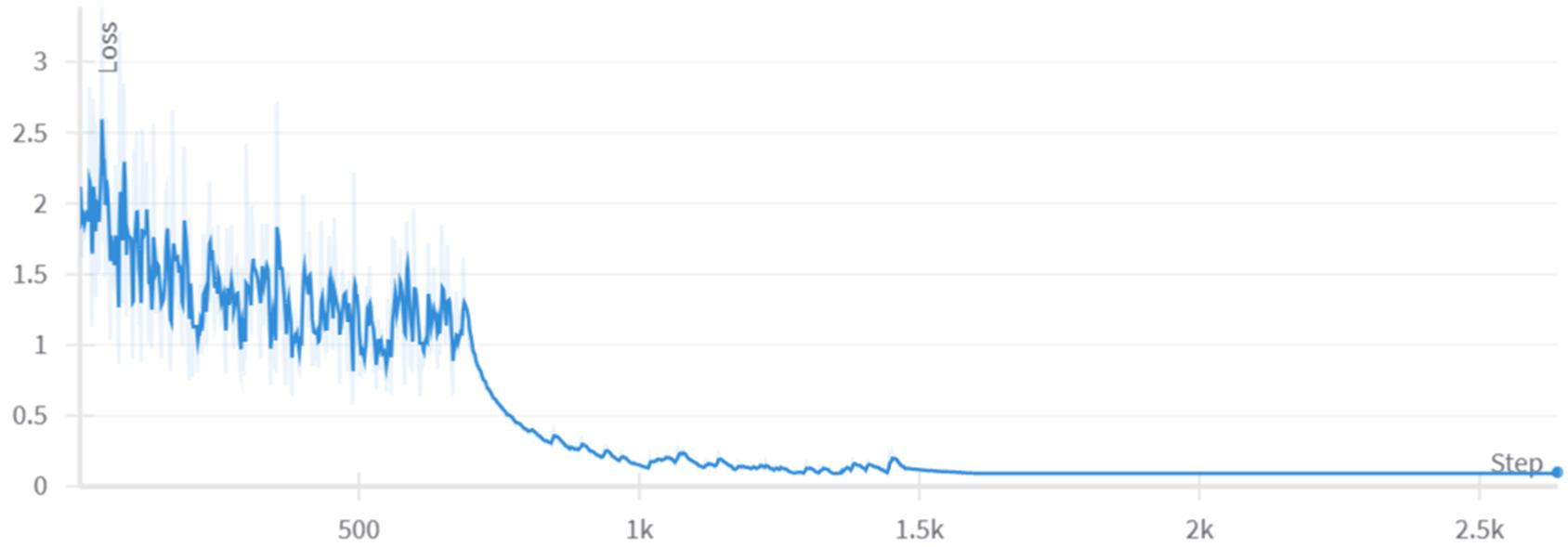}
    \caption{Total variation loss ($L_{abtv}$)}
  \end{subfigure}
  \begin{subfigure}{0.49\linewidth}
    \includegraphics[width=1\linewidth]{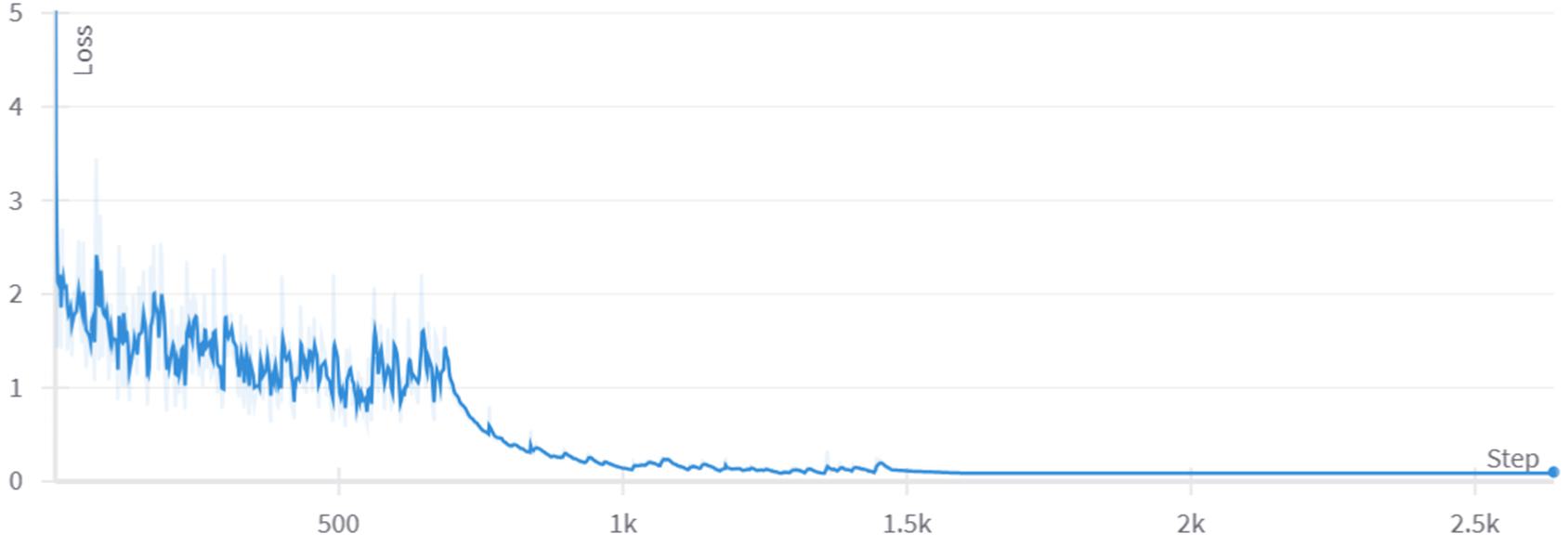}
    \caption{Total loss ($L$)}
  \end{subfigure}
  \caption{Empirical demonstration of background attack loss convergence. Please zoom in for details.}
  \label{fig_converged_loss}
\end{figure}

As a consequence, our formulated background adversarial attack is mathematically convergent with mild sufficient conditions, which is also demonstrated with experimental results as shown in Fig. \ref{fig_converged_loss}.
The loss functions are detailed in Sec. \ref{subsection_objective_loss}.
Through the above convergence analysis, we made a positive step toward understanding the theoretical behavior of the proposed background attack methods.

\section{Experiments}
\label{sec_experiments}

In this section, we present the experimental settings in \ref{subsec_experimental_settings}.
Then, we demonstrate the effectiveness of attacking anything in both digital and physical domains in \ref{subsec_digital_background_attacks} and \ref{subsec_physical_background_attacks}, respectively.
Furthermore, we compare the proposed background adversarial attack with SOTA physical attacks in \ref{subsec_physical_attack_comparison}.
Next, we showcase the effectiveness of attacking anything across different objects, models, and tasks in \ref{subsec_attack_anything}.
Finally, we conduct an ablation study to verify the effectiveness of the proposed ensemble strategy and novel smoothness loss in \ref{subsec_ablation_study}.

\subsection{Experimental Settings}
\label{subsec_experimental_settings}

\subsubsection{Models}
We use several canonical or SOTA object detectors as victim models, including YOLOv3 \cite{redmon2018yolov3}, YOLOv5 \cite{glenn_jocher_2020_4154370}, SSD \cite{liu2016ssd}, Faster R-CNN \cite{ren2015faster}, Swin Transformer \cite{liu2021swin}, Cascade R-CNN \cite{cai2019cascade}, RetinaNet \cite{lin2017focal}, Mask R-CNN \cite{he2017mask}, FoveaBox \cite{kong2020foveabox}, FreeAnchor \cite{zhang2019freeanchor}, FSAF \cite{zhu2019feature}, RepPoints \cite{yang2019reppoints}, TOOD \cite{feng2021tood}, ATSS \cite{zhang2020bridging}, and VarifocalNet \cite{zhang2021varifocalnet}.

\subsubsection{Datasets}
Two public datasets: COCO \cite{lin2014microsoft} and DOTA \cite{xia2018dota} are involved in the experiments.
Specifically, we adopt the training set and validation set from COCO to train and validate background perturbations, respectively, and we use DOTA to train aerial detectors.

\subsubsection{Metrics}
Mean average precision (mAP) and detection rate (DR) \cite{wu2020making} are adopted as the metrics of detection performance under digital and physical attacks, respectively.
The default threshold of confidence score and intersection over union (IOU) are set as 0.25 and 0.5, respectively.
Attack successful rate (ASR) is used for the measurement of attack performance. 
We detail the mathematical description of these metrics in the Appendix.

\subsubsection{Implementations}
Initial perturbation $\boldsymbol{P}^{(0)}$ is randomly initialized.
Hyperparameters $\eta, \lambda$, start learning rate, and max epoch are set as 9, 0.01, 0.03, and 50, respectively.
YOLOv3 and YOLOv5 are trained by \cite{glenn_jocher_2020_4154370}, and the rest detectors are from MMDetection \cite{mmdetection}.
The default settings of detectors are adopted in perturbation optimization.
We conduct the experiments based on Pytorch on NVIDIA RTX 3090 24GB GPUs.

\begin{figure}[!htbp]
  \centering
  \begin{subfigure}{0.99\linewidth}
    \includegraphics[width=1\linewidth]{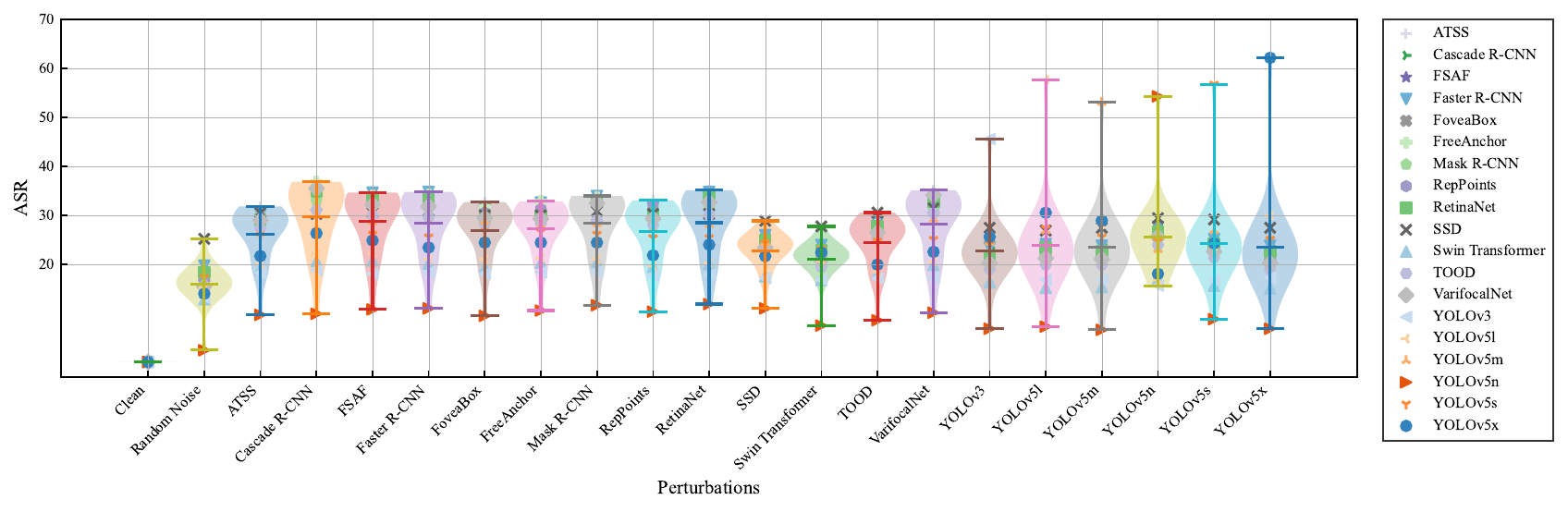}
    \caption{Attack perspective}
  \end{subfigure}
  \begin{subfigure}{0.99\linewidth}
    \includegraphics[width=1\linewidth]{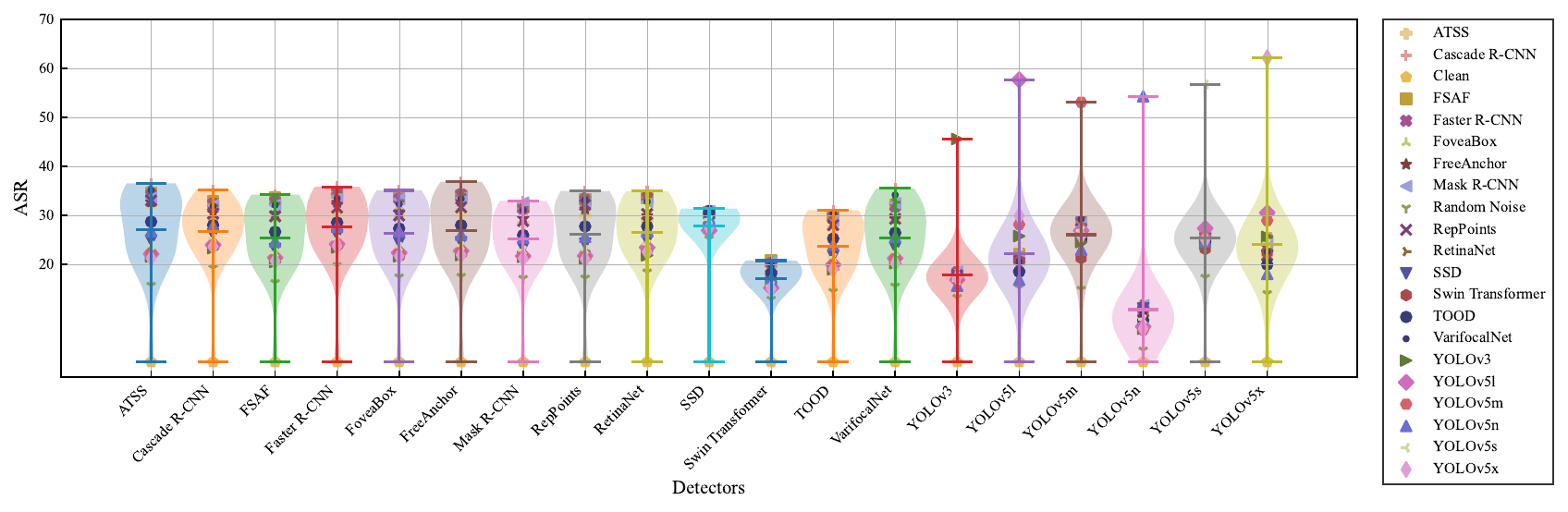}
    \caption{Detection perspective}
  \end{subfigure}
  \caption{The figures visualize the quantitative experimental results of digital background attacks from the perspectives of detection and attack in the metric of ASR. Please zoom in for a better view.}
  \label{fig_digital_attack_visualization}
  \end{figure}

\subsection{Digital Background Attacks}
\label{subsec_digital_background_attacks}

We perform digital background attacks with numerous mainstream object detection methods.
Specifically, we use the validation set of COCO to verify digital attack efficacy by replacing the objects' backgrounds with the elaborated adversarial perturbations.
We report the quantitative experimental results in Table \ref{table_digital_attack}, and the metric is mAP0.5.
In addition, we also visualize the quantitative experimental results from the perspective of detection and attack in terms of ASR in Fig. \ref{fig_digital_attack_visualization}.
It is demonstrated that:

\begin{itemize}
  \item We can easily fool SOTA object detectors by only manipulating background features with a universal background perturbation.
  \item The mAP0.5 of SOTA detectors has decreased significantly up to 62.1\% (0.689 to 0.261 of YOLOv5x) even background perturbation undergoes multi-scale objects and unbalanced categories, which confirms the significant role of background features in visual perception based on DNNs.
  \item The attack efficacy can transfer well between different models with different neural network structures, such as convolutional neural networks and transformers, which demonstrates the general mechanism weakness of DNNs.
\end{itemize}

The qualitative experimental results are shown in Fig. \ref{fig_background_attack_exhibition} (c).
It is observed that most objects have been successfully hidden under our digital background attack.
Please refer to the Appendix for more experimental results in the metric of mAP0.5:0.95.

\begin{figure}[!htbp]
  \centering
  \begin{subfigure}{0.49\linewidth}
    \includegraphics[width=1\linewidth]{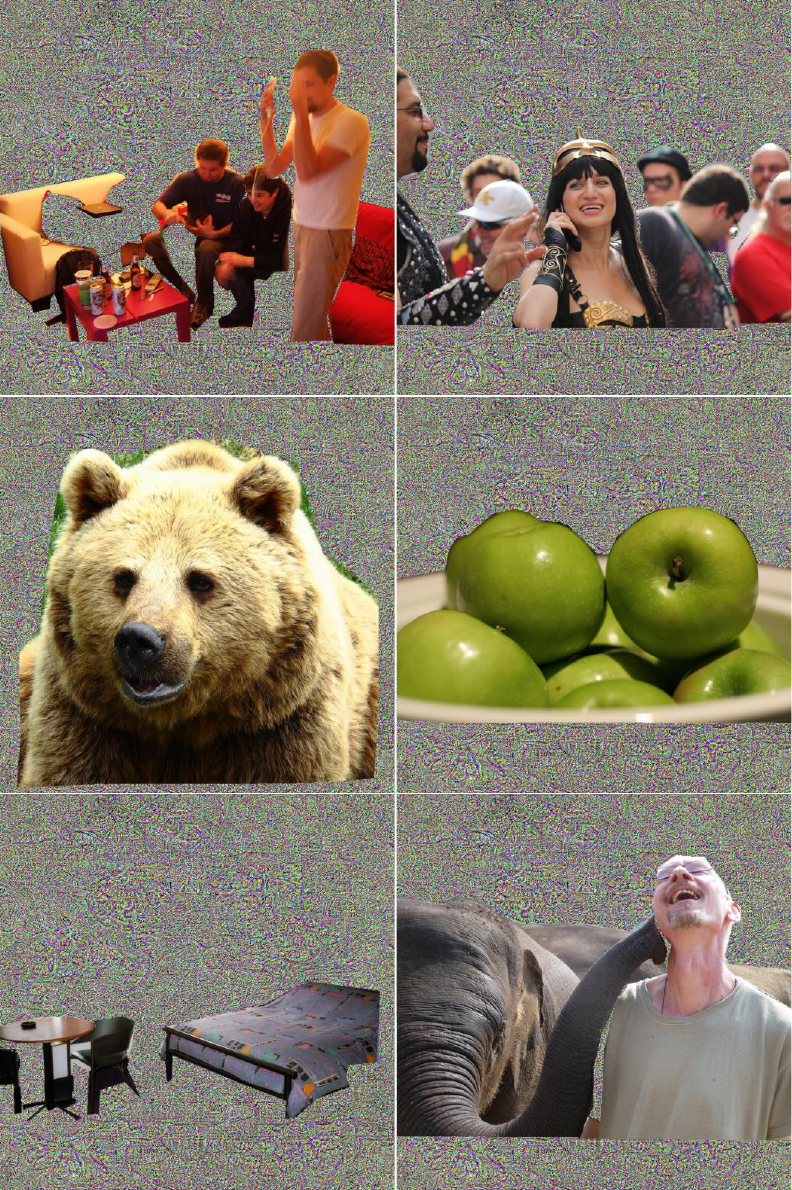}
    \caption{Successful attacks}
    \label{fig_digital_attack_successful}
  \end{subfigure}
  \begin{subfigure}{0.49\linewidth}
    \includegraphics[width=1\linewidth]{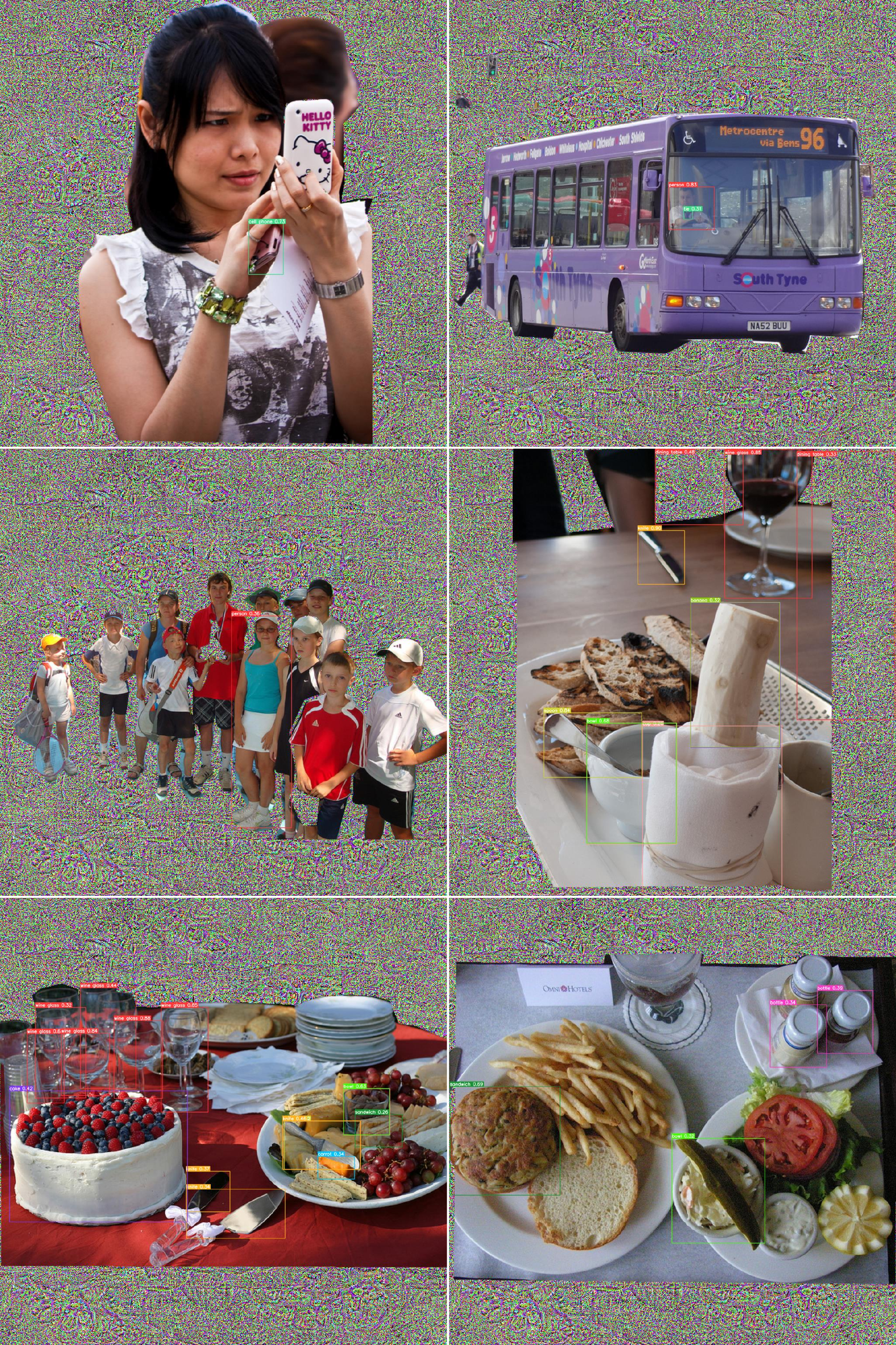}
    \caption{Failed attacks}
    \label{fig_digital_attack_failed}
  \end{subfigure}
  \caption{Successful and failed attacks.  Please zoom in for better visualization.}
  \label{fig_digital_attack_successful_failed}
\end{figure}

\begin{table*}[!htbp]
  \centering
  \tiny
  \setlength{\tabcolsep}{1.224mm}
  \begin{tabular*}{\hsize}{cccccccccccccccccccc}
  \hline
  & \cellcolor[HTML]{EBF1DE}\rotatebox{75}{SSD}   & \cellcolor[HTML]{EBF1DE}\rotatebox{75}{Faster R-CNN} & \cellcolor[HTML]{EBF1DE}\rotatebox{75}{Swin Transformer} & \cellcolor[HTML]{EBF1DE}\rotatebox{75}{YOLOv3} & \cellcolor[HTML]{EBF1DE}\rotatebox{75}{YOLOv5n} & \cellcolor[HTML]{EBF1DE}\rotatebox{75}{YOLOv5s} & \cellcolor[HTML]{EBF1DE}\rotatebox{75}{YOLOv5m} & \cellcolor[HTML]{EBF1DE}\rotatebox{75}{YOLOv5l} & \cellcolor[HTML]{EBF1DE}\rotatebox{75}{YOLOv5x} & \cellcolor[HTML]{EBF1DE}\rotatebox{75}{Cascade R-CNN} & \cellcolor[HTML]{EBF1DE}\rotatebox{75}{RetinaNet} & \cellcolor[HTML]{EBF1DE}\rotatebox{75}{Mask R-CNN}& \cellcolor[HTML]{EBF1DE}\rotatebox{75}{FreeAnchor} & \cellcolor[HTML]{EBF1DE}\rotatebox{75}{FSAF}  & \cellcolor[HTML]{EBF1DE}\rotatebox{75}{RepPoints} & \cellcolor[HTML]{EBF1DE}\rotatebox{75}{TOOD}  & \cellcolor[HTML]{EBF1DE}\rotatebox{75}{ATSS}  & \cellcolor[HTML]{EBF1DE}\rotatebox{75}{FoveaBox} & \cellcolor[HTML]{EBF1DE}\rotatebox{75}{VarifocalNet} \\ \hline
  \multicolumn{1}{c}{\cellcolor[HTML]{E2EFDA}Clean}            & \multicolumn{1}{c}{\cellcolor[HTML]{FED6DE}0.394}          & \multicolumn{1}{c}{\cellcolor[HTML]{00B0F0}1.000}          & \multicolumn{1}{c}{\cellcolor[HTML]{00B0F0}1.000}            & \multicolumn{1}{c}{\cellcolor[HTML]{00B0F0}1.000}          & \multicolumn{1}{c}{\cellcolor[HTML]{60CEF6}0.813}          & \multicolumn{1}{c}{\cellcolor[HTML]{07B3F1}0.987}          & \multicolumn{1}{c}{\cellcolor[HTML]{00B0F0}1.000}          & \multicolumn{1}{c}{\cellcolor[HTML]{00B0F0}1.000}          & \multicolumn{1}{c}{\cellcolor[HTML]{00B0F0}1.000}          & \multicolumn{1}{c}{\cellcolor[HTML]{07B3F1}0.987}          & \multicolumn{1}{c}{\cellcolor[HTML]{00B0F0}1.000}          & \multicolumn{1}{c}{\cellcolor[HTML]{00B0F0}1.000}          & \multicolumn{1}{c}{\cellcolor[HTML]{00B0F0}1.000}          & \multicolumn{1}{c}{\cellcolor[HTML]{00B0F0}1.000}          & \multicolumn{1}{c}{\cellcolor[HTML]{00B0F0}1.000}          & \multicolumn{1}{c}{\cellcolor[HTML]{00B0F0}1.000}          & \multicolumn{1}{c}{\cellcolor[HTML]{04B2F1}0.994}          & \multicolumn{1}{c}{\cellcolor[HTML]{00B0F0}1.000}          & \multicolumn{1}{c}{\cellcolor[HTML]{00B0F0}1.000}          \\
  \multicolumn{1}{c}{\cellcolor[HTML]{E2EFDA}Random Noise}     & \multicolumn{1}{c}{\cellcolor[HTML]{FE6681}0.093}          & \multicolumn{1}{c}{\cellcolor[HTML]{FEE5EA}0.433}          & \multicolumn{1}{c}{\cellcolor[HTML]{04B2F1}0.993}            & \multicolumn{1}{c}{\cellcolor[HTML]{FEF2F4}0.467}          & \multicolumn{1}{c}{\cellcolor[HTML]{4FC9F5}0.847}          & \multicolumn{1}{c}{\cellcolor[HTML]{26BCF3}0.927}          & \multicolumn{1}{c}{\cellcolor[HTML]{26BCF3}0.927}          & \multicolumn{1}{c}{\cellcolor[HTML]{63CFF6}0.807}          & \multicolumn{1}{c}{\cellcolor[HTML]{4BC8F5}0.853}          & \multicolumn{1}{c}{\cellcolor[HTML]{DDF5FD}0.567}          & \multicolumn{1}{c}{\cellcolor[HTML]{E1F6FE}0.560}          & \multicolumn{1}{c}{\cellcolor[HTML]{66D0F6}0.800}          & \multicolumn{1}{c}{\cellcolor[HTML]{00B0F0}1.000}          & \multicolumn{1}{c}{\cellcolor[HTML]{FED9E0}0.400}          & \multicolumn{1}{c}{\cellcolor[HTML]{BCEAFC}0.633}          & \multicolumn{1}{c}{\cellcolor[HTML]{4BC8F5}0.853}          & \multicolumn{1}{c}{\cellcolor[HTML]{FEDEE4}0.413}          & \multicolumn{1}{c}{\cellcolor[HTML]{E1F6FE}0.560}          & \multicolumn{1}{c}{\cellcolor[HTML]{5CCDF6}0.820}          \\
  \multicolumn{1}{c}{\cellcolor[HTML]{E2EFDA}SSD}              & \multicolumn{1}{c}{\cellcolor[HTML]{FE4365}\textbf{0.000}} & \multicolumn{1}{c}{\cellcolor[HTML]{06B2F1}0.989}          & \multicolumn{1}{c}{\cellcolor[HTML]{29BDF3}0.921}            & \multicolumn{1}{c}{\cellcolor[HTML]{00B0F0}1.000}          & \multicolumn{1}{c}{\cellcolor[HTML]{FEAFBD}0.288}          & \multicolumn{1}{c}{\cellcolor[HTML]{74D4F7}0.774}          & \multicolumn{1}{c}{\cellcolor[HTML]{7CD7F8}0.757}          & \multicolumn{1}{c}{\cellcolor[HTML]{06B2F1}0.989}          & \multicolumn{1}{c}{\cellcolor[HTML]{04B1F1}0.994}          & \multicolumn{1}{c}{\cellcolor[HTML]{00B0F0}1.000}          & \multicolumn{1}{c}{\cellcolor[HTML]{29BDF3}0.921}          & \multicolumn{1}{c}{\cellcolor[HTML]{12B6F2}0.966}          & \multicolumn{1}{c}{\cellcolor[HTML]{00B0F0}1.000}          & \multicolumn{1}{c}{\cellcolor[HTML]{17B8F2}0.955}          & \multicolumn{1}{c}{\cellcolor[HTML]{0FB5F1}0.972}          & \multicolumn{1}{c}{\cellcolor[HTML]{3AC2F4}0.887}          & \multicolumn{1}{c}{\cellcolor[HTML]{1BB9F2}0.949}          & \multicolumn{1}{c}{\cellcolor[HTML]{29BDF3}0.921}          & \multicolumn{1}{c}{\cellcolor[HTML]{0CB4F1}0.977}          \\
  \multicolumn{1}{c}{\cellcolor[HTML]{E2EFDA}Faster R-CNN}     & \multicolumn{1}{c}{\cellcolor[HTML]{FE4365}0.000}          & \multicolumn{1}{c}{\cellcolor[HTML]{FE768F}\textbf{0.138}} & \multicolumn{1}{c}{\cellcolor[HTML]{FE5372}0.043}            & \multicolumn{1}{c}{\cellcolor[HTML]{FE4567}0.007}          & \multicolumn{1}{c}{\cellcolor[HTML]{FE4365}0.000}          & \multicolumn{1}{c}{\cellcolor[HTML]{FE4365}0.000}          & \multicolumn{1}{c}{\cellcolor[HTML]{FE4365}0.000}          & \multicolumn{1}{c}{\cellcolor[HTML]{FE4567}0.007}          & \multicolumn{1}{c}{\cellcolor[HTML]{FE4E6E}0.030}          & \multicolumn{1}{c}{\cellcolor[HTML]{FE5C7A}0.069}          & \multicolumn{1}{c}{\cellcolor[HTML]{FE7D94}0.155}          & \multicolumn{1}{c}{\cellcolor[HTML]{FE7A92}0.148}          & \multicolumn{1}{c}{\cellcolor[HTML]{FEDDE3}0.411}          & \multicolumn{1}{c}{\cellcolor[HTML]{FE728B}0.125}          & \multicolumn{1}{c}{\cellcolor[HTML]{FEA5B6}0.263}          & \multicolumn{1}{c}{\cellcolor[HTML]{FEB3C1}0.299}          & \multicolumn{1}{c}{\cellcolor[HTML]{FE6480}0.089}          & \multicolumn{1}{c}{\cellcolor[HTML]{FE6984}0.102}          & \multicolumn{1}{c}{\cellcolor[HTML]{FEAEBD}0.286}          \\
  \multicolumn{1}{c}{\cellcolor[HTML]{E2EFDA}Swin Transformer} & \multicolumn{1}{c}{\cellcolor[HTML]{FE4365}0.000}          & \multicolumn{1}{c}{\cellcolor[HTML]{FE4768}0.011}          & \multicolumn{1}{c}{\cellcolor[HTML]{FE5372}\textbf{0.043}}   & \multicolumn{1}{c}{\cellcolor[HTML]{FE4365}0.000}          & \multicolumn{1}{c}{\cellcolor[HTML]{FE4365}0.000}          & \multicolumn{1}{c}{\cellcolor[HTML]{FE95A8}0.219}          & \multicolumn{1}{c}{\cellcolor[HTML]{FEA1B2}0.251}          & \multicolumn{1}{c}{\cellcolor[HTML]{FECFD8}0.374}          & \multicolumn{1}{c}{\cellcolor[HTML]{FEF1F4}0.465}          & \multicolumn{1}{c}{\cellcolor[HTML]{FE4365}0.000}          & \multicolumn{1}{c}{\cellcolor[HTML]{FE4768}0.011}          & \multicolumn{1}{c}{\cellcolor[HTML]{FE4365}0.000}          & \multicolumn{1}{c}{\cellcolor[HTML]{FE5372}0.043}          & \multicolumn{1}{c}{\cellcolor[HTML]{FE4365}0.000}          & \multicolumn{1}{c}{\cellcolor[HTML]{FE4466}0.005}          & \multicolumn{1}{c}{\cellcolor[HTML]{FEB3C1}0.299}          & \multicolumn{1}{c}{\cellcolor[HTML]{FE4365}0.000}          & \multicolumn{1}{c}{\cellcolor[HTML]{FE4365}0.000}          & \multicolumn{1}{c}{\cellcolor[HTML]{FE637F}0.086}          \\
  \multicolumn{1}{c}{\cellcolor[HTML]{E2EFDA}YOLOv3}           & \multicolumn{1}{c}{\cellcolor[HTML]{FE4365}0.000}          & \multicolumn{1}{c}{\cellcolor[HTML]{FECED7}0.372}          & \multicolumn{1}{c}{\cellcolor[HTML]{FE4B6C}0.023}            & \multicolumn{1}{c}{\cellcolor[HTML]{FE5372}\textbf{0.045}} & \multicolumn{1}{c}{\cellcolor[HTML]{FE4365}0.000}          & \multicolumn{1}{c}{\cellcolor[HTML]{FE4B6C}0.023}          & \multicolumn{1}{c}{\cellcolor[HTML]{FE4D6D}0.029}          & \multicolumn{1}{c}{\cellcolor[HTML]{FE8399}0.171}          & \multicolumn{1}{c}{\cellcolor[HTML]{FEA6B6}0.265}          & \multicolumn{1}{c}{\cellcolor[HTML]{FEC0CB}0.333}          & \multicolumn{1}{c}{\cellcolor[HTML]{FEB9C5}0.314}          & \multicolumn{1}{c}{\cellcolor[HTML]{FEE3E8}0.427}          & \multicolumn{1}{c}{\cellcolor[HTML]{D2F1FD}0.589}          & \multicolumn{1}{c}{\cellcolor[HTML]{FEE6EA}0.434}          & \multicolumn{1}{c}{\cellcolor[HTML]{FAFEFF}0.511}          & \multicolumn{1}{c}{\cellcolor[HTML]{FEDEE4}0.414}          & \multicolumn{1}{c}{\cellcolor[HTML]{FE9AAC}0.233}          & \multicolumn{1}{c}{\cellcolor[HTML]{FE8399}0.171}          & \multicolumn{1}{c}{\cellcolor[HTML]{FEFEFE}0.498}          \\
  \multicolumn{1}{c}{\cellcolor[HTML]{E2EFDA}YOLOv5n}          & \multicolumn{1}{c}{\cellcolor[HTML]{FE4365}0.000}          & \multicolumn{1}{c}{\cellcolor[HTML]{2CBEF3}0.914}          & \multicolumn{1}{c}{\cellcolor[HTML]{24BCF3}0.930}            & \multicolumn{1}{c}{\cellcolor[HTML]{D7F3FD}0.579}          & \multicolumn{1}{c}{\cellcolor[HTML]{FE4365}\textbf{0.000}} & \multicolumn{1}{c}{\cellcolor[HTML]{FE5A78}0.063}          & \multicolumn{1}{c}{\cellcolor[HTML]{FE99AC}0.231}          & \multicolumn{1}{c}{\cellcolor[HTML]{FEFAFA}0.487}          & \multicolumn{1}{c}{\cellcolor[HTML]{CFF0FD}0.595}          & \multicolumn{1}{c}{\cellcolor[HTML]{49C7F5}0.858}          & \multicolumn{1}{c}{\cellcolor[HTML]{8DDCF9}0.725}          & \multicolumn{1}{c}{\cellcolor[HTML]{46C6F5}0.864}          & \multicolumn{1}{c}{\cellcolor[HTML]{1AB8F2}0.950}          & \multicolumn{1}{c}{\cellcolor[HTML]{5ECEF6}0.816}          & \multicolumn{1}{c}{\cellcolor[HTML]{4EC9F5}0.848}          & \multicolumn{1}{c}{\cellcolor[HTML]{21BAF2}0.937}          & \multicolumn{1}{c}{\cellcolor[HTML]{64CFF6}0.804}          & \multicolumn{1}{c}{\cellcolor[HTML]{4EC9F5}0.848}          & \multicolumn{1}{c}{\cellcolor[HTML]{1EBAF2}0.943}          \\
  \multicolumn{1}{c}{\cellcolor[HTML]{E2EFDA}YOLOv5s}          & \multicolumn{1}{c}{\cellcolor[HTML]{FE4365}0.000}          & \multicolumn{1}{c}{\cellcolor[HTML]{C8EEFC}0.609}          & \multicolumn{1}{c}{\cellcolor[HTML]{FED1D9}0.379}            & \multicolumn{1}{c}{\cellcolor[HTML]{FE4667}0.009}          & \multicolumn{1}{c}{\cellcolor[HTML]{FE4365}0.000}          & \multicolumn{1}{c}{\cellcolor[HTML]{FE4465}\textbf{0.003}} & \multicolumn{1}{c}{\cellcolor[HTML]{FE5473}0.047}          & \multicolumn{1}{c}{\cellcolor[HTML]{FE4566}0.006}          & \multicolumn{1}{c}{\cellcolor[HTML]{FED1D9}0.379}          & \multicolumn{1}{c}{\cellcolor[HTML]{BDEBFC}0.630}          & \multicolumn{1}{c}{\cellcolor[HTML]{BCEAFC}0.633}          & \multicolumn{1}{c}{\cellcolor[HTML]{BDEBFC}0.630}          & \multicolumn{1}{c}{\cellcolor[HTML]{25BCF3}0.929}          & \multicolumn{1}{c}{\cellcolor[HTML]{B0E7FB}0.655}          & \multicolumn{1}{c}{\cellcolor[HTML]{A2E3FA}0.683}          & \multicolumn{1}{c}{\cellcolor[HTML]{33C0F3}0.901}          & \multicolumn{1}{c}{\cellcolor[HTML]{CBEFFC}0.602}          & \multicolumn{1}{c}{\cellcolor[HTML]{E1F6FE}0.559}          & \multicolumn{1}{c}{\cellcolor[HTML]{7AD6F8}0.761}          \\
  \multicolumn{1}{c}{\cellcolor[HTML]{E2EFDA}YOLOv5m}          & \multicolumn{1}{c}{\cellcolor[HTML]{FE4465}0.003}          & \multicolumn{1}{c}{\cellcolor[HTML]{CBEFFC}0.603}          & \multicolumn{1}{c}{\cellcolor[HTML]{FFFFFF}0.500}            & \multicolumn{1}{c}{\cellcolor[HTML]{FE6F89}0.118}          & \multicolumn{1}{c}{\cellcolor[HTML]{FE5372}0.045}          & \multicolumn{1}{c}{\cellcolor[HTML]{FE6681}0.094}          & \multicolumn{1}{c}{\cellcolor[HTML]{FE4365}\textbf{0.000}} & \multicolumn{1}{c}{\cellcolor[HTML]{FE4566}0.006}          & \multicolumn{1}{c}{\cellcolor[HTML]{FE8A9F}0.191}          & \multicolumn{1}{c}{\cellcolor[HTML]{E5F7FE}0.551}          & \multicolumn{1}{c}{\cellcolor[HTML]{D4F2FD}0.585}          & \multicolumn{1}{c}{\cellcolor[HTML]{56CBF6}0.833}          & \multicolumn{1}{c}{\cellcolor[HTML]{30BFF3}0.906}          & \multicolumn{1}{c}{\cellcolor[HTML]{C9EFFC}0.606}          & \multicolumn{1}{c}{\cellcolor[HTML]{A0E2FA}0.688}          & \multicolumn{1}{c}{\cellcolor[HTML]{49C7F5}0.858}          & \multicolumn{1}{c}{\cellcolor[HTML]{CEF0FD}0.597}          & \multicolumn{1}{c}{\cellcolor[HTML]{E0F6FE}0.561}          & \multicolumn{1}{c}{\cellcolor[HTML]{77D5F7}0.767}          \\
  \multicolumn{1}{c}{\cellcolor[HTML]{E2EFDA}YOLOv5l}          & \multicolumn{1}{c}{\cellcolor[HTML]{FE4465}0.003}          & \multicolumn{1}{c}{\cellcolor[HTML]{84D9F8}0.743}          & \multicolumn{1}{c}{\cellcolor[HTML]{D8F3FD}0.578}            & \multicolumn{1}{c}{\cellcolor[HTML]{FE5E7B}0.073}          & \multicolumn{1}{c}{\cellcolor[HTML]{FE506F}0.035}          & \multicolumn{1}{c}{\cellcolor[HTML]{FE6984}0.102}          & \multicolumn{1}{c}{\cellcolor[HTML]{FE5674}0.051}          & \multicolumn{1}{c}{\cellcolor[HTML]{FE5876}\textbf{0.057}} & \multicolumn{1}{c}{\cellcolor[HTML]{FEB7C4}0.311}          & \multicolumn{1}{c}{\cellcolor[HTML]{78D6F8}0.765}          & \multicolumn{1}{c}{\cellcolor[HTML]{DEF5FE}0.565}          & \multicolumn{1}{c}{\cellcolor[HTML]{70D3F7}0.781}          & \multicolumn{1}{c}{\cellcolor[HTML]{00B0F0}1.000}          & \multicolumn{1}{c}{\cellcolor[HTML]{49C7F5}0.857}          & \multicolumn{1}{c}{\cellcolor[HTML]{55CBF5}0.835}          & \multicolumn{1}{c}{\cellcolor[HTML]{63CFF6}0.806}          & \multicolumn{1}{c}{\cellcolor[HTML]{FEC9D3}0.359}          & \multicolumn{1}{c}{\cellcolor[HTML]{92DEF9}0.714}          & \multicolumn{1}{c}{\cellcolor[HTML]{2EBFF3}0.911}          \\
  \multicolumn{1}{c}{\cellcolor[HTML]{E2EFDA}YOLOv5x}          & \multicolumn{1}{c}{\cellcolor[HTML]{FE4365}0.000}          & \multicolumn{1}{c}{\cellcolor[HTML]{A2E3FA}0.683}          & \multicolumn{1}{c}{\cellcolor[HTML]{FE718B}0.124}            & \multicolumn{1}{c}{\cellcolor[HTML]{FE4F6E}0.032}          & \multicolumn{1}{c}{\cellcolor[HTML]{FE4466}0.005}          & \multicolumn{1}{c}{\cellcolor[HTML]{FE4969}0.016}          & \multicolumn{1}{c}{\cellcolor[HTML]{FE5775}0.054}          & \multicolumn{1}{c}{\cellcolor[HTML]{FE4365}0.000}          & \multicolumn{1}{c}{\cellcolor[HTML]{FE4365}\textbf{0.000}} & \multicolumn{1}{c}{\cellcolor[HTML]{FE8DA2}0.199}          & \multicolumn{1}{c}{\cellcolor[HTML]{FE6782}0.097}          & \multicolumn{1}{c}{\cellcolor[HTML]{D4F2FD}0.586}          & \multicolumn{1}{c}{\cellcolor[HTML]{60CEF6}0.812}          & \multicolumn{1}{c}{\cellcolor[HTML]{FE4969}0.016}          & \multicolumn{1}{c}{\cellcolor[HTML]{E7F8FE}0.548}          & \multicolumn{1}{c}{\cellcolor[HTML]{C3EDFC}0.618}          & \multicolumn{1}{c}{\cellcolor[HTML]{E1F6FE}0.559}          & \multicolumn{1}{c}{\cellcolor[HTML]{FE4768}0.011}          & \multicolumn{1}{c}{\cellcolor[HTML]{87DAF8}0.737}          \\
  \multicolumn{1}{c}{\cellcolor[HTML]{E2EFDA}Cascade R-CNN}    & \multicolumn{1}{c}{\cellcolor[HTML]{FE4365}0.000}          & \multicolumn{1}{c}{\cellcolor[HTML]{FE9BAD}0.236}          & \multicolumn{1}{c}{\cellcolor[HTML]{FE4C6C}0.024}            & \multicolumn{1}{c}{\cellcolor[HTML]{FE4465}0.003}          & \multicolumn{1}{c}{\cellcolor[HTML]{FE4365}0.000}          & \multicolumn{1}{c}{\cellcolor[HTML]{FE4566}0.006}          & \multicolumn{1}{c}{\cellcolor[HTML]{FE4869}0.015}          & \multicolumn{1}{c}{\cellcolor[HTML]{FE4465}0.003}          & \multicolumn{1}{c}{\cellcolor[HTML]{FE5977}0.061}          & \multicolumn{1}{c}{\cellcolor[HTML]{FE6884}\textbf{0.101}} & \multicolumn{1}{c}{\cellcolor[HTML]{FE7D95}0.156}          & \multicolumn{1}{c}{\cellcolor[HTML]{FE9BAD}0.236}          & \multicolumn{1}{c}{\cellcolor[HTML]{FEC4CE}0.344}          & \multicolumn{1}{c}{\cellcolor[HTML]{FE748D}0.132}          & \multicolumn{1}{c}{\cellcolor[HTML]{FE8DA2}0.199}          & \multicolumn{1}{c}{\cellcolor[HTML]{FEB5C2}0.304}          & \multicolumn{1}{c}{\cellcolor[HTML]{FE728B}0.126}          & \multicolumn{1}{c}{\cellcolor[HTML]{FE5B78}0.064}          & \multicolumn{1}{c}{\cellcolor[HTML]{FE97A9}0.224}          \\
  \multicolumn{1}{c}{\cellcolor[HTML]{E2EFDA}RetinaNet}        & \multicolumn{1}{c}{\cellcolor[HTML]{FE4365}0.000}          & \multicolumn{1}{c}{\cellcolor[HTML]{FE5B79}0.066}          & \multicolumn{1}{c}{\cellcolor[HTML]{FE4667}0.009}            & \multicolumn{1}{c}{\cellcolor[HTML]{FE4667}0.009}          & \multicolumn{1}{c}{\cellcolor[HTML]{FE4365}0.000}          & \multicolumn{1}{c}{\cellcolor[HTML]{FE4465}0.003}          & \multicolumn{1}{c}{\cellcolor[HTML]{FE4365}0.000}          & \multicolumn{1}{c}{\cellcolor[HTML]{FE4465}0.003}          & \multicolumn{1}{c}{\cellcolor[HTML]{FE4465}0.003}          & \multicolumn{1}{c}{\cellcolor[HTML]{FE4A6A}0.019}          & \multicolumn{1}{c}{\cellcolor[HTML]{FE5271}\textbf{0.041}} & \multicolumn{1}{c}{\cellcolor[HTML]{FE5473}0.047}          & \multicolumn{1}{c}{\cellcolor[HTML]{FEA4B4}0.259}          & \multicolumn{1}{c}{\cellcolor[HTML]{FE5271}0.041}          & \multicolumn{1}{c}{\cellcolor[HTML]{FE5574}0.050}          & \multicolumn{1}{c}{\cellcolor[HTML]{FE6D88}0.114}          & \multicolumn{1}{c}{\cellcolor[HTML]{FE4566}0.006}          & \multicolumn{1}{c}{\cellcolor[HTML]{FE4465}0.003}          & \multicolumn{1}{c}{\cellcolor[HTML]{FE5977}0.060}          \\
  \multicolumn{1}{c}{\cellcolor[HTML]{E2EFDA}Mask R-CNN}       & \multicolumn{1}{c}{\cellcolor[HTML]{FE4365}0.000}          & \multicolumn{1}{c}{\cellcolor[HTML]{FE9DAE}0.240}          & \multicolumn{1}{c}{\cellcolor[HTML]{FE5171}0.039}            & \multicolumn{1}{c}{\cellcolor[HTML]{FE5F7C}0.075}          & \multicolumn{1}{c}{\cellcolor[HTML]{FE4567}0.007}          & \multicolumn{1}{c}{\cellcolor[HTML]{FE4C6C}0.025}          & \multicolumn{1}{c}{\cellcolor[HTML]{FE4768}0.011}          & \multicolumn{1}{c}{\cellcolor[HTML]{FE5876}0.057}          & \multicolumn{1}{c}{\cellcolor[HTML]{FE728B}0.125}          & \multicolumn{1}{c}{\cellcolor[HTML]{FE8EA2}0.201}          & \multicolumn{1}{c}{\cellcolor[HTML]{FE738C}0.129}          & \multicolumn{1}{c}{\cellcolor[HTML]{FE9DAE}\textbf{0.240}} & \multicolumn{1}{c}{\cellcolor[HTML]{FBFEFF}0.509}          & \multicolumn{1}{c}{\cellcolor[HTML]{FEA1B2}0.251}          & \multicolumn{1}{c}{\cellcolor[HTML]{FEC8D2}0.355}          & \multicolumn{1}{c}{\cellcolor[HTML]{FEBCC8}0.323}          & \multicolumn{1}{c}{\cellcolor[HTML]{FE708A}0.122}          & \multicolumn{1}{c}{\cellcolor[HTML]{FE7790}0.140}          & \multicolumn{1}{c}{\cellcolor[HTML]{FEACBB}0.280}          \\
  \multicolumn{1}{c}{\cellcolor[HTML]{E2EFDA}FreeAnchor}       & \multicolumn{1}{c}{\cellcolor[HTML]{FE4365}0.000}          & \multicolumn{1}{c}{\cellcolor[HTML]{FEA9B8}0.272}          & \multicolumn{1}{c}{\cellcolor[HTML]{FE5C79}0.067}            & \multicolumn{1}{c}{\cellcolor[HTML]{FE5C79}0.067}          & \multicolumn{1}{c}{\cellcolor[HTML]{FE4365}0.000}          & \multicolumn{1}{c}{\cellcolor[HTML]{FE4668}0.010}          & \multicolumn{1}{c}{\cellcolor[HTML]{FE4465}0.003}          & \multicolumn{1}{c}{\cellcolor[HTML]{FE4365}0.000}          & \multicolumn{1}{c}{\cellcolor[HTML]{FE4C6D}0.026}          & \multicolumn{1}{c}{\cellcolor[HTML]{FE728B}0.125}          & \multicolumn{1}{c}{\cellcolor[HTML]{FE8DA1}0.198}          & \multicolumn{1}{c}{\cellcolor[HTML]{FE768F}0.137}          & \multicolumn{1}{c}{\cellcolor[HTML]{FEAEBD}\textbf{0.287}} & \multicolumn{1}{c}{\cellcolor[HTML]{FE5674}0.051}          & \multicolumn{1}{c}{\cellcolor[HTML]{FE8FA3}0.204}          & \multicolumn{1}{c}{\cellcolor[HTML]{FEF1F3}0.463}          & \multicolumn{1}{c}{\cellcolor[HTML]{FE8299}0.169}          & \multicolumn{1}{c}{\cellcolor[HTML]{FE5D7A}0.070}          & \multicolumn{1}{c}{\cellcolor[HTML]{FE8FA3}0.204}          \\
  \rowcolor[HTML]{FE4365} 
  \multicolumn{1}{c}{\cellcolor[HTML]{E2EFDA}FSAF}             & \multicolumn{1}{c}{\cellcolor[HTML]{FE4365}0.000}          & \multicolumn{1}{c}{\cellcolor[HTML]{FE4365}0.000}          & \multicolumn{1}{c}{\cellcolor[HTML]{FE4365}0.000}            & \multicolumn{1}{c}{\cellcolor[HTML]{FE4365}0.000}          & \multicolumn{1}{c}{\cellcolor[HTML]{FE4365}0.000}          & \multicolumn{1}{c}{\cellcolor[HTML]{FE4365}0.000}          & \multicolumn{1}{c}{\cellcolor[HTML]{FE4365}0.000}          & \multicolumn{1}{c}{\cellcolor[HTML]{FE4365}0.000}          & \multicolumn{1}{c}{\cellcolor[HTML]{FE4365}0.000}          & \multicolumn{1}{c}{\cellcolor[HTML]{FE4768}0.011}          & \multicolumn{1}{c}{\cellcolor[HTML]{FE4466}0.005}          & \multicolumn{1}{c}{\cellcolor[HTML]{FE4969}0.016}          & \multicolumn{1}{c}{\cellcolor[HTML]{FEA4B4}0.258}          & \multicolumn{1}{c}{\cellcolor[HTML]{FE4466}\textbf{0.005}} & \multicolumn{1}{c}{\cellcolor[HTML]{FE5372}0.043}          & \multicolumn{1}{c}{\cellcolor[HTML]{FE4466}0.005}          & \multicolumn{1}{c}{\cellcolor[HTML]{FE4365}0.000}          & \multicolumn{1}{c}{\cellcolor[HTML]{FE4466}0.005}          & \multicolumn{1}{c}{\cellcolor[HTML]{FE758E}0.134}          \\
  \rowcolor[HTML]{FE4365} 
  \multicolumn{1}{c}{\cellcolor[HTML]{E2EFDA}RepPoints}        & \multicolumn{1}{c}{\cellcolor[HTML]{FE4365}0.000}          & \multicolumn{1}{c}{\cellcolor[HTML]{FE4365}0.000}          & \multicolumn{1}{c}{\cellcolor[HTML]{FE4466}0.005}            & \multicolumn{1}{c}{\cellcolor[HTML]{FE4365}0.000}          & \multicolumn{1}{c}{\cellcolor[HTML]{FE4365}0.000}          & \multicolumn{1}{c}{\cellcolor[HTML]{FE4365}0.000}          & \multicolumn{1}{c}{\cellcolor[HTML]{FE4365}0.000}          & \multicolumn{1}{c}{\cellcolor[HTML]{FE4A6B}0.021}          & \multicolumn{1}{c}{\cellcolor[HTML]{FE6581}0.091}          & \multicolumn{1}{c}{\cellcolor[HTML]{FE4365}0.000}          & \multicolumn{1}{c}{\cellcolor[HTML]{FE4365}0.000}          & \multicolumn{1}{c}{\cellcolor[HTML]{FE4365}0.000}          & \multicolumn{1}{c}{\cellcolor[HTML]{FE6782}0.097}          & \multicolumn{1}{c}{\cellcolor[HTML]{FE4365}0.000}          & \multicolumn{1}{c}{\cellcolor[HTML]{FE4365}\textbf{0.000}} & \multicolumn{1}{c}{\cellcolor[HTML]{FE4A6B}0.021}          & \multicolumn{1}{c}{\cellcolor[HTML]{FE4365}0.000}          & \multicolumn{1}{c}{\cellcolor[HTML]{FE4365}0.000}          & \multicolumn{1}{c}{\cellcolor[HTML]{FE4365}0.000}          \\
  \rowcolor[HTML]{FE4365} 
  \multicolumn{1}{c}{\cellcolor[HTML]{E2EFDA}TOOD}             & \multicolumn{1}{c}{\cellcolor[HTML]{FE4365}0.000}          & \multicolumn{1}{c}{\cellcolor[HTML]{FE4365}0.000}          & \multicolumn{1}{c}{\cellcolor[HTML]{FE4365}0.000}            & \multicolumn{1}{c}{\cellcolor[HTML]{FE4365}0.000}          & \multicolumn{1}{c}{\cellcolor[HTML]{FE4365}0.000}          & \multicolumn{1}{c}{\cellcolor[HTML]{FE4365}0.000}          & \multicolumn{1}{c}{\cellcolor[HTML]{FE4365}0.000}          & \multicolumn{1}{c}{\cellcolor[HTML]{FE4365}0.000}          & \multicolumn{1}{c}{\cellcolor[HTML]{FE4365}0.000}          & \multicolumn{1}{c}{\cellcolor[HTML]{FE4365}0.000}          & \multicolumn{1}{c}{\cellcolor[HTML]{FE4A6B}0.021}          & \multicolumn{1}{c}{\cellcolor[HTML]{FE4365}0.000}          & \multicolumn{1}{c}{\cellcolor[HTML]{FE8EA2}0.201}          & \multicolumn{1}{c}{\cellcolor[HTML]{FE4365}0.000}          & \multicolumn{1}{c}{\cellcolor[HTML]{FE4A6B}0.021}          & \multicolumn{1}{c}{\cellcolor[HTML]{FE4C6D}\textbf{0.026}} & \multicolumn{1}{c}{\cellcolor[HTML]{FE4365}0.000}          & \multicolumn{1}{c}{\cellcolor[HTML]{FE4365}0.000}          & \multicolumn{1}{c}{\cellcolor[HTML]{FE4365}0.000}          \\
  \multicolumn{1}{c}{\cellcolor[HTML]{E2EFDA}ATSS}             & \multicolumn{1}{c}{\cellcolor[HTML]{FE4365}0.000}          & \multicolumn{1}{c}{\cellcolor[HTML]{FEBCC8}0.323}          & \multicolumn{1}{c}{\cellcolor[HTML]{FE6984}0.103}            & \multicolumn{1}{c}{\cellcolor[HTML]{FE748D}0.132}          & \multicolumn{1}{c}{\cellcolor[HTML]{FE4365}0.000}          & \multicolumn{1}{c}{\cellcolor[HTML]{FE6883}0.100}          & \multicolumn{1}{c}{\cellcolor[HTML]{FE5D7A}0.071}          & \multicolumn{1}{c}{\cellcolor[HTML]{FE5876}0.058}          & \multicolumn{1}{c}{\cellcolor[HTML]{FE99AB}0.229}          & \multicolumn{1}{c}{\cellcolor[HTML]{FE8EA2}0.200}          & \multicolumn{1}{c}{\cellcolor[HTML]{FE9BAD}0.235}          & \multicolumn{1}{c}{\cellcolor[HTML]{FE94A7}0.216}          & \multicolumn{1}{c}{\cellcolor[HTML]{BFECFC}0.626}          & \multicolumn{1}{c}{\cellcolor[HTML]{FEAAB9}0.274}          & \multicolumn{1}{c}{\cellcolor[HTML]{FEC2CD}0.339}          & \multicolumn{1}{c}{\cellcolor[HTML]{FEEEF1}0.455}          & \multicolumn{1}{c}{\cellcolor[HTML]{FE8FA3}\textbf{0.203}} & \multicolumn{1}{c}{\cellcolor[HTML]{FE718A}0.123}          & \multicolumn{1}{c}{\cellcolor[HTML]{FEB0BE}0.290}          \\
  \multicolumn{1}{c}{\cellcolor[HTML]{E2EFDA}FoveaBox}         & \multicolumn{1}{c}{\cellcolor[HTML]{FE4365}0.000}          & \multicolumn{1}{c}{\cellcolor[HTML]{FE4466}0.005}          & \multicolumn{1}{c}{\cellcolor[HTML]{FE4365}0.000}            & \multicolumn{1}{c}{\cellcolor[HTML]{FE4D6D}0.027}          & \multicolumn{1}{c}{\cellcolor[HTML]{FE4365}0.000}          & \multicolumn{1}{c}{\cellcolor[HTML]{FE4969}0.016}          & \multicolumn{1}{c}{\cellcolor[HTML]{FE4D6D}0.027}          & \multicolumn{1}{c}{\cellcolor[HTML]{FE4D6D}0.027}          & \multicolumn{1}{c}{\cellcolor[HTML]{FE849A}0.175}          & \multicolumn{1}{c}{\cellcolor[HTML]{FE4365}0.000}          & \multicolumn{1}{c}{\cellcolor[HTML]{FE4466}0.005}          & \multicolumn{1}{c}{\cellcolor[HTML]{FE4768}0.011}          & \multicolumn{1}{c}{\cellcolor[HTML]{FE4B6B}0.022}          & \multicolumn{1}{c}{\cellcolor[HTML]{FE4365}0.000}          & \multicolumn{1}{c}{\cellcolor[HTML]{FE4365}0.000}          & \multicolumn{1}{c}{\cellcolor[HTML]{FE5F7C}0.076}          & \multicolumn{1}{c}{\cellcolor[HTML]{FE4365}0.000}          & \multicolumn{1}{c}{\cellcolor[HTML]{FE4365}\textbf{0.000}} & \multicolumn{1}{c}{\cellcolor[HTML]{FE4B6B}0.022}          \\
  \rowcolor[HTML]{FE4365} 
  \multicolumn{1}{c}{\cellcolor[HTML]{E2EFDA}VarifocalNet}     & \multicolumn{1}{c}{\cellcolor[HTML]{FE4365}0.000}          & \multicolumn{1}{c}{\cellcolor[HTML]{FE4365}0.000}          & \multicolumn{1}{c}{\cellcolor[HTML]{FE4365}0.000}            & \multicolumn{1}{c}{\cellcolor[HTML]{FE4365}0.000}          & \multicolumn{1}{c}{\cellcolor[HTML]{FE4365}0.000}          & \multicolumn{1}{c}{\cellcolor[HTML]{FE4365}0.000}          & \multicolumn{1}{c}{\cellcolor[HTML]{FE4365}0.000}          & \multicolumn{1}{c}{\cellcolor[HTML]{FE4768}0.011}          & \multicolumn{1}{c}{\cellcolor[HTML]{FE4F6E}0.032}          & \multicolumn{1}{c}{\cellcolor[HTML]{FE4365}0.000}          & \multicolumn{1}{c}{\cellcolor[HTML]{FE4768}0.011}          & \multicolumn{1}{c}{\cellcolor[HTML]{FE4365}0.000}          & \multicolumn{1}{c}{\cellcolor[HTML]{FE92A6}0.212}          & \multicolumn{1}{c}{\cellcolor[HTML]{FE4365}0.000}          & \multicolumn{1}{c}{\cellcolor[HTML]{FE4365}0.000}          & \multicolumn{1}{c}{\cellcolor[HTML]{FE4365}0.000}          & \multicolumn{1}{c}{\cellcolor[HTML]{FE4365}0.000}          & \multicolumn{1}{c}{\cellcolor[HTML]{FE4365}0.000}          & \multicolumn{1}{c}{\cellcolor[HTML]{FE4A6B}\textbf{0.021}} \\
   
  \hline
  \end{tabular*}
  \caption{
    Experimental results of physical background attack in the metric of DR, where white-box attacks are highlighted in bold and the rest are black-box attacks.
  The \textbf{redder} the cell, the \textbf{higher} the \textbf{attack efficacy}.
  The \textbf{bluer} the cell, the \textbf{lower} the \textbf{attack efficacy}.
  Clean and Random Noise mean experiments on clean images and images with random noise, respectively.
The 19 detectors of the first row and the first column are for detection and perturbation optimization, respectively.  }
  \label{table_physical_attack}
  \end{table*}

For digital attacks, a notable reduction is evident in both mAP0.5 and mAP0.5:0.95. 
Interestingly, the mAP of several experimental outcomes tends to decline only up to a certain threshold, approximately reaching 0.250 for mAP0.5 and 0.160 for mAP0.5:0.95. 
This observation appears to deviate from our qualitative experimental findings and prompts a deeper investigation. 
Our exploration involved an extensive examination of qualitative experimental outcomes. 
These examinations encompassed representative instances of successful and unsuccessful attack attempts, as illustrated in Fig. \ref{fig_digital_attack_successful_failed}.
It is observed that:

\begin{itemize}
    \item The background perturbations crafted for digital attacks exhibit robust attack efficacy across a wide spectrum of objects as shown in Fig. \ref{fig_digital_attack_successful_failed} (a), encompassing entities like individuals, animals, and fruits, while even accommodating multi-scale objects and imbalanced categories.
    \item In the context of unsuccessful attack attempts, as illustrated in Fig. \ref{fig_digital_attack_successful_failed} (b), a clear trend emerges, revealing that objects that resist concealment are predominantly those situated amidst other objects. 
    Instances include scenarios such as a mobile phone placed in front of individuals, people within a bus, or various items scattered across a table.
\end{itemize}

In summary, when considering dispersed objects as the target of concealment, the proposed approach presented in this study exhibits a notably elevated level of attack performance, as evidenced by the experimental results, which also partially explain the performance discrepancy between the physical and the digital attacks.

\subsection{Physical Background Attacks}
\label{subsec_physical_background_attacks}

We conduct physical background attacks with various SOTA object detection methods same as digital attacks.
Please note that if there are no additional instructions, the detector and target we use by default are YOLOv5 and bottle (please refer to the attached file for the video demo), and the confidence score is set as 0.25.
The reason for choosing a bottle of cola as the tarted object is that it is a common object in daily life and easier to control for a more comprehensive evaluation in comparison with person, vehicle, etc.
Technically, we use an LED screen to display background perturbations and then place objects in front of the screen, followed by video recording and detection.

\begin{figure}[!h]
  \centering
  \begin{subfigure}{0.99\linewidth}
    \includegraphics[width=1\linewidth]{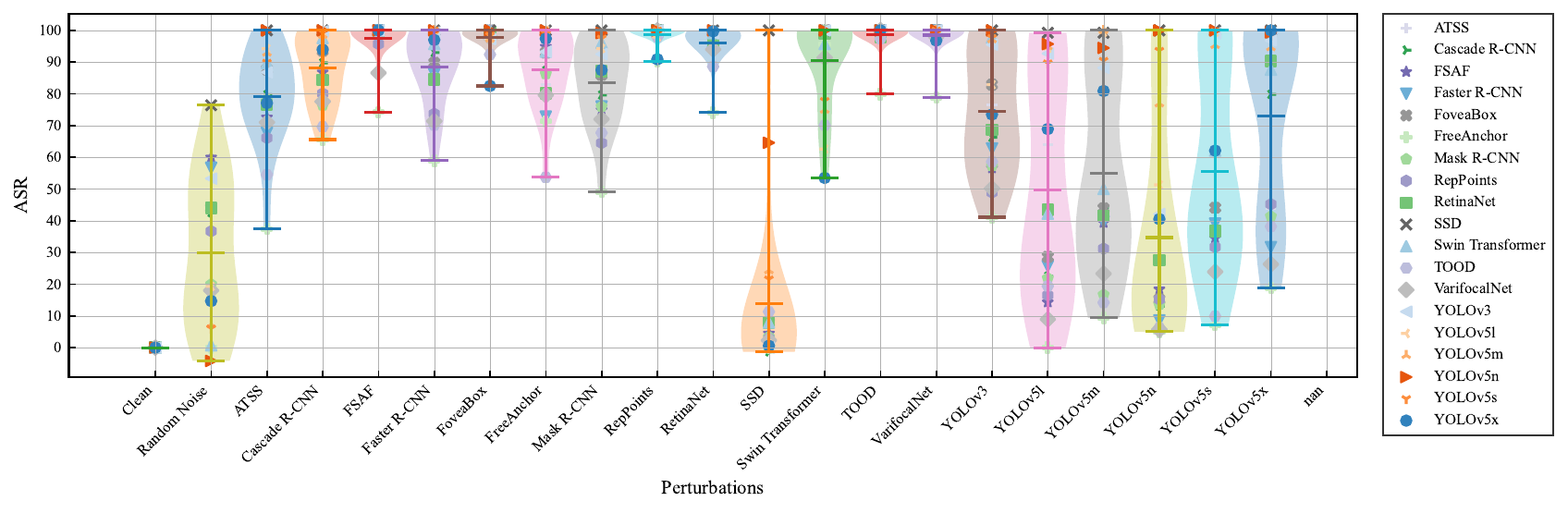}
    \caption{Attack perspective}
  \end{subfigure}
  \begin{subfigure}{0.99\linewidth}
    \includegraphics[width=1\linewidth]{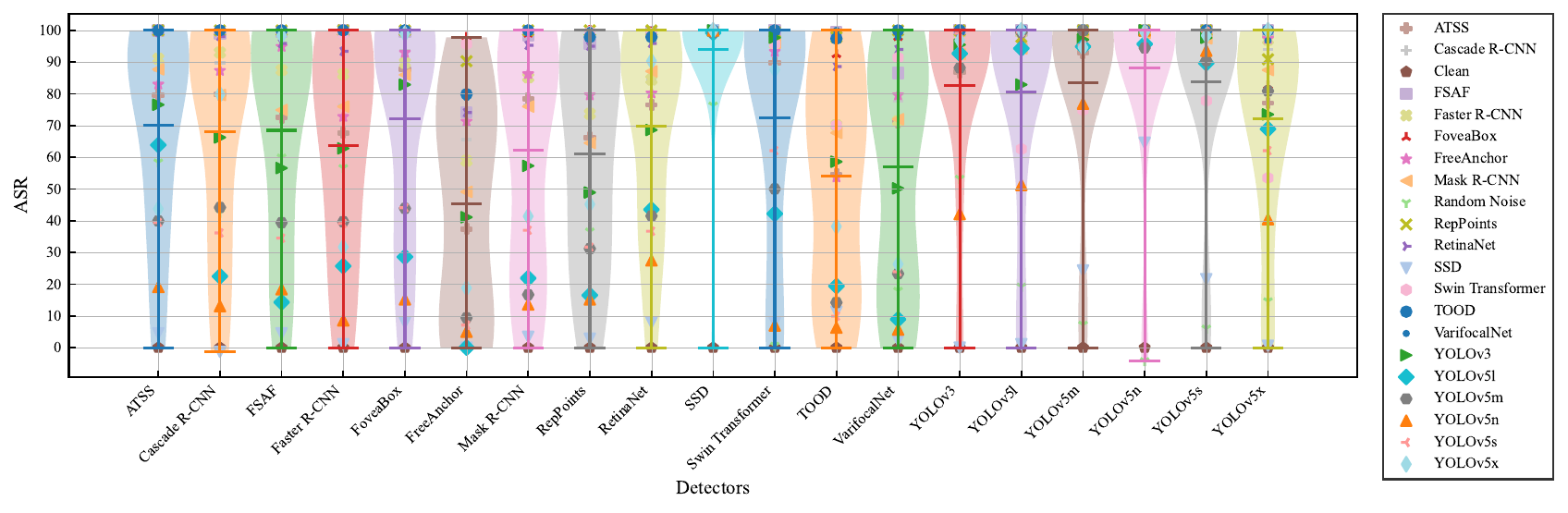}
    \caption{Detection perspective}
  \end{subfigure}
  \caption{The figures visualize the quantitative experimental results of physical background attacks from the perspectives of detection and attack in the metric of ASR. Please zoom in for a better view.}
  \label{fig_physical_attack_visualization}
  \end{figure}

We report the quantitative experimental results in Table \ref{table_physical_attack}, and the metric is DR.
In addition, we also visualize the quantitative experimental results from the perspective of detection and attack in terms of ASR in Fig. \ref{fig_physical_attack_visualization}.
It is concluded that:

\begin{itemize}
  \item The attack efficacy of background perturbations can be fluently extended to physical attacks with ASR up to 100\%, i.e., the elaborated background perturbations remain undistorted after cross-domain transformation, which not only strengthens the key value of background features but also reveals their resilience.
  \item The physical attack efficacy can also transfer well between different models under black box conditions, which poses significant concerns for the applications of DNNs in safety-critical scenarios.
\end{itemize}


The qualitative experimental results are shown in Fig. \ref{fig_background_attack_exhibition} (d).
It is observed that the objects in front of our elaborated background perturbations are successfully hidden from being detected.

\begin{figure}[!h]
  \centering
  \begin{subfigure}{0.99\linewidth}
    \includegraphics[width=1\linewidth]{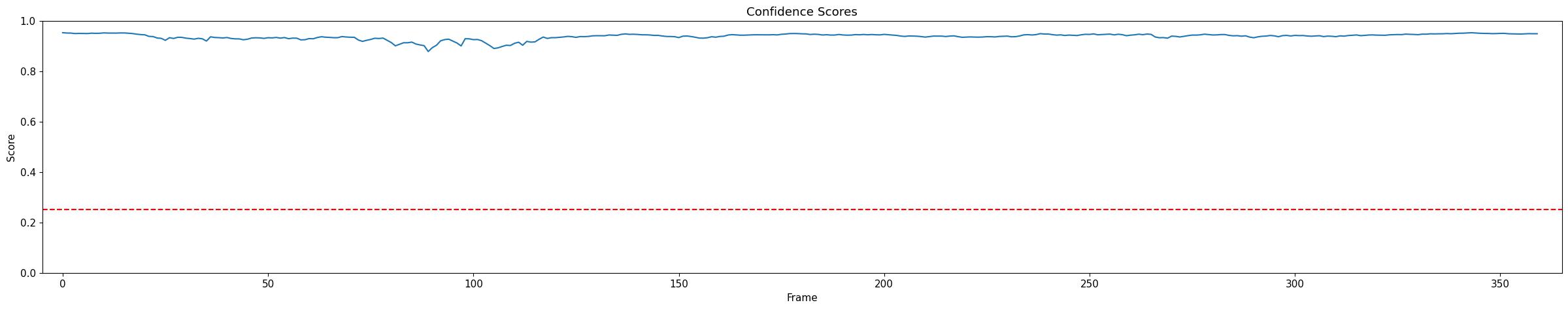}
    \caption{Clean}
  \end{subfigure}
  \begin{subfigure}{0.99\linewidth}
    \includegraphics[width=1\linewidth]{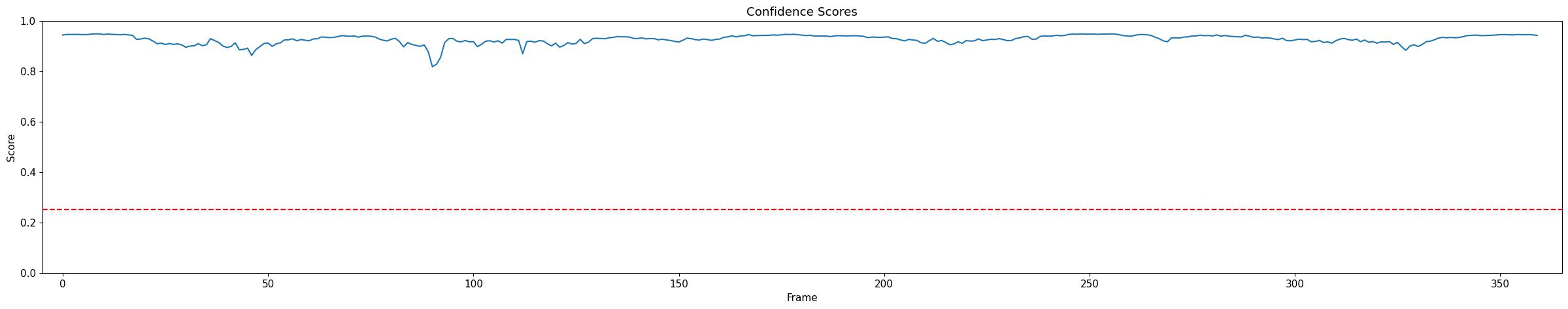}
    \caption{Random}
  \end{subfigure}
  \begin{subfigure}{0.99\linewidth}
    \includegraphics[width=1\linewidth]{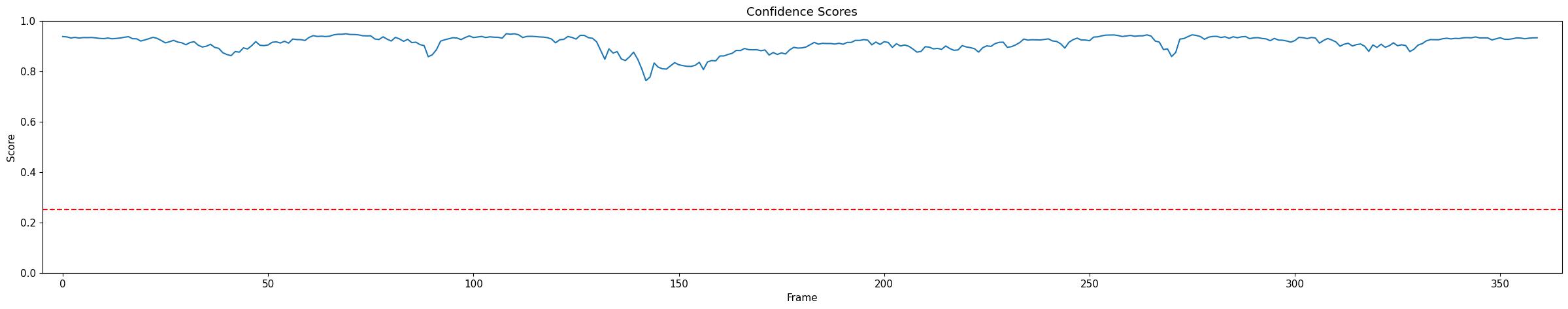}
    \caption{DTA}
  \end{subfigure}
  \begin{subfigure}{0.99\linewidth}
    \includegraphics[width=1\linewidth]{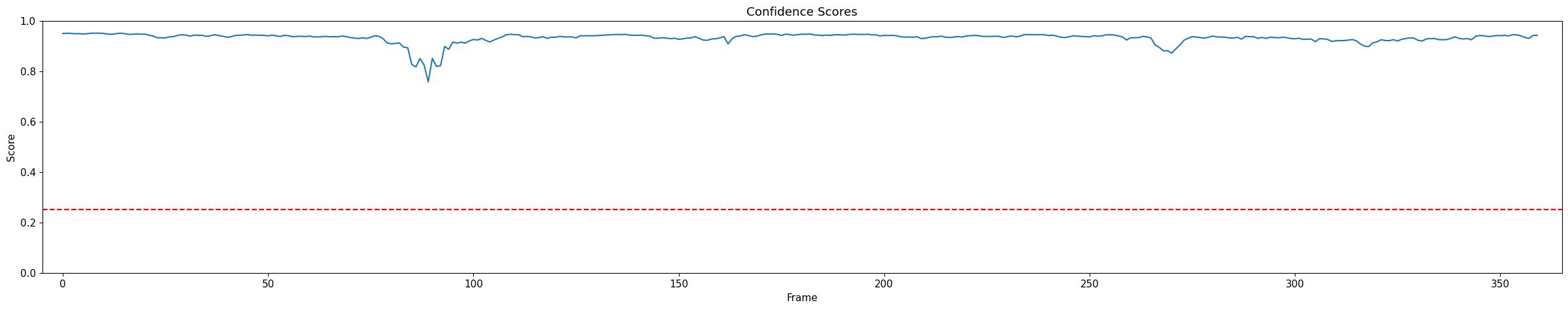}
    \caption{FCA}
  \end{subfigure}
  \begin{subfigure}{0.99\linewidth}
      \includegraphics[width=1\linewidth]{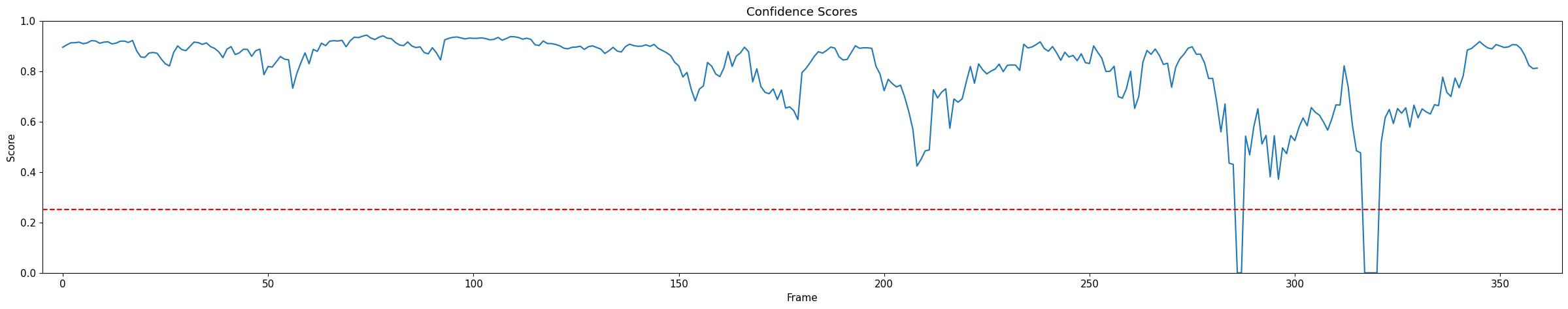}
      \caption{ACTIVE}
    \end{subfigure}
    \begin{subfigure}{0.99\linewidth}
      \includegraphics[width=1\linewidth]{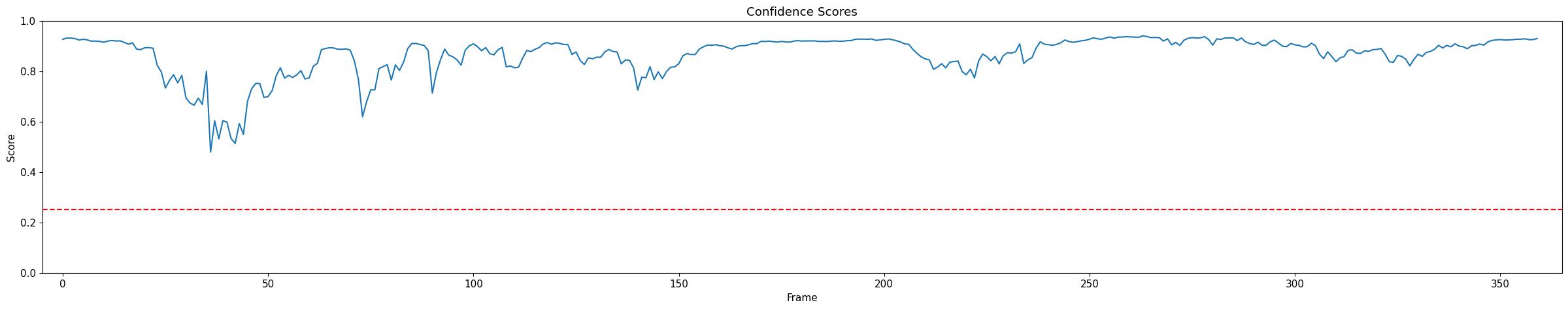}
      \caption{AA-fg}
    \end{subfigure}
    \begin{subfigure}{0.99\linewidth}
      \includegraphics[width=1\linewidth]{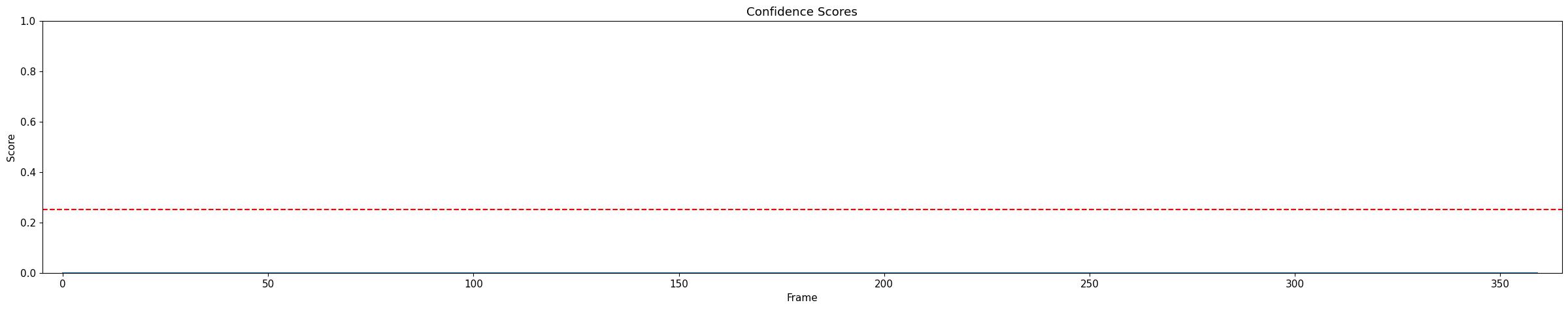}
      \caption{AA-bg}
    \end{subfigure}
    \begin{subfigure}{0.99\linewidth}
      \includegraphics[width=1\linewidth]{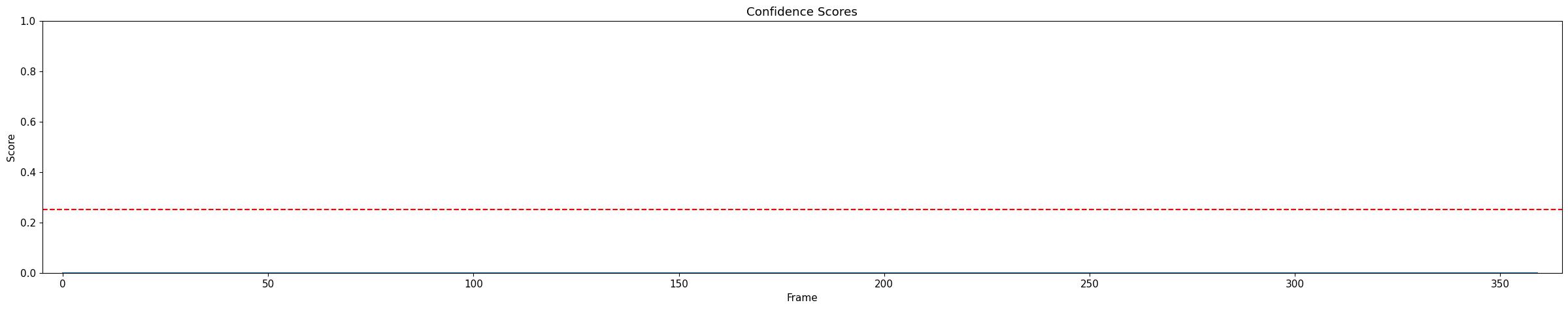}
      \caption{AA-bf}
    \end{subfigure}
  \caption{The comparison of confidence scores (depicted by the blue line) for car detection within a physically-based simulation, utilizing YOLOv3 as the victim model. A confidence threshold, represented by the red dashed line, is established at 0.25. This implies that any confidence score below 0.25 is set as 0 and interpreted as a failure to detect anything. Please zoom in for a better view.}
  \label{fig_scores_line_yolov3_car}
  \end{figure}
  
  \begin{figure}[!h]
  \centering
  \begin{subfigure}{0.99\linewidth}
    \includegraphics[width=1\linewidth]{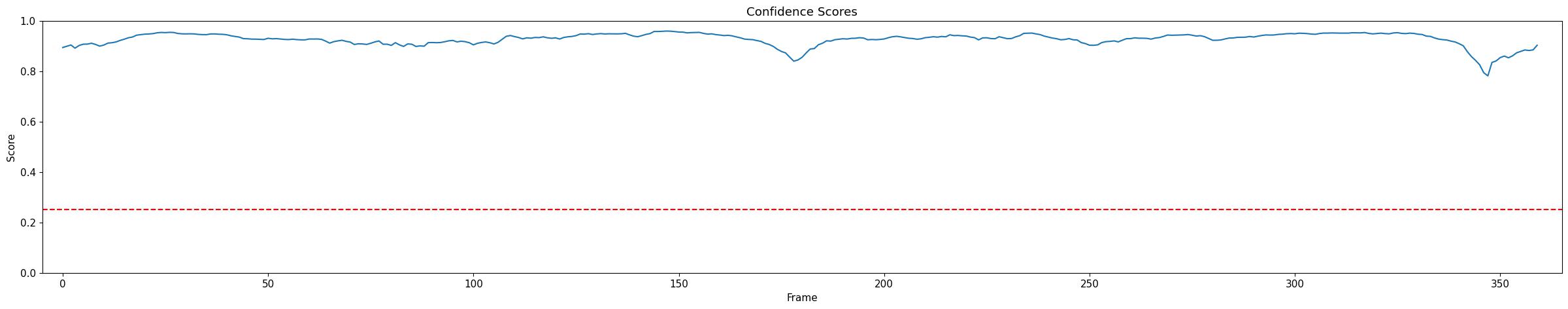}
    \caption{Clean}
  \end{subfigure}
  \begin{subfigure}{0.99\linewidth}
    \includegraphics[width=1\linewidth]{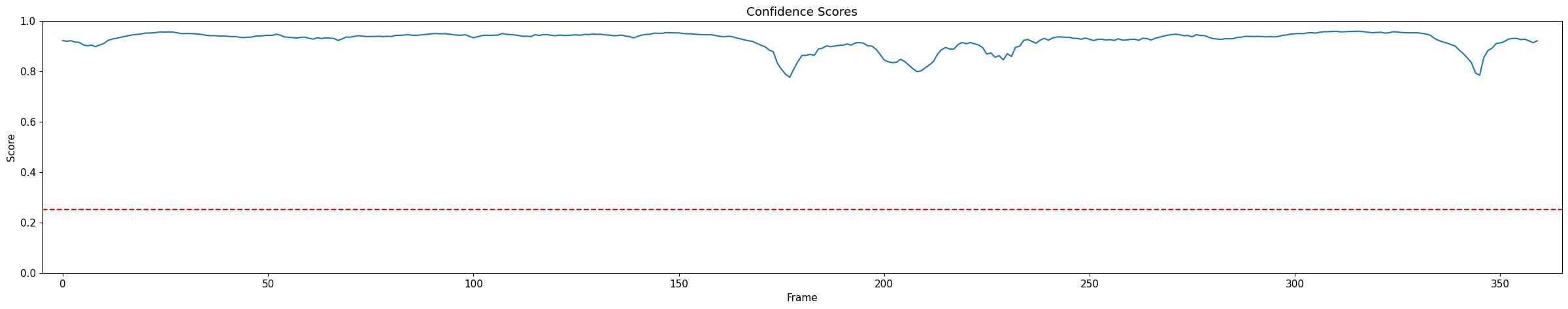}
    \caption{Random}
  \end{subfigure}
  \begin{subfigure}{0.99\linewidth}
    \includegraphics[width=1\linewidth]{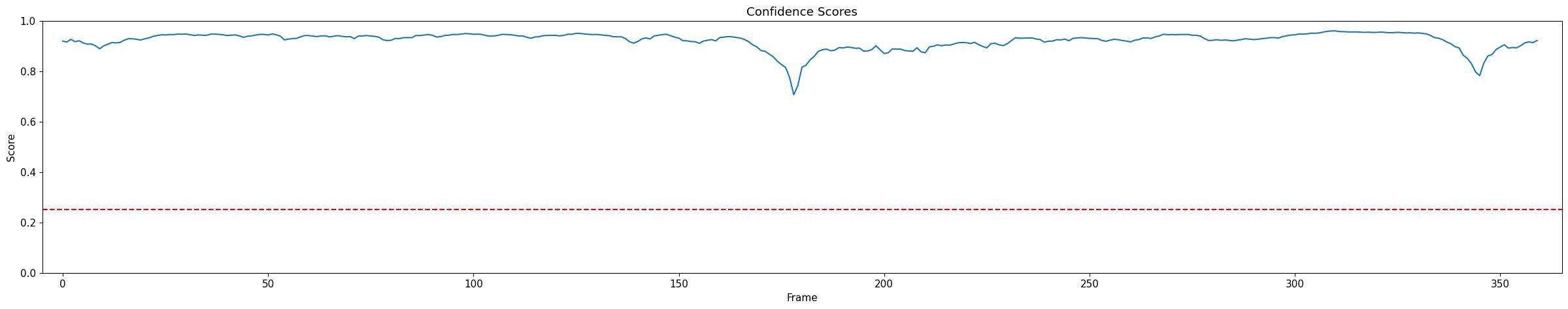}
    \caption{DTA}
  \end{subfigure}
  \begin{subfigure}{0.99\linewidth}
    \includegraphics[width=1\linewidth]{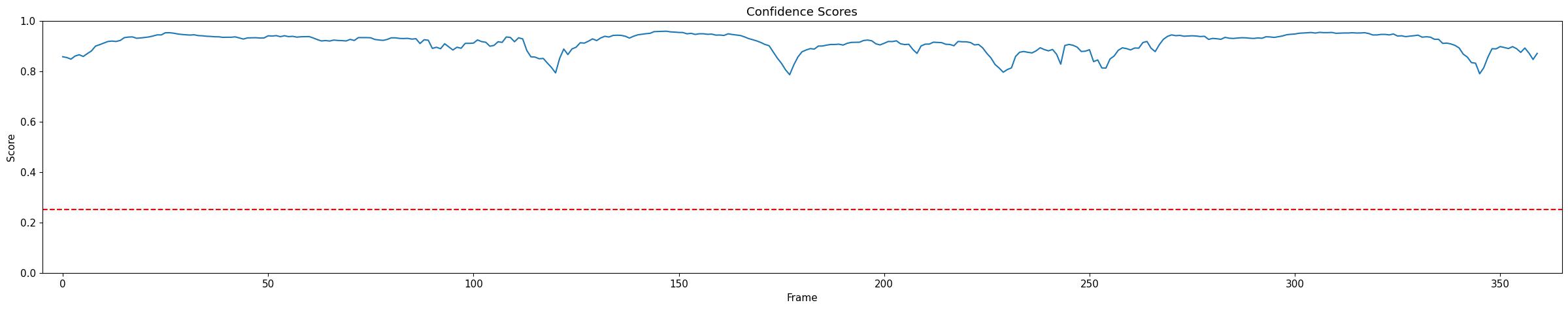}
    \caption{FCA}
  \end{subfigure}
  \begin{subfigure}{0.99\linewidth}
      \includegraphics[width=1\linewidth]{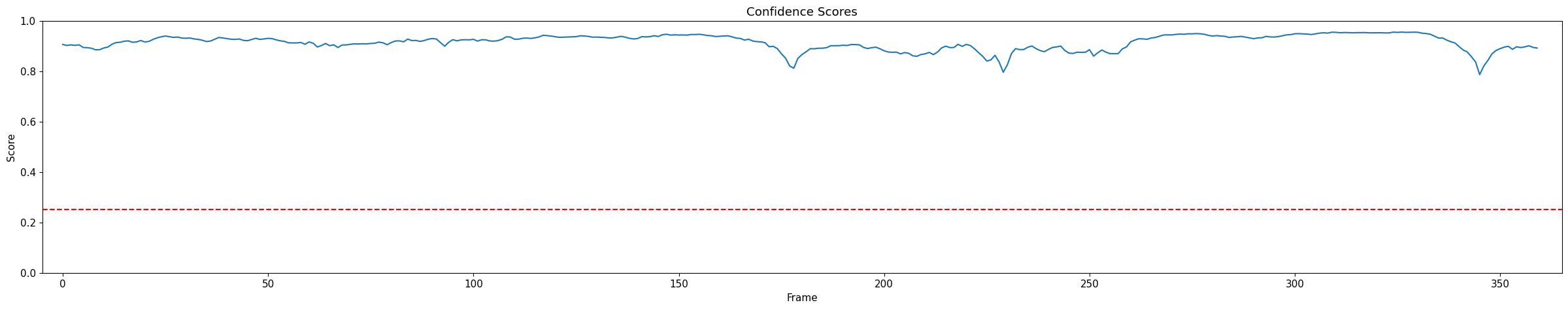}
      \caption{ACTIVE}
    \end{subfigure}
    \begin{subfigure}{0.99\linewidth}
      \includegraphics[width=1\linewidth]{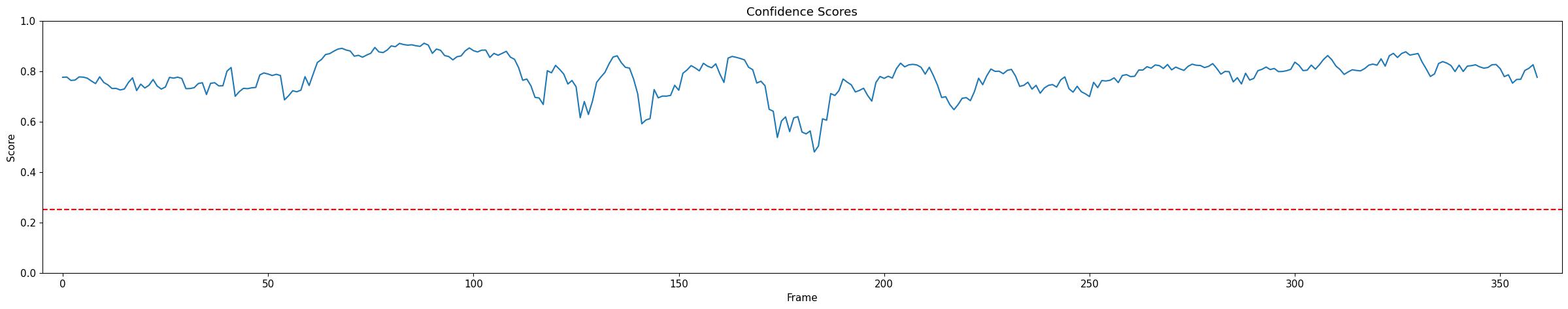}
      \caption{AA-fg}
    \end{subfigure}
    \begin{subfigure}{0.99\linewidth}
      \includegraphics[width=1\linewidth]{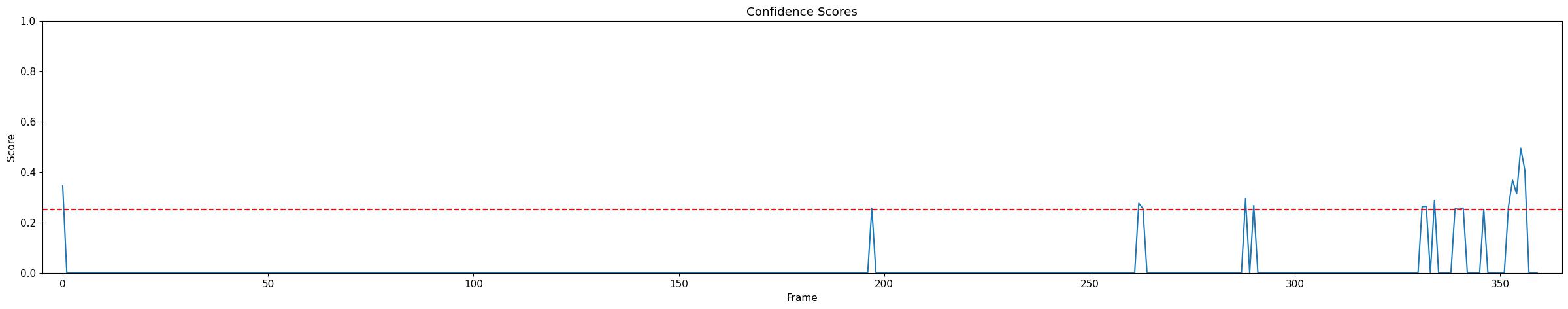}
      \caption{AA-bg}
    \end{subfigure}
    \begin{subfigure}{0.99\linewidth}
      \includegraphics[width=1\linewidth]{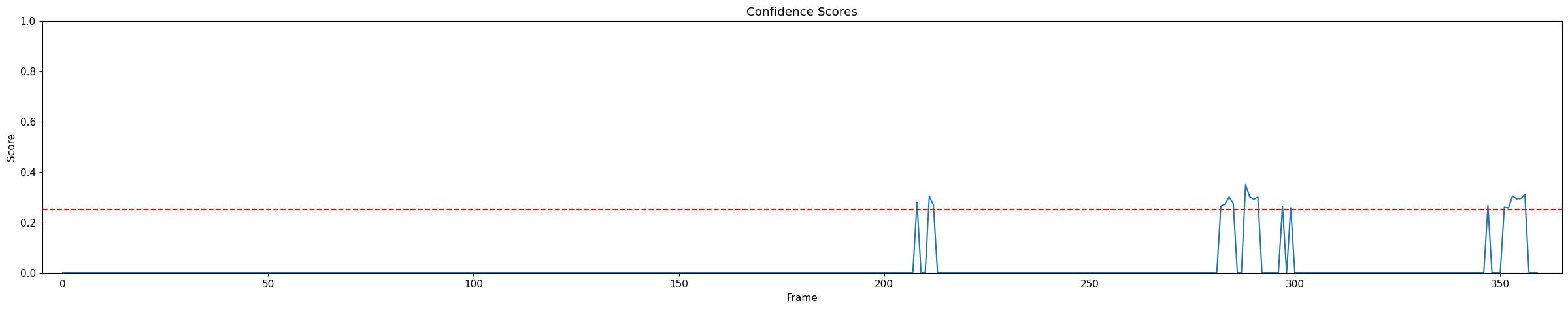}
      \caption{AA-bf}
    \end{subfigure}
  \caption{The comparison of confidence scores (depicted by the blue line) for person detection within a physically-based simulation, utilizing YOLOv3 as the victim model. A confidence threshold, represented by the red dashed line, is established at 0.25. This implies that any confidence score below 0.25 is set as 0 and interpreted as a failure to detect anything. Please zoom in for a better view.}
  \label{fig_scores_line_yolov3_person}
  \end{figure}

\begin{table}[!htbp]
  \setlength{\tabcolsep}{0.8mm}
  \tiny
  \centering
  \begin{tabular}{c
    >{\columncolor[HTML]{EDEDED}}c 
    >{\columncolor[HTML]{00B0F0}}c 
    >{\columncolor[HTML]{00B0F0}}c 
    >{\columncolor[HTML]{00B0F0}}c 
    >{\columncolor[HTML]{00B0F0}}c c
    >{\columncolor[HTML]{00B0F0}}c cc}
    \hline
                                                            & TH   & \cellcolor[HTML]{FFF2CC}Clean & \cellcolor[HTML]{FFF2CC}RD    & \cellcolor[HTML]{FFF2CC}DTA   & \cellcolor[HTML]{FFF2CC}FCA   & \cellcolor[HTML]{FFF2CC}ACTIVE         & \cellcolor[HTML]{FFF2CC}AA-fg & \cellcolor[HTML]{FFF2CC}AA-bg          & \cellcolor[HTML]{FFF2CC}AA-bf          \\\hline
    \cellcolor[HTML]{FCE4D6}                                & 0.25 & \cellcolor[HTML]{3DC3F4}0.881 & \cellcolor[HTML]{43C5F4}0.869 & \cellcolor[HTML]{F3FCFF}0.525 & \cellcolor[HTML]{D1F1FD}0.592 & \cellcolor[HTML]{FA9FA0}\textbf{0.181} & \cellcolor[HTML]{32C0F3}0.903 & \cellcolor[HTML]{23BBF3}0.933          & \cellcolor[HTML]{1EBAF2}0.942          \\
    \cellcolor[HTML]{FCE4D6}                                & 0.35 & \cellcolor[HTML]{44C6F4}0.867 & \cellcolor[HTML]{5ECDF6}0.817 & \cellcolor[HTML]{FDDBDB}0.381 & \cellcolor[HTML]{FEEAEA}0.431 & \cellcolor[HTML]{F98C8D}\textbf{0.117} & \cellcolor[HTML]{39C2F4}0.889 & \cellcolor[HTML]{2ABDF3}0.919          & \cellcolor[HTML]{24BBF3}0.931          \\
    \cellcolor[HTML]{FCE4D6}                                & 0.45 & \cellcolor[HTML]{4CC8F5}0.853 & \cellcolor[HTML]{87DAF8}0.736 & \cellcolor[HTML]{FBB4B5}0.253 & \cellcolor[HTML]{FCC3C3}0.300 & \cellcolor[HTML]{F87D7F}\textbf{0.069} & \cellcolor[HTML]{3CC3F4}0.883 & \cellcolor[HTML]{31BFF3}0.906          & \cellcolor[HTML]{25BCF3}0.928          \\
    \multirow{-4}{*}{\cellcolor[HTML]{FCE4D6}SSD}           & 0.55 & \cellcolor[HTML]{72D4F7}0.778 & \cellcolor[HTML]{CAEFFC}0.606 & \cellcolor[HTML]{FA9D9E}0.175 & \cellcolor[HTML]{FAA7A8}0.208 & \cellcolor[HTML]{F87375}\textbf{0.036} & \cellcolor[HTML]{4DC8F5}0.850 & \cellcolor[HTML]{35C1F4}0.897          & \cellcolor[HTML]{2ABDF3}0.919          \\\hline
    \cellcolor[HTML]{FCE4D6}                                & 0.25 & 1.000                         & 1.000                         & 1.000                         & 1.000                         & \cellcolor[HTML]{66D0F6}\textbf{0.800} & 1.000                         & \cellcolor[HTML]{0FB5F1}0.972          & \cellcolor[HTML]{03B1F1}0.994          \\
    \cellcolor[HTML]{FCE4D6}                                & 0.35 & 1.000                         & 1.000                         & 1.000                         & 1.000                         & \cellcolor[HTML]{86DAF8}\textbf{0.739} & 1.000                         & \cellcolor[HTML]{13B6F2}0.964          & \cellcolor[HTML]{05B2F1}0.992          \\
    \cellcolor[HTML]{FCE4D6}                                & 0.45 & 1.000                         & 1.000                         & 1.000                         & 1.000                         & \cellcolor[HTML]{A2E3FA}\textbf{0.683} & 1.000                         & \cellcolor[HTML]{16B7F2}0.958          & \cellcolor[HTML]{0AB4F1}0.981          \\
    \multirow{-4}{*}{\cellcolor[HTML]{FCE4D6}Faster R-CNN}  & 0.55 & 1.000                         & 1.000                         & 1.000                         & 1.000                         & \cellcolor[HTML]{BCEAFC}\textbf{0.633} & 1.000                         & \cellcolor[HTML]{1AB8F2}0.950          & \cellcolor[HTML]{0DB4F1}0.975          \\\hline
    \cellcolor[HTML]{FCE4D6}                                & 0.25 & 1.000                         & 1.000                         & 1.000                         & \cellcolor[HTML]{0AB4F1}0.981 & \cellcolor[HTML]{17B8F2}\textbf{0.956} & 1.000                         & \cellcolor[HTML]{00B0F0}1.000          & \cellcolor[HTML]{0CB4F1}0.978          \\
    \cellcolor[HTML]{FCE4D6}                                & 0.35 & 1.000                         & \cellcolor[HTML]{02B1F1}0.997 & 1.000                         & \cellcolor[HTML]{14B7F2}0.961 & \cellcolor[HTML]{23BBF3}\textbf{0.933} & 1.000                         & \cellcolor[HTML]{06B2F1}0.989          & \cellcolor[HTML]{11B6F1}0.967          \\
    \cellcolor[HTML]{FCE4D6}                                & 0.45 & 1.000                         & \cellcolor[HTML]{02B1F1}0.997 & \cellcolor[HTML]{02B1F1}0.997 & \cellcolor[HTML]{1EBAF2}0.942 & \cellcolor[HTML]{36C1F4}\textbf{0.894} & 1.000                         & \cellcolor[HTML]{0DB4F1}0.975          & \cellcolor[HTML]{17B8F2}0.956          \\
    \multirow{-4}{*}{\cellcolor[HTML]{FCE4D6}Swin}          & 0.55 & \cellcolor[HTML]{03B1F1}0.994 & \cellcolor[HTML]{05B2F1}0.992 & \cellcolor[HTML]{05B2F1}0.992 & \cellcolor[HTML]{25BCF3}0.928 & \cellcolor[HTML]{4AC7F5}\textbf{0.856} & \cellcolor[HTML]{02B1F1}0.997 & \cellcolor[HTML]{13B6F2}0.964          & \cellcolor[HTML]{20BAF2}0.939          \\\hline
    \cellcolor[HTML]{FCE4D6}                                & 0.25 & 1.000                         & 1.000                         & 1.000                         & 1.000                         & \cellcolor[HTML]{09B3F1}0.983          & 1.000                         & \cellcolor[HTML]{F8696B}\textbf{0.000} & \cellcolor[HTML]{F8696B}\textbf{0.000} \\
    \cellcolor[HTML]{FCE4D6}                                & 0.35 & 1.000                         & 1.000                         & 1.000                         & 1.000                         & \cellcolor[HTML]{09B3F1}0.983          & 1.000                         & \cellcolor[HTML]{F8696B}\textbf{0.000} & \cellcolor[HTML]{F8696B}\textbf{0.000} \\
    \cellcolor[HTML]{FCE4D6}                                & 0.45 & 1.000                         & 1.000                         & 1.000                         & 1.000                         & \cellcolor[HTML]{10B5F1}0.969          & 1.000                         & \cellcolor[HTML]{F8696B}\textbf{0.000} & \cellcolor[HTML]{F8696B}\textbf{0.000} \\
    \multirow{-4}{*}{\cellcolor[HTML]{FCE4D6}YOLOv3}        & 0.55 & 1.000                         & 1.000                         & 1.000                         & 1.000                         & \cellcolor[HTML]{25BCF3}0.928          & \cellcolor[HTML]{08B3F1}0.986 & \cellcolor[HTML]{F8696B}\textbf{0.000} & \cellcolor[HTML]{F8696B}\textbf{0.000} \\\hline
    \cellcolor[HTML]{FCE4D6}                                & 0.25 & \cellcolor[HTML]{38C2F4}0.892 & \cellcolor[HTML]{10B5F1}0.969 & \cellcolor[HTML]{40C4F4}0.875 & \cellcolor[HTML]{4CC8F5}0.853 & \cellcolor[HTML]{FA999A}\textbf{0.161} & \cellcolor[HTML]{19B8F2}0.953 & \cellcolor[HTML]{2EBFF3}0.911          & \cellcolor[HTML]{40C4F4}0.875          \\
    \cellcolor[HTML]{FCE4D6}                                & 0.35 & \cellcolor[HTML]{4EC9F5}0.847 & \cellcolor[HTML]{1DB9F2}0.944 & \cellcolor[HTML]{6CD2F7}0.789 & \cellcolor[HTML]{6CD2F7}0.789 & \cellcolor[HTML]{F98486}\textbf{0.092} & \cellcolor[HTML]{32C0F3}0.903 & \cellcolor[HTML]{8DDCF9}0.725          & \cellcolor[HTML]{97DFF9}0.706          \\
    \cellcolor[HTML]{FCE4D6}                                & 0.45 & \cellcolor[HTML]{7FD8F8}0.753 & \cellcolor[HTML]{32C0F3}0.903 & \cellcolor[HTML]{BCEAFC}0.633 & \cellcolor[HTML]{91DDF9}0.717 & \cellcolor[HTML]{F8797B}\textbf{0.056} & \cellcolor[HTML]{4DC8F5}0.850 & \cellcolor[HTML]{FEF5F5}0.467          & \cellcolor[HTML]{FEF5F5}0.467          \\
    \multirow{-4}{*}{\cellcolor[HTML]{FCE4D6}YOLOv5n}       & 0.55 & \cellcolor[HTML]{D8F3FD}0.578 & \cellcolor[HTML]{5ACCF6}0.825 & \cellcolor[HTML]{FEF0F1}0.453 & \cellcolor[HTML]{FEEAEA}0.431 & \cellcolor[HTML]{F87274}\textbf{0.031} & \cellcolor[HTML]{9CE1FA}0.694 & \cellcolor[HTML]{FBB9B9}0.267          & \cellcolor[HTML]{FBBABB}0.272          \\\hline
    \cellcolor[HTML]{FCE4D6}                                & 0.25 & \cellcolor[HTML]{32C0F3}0.903 & \cellcolor[HTML]{10B5F1}0.969 & \cellcolor[HTML]{65D0F6}0.803 & \cellcolor[HTML]{5BCDF6}0.822 & \cellcolor[HTML]{FDE6E6}\textbf{0.417} & 1.000                         & \cellcolor[HTML]{65D0F6}0.803          & \cellcolor[HTML]{5ACCF6}0.825          \\
    \cellcolor[HTML]{FCE4D6}                                & 0.35 & \cellcolor[HTML]{39C2F4}0.889 & \cellcolor[HTML]{13B6F2}0.964 & \cellcolor[HTML]{83D9F8}0.744 & \cellcolor[HTML]{6FD3F7}0.783 & \cellcolor[HTML]{FBBCBD}\textbf{0.278} & 1.000                         & \cellcolor[HTML]{81D8F8}0.747          & \cellcolor[HTML]{91DDF9}0.717          \\
    \cellcolor[HTML]{FCE4D6}                                & 0.45 & \cellcolor[HTML]{44C6F4}0.867 & \cellcolor[HTML]{1EBAF2}0.942 & \cellcolor[HTML]{ACE6FB}0.664 & \cellcolor[HTML]{90DDF9}0.719 & \cellcolor[HTML]{FA9D9E}\textbf{0.175} & 1.000                         & \cellcolor[HTML]{FAFEFF}0.511          & \cellcolor[HTML]{FEFBFB}0.489          \\
    \multirow{-4}{*}{\cellcolor[HTML]{FCE4D6}YOLOv5s}       & 0.55 & \cellcolor[HTML]{53CAF5}0.839 & \cellcolor[HTML]{32C0F3}0.903 & \cellcolor[HTML]{DCF5FD}0.569 & \cellcolor[HTML]{A8E4FA}0.672 & \cellcolor[HTML]{F87778}\textbf{0.047} & \cellcolor[HTML]{05B2F1}0.992 & \cellcolor[HTML]{FBB4B5}0.253          & \cellcolor[HTML]{FBB1B2}0.242          \\\hline
    \cellcolor[HTML]{FCE4D6}                                & 0.25 & 1.000                         & 1.000                         & 1.000                         & 1.000                         & \cellcolor[HTML]{08B3F1}0.986          & 1.000                         & \cellcolor[HTML]{4CC8F5}\textbf{0.853} & \cellcolor[HTML]{43C5F4}0.869          \\
    \cellcolor[HTML]{FCE4D6}                                & 0.35 & 1.000                         & 1.000                         & 1.000                         & 1.000                         & \cellcolor[HTML]{10B5F1}0.969          & 1.000                         & \cellcolor[HTML]{83D9F8}\textbf{0.744} & \cellcolor[HTML]{83D9F8}\textbf{0.744} \\
    \cellcolor[HTML]{FCE4D6}                                & 0.45 & 1.000                         & 1.000                         & 1.000                         & 1.000                         & \cellcolor[HTML]{16B7F2}0.958          & 1.000                         & \cellcolor[HTML]{C7EEFC}0.611          & \cellcolor[HTML]{F6FCFF}\textbf{0.519} \\
    \multirow{-4}{*}{\cellcolor[HTML]{FCE4D6}YOLOv5m}       & 0.55 & 1.000                         & 1.000                         & 1.000                         & \cellcolor[HTML]{02B1F1}0.997 & \cellcolor[HTML]{32C0F3}0.903          & 1.000                         & \cellcolor[HTML]{FEF0F0}0.450          & \cellcolor[HTML]{FDD5D5}\textbf{0.361} \\\hline
    \cellcolor[HTML]{FCE4D6}                                & 0.25 & 1.000                         & 1.000                         & 1.000                         & 1.000                         & \cellcolor[HTML]{00B0F0}1.000          & 1.000                         & \cellcolor[HTML]{00B0F0}1.000          & \cellcolor[HTML]{00B0F0}1.000          \\
    \cellcolor[HTML]{FCE4D6}                                & 0.35 & 1.000                         & 1.000                         & 1.000                         & 1.000                         & \cellcolor[HTML]{00B0F0}1.000          & 1.000                         & \cellcolor[HTML]{00B0F0}1.000          & \cellcolor[HTML]{00B0F0}1.000          \\
    \cellcolor[HTML]{FCE4D6}                                & 0.45 & 1.000                         & 1.000                         & 1.000                         & 1.000                         & \cellcolor[HTML]{00B0F0}1.000          & 1.000                         & \cellcolor[HTML]{00B0F0}1.000          & \cellcolor[HTML]{00B0F0}1.000          \\
    \multirow{-4}{*}{\cellcolor[HTML]{FCE4D6}YOLOv5l}       & 0.55 & 1.000                         & 1.000                         & 1.000                         & 1.000                         & \cellcolor[HTML]{00B0F0}1.000          & 1.000                         & \cellcolor[HTML]{05B2F1}\textbf{0.992} & \cellcolor[HTML]{02B1F1}0.997          \\\hline
    \cellcolor[HTML]{FCE4D6}                                & 0.25 & 1.000                         & 1.000                         & 1.000                         & 1.000                         & \cellcolor[HTML]{02B1F1}\textbf{0.997} & 1.000                         & \cellcolor[HTML]{00B0F0}1.000          & \cellcolor[HTML]{02B1F1}\textbf{0.997} \\
    \cellcolor[HTML]{FCE4D6}                                & 0.35 & 1.000                         & 1.000                         & 1.000                         & 1.000                         & \cellcolor[HTML]{02B1F1}\textbf{0.997} & 1.000                         & \cellcolor[HTML]{00B0F0}1.000          & \cellcolor[HTML]{02B1F1}\textbf{0.997} \\
    \cellcolor[HTML]{FCE4D6}                                & 0.45 & 1.000                         & 1.000                         & 1.000                         & 1.000                         & \cellcolor[HTML]{02B1F1}0.997          & 1.000                         & \cellcolor[HTML]{00B0F0}1.000          & \cellcolor[HTML]{05B2F1}\textbf{0.992} \\
    \multirow{-4}{*}{\cellcolor[HTML]{FCE4D6}YOLOv5x}       & 0.55 & 1.000                         & 1.000                         & 1.000                         & 1.000                         & \cellcolor[HTML]{02B1F1}0.997          & 1.000                         & \cellcolor[HTML]{00B0F0}1.000          & \cellcolor[HTML]{13B6F2}\textbf{0.964} \\\hline
    \cellcolor[HTML]{FCE4D6}                                & 0.25 & 1.000                         & 1.000                         & 1.000                         & 1.000                         & \cellcolor[HTML]{21BBF2}\textbf{0.936} & 1.000                         & \cellcolor[HTML]{0AB4F1}0.981          & \cellcolor[HTML]{0FB5F1}0.972          \\
    \cellcolor[HTML]{FCE4D6}                                & 0.35 & 1.000                         & 1.000                         & 1.000                         & 1.000                         & \cellcolor[HTML]{27BCF3}\textbf{0.925} & 1.000                         & \cellcolor[HTML]{0DB4F1}0.975          & \cellcolor[HTML]{03B1F1}0.994          \\
    \cellcolor[HTML]{FCE4D6}                                & 0.45 & 1.000                         & 1.000                         & 1.000                         & 1.000                         & \cellcolor[HTML]{2BBEF3}\textbf{0.917} & 1.000                         & \cellcolor[HTML]{11B6F1}0.967          & \cellcolor[HTML]{05B2F1}0.992          \\
    \multirow{-4}{*}{\cellcolor[HTML]{FCE4D6}Cascade R-CNN} & 0.55 & 1.000                         & 1.000                         & 1.000                         & 1.000                         & \cellcolor[HTML]{32C0F3}\textbf{0.903} & 1.000                         & \cellcolor[HTML]{16B7F2}0.958          & \cellcolor[HTML]{06B2F1}0.989       
             
  \\\hline       
  \end{tabular}
  \caption{Quantitative attack comparison of car detection in physically-based simulation in the metric of DR, where the best results are highlighted in bold. The \textbf{redder} the cell, the \textbf{higher} the \textbf{attack efficacy}. The \textbf{bluer} the cell, the \textbf{lower} the \textbf{attack efficacy}. TH and RN mean the threshold of confidence score and random noise, respectively. ``fg", ``bg", and ``bf" represent the perturbation on foreground, background, and both, respectively.}
  \label{table_attack_comparison_car}
\end{table}

\begin{table}[!h]
  \setlength{\tabcolsep}{0.8mm}
  \tiny
  \centering
  \begin{tabular}{c
    >{\columncolor[HTML]{EDEDED}}c 
    >{\columncolor[HTML]{00B0F0}}c 
    >{\columncolor[HTML]{00B0F0}}c 
    >{\columncolor[HTML]{00B0F0}}c 
    >{\columncolor[HTML]{00B0F0}}c 
    >{\columncolor[HTML]{00B0F0}}c 
    >{\columncolor[HTML]{00B0F0}}c 
    >{\columncolor[HTML]{00B0F0}}c 
    >{\columncolor[HTML]{00B0F0}}c }
    \hline
                                                            & TH   & \cellcolor[HTML]{FFF2CC}Clean & \cellcolor[HTML]{FFF2CC}RN    & \cellcolor[HTML]{FFF2CC}DTA   & \cellcolor[HTML]{FFF2CC}FCA            & \cellcolor[HTML]{FFF2CC}ACTIVE & \cellcolor[HTML]{FFF2CC}AA-fg & \cellcolor[HTML]{FFF2CC}AA-bg          & \cellcolor[HTML]{FFF2CC}AA-bf          \\\hline
    \cellcolor[HTML]{FCE4D6}                                & 0.25 & 1.000                         & 1.000                         & 1.000                         & 1.000                                  & 1.000                          & 1.000                         & 1.000                                  & 1.000                                  \\
    \cellcolor[HTML]{FCE4D6}                                & 0.35 & 1.000                         & 1.000                         & 1.000                         & 1.000                                  & 1.000                          & 1.000                         & 1.000                                  & 1.000                                  \\
    \cellcolor[HTML]{FCE4D6}                                & 0.45 & 1.000                         & 1.000                         & 1.000                         & 1.000                                  & 1.000                          & 1.000                         & 1.000                                  & 1.000                                  \\
    \multirow{-4}{*}{\cellcolor[HTML]{FCE4D6}SSD}           & 0.55 & 1.000                         & 1.000                         & 1.000                         & 1.000                                  & 1.000                          & 1.000                         & 1.000                                  & 1.000                                  \\\hline
    \cellcolor[HTML]{FCE4D6}                                & 0.25 & 1.000                         & 1.000                         & 1.000                         & 1.000                                  & 1.000                          & 1.000                         & 1.000                                  & 1.000                                  \\
    \cellcolor[HTML]{FCE4D6}                                & 0.35 & 1.000                         & 1.000                         & 1.000                         & 1.000                                  & 1.000                          & 1.000                         & 1.000                                  & 1.000                                  \\
    \cellcolor[HTML]{FCE4D6}                                & 0.45 & 1.000                         & 1.000                         & 1.000                         & 1.000                                  & 1.000                          & 1.000                         & 1.000                                  & 1.000                                  \\
    \multirow{-4}{*}{\cellcolor[HTML]{FCE4D6}Faster R-CNN}  & 0.55 & 1.000                         & 1.000                         & 1.000                         & 1.000                                  & 1.000                          & 1.000                         & 1.000                                  & 1.000                                  \\\hline
    \cellcolor[HTML]{FCE4D6}                                & 0.25 & 1.000                         & 1.000                         & 1.000                         & 1.000                                  & 1.000                          & 1.000                         & 1.000                                  & 1.000                                  \\
    \cellcolor[HTML]{FCE4D6}                                & 0.35 & 1.000                         & 1.000                         & 1.000                         & 1.000                                  & 1.000                          & 1.000                         & 1.000                                  & 1.000                                  \\
    \cellcolor[HTML]{FCE4D6}                                & 0.45 & 1.000                         & 1.000                         & 1.000                         & 1.000                                  & 1.000                          & 1.000                         & 1.000                                  & 1.000                                  \\
    \multirow{-4}{*}{\cellcolor[HTML]{FCE4D6}Swin}          & 0.55 & 1.000                         & 1.000                         & 1.000                         & 1.000                                  & 1.000                          & 1.000                         & 1.000                                  & 1.000                                  \\\hline
    \cellcolor[HTML]{FCE4D6}                                & 0.25 & 1.000                         & 1.000                         & 1.000                         & 1.000                                  & 1.000                          & 1.000                         & \cellcolor[HTML]{F87879}\textbf{0.050} & \cellcolor[HTML]{F8797B}0.056          \\
    \cellcolor[HTML]{FCE4D6}                                & 0.35 & 1.000                         & 1.000                         & 1.000                         & 1.000                                  & 1.000                          & 1.000                         & \cellcolor[HTML]{F86B6D}0.008          & \cellcolor[HTML]{F8696B}\textbf{0.003} \\
    \cellcolor[HTML]{FCE4D6}                                & 0.45 & 1.000                         & 1.000                         & 1.000                         & 1.000                                  & 1.000                          & 1.000                         & \cellcolor[HTML]{F8696B}0.003          & \cellcolor[HTML]{F8696B}\textbf{0.000} \\
    \multirow{-4}{*}{\cellcolor[HTML]{FCE4D6}YOLOv3}        & 0.55 & 1.000                         & 1.000                         & 1.000                         & 1.000                                  & 1.000                          & \cellcolor[HTML]{05B2F1}0.992 & \cellcolor[HTML]{F8696B}\textbf{0.000} & \cellcolor[HTML]{F8696B}\textbf{0.000} \\\hline
    \cellcolor[HTML]{FCE4D6}                                & 0.25 & \cellcolor[HTML]{02B1F1}0.997 & \cellcolor[HTML]{31BFF3}0.906 & \cellcolor[HTML]{2ABDF3}0.919 & \cellcolor[HTML]{36C1F4}0.894          & \cellcolor[HTML]{1DB9F2}0.944  & \cellcolor[HTML]{03B1F1}0.994 & \cellcolor[HTML]{C4EDFC}0.617          & \cellcolor[HTML]{FEF0F1}\textbf{0.453} \\
    \cellcolor[HTML]{FCE4D6}                                & 0.35 & \cellcolor[HTML]{2EBFF3}0.911 & \cellcolor[HTML]{5ECDF6}0.817 & \cellcolor[HTML]{79D6F8}0.764 & \cellcolor[HTML]{B4E8FB}0.647          & \cellcolor[HTML]{86DAF8}0.739  & \cellcolor[HTML]{56CBF6}0.833 & \cellcolor[HTML]{FA9FA0}0.181          & \cellcolor[HTML]{FA9D9E}\textbf{0.175} \\
    \cellcolor[HTML]{FCE4D6}                                & 0.45 & \cellcolor[HTML]{9FE2FA}0.689 & \cellcolor[HTML]{B7E9FB}0.642 & \cellcolor[HTML]{E0F6FE}0.561 & \cellcolor[HTML]{FBFEFF}0.508          & \cellcolor[HTML]{D4F2FD}0.586  & \cellcolor[HTML]{FEFFFF}0.503 & \cellcolor[HTML]{F86E6F}\textbf{0.017} & \cellcolor[HTML]{F86E6F}\textbf{0.017} \\
    \multirow{-4}{*}{\cellcolor[HTML]{FCE4D6}YOLOv5n}       & 0.55 & \cellcolor[HTML]{EFFAFF}0.533 & \cellcolor[HTML]{FFFFFF}0.500 & \cellcolor[HTML]{FDDCDD}0.386 & \cellcolor[HTML]{FDDCDD}0.386          & \cellcolor[HTML]{FEF5F5}0.467  & \cellcolor[HTML]{FAA7A8}0.208 & \cellcolor[HTML]{F8696B}\textbf{0.000} & \cellcolor[HTML]{F8696B}\textbf{0.000} \\\hline
    \cellcolor[HTML]{FCE4D6}                                & 0.25 & 1.000                         & \cellcolor[HTML]{02B1F1}0.997 & \cellcolor[HTML]{02B1F1}0.997 & \cellcolor[HTML]{17B8F2}0.956          & \cellcolor[HTML]{25BCF3}0.928  & \cellcolor[HTML]{33C0F3}0.900 & \cellcolor[HTML]{2BBEF3}0.917          & \cellcolor[HTML]{5ACCF6}\textbf{0.825} \\
    \cellcolor[HTML]{FCE4D6}                                & 0.35 & 1.000                         & \cellcolor[HTML]{03B1F1}0.994 & \cellcolor[HTML]{2FBFF3}0.908 & \cellcolor[HTML]{6FD3F7}0.783          & \cellcolor[HTML]{46C6F5}0.864  & \cellcolor[HTML]{D4F2FD}0.586 & \cellcolor[HTML]{FEFEFE}0.497          & \cellcolor[HTML]{FDD7D7}\textbf{0.367} \\
    \cellcolor[HTML]{FCE4D6}                                & 0.45 & \cellcolor[HTML]{03B1F1}0.994 & \cellcolor[HTML]{31BFF3}0.906 & \cellcolor[HTML]{95DFF9}0.708 & \cellcolor[HTML]{BDEBFC}0.631          & \cellcolor[HTML]{7DD7F8}0.756  & \cellcolor[HTML]{FEF0F0}0.450 & \cellcolor[HTML]{F99091}0.131          & \cellcolor[HTML]{F98687}\textbf{0.097} \\
    \multirow{-4}{*}{\cellcolor[HTML]{FCE4D6}YOLOv5s}       & 0.55 & \cellcolor[HTML]{27BCF3}0.925 & \cellcolor[HTML]{B0E7FB}0.656 & \cellcolor[HTML]{FEFDFD}0.494 & \cellcolor[HTML]{FEF9FA}0.483          & \cellcolor[HTML]{ECF9FE}0.539  & \cellcolor[HTML]{FDDADA}0.378 & \cellcolor[HTML]{F8696B}\textbf{0.003} & \cellcolor[HTML]{F87072}0.025          \\\hline
    \cellcolor[HTML]{FCE4D6}                                & 0.25 & 1.000                         & 1.000                         & 1.000                         & 1.000                                  & 1.000                          & 1.000                         & 1.000                                  & 1.000                                  \\
    \cellcolor[HTML]{FCE4D6}                                & 0.35 & 1.000                         & \cellcolor[HTML]{0CB4F1}0.978 & 1.000                         & 1.000                                  & 1.000                          & \cellcolor[HTML]{05B2F1}0.992 & \cellcolor[HTML]{13B6F2}0.964          & \cellcolor[HTML]{4DC8F5}\textbf{0.850} \\
    \cellcolor[HTML]{FCE4D6}                                & 0.45 & 1.000                         & \cellcolor[HTML]{10B5F1}0.969 & 1.000                         & 1.000                                  & 1.000                          & \cellcolor[HTML]{2BBEF3}0.917 & \cellcolor[HTML]{C1ECFC}0.622          & \cellcolor[HTML]{FEF9F9}\textbf{0.481} \\
    \multirow{-4}{*}{\cellcolor[HTML]{FCE4D6}YOLOv5m}       & 0.55 & 1.000                         & \cellcolor[HTML]{13B6F2}0.964 & \cellcolor[HTML]{05B2F1}0.992 & 1.000                                  & \cellcolor[HTML]{06B2F1}0.989  & \cellcolor[HTML]{68D1F7}0.797 & \cellcolor[HTML]{FCD1D1}0.347          & \cellcolor[HTML]{FAA1A2}\textbf{0.189} \\\hline
    \cellcolor[HTML]{FCE4D6}                                & 0.25 & 1.000                         & 1.000                         & 1.000                         & 1.000                                  & 1.000                          & 1.000                         & 1.000                                  & \cellcolor[HTML]{03B1F1}\textbf{0.994} \\
    \cellcolor[HTML]{FCE4D6}                                & 0.35 & 1.000                         & 1.000                         & 1.000                         & 1.000                                  & 1.000                          & \cellcolor[HTML]{06B2F1}0.989 & 1.000                                  & \cellcolor[HTML]{14B7F2}\textbf{0.961} \\
    \cellcolor[HTML]{FCE4D6}                                & 0.45 & 1.000                         & 1.000                         & 1.000                         & 1.000                                  & 1.000                          & \cellcolor[HTML]{1EBAF2}0.942 & \cellcolor[HTML]{11B6F1}0.967          & \cellcolor[HTML]{33C0F3}\textbf{0.900} \\
    \multirow{-4}{*}{\cellcolor[HTML]{FCE4D6}YOLOv5l}       & 0.55 & 1.000                         & 1.000                         & 1.000                         & 1.000                                  & 1.000                          & \cellcolor[HTML]{92DEF9}0.714 & \cellcolor[HTML]{62CFF6}0.808          & \cellcolor[HTML]{9BE0FA}\textbf{0.697} \\\hline
    \cellcolor[HTML]{FCE4D6}                                & 0.25 & 1.000                         & 1.000                         & 1.000                         & 1.000                                  & 1.000                          & 1.000                         & \cellcolor[HTML]{06B2F1}0.989          & \cellcolor[HTML]{16B7F2}\textbf{0.958} \\
    \cellcolor[HTML]{FCE4D6}                                & 0.35 & 1.000                         & 1.000                         & 1.000                         & 1.000                                  & 1.000                          & \cellcolor[HTML]{03B1F1}0.994 & \cellcolor[HTML]{49C7F5}0.858          & \cellcolor[HTML]{6BD1F7}\textbf{0.792} \\
    \cellcolor[HTML]{FCE4D6}                                & 0.45 & 1.000                         & 1.000                         & 1.000                         & 1.000                                  & 1.000                          & \cellcolor[HTML]{2ABDF3}0.919 & \cellcolor[HTML]{B6E9FB}0.644          & \cellcolor[HTML]{C4EDFC}\textbf{0.617} \\
    \multirow{-4}{*}{\cellcolor[HTML]{FCE4D6}YOLOv5x}       & 0.55 & 1.000                         & 1.000                         & 1.000                         & 1.000                                  & 1.000                          & \cellcolor[HTML]{86DAF8}0.739 & \cellcolor[HTML]{F7FDFF}0.517          & \cellcolor[HTML]{FEFFFF}\textbf{0.503} \\\hline
    \cellcolor[HTML]{FCE4D6}                                & 0.25 & 1.000                         & 1.000                         & 1.000                         & 1.000                                  & 1.000                          & 1.000                         & 1.000                                  & 1.000                                  \\
    \cellcolor[HTML]{FCE4D6}                                & 0.35 & 1.000                         & 1.000                         & 1.000                         & 1.000                                  & 1.000                          & 1.000                         & 1.000                                  & 1.000                                  \\
    \cellcolor[HTML]{FCE4D6}                                & 0.45 & 1.000                         & 1.000                         & 1.000                         & 1.000                                  & 1.000                          & 1.000                         & 1.000                                  & 1.000                                  \\
    \multirow{-4}{*}{\cellcolor[HTML]{FCE4D6}Cascade R-CNN} & 0.55 & 1.000                         & 1.000                         & 1.000                         & 1.000                                  & 1.000                          & 1.000                         & 1.000                                  & 1.000                                  
                                  \\\hline
  \end{tabular}
  \caption{Quantitative attack comparison of person detection in physically-based simulation in the metric of DR, where the best results are highlighted in bold. The \textbf{redder} the cell, the \textbf{higher} the \textbf{attack efficacy}. The \textbf{bluer} the cell, the \textbf{lower} the \textbf{attack efficacy}. TH and RN mean the threshold of confidence score and random noise, respectively. ``fg", ``bg", and ``bf" represent the perturbation on foreground, background, and both, respectively.}
  \label{table_attack_comparison_person}
\end{table}

  \begin{figure}[!htbp]
    \centering
    \begin{subfigure}{0.49\linewidth}
      \includegraphics[width=1\linewidth]{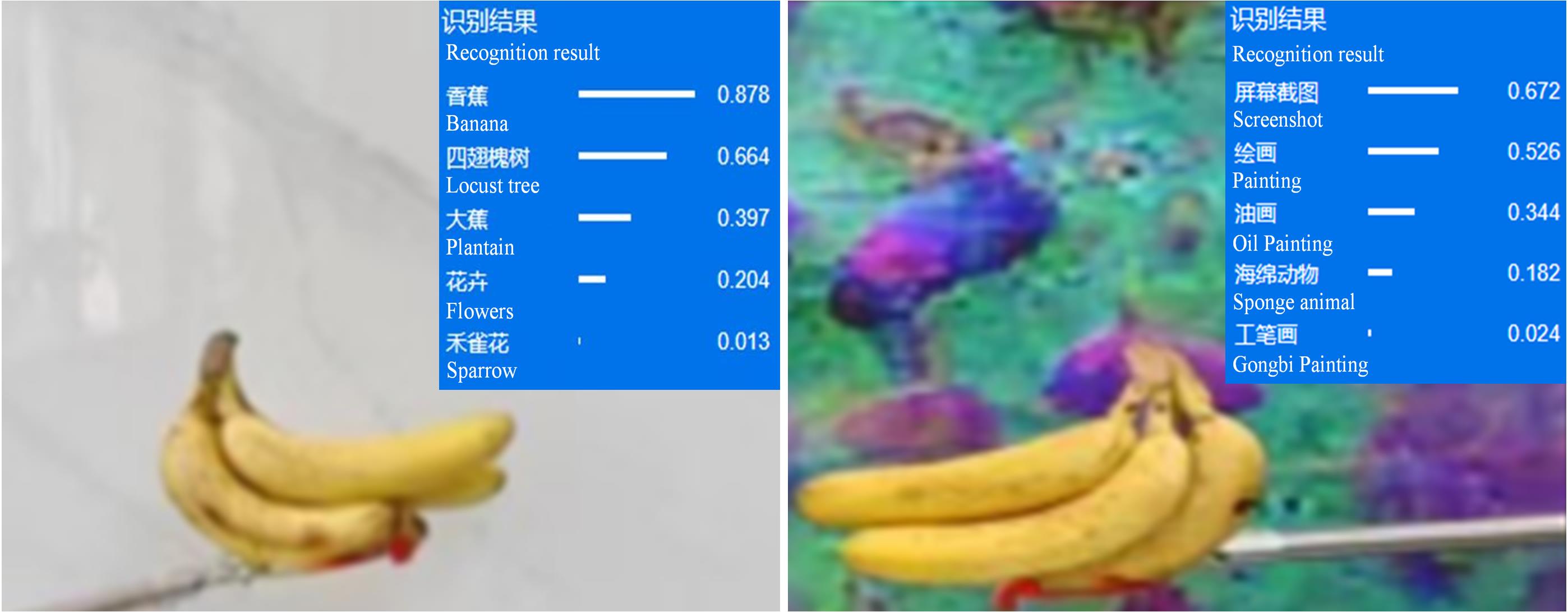}
      \caption{Image classification}
    \end{subfigure}
    \begin{subfigure}{0.49\linewidth}
      \includegraphics[width=1\linewidth]{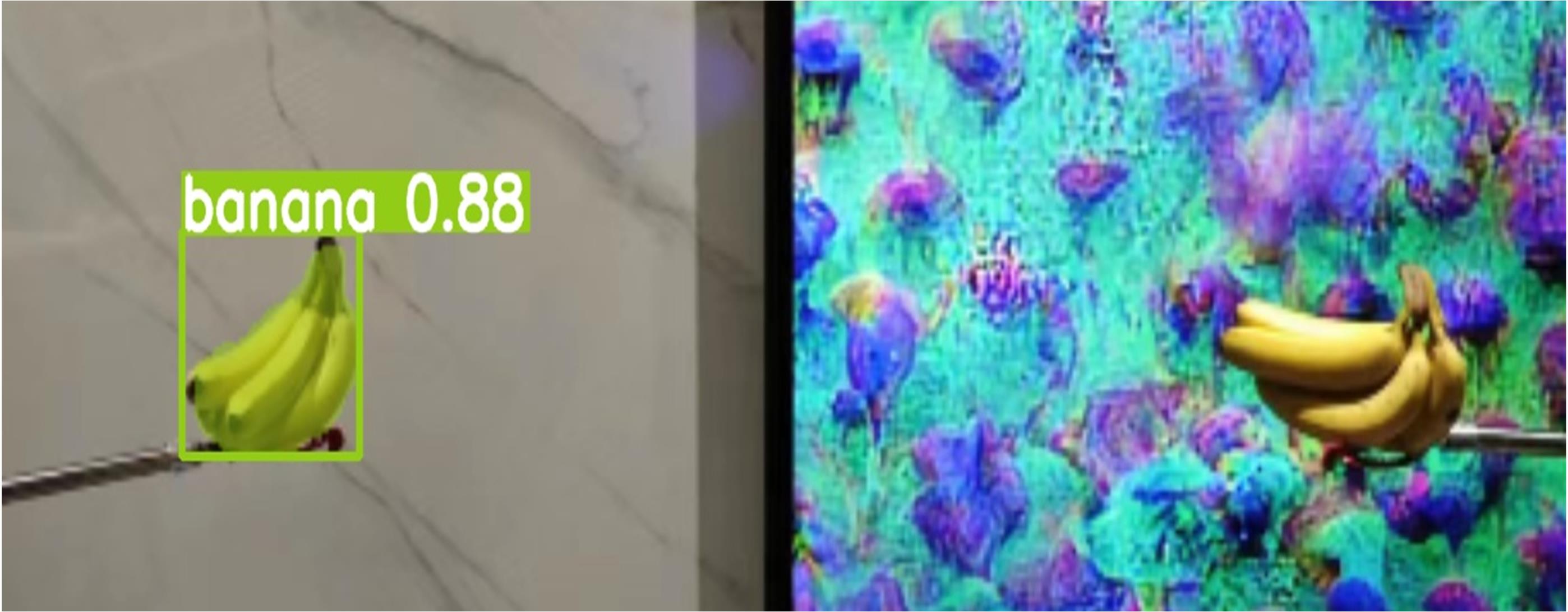}
      \caption{Detection and segmentation}
    \end{subfigure}
    \begin{subfigure}{0.49\linewidth}
      \includegraphics[width=1\linewidth]{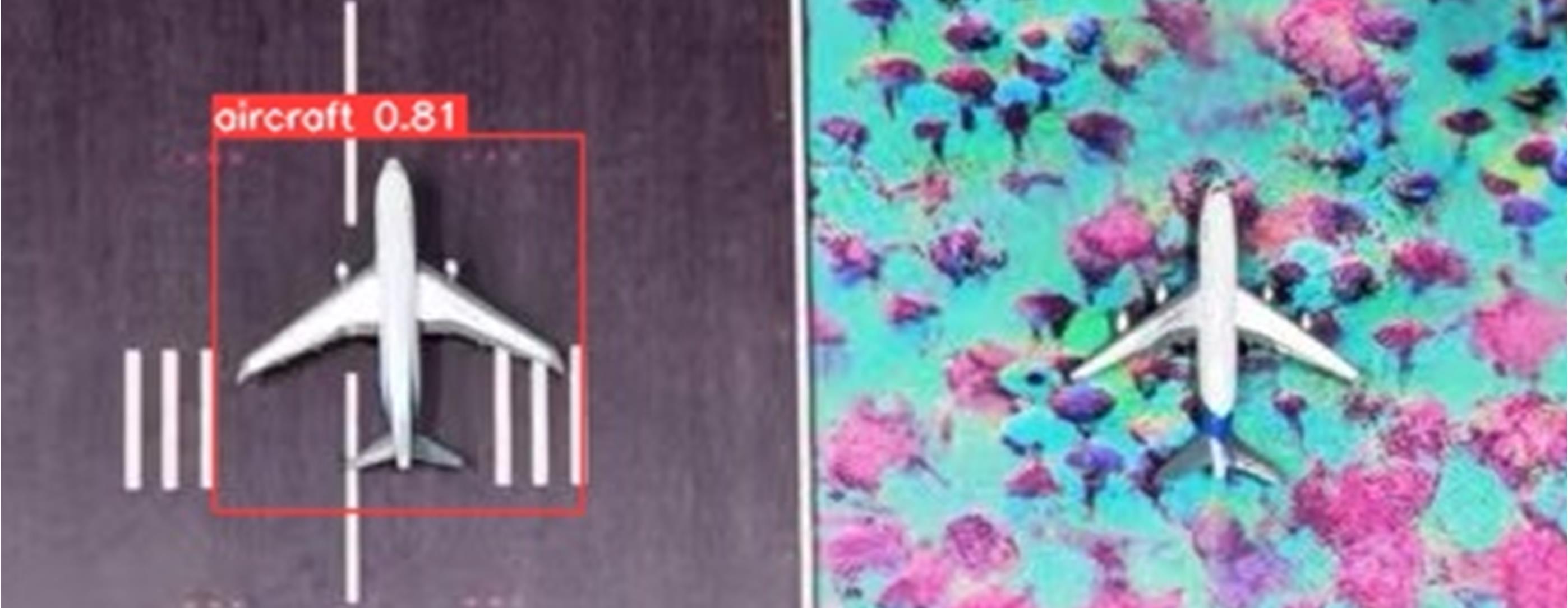}
      \caption{Aerial detection}
    \end{subfigure}
    \begin{subfigure}{0.49\linewidth}
      \includegraphics[width=1\linewidth]{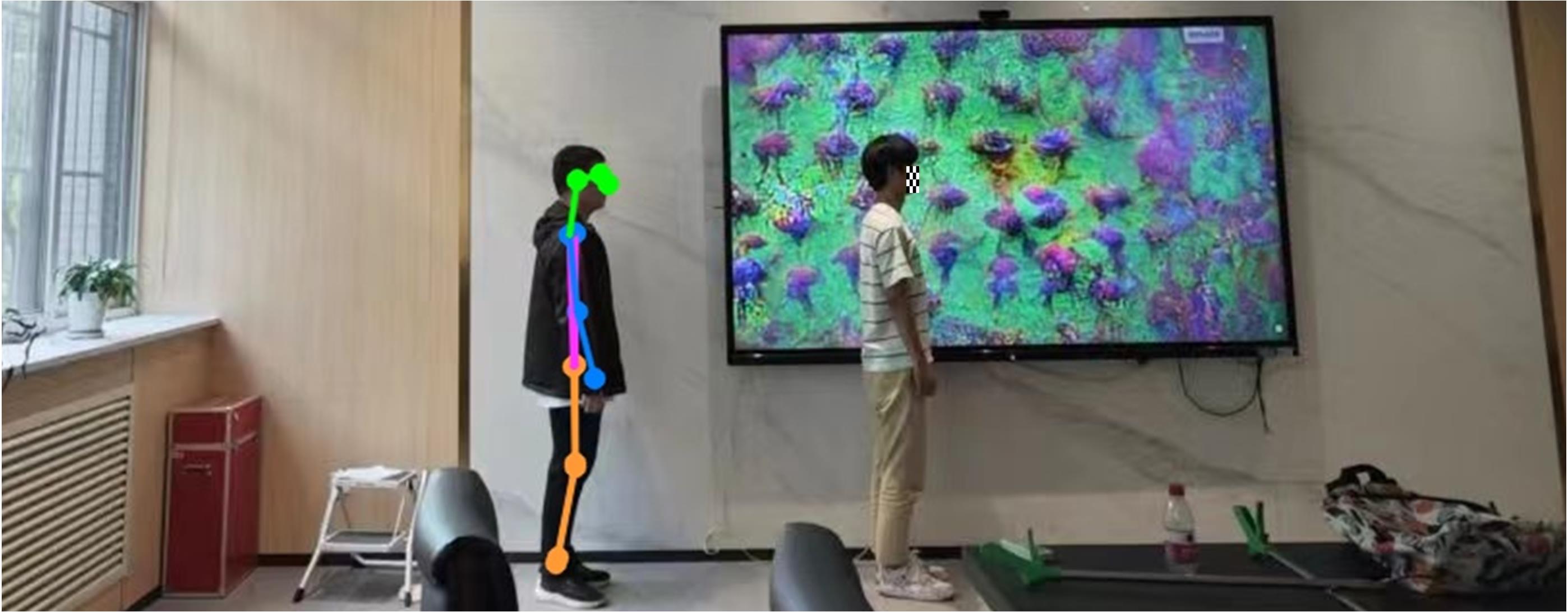}
      \caption{Pose estimation}
    \end{subfigure}
    \caption{Physical attack against different tasks under black box conditions. We demonstrate the effectiveness of background attacks by comparing the detection results of the same targets under clean and adversarial backgrounds. The confidence threshold is set to 0.25. Please note that the aerial detector (YOLOv5) is trained on the aerial detection dataset DOTA.}
    \label{fig_different_tasks}
  \end{figure}

  \begin{figure*}[!h]
  \centering
  \includegraphics[width=0.99\linewidth]{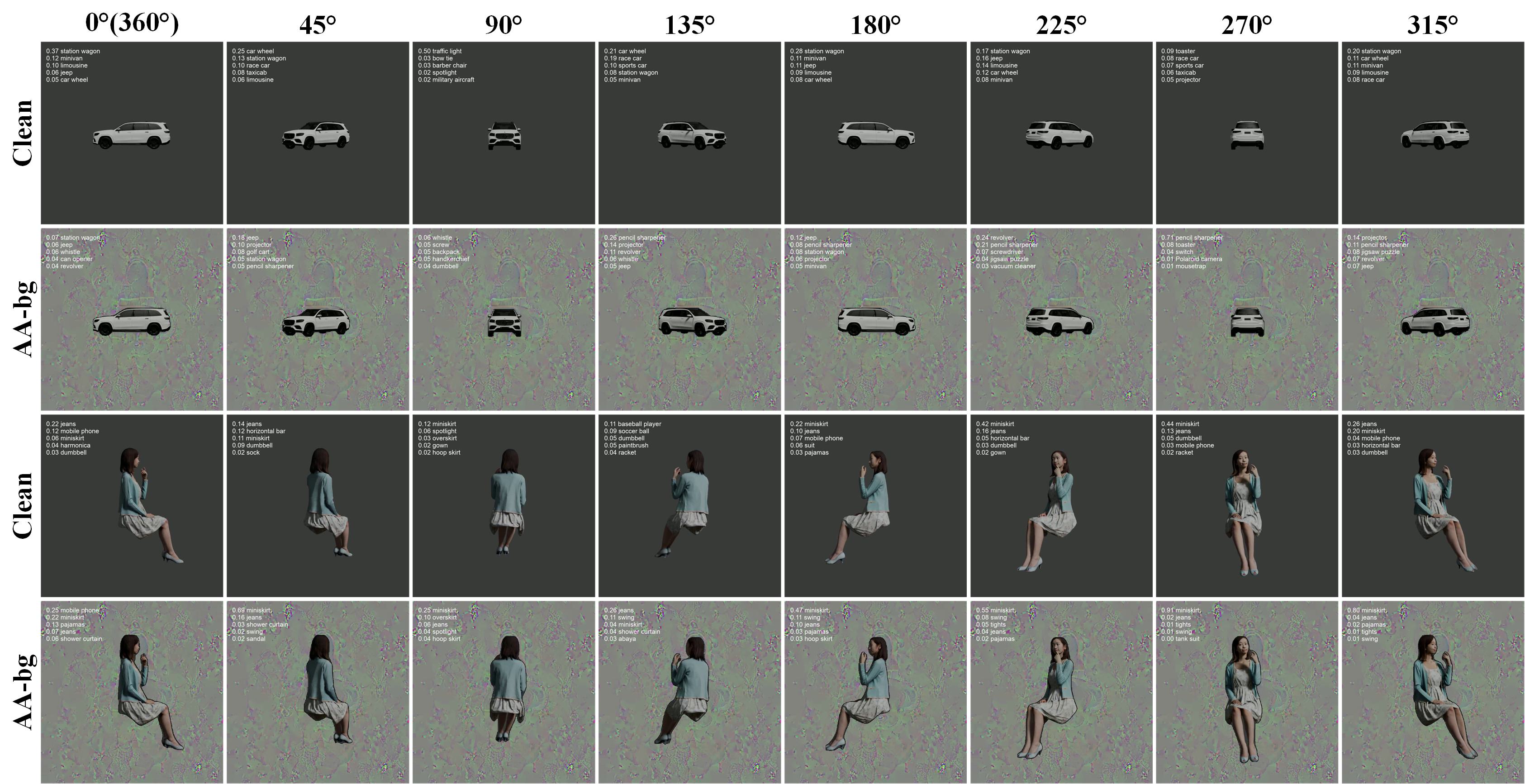} 
  \caption{Trasfer attack against image classification model in physically-based simulation and the victim model is YOLOv5x-cls. Please zoom in for better visualization.}
  \label{fig_attack_yolov5x-cls}
  \end{figure*}

  \begin{figure*}[!h]
    \centering
    \includegraphics[width=0.99\linewidth]{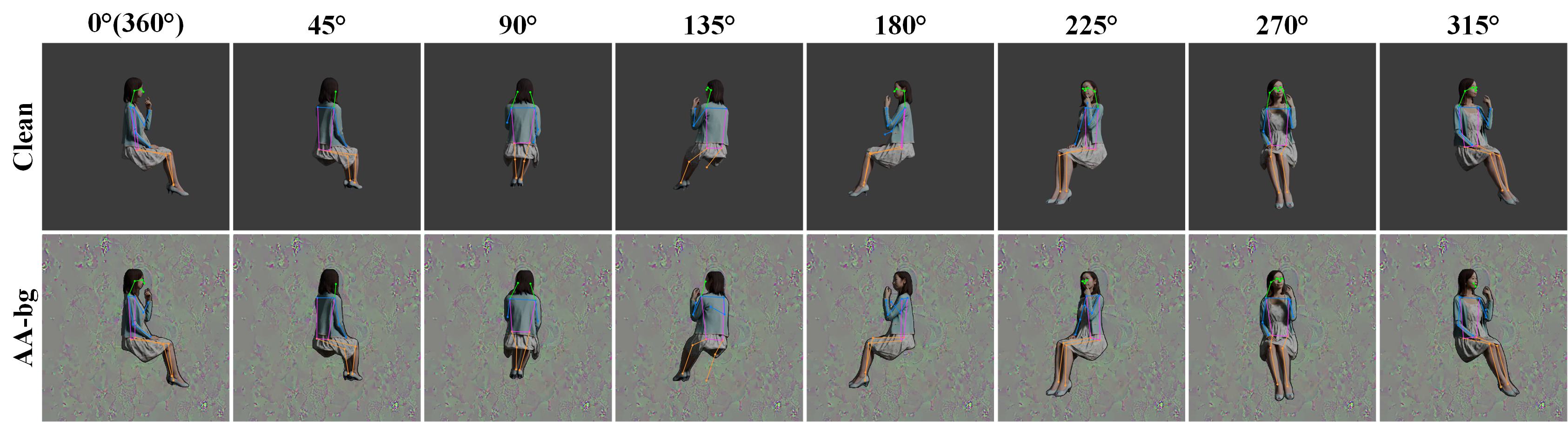} 
    \caption{Trasfer attack against image segmentation model in physically-based simulation and the victim model is YOLOv8s-pose. Please zoom in for better visualization.}
    \label{fig_attack_yolov8s-pose}
    \end{figure*}

\begin{table*}[!h]
  \centering
  \tiny
  \setlength{\tabcolsep}{1.141mm}
  \begin{tabular*}{\hsize}{cccccccccccccccccccc}
  \hline
  & \cellcolor[HTML]{EBF1DE}\rotatebox{75}{SSD}   & \cellcolor[HTML]{EBF1DE}\rotatebox{75}{Faster R-CNN} & \cellcolor[HTML]{EBF1DE}\rotatebox{75}{Swin Transformer} & \cellcolor[HTML]{EBF1DE}\rotatebox{75}{YOLOv3} & \cellcolor[HTML]{EBF1DE}\rotatebox{75}{YOLOv5n} & \cellcolor[HTML]{EBF1DE}\rotatebox{75}{YOLOv5s} & \cellcolor[HTML]{EBF1DE}\rotatebox{75}{YOLOv5m} & \cellcolor[HTML]{EBF1DE}\rotatebox{75}{YOLOv5l} & \cellcolor[HTML]{EBF1DE}\rotatebox{75}{YOLOv5x} & \cellcolor[HTML]{EBF1DE}\rotatebox{75}{Cascade R-CNN} & \cellcolor[HTML]{EBF1DE}\rotatebox{75}{RetinaNet} & \cellcolor[HTML]{EBF1DE}\rotatebox{75}{Mask R-CNN}& \cellcolor[HTML]{EBF1DE}\rotatebox{75}{FreeAnchor} & \cellcolor[HTML]{EBF1DE}\rotatebox{75}{FSAF}  & \cellcolor[HTML]{EBF1DE}\rotatebox{75}{RepPoints} & \cellcolor[HTML]{EBF1DE}\rotatebox{75}{TOOD}  & \cellcolor[HTML]{EBF1DE}\rotatebox{75}{ATSS}  & \cellcolor[HTML]{EBF1DE}\rotatebox{75}{FoveaBox} & \cellcolor[HTML]{EBF1DE}\rotatebox{75}{VarifocalNet} \\ \hline
  \cellcolor[HTML]{E2EFDA}Swin Transformer         & \cellcolor[HTML]{FE4365}0.000 & \cellcolor[HTML]{FE4768}0.011 & \cellcolor[HTML]{FE5372}\textbf{0.043} & \cellcolor[HTML]{FE4365}0.000 & \cellcolor[HTML]{FE4365}0.000 & \cellcolor[HTML]{FE95A8}0.219 & \cellcolor[HTML]{FEA1B2}0.251          & \cellcolor[HTML]{FECFD8}0.374 & \cellcolor[HTML]{FEF1F4}0.465 & \cellcolor[HTML]{FE4365}0.000 & \cellcolor[HTML]{FE4768}0.011 & \cellcolor[HTML]{FE4365}0.000          & \cellcolor[HTML]{FE5372}0.043 & \cellcolor[HTML]{FE4365}0.000 & \cellcolor[HTML]{FE4466}0.005 & \cellcolor[HTML]{FEB3C1}0.299 & \cellcolor[HTML]{FE4365}0.000 & \cellcolor[HTML]{FE4365}0.000 & \cellcolor[HTML]{FE637F}0.086 \\
  \cellcolor[HTML]{E2EFDA}YOLOv5m                  & \cellcolor[HTML]{FE4465}0.003 & \cellcolor[HTML]{CBEFFC}0.603 & \cellcolor[HTML]{FFFFFF}0.500          & \cellcolor[HTML]{FE6F89}0.118 & \cellcolor[HTML]{FE5372}0.045 & \cellcolor[HTML]{FE6681}0.094 & \cellcolor[HTML]{FE4365}\textbf{0.000} & \cellcolor[HTML]{FE4566}0.006 & \cellcolor[HTML]{FE8A9F}0.191 & \cellcolor[HTML]{E5F7FE}0.551 & \cellcolor[HTML]{D4F2FD}0.585 & \cellcolor[HTML]{56CBF6}0.833          & \cellcolor[HTML]{30BFF3}0.906 & \cellcolor[HTML]{C9EFFC}0.606 & \cellcolor[HTML]{A0E2FA}0.688 & \cellcolor[HTML]{49C7F5}0.858 & \cellcolor[HTML]{CEF0FD}0.597 & \cellcolor[HTML]{E0F6FE}0.561 & \cellcolor[HTML]{77D5F7}0.767 \\
  \cellcolor[HTML]{E2EFDA}Mask R-CNN               & \cellcolor[HTML]{FE4365}0.000 & \cellcolor[HTML]{FE9DAE}0.240 & \cellcolor[HTML]{FE5171}0.039          & \cellcolor[HTML]{FE5F7C}0.075 & \cellcolor[HTML]{FE4567}0.007 & \cellcolor[HTML]{FE4C6C}0.025 & \cellcolor[HTML]{FE4768}0.011          & \cellcolor[HTML]{FE5876}0.057 & \cellcolor[HTML]{FE728B}0.125 & \cellcolor[HTML]{FE8EA2}0.201 & \cellcolor[HTML]{FE738C}0.129 & \cellcolor[HTML]{FE9DAE}\textbf{0.240} & \cellcolor[HTML]{FBFEFF}0.509 & \cellcolor[HTML]{FEA1B2}0.251 & \cellcolor[HTML]{FEC8D2}0.355 & \cellcolor[HTML]{FEBCC8}0.323 & \cellcolor[HTML]{FE708A}0.122 & \cellcolor[HTML]{FE7790}0.140 & \cellcolor[HTML]{FEACBB}0.280 \\
  \cellcolor[HTML]{E2EFDA}Swin Transformer+YOLOv5m & \cellcolor[HTML]{FE4365}0.000 & \cellcolor[HTML]{FE4969}0.016 & \cellcolor[HTML]{FE4668}\textbf{0.010} & \cellcolor[HTML]{FE4466}0.005 & \cellcolor[HTML]{FE4365}0.000 & \cellcolor[HTML]{FE4365}0.000 & \cellcolor[HTML]{FE4365}\textbf{0.000} & \cellcolor[HTML]{FE4668}0.010 & \cellcolor[HTML]{FE4969}0.016 & \cellcolor[HTML]{FE4365}0.000 & \cellcolor[HTML]{FE4969}0.016 & \cellcolor[HTML]{FE5271}0.042          & \cellcolor[HTML]{FE6E88}0.116 & \cellcolor[HTML]{FE4365}0.000 & \cellcolor[HTML]{FE4365}0.000 & \cellcolor[HTML]{FE5E7B}0.074 & \cellcolor[HTML]{FE4365}0.000 & \cellcolor[HTML]{FE5271}0.042 & \cellcolor[HTML]{FE5A78}0.063 \\
  \rowcolor[HTML]{FE4365} 
  \cellcolor[HTML]{E2EFDA}YOLOv5m+Mask R-CNN       & 0.000                         & \cellcolor[HTML]{FE5473}0.047 & \cellcolor[HTML]{FE4A6B}0.021          & 0.000                         & 0.000                         & 0.000                         & \textbf{0.000}                         & 0.000                         & 0.000                         & \cellcolor[HTML]{FE4869}0.015 & \cellcolor[HTML]{FE4C6D}0.026 & \textbf{0.000}                         & \cellcolor[HTML]{FEBDC9}0.326 & 0.000                         & \cellcolor[HTML]{FE5876}0.057 & \cellcolor[HTML]{FE4869}0.015 & 0.000                         & \cellcolor[HTML]{FE758E}0.135 & \cellcolor[HTML]{FE5C79}0.067   \\
  \hline
  \end{tabular*}
  \caption{
  Ablation study on ensemble strategy (``Swin Transformer+YOLOv5m" and ``YOLOv5m+Mask R-CNN") in physical attack settings in the metric of DR, where white-box attacks are highlighted in bold and the rest are black-box attacks.
  The \textbf{redder} the cell, the \textbf{higher} the \textbf{attack efficacy}.
  The \textbf{bluer} the cell, the \textbf{lower} the \textbf{attack efficacy}.
The 19 detectors of the first row and the first column are for detection and perturbation optimization, respectively.  }
  \label{table_ensemble_attack}
  \end{table*}

\begin{table}[!h]
  \footnotesize
  \setlength{\tabcolsep}{3mm}
  \centering
  \begin{tabular}{cccccc}
  \hline
  Threshold & 0.15 & 0.25 & 0.35 & 0.45 & 0.55 \\ \hline
  Bottle & 0.000         & 0.000         & 0.000         & 0.000         & 0.000         \\
  Person & 0.132          & 0.016          & 0.000         & 0.000         & 0.000         \\
  Apple  & 0.070          & 0.020          & 0.020          & 0.000         & 0.000         \\
  Banana & 0.000         & 0.000         & 0.000         & 0.000         & 0.000         \\
  Orange & 0.000         & 0.000         & 0.000         & 0.000         & 0.000         \\
  Cup    & 0.000         & 0.000         & 0.000         & 0.000         & 0.000         \\
  Car    & 0.168          & 0.045          & 0.018          & 0.000         & 0.000       \\ \hline 
  \end{tabular}
  \caption{
  The physical attack performance across different objects in the metric of DR with different thresholds of confidence score.
  }
  \label{table_physical_attack_different_objects}
  \end{table}

\begin{table}[!h]
  \footnotesize
  \setlength{\tabcolsep}{3mm}
  \centering
  \begin{tabular}{cccccc}
  \hline
  Threshold & 0.15 & 0.25 & 0.35 & 0.45 & 0.55 \\ \hline
  Unsmooth              & 0.762          & 0.573          & 0.420          & 0.322          & 0.185          \\
  Smooth             & 0.000         & 0.000         & 0.000         & 0.000         & 0.000         \\ \hline 
  \end{tabular}
  \caption{
  Ablation study on smoothness setting with various thresholds of confidence score in the metric of DR.
  }
  \label{table_ablation_study_tv}
\end{table}
  
\begin{figure*}[!htbp]
\centering
\includegraphics[width=0.99\linewidth]{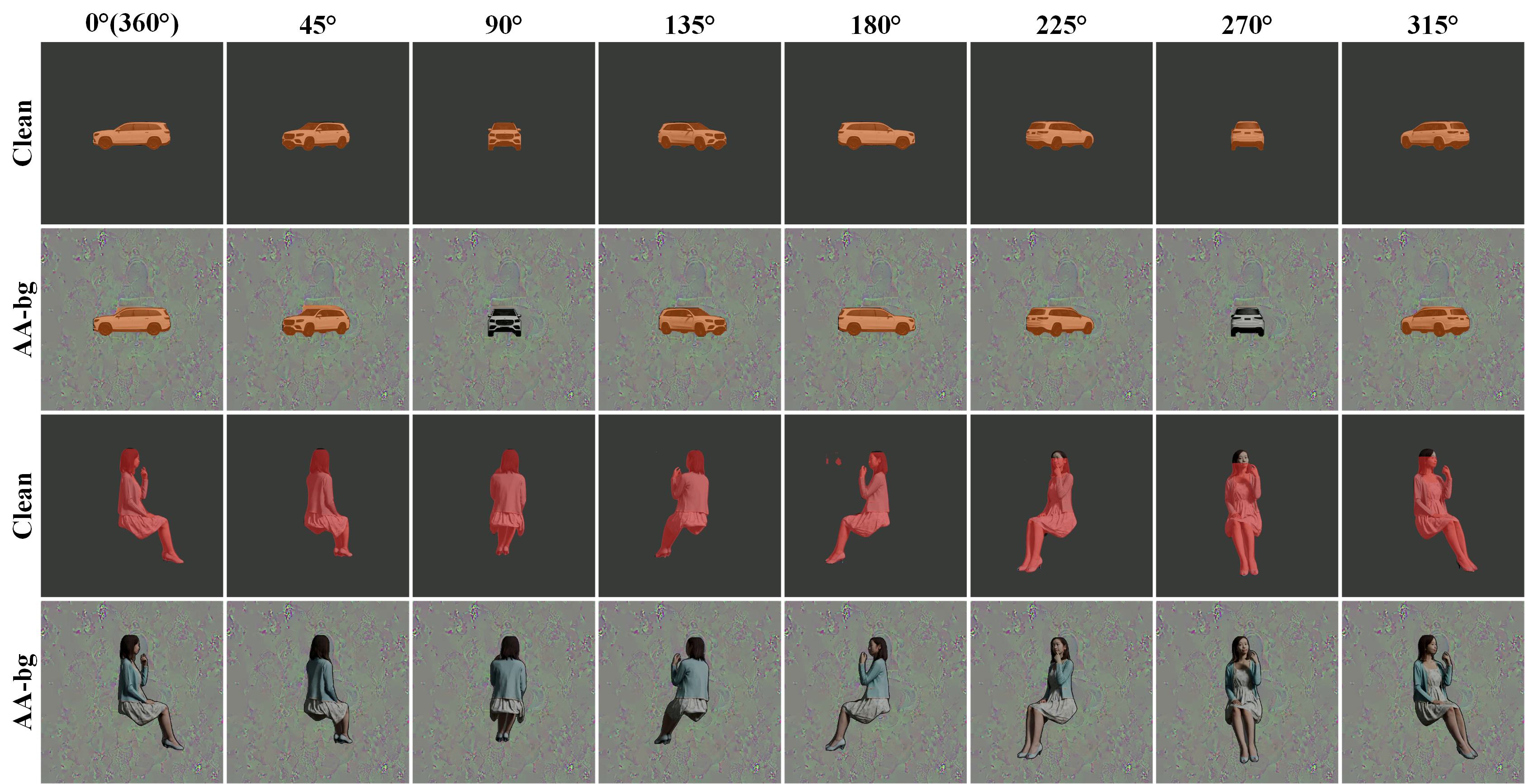} 
\caption{Trasfer attack against image segmentation model in physically-based simulation and the victim model is YOLOv5x-seg. Please zoom in for better visualization.}
\label{fig_attack_yolov5x-seg}
\end{figure*}

\subsection{Physical Attack Comparison}
\label{subsec_physical_attack_comparison}

We conduct comparison experiments with several SOTA physical attack methods on the object detection task, such as ACTIVE \cite{suryanto2023active}, FCA \cite{wang2022fca} and DTA \cite{suryanto2022dta}. 
We compare the attack efficacy and transferability by adopting objects on a clean background (pure gray) to suppress background discrepancy.
To control physical dynamics, we use 3D simulation to parameterize these factors, such as the rotation angle, the distance between the camera and the object, and the light intensity, which can not be fairly guaranteed in real-world scenarios.
Technically, we use Blender 4.0, a 3D modeling software, to generate 3D adversarial objects by directly rendering the physical perturbation on the targeted objects.
To emphasize the background attack effectiveness, we attach our elaborated background perturbations to the targeted objects (AA-fg), background (AA-bg), and both (AA-bf), respectively.
Then, we export the rotation of the 3D object to a video clip in mp4 format, which consists of 360 frames corresponding to 360 degrees with a resolution of 1024*1024 space.
These video clips are fed into various mainstream object detectors to compare the performance of different attack methods.
Detection rate (DR), i.e. the percentage of frames where the object is successfully detected, is adopted as the metric.

The quantitative experimental results of car and person detection are shown in Table \ref{table_attack_comparison_car} and Table \ref{table_attack_comparison_person}, respectively.
In addition, we also display the qualitative experimental results of car and person detection in the Appendix.
The confidence score lines of the correct detection are shown in Fig. \ref{fig_scores_line_yolov3_car} and Fig. \ref{fig_scores_line_yolov3_person} to further illustrate the attack performance.
It is observed that:
\begin{itemize}
  \item Our elaborated background perturbations can effectively sway the detection performance of SOTA object detection methods even under black-box conditions, which demonstrates the significance of background features beyond our original expectations.
  \item In comparison with other physical attack methods, our elaborated background perturbations achieve comparable performance, and even better attack efficacy and transferability without ensemble strategy.
\end{itemize}
Please refer to the Appendix for more experimental details for other object detection methods.

\subsection{Attack Anything}
\label{subsec_attack_anything}

\subsubsection{Across Different Models}

As shown in Table \ref{table_digital_attack} and \ref{table_physical_attack}, the method generalizes well across various models in the white box and black box conditions for most cases.
However, some perturbations generated by detectors with similar structures may transfer well between each other, while it is hard to generalize to other models as shown in Table \ref{table_physical_attack}.
The devised ensemble attack may properly resolve the above issues.
The experimental results are shown in Table. \ref{table_ensemble_attack}.
It is observed that the attack transferability is significantly improved by our designed ensemble strategy.

\subsubsection{Across Different Objects}

We conduct physical attacks on YOLOv5 with different objects, such as a bottle, person, cup, car, and several kinds of fruits.
The quantitative experimental results are exhibited in Table \ref{table_physical_attack_different_objects}.
We can observe that the proposed attack anything framework generalizes well between various objects with DR decreasing to 0 for most cases.
The qualitative experimental results as shown in Fig. \ref{fig_background_attack_exhibition} (d).

\subsubsection{Across Different Tasks}

To verify the attack effectiveness across different tasks, we perform physical attacks on image classification and image segmentation in addition to object detection.
We exhibit the attack results in real-world scenarios as shown in Fig. \ref{fig_different_tasks}.
Furthermore, we also conduct experiments with data generated by 3D modeling simulation to control physical dynamic factors.
The experimental results of attacking image classification, segmentation, and pose estimation are shown in Fig. \ref{fig_attack_yolov5x-cls}, \ref{fig_attack_yolov8s-pose}, and \ref{fig_attack_yolov5x-seg}, respectively, which demonstrate that our elaborated background perturbations with significant generalizability between various vision tasks.
Please refer to the Appendix for more experimental results on other image classification, segmentation, and pose estimation methods.

\subsection{Ablation Study}
\label{subsec_ablation_study}

To verify the effectiveness of the proposed ensemble attack strategy, we compare the attack performance of the ensemble attack with the single attack.
The experimental results are shown in Table. \ref{table_ensemble_attack}.
It is observed that the attack transferability is significantly improved by the proposed ensemble strategy.

To verify the key value of smoothness loss for physical attacks, we compare the attack efficacy of smooth and smooth-less perturbations under various thresholds of confidence score as shown in Table \ref{table_ablation_study_tv}.
The experimental results demonstrate that smoothness loss plays an indispensable role in conducting physical attacks.

\section{Discussion}
\label{sec_discussion}

The proposed background adversarial attack framework represents a paradigm shift in adversarial attacks by targeting the background rather than the primary object of interest. 
This method achieves remarkable generalization and robustness across different objects, models, and tasks, indicating that background features play a critical role in DNNs' decision-making processes. 
The theoretical analysis demonstrates the convergence of the background attack under certain conditions, which is a significant step towards understanding the underlying dynamics of DNNs and adversarial phenomena.
The experimental results validate the effectiveness of the attack in both digital and physical domains, showcasing its potential to disrupt AI applications in real-world scenarios. 

\section{Conclusion}
\label{sec_conclusion}

In this paper, we have innovated a comprehensive framework for mounting background adversarial attacks, displaying exceptional versatility and potency across a broad spectrum of objects, models, and tasks. 
From a mathematical standpoint, our approach formulates background adversarial attacks as an iterative optimization problem, akin to the training process of DNNs. 
We substantiate the theoretical convergence of our method under a set of mild yet sufficient conditions, ensuring its mathematical and practical applicability.
Moreover, we introduce an ensemble strategy specifically tailored to adversarial perturbations, enhancing both the effectiveness and transferability of attacks. 
Accompanying this, we have devised a novel smoothness constraint mechanism, which ensures the perturbations are seamlessly incorporated into the background.
Through an extensive series of experiments conducted under varied conditions, including digital and physical domains, as well as white-box and black-box scenarios, we have empirically validated the superior performance of our framework. 
The results demonstrate the efficacy of our ``attack anything" paradigm by only manipulating background.
Our work underscores the pivotal role of background features in adversarial attacks and DNNs-based visual perception, which calls for a comprehensive reevaluation and augmentation of DNNs' robustness.
This research stands as a critical revelation in the field of DNNs and adversarial threats, shedding light on new dimensions of alignment between human and machine vision in terms of background variations. 

\section*{CRediT authorship contribution statement}
\textbf{Jiawei Lian}: Conceptualization, Data curation, Formal analysis, Writing - original draft, Writing - review and editing.
\textbf{Shaohui Mei}: Supervision, Writing - review and editing.
\textbf{Xiaofei Wang}: Data curation, Formal analysis, Writing - original draft.
\textbf{Lefan Wang} and \textbf{Yingjie Lu}: Formal analysis, Writing - original draft.
\textbf{Yi Wang}, \textbf{Mingyang Ma} and \textbf{Lap-Pui Chau}: Writing - review and editing.

\section*{Declaration of competing interest}
The authors declare that they have no known competing financial interests or personal relationships that could have appeared to influence the work reported in this paper.

\section*{Acknowledgements}
This work was supported by the National Natural Science Foundation of China (62171381 and 62201445).

\section*{Data availability}
The datasets used in this study are publicly available and can be accessed through \url{https://github.com/JiaweiLian/Attack_Anything}.

\bibliographystyle{elsarticle-num} 
\bibliography{references}

\clearpage

\appendix

\section{Proof of the Convergence Analysis}
\label{appendix_convergence_analysis}

In this section, we provide detailed proof of the convergence analysis of the proposed background adversarial attack framework.
The convergence analysis is based on previous works \cite{yang2016unified,zhou2018convergence,reddi2018convergence,chen2019convergence}, which have been widely used in the convergence analysis of optimization algorithms.
Please refer to these works for more basic mathematical principles and prerequisites.

Firstly, $\boldsymbol{\xi}^{(t)}$ is defined as:
\begin{equation}
    \boldsymbol{\xi}^{(t)}=\begin{cases} \boldsymbol{P}^{(t)} & t=1\\ \boldsymbol{P}^{(t)}+\frac{\beta_{1}}{1-\beta_{1}}\left(\boldsymbol{P}^{(t)}-\boldsymbol{P}^{(t-1)}\right) & t\geq2 \end{cases}.
\end{equation}
Because $f$ is an L-smooth function, it satisfies Eq. \ref{lpc_ctn_1} and \ref{lpc_ctn_2}, i.e., 
\begin{equation}
    \begin{split}
        f\left(\boldsymbol{\xi}^{(t+1)}\right)\leq f\left(\boldsymbol{\xi}^{(t)}\right)+\left\langle \nabla f\left(\boldsymbol{\xi}^{(t)}\right),\boldsymbol{\xi}^{(t+1)}-\boldsymbol{\xi}^{(t)}\right\rangle \\ 
        +\frac{L}{2}\left\Vert \boldsymbol{\xi}^{(t+1)}-\boldsymbol{\xi}^{(t)}\right\Vert _{2}^{2}
    \end{split}
\end{equation}
and 
\begin{equation}
    \left\Vert \nabla f\left(\boldsymbol{\xi}^{(t)}\right)-\nabla f\left(\boldsymbol{P}^{(t)}\right)\right\Vert _{2}^{2}\leq L^{2}\left\Vert \boldsymbol{\xi}^{(t)}-\boldsymbol{P}^{(t)}\right\Vert _{2}^{2},
\end{equation}
which yields
\begin{equation}
    \begin{aligned} & f\left(\boldsymbol{\xi}^{(t+1)}\right)-f\left(\boldsymbol{\xi}^{(t)}\right)\\ \leq & \left\langle \nabla f\left(\boldsymbol{\xi}^{(t)}\right),\boldsymbol{\xi}^{(t+1)}-\boldsymbol{\xi}^{(t)}\right\rangle +\frac{L}{2}\left\Vert \boldsymbol{\xi}^{(t+1)}-\boldsymbol{\xi}^{(t)}\right\Vert _{2}^{2}\\ = & \left\langle \frac{1}{\sqrt{L}}\left(\nabla f\left(\boldsymbol{\xi}^{(t)}\right)-\nabla f\left(\boldsymbol{P}^{(t)}\right)\right), \sqrt{L}\left(\boldsymbol{\xi}^{(t+1)}-\boldsymbol{\xi}^{(t)}\right)\right\rangle \\ & +\left\langle \nabla f\left(\boldsymbol{P}^{(t)}\right),\boldsymbol{\xi}^{(t+1)}-\boldsymbol{\xi}^{(t)}\right\rangle +\frac{L}{2}\left\Vert \boldsymbol{\xi}^{(t+1)}-\boldsymbol{\xi}^{(t)}\right\Vert _{2}^{2}\\ \leq & \frac{1}{2}\left(\frac{1}{L}\left\Vert \nabla f\left(\boldsymbol{\xi}^{(t)}\right)-\nabla f\left(\boldsymbol{P}^{(t)}\right)\right\Vert _{ 2}^{2}+L\left\Vert \boldsymbol{\xi}^{(t+1)}-\boldsymbol{\xi}^{(t)}\right\Vert _{2}^{2}\right)\\ & +\left\langle \nabla f\left(\boldsymbol{P}^{(t)}\right),\boldsymbol{\xi}^{(t+1)}-\boldsymbol{\xi}^{(t)}\right\rangle +\frac{L}{2}\left\Vert \boldsymbol{\xi}^{(t+1)}-\boldsymbol{\xi}^{(t)}\right\Vert _{2}^{2}\\ \leq & \frac{1}{2L}\left\Vert \nabla f\left(\boldsymbol{\xi}^{(t)}\right)-\nabla f\left(\boldsymbol{P}^{(t)}\right)\right\Vert _{ 2}^{2}+L\left\Vert \boldsymbol{\xi}^{(t+1)}-\boldsymbol{\xi}^{(t)}\right\Vert _{2}^{2}\\ & +\left\langle \nabla f\left(\boldsymbol{P}^{(t)}\right),\boldsymbol{\xi}^{(t+1)}-\boldsymbol{\xi}^{(t)}\right\rangle \\ \leq & \frac{1}{2L}L^{2}\left\Vert \boldsymbol{\xi}^{(t)}-\boldsymbol{P}^{(t)}\right\Vert _{2}^{2}+L\left\Vert \boldsymbol{\xi}^{(t+1)}-\boldsymbol{\xi}^{(t)}\right\Vert _{2}^{2}\\ & +\left\langle \nabla f\left(\boldsymbol{P}^{(t)}\right),\boldsymbol{\xi}^{(t+1)}-\boldsymbol{\xi}^{(t)}\right\rangle \\ = & \frac{L}{2}\underset{\left(1\right)}{\underbrace{\left\Vert \boldsymbol{\xi}^{(t)}-\boldsymbol{P}^{(t)}\right\Vert _{2}^{2}}}+L\underset{\left(2\right)}{\underbrace{\left\Vert \boldsymbol{\xi}^{(t+1)}-\boldsymbol{\xi}^{(t)}\right\Vert _{2}^{2}}} \\ & +\underset{\left(3\right)}{\underbrace{\left\langle \nabla f\left(\boldsymbol{P}^{(t)}\right), \boldsymbol{\xi}^{(t+1)}-\boldsymbol{\xi}^{(t)}\right\rangle }} \end{aligned}.
\end{equation}
Then we process the above three parts respectively.

For \textbf{part (1)}, when $t=1$, 
\begin{equation}
    \left\Vert \boldsymbol{\xi}^{(t)}-\boldsymbol{P}^{(t)}\right\Vert _{2}^{2}=0,
\end{equation}
when $t\geq2$,
\begin{equation}
    \begin{aligned} & \left\Vert \boldsymbol{\xi}^{(t)}-\boldsymbol{P}^{(t)}\right\Vert _{2}^{2}\\=&\left\Vert \frac{\beta_{1}}{1-\beta_{1}}\left(\boldsymbol{P}^{(t)}-\boldsymbol{P}^{(t-1)}\right)\right\Vert _{2}^{2}\\ =& \frac{\beta_{1}^{2}}{\left(1-\beta_{1}\right)^{2}}\alpha_{t-1}^{2}\left\Vert \hat{\mathbf{m}}^{(t-1)}/\sqrt{\hat{\mathbf{v}}^{(t-1)}}\right\Vert _{2}^{2}\\ =& \frac{\beta_{1}^{2}}{\left(1-\beta_{1}\right)^{2}}\alpha_{t-1}^{2}\sum_{i=1}^{d}\left(\hat{m}_{i}^{(t-1)}\right)^{2}/\hat{v}_{i}^{(t-1)}\\ \overset{\left(a\right)}{\leq}& \frac{\beta_{1}^{2}}{\left(1-\beta_{1}\right)^{2}}\alpha_{t-1}^{2}\sum_{i=1}^{d}G_{i}^{2}/c^{2} \end{aligned},
\end{equation}
where (a) means that for any $t$, it satisfies
\begin{equation}
    \begin{aligned}
        \left|\hat{m}_{i}^{(t)}\right|  & \leq\frac{1}{1-\beta_{1}^{t}}\sum_{s=1}^{t}\left(1-\beta_{1}\right)\beta_{1}^{t-s}\left|g_{s,i}\right| \\ & \leq\frac{1}{1-\beta_{1}^{t}}\sum_{s=1}^{t}\left(1-\beta_{1}\right)\beta_{1}^{t-s}G_{i}=G_{i}    
    \end{aligned}
\end{equation}
and
\begin{equation}
    \begin{aligned}
        \hat{v}_{i}^{(t)} & =\max\left(\hat{v}_{i}^{(t-1)},\frac{v^{(t)}}{1-\beta_{2}^{t}}\right) \\ & \geq\hat{v}_{i}^{(t-1)}\geq\ldots\geq\hat{v}_{i}^{(1)}=\frac{\left(1-\beta_{2}\right)g_{1,i}^{2}}{1-\beta_{2}}\geq c^{2}    
    \end{aligned}
\end{equation}

For \textbf{part (2)}, when $t=1$,
\begin{equation}
    \begin{aligned} & \boldsymbol{\xi}^{(t+1)}-\boldsymbol{\xi}^{(t)} \\= & \boldsymbol{P}^{(t+1)}+\frac{\beta_{1}}{1-\beta_{1}}\left(\boldsymbol{P}^{(t+1)}-\boldsymbol{P}^{(t)}\right)-\boldsymbol{P}^{(t)}\\ = & \frac{1}{1-\beta_{1}}\left(\boldsymbol{P}^{(t+1)}-\boldsymbol{P}^{(t)}\right)\\ = & -\frac{\alpha_{t}}{1-\beta_{1}}\frac{1}{1-\beta_{1}^{t}}\left(\mathbf{m}^{(t)}/\sqrt{\hat{\mathbf{v}}^{(t)}}\right)\\ = & -\frac{\alpha_{t}}{1-\beta_{1}}\frac{1}{1-\beta_{1}^{t}}\left(\left(\beta_{1}\cancelto{\mathbf{0}}{\mathbf{m}^{(t-1)}}+\left(1-\beta_{1}\right)\mathbf{g}_{t}\right)/\sqrt{\hat{\mathbf{v}}^{(t)}}\right)\\ = & -\frac{\alpha_{t}}{1-\beta_{1}^{t}}\mathbf{g}_{t}/\sqrt{\hat{\mathbf{v}}^{(t)}} \end{aligned}
\end{equation}
\begin{equation}
    \begin{aligned}&\left\Vert\boldsymbol{\xi}^{(t+1)}-\boldsymbol{\xi}^{(t)}\right\Vert _{2}^{2}\\= & \frac{\alpha_{t}^{2}}{\left(1-\beta_{1}^{t}\right)^{2}}\left\Vert \mathbf{g}_{t}/\sqrt{\hat{\mathbf{v}}^{(t)}}\right\Vert _{2}^{2}\\ = & \frac{\alpha_{t}^{2}}{\left(1-\beta_{1}^{t}\right)^{2}}\sum_{i=1}^{d}g_{t, i}^{2}/\hat{v}_{i}^{(t)}\\ \leq & \frac{\alpha_{t}^{2}}{\left(1-\beta_{1}^{t}\right)^{2}}\sum_{i=1}^{d}G_{i}^{2}/c^{2} \end{aligned}
\end{equation}
when $t\geq2$,
\begin{equation}
    \begin{aligned}&\boldsymbol{\xi}^{(t+1)}-\boldsymbol{\xi}^{(t)}\\= & \boldsymbol{P}^{(t+1)}+\frac{\beta_{1}}{1-\beta_{1}}\left(\boldsymbol{P}^{(t+1)}-\boldsymbol{P}^{(t)}\right)-\boldsymbol{P}^{(t)}\\ &-\frac{\beta_{1}}{1-\beta_{1}}\left(\boldsymbol{P}^{(t)}-\boldsymbol{P}^{(t-1)}\right)\\ = & \frac{1}{1-\beta_{1}}\left(\boldsymbol{P}^{(t+1)}-\boldsymbol{P}^{(t)}\right)-\frac{\beta_{1}}{1-\beta_{1}}\left(\boldsymbol{P}^{(t)}-\boldsymbol{P}^{(t-1)}\right) \end{aligned}
\end{equation}
since 
\begin{equation}
    \begin{aligned}&\boldsymbol{P}^{(t+1)}-\boldsymbol{P}^{(t)}\\= & -\alpha_{t}\hat{\mathbf{m}}^{(t)}/\sqrt{\hat{\mathbf{v}}^{(t)}}\\ = & -\frac{\alpha_{t}}{1-\beta_{1}^{t}}\mathbf{m}^{(t)}/\sqrt{\hat{\mathbf{v}}^{(t)}}\\ = & -\frac{\alpha_{t}}{1-\beta_{1}^{t}}\left(\beta_{1}\mathbf{m}^{(t-1)}+\left(1-\beta_{1}\right)\mathbf{g}_{t}\right)/\sqrt{\hat{\mathbf{v}}^{(t)}} \end{aligned}
\end{equation}
leading to 
\begin{equation}
    \begin{aligned} & \boldsymbol{\xi}^{(t+1)}-\boldsymbol{\xi}^{(t)}\\ = & \frac{1}{1-\beta_{1}}\left(-\frac{\alpha_{t}}{1-\beta_{1}^{t}}\left(\beta_{1}\mathbf{m}^{(t-1)}+\left(1-\beta_{1}\right)\mathbf{g}_{t}\right)/\sqrt{\hat{\mathbf{v}}^{(t)}}\right)\\ & -\frac{\beta_{1}}{1-\beta_{1}}\left(-\frac{\alpha_{t-1}}{1-\beta_{1}^{t-1}}\mathbf{m}^{(t-1)}/\sqrt{\hat{\mathbf{v}}^{(t-1)}}\right)\\ = & -\frac{\beta_{1}}{1-\beta_{1}}\mathbf{m}^{(t-1)}\odot\left(\frac{\alpha_{t}}{1-\beta_{1}^{t}}/\sqrt{\hat{\mathbf{v}}^{(t)}}-\frac{\alpha_{t-1}}{1-\beta_{1}^{t-1}}/\sqrt{\hat{\mathbf{v}}^{(t-1)}}\right)\\ & -\frac{\alpha_{t}}{1-\beta_{1}^{t}}\mathbf{g}_{t}/\sqrt{\hat{\mathbf{v}}^{(t)}} \end{aligned}
\end{equation}
and 
\begin{equation}
    \begin{aligned} & \left\Vert \boldsymbol{\xi}^{(t+1)}-\boldsymbol{\xi}^{(t)}\right\Vert _{2}^{2}
    \\\leq&2\left\Vert -\frac{\beta_{1}}{1-\beta_{1}}\mathbf{m}^{(t-1)}\odot\Bigg(\frac{\alpha_{t}}{1-\beta_{1}^{t}}/\sqrt{\hat{\mathbf{v}}^{(t)}}\right.
    \\ & \left.-\frac{\alpha_{t-1}}{1-\beta_{1}^{t-1}}/\sqrt{\hat{\mathbf{v}}^{(t-1)}}\Bigg)\right\Vert _{2}^{2}+2\left\Vert -\frac{\alpha_{t}}{1-\beta_{1}^{t}}\mathbf{g}_{t}/\sqrt{\hat{\mathbf{v}}^{(t)}}\right\Vert _{2}^{2}
    \\  \leq&2\frac{\beta_{1}^{2}}{\left(1-\beta_{1}\right)^{2}}\left\Vert \mathbf{m}^{(t-1)}\right\Vert _{\infty}^{2}\left\Vert \frac{\alpha_{t}}{1-\beta_{1}^{t}}/\sqrt{\hat{\mathbf{v}}^{(t)}} \right.
    \\ & \left.-\frac{\alpha_{t-1}}{1-\beta_{1}^{t-1}}/\sqrt{\hat{\mathbf{v}}^{(t-1)}}\right\Vert _{\infty}\cdot \left\Vert \frac{\alpha_{t}}{1-\beta_{1}^{t}}/\sqrt{\hat{\mathbf{v}}^{(t)}}\right.
    \\&\left.-\frac{\alpha_{t-1}}{1-\beta_{1}^{t-1}}/\sqrt{\hat{\mathbf{v}}^{(t-1)}}\right\Vert _{1}+2\frac{\alpha_{t}^{2}}{\left(1-\beta_{1}^{t}\right)^{2}}\left\Vert \mathbf{g}_{t}/\sqrt{\hat{\mathbf{v}}^{(t)}}\right\Vert _{2}^{2} \end{aligned}.
\end{equation}
Since
\begin{equation}
    \begin{aligned}
        \left|m_{i}^{(t-1)}\right|&=\left(1-\beta_{1}^{t}\right)\left|\hat{m}_{i}^{(t)}\right|
        \\&\leq\left|\hat{m}_{i}^{(t)}\right|\leq G_{i},\left\Vert \mathbf{m}^{(t-1)}\right\Vert _{\infty}^{2}\leq\left(\max_{i}G_{i}\right)^{2},
    \end{aligned}
\end{equation}
\begin{equation}
    \left\Vert \mathbf{g}_{t}/\sqrt{\hat{\mathbf{v}}^{(t)}}\right\Vert _{2}^{2}=\sum_{i=1}^{d}g_{t,i}^{2}/\hat{v}_{i}^{(t)}\leq\sum_{i=1}^{d}G_{i}^{2}/c^{2},
\end{equation}
\begin{equation}
    \begin{aligned}
        &\left\Vert \frac{\alpha_{t}}{1-\beta_{1}^{t}}/\sqrt{\hat{\mathbf{v}}^{(t)}}-\frac{\alpha_{t-1}}{1-\beta_{1}^{t-1}}/\sqrt{\hat{\mathbf{v}}^{(t-1)}}\right\Vert _{\infty}
        \\=&\max_{i}\left|\frac{\alpha_{t}}{1-\beta_{1}^{t}}/\sqrt{\hat{v}_{i}^{(t)}}-\frac{\alpha_{t-1}}{1-\beta_{1}^{t-1}}/\sqrt{\hat{v}_{i}^{(t-1)}}\right|,
    \end{aligned}
\end{equation}
where 
\begin{equation}
    \begin{aligned} & \alpha_{t}/\left(1-\beta_{1}^{t}\right)/\sqrt{\hat{v}_{i}^{(t)}}\geq0,\thinspace\thinspace\alpha_{t-1}/\left(1-\beta_{1}^{t-1}\right)/\sqrt{\hat{v}_{i}^{(t-1)}}\geq0\\ & \alpha_{t}\leq\alpha_{t-1},\thinspace\thinspace\frac{1}{1-\beta_{1}^{t}}\leq\frac{1}{1-\beta_{1}^{t-1}},\thinspace\thinspace\frac{1}{\sqrt{\hat{v}_{i}^{(t)}}}\leq\frac{1}{\sqrt{\hat{v}_{i}^{(t-1)}}}\\ \Longrightarrow & \frac{\alpha_{t}}{1-\beta_{1}^{t}}/\sqrt{\hat{\mathbf{v}}^{(t)}}\leq\frac{\alpha_{t-1}}{1-\beta_{1}^{t-1}}/\sqrt{\hat{\mathbf{v}}^{(t-1)}}\\ \Longrightarrow & \left|\frac{\alpha_{t}}{1-\beta_{1}^{t}}/\sqrt{\hat{v}_{i}^{(t)}}-\frac{\alpha_{t-1}}{1-\beta_{1}^{t-1}}/\sqrt{\hat{v}_{i}^{(t-1)}}\right|\\ & =\alpha_{t-1}/\left(1-\beta_{1}^{t-1}\right)/\sqrt{\hat{v}_{i}^{(t-1)}}-\alpha_{t}/\left(1-\beta_{1}^{t}\right)/\sqrt{\hat{v}_{i}^{(t)}}\\ & \leq\alpha_{t-1}/\left(1-\beta_{1}^{t-1}\right)/\sqrt{\hat{v}_{i}^{(t-1)}}\leq\alpha_{1}/\left(1-\beta_{1}\right)/c\\ \Longrightarrow & \left\Vert \frac{\alpha_{t}}{1-\beta_{1}^{t}}/\sqrt{\hat{\mathbf{v}}^{(t)}}-\frac{\alpha_{t-1}}{1-\beta_{1}^{t-1}}/\sqrt{\hat{\mathbf{v}}^{(t-1)}}\right\Vert _{\infty}\leq\frac{\alpha_{1}}{\left(1-\beta_{1}\right)c} \end{aligned},
\end{equation}
and 
\begin{equation}
    \begin{aligned}
        &\left\Vert \frac{\alpha_{t}}{1-\beta_{1}^{t}}/\sqrt{\hat{\mathbf{v}}^{(t)}}-\frac{\alpha_{t-1}}{1-\beta_{1}^{t-1}}/\sqrt{\hat{\mathbf{v}}^{(t-1)}}\right\Vert _{1}
        \\=&\sum_{i=1}^{d}\left(\alpha_{t-1}/\left(1-\beta_{1}^{t-1}\right)/\sqrt{\hat{v}_{i}^{(t-1)}}-\alpha_{t}/\left(1-\beta_{1}^{t}\right)/\sqrt{\hat{v}_{i}^{(t)}}\right),
    \end{aligned}
\end{equation}
so 
\begin{equation}
    \begin{aligned} & \left\Vert \boldsymbol{\xi}^{(t+1)}-\boldsymbol{\xi}^{(t)}\right\Vert _{2}^{2}\\ \leq & 2\frac{\beta_{1}^{2}}{\left(1-\beta_{1}\right)^{2}}\left(\max_{i}G_{i}\right)^{2}\frac{\alpha_{1}}{\left(1-\beta_{1}\right)c}\cdot\\ & \sum_{i=1}^{d}\left(\frac{\alpha_{t-1}}{\left(1-\beta_{1}^{t-1}\right)\sqrt{\hat{v}_{i}^{(t-1)}}}-\frac{\alpha_{t}}{\left(1-\beta_{1}^{t}\right)\sqrt{\hat{v}_{i}^{(t)}}}\right)\\ & +2\frac{\alpha_{t}^{2}}{\left(1-\beta_{1}^{t}\right)^{2}}\sum_{i=1}^{d}G_{i}^{2}/c^{2} \end{aligned}
\end{equation}

For \textbf{part (3)}, when $t=1$ and $\mathbf{g}_{t}=\nabla f\left(\boldsymbol{P}^{(t)}\right)+\mathbf{n}_{t}$
\begin{equation}
    \begin{aligned} & \left\langle \nabla f\left(\boldsymbol{P}^{(t)}\right),\boldsymbol{\xi}^{(t+1)}-\boldsymbol{\xi}^{(t)}\right\rangle \\ = & \left\langle \nabla f\left(\boldsymbol{P}^{(t)}\right),-\frac{\alpha_{t}}{1-\beta_{1}^{t}}\mathbf{g}_{t}/\sqrt{\hat{\mathbf{v}}^{(t)}}\right\rangle \\ = & \left\langle \nabla f\left(\boldsymbol{P}^{(t)}\right),-\frac{\alpha_{t}}{1-\beta_{1}^{t}}\nabla f\left(\boldsymbol{P}^{(t)}\right)/\sqrt{\hat{\mathbf{v}}^{(t)}}\right\rangle \\ & +\left\langle \nabla f\left(\boldsymbol{P}^{(t)}\right),-\frac{\alpha_{t}}{1-\beta_{1}^{t}}\mathbf{n}_{t}/\sqrt{\hat{\mathbf{v}}^{(t)}}\right\rangle \end{aligned},
\end{equation}
where 
\begin{equation}
    \begin{aligned} & \left\langle \nabla f\left(\boldsymbol{P}^{(t)}\right),-\frac{\alpha_{t}}{1-\beta_{1}^{t}}\nabla f\left(\boldsymbol{P}^{(t)}\right)/\sqrt{\hat{\mathbf{v}}^{(t)}}\right\rangle \\ = & -\frac{\alpha_{t}}{1-\beta_{1}^{t}}\sum_{i=1}^{d}\left[\nabla f\left(\boldsymbol{P}^{(t)}\right)\right]_{i}^{2}/\sqrt{\hat{v}_{i}^{(t)}}\\ \leq & -\frac{\alpha_{t}}{1-\beta_{1}^{t}}\sum_{i=1}^{d}\left[\nabla f\left(\boldsymbol{P}^{(t)}\right)\right]_{i}^{2}/\max_{i}\left(G_{i}\right)\\ = & -\frac{\alpha_{t}}{\left(1-\beta_{1}^{t}\right)\max_{i}\left(G_{i}\right)}\left\Vert \nabla f\left(\boldsymbol{P}^{(t)}\right)\right\Vert _{2}^{2} \end{aligned}
\end{equation}
and 
\begin{equation}
    \begin{aligned} & \left\langle \nabla f\left(\boldsymbol{P}^{(t)}\right),-\frac{\alpha_{t}}{1-\beta_{1}^{t}}\mathbf{n}_{t}/\sqrt{\hat{\mathbf{v}}^{(t)}}\right\rangle \\ \leq & \frac{\alpha_{t}}{1-\beta_{1}^{t}}\left\Vert \nabla f\left(\boldsymbol{P}^{(t)}\right)\right\Vert _{\infty}\left\Vert \mathbf{n}_{t}\right\Vert _{\infty}\left\Vert 1/\sqrt{\hat{\mathbf{v}}^{(t)}}\right\Vert _{1}\\ = & \frac{\alpha_{t}}{1-\beta_{ 1}^{t}}\left\Vert \nabla f\left(\boldsymbol{P}^{(t)}\right)\right\Vert _{\infty}\left\Vert \mathbf{g}_{t}-\nabla f\left(\boldsymbol{P}^{(t)}\right)\right\Vert _{\infty}\left\Vert 1/\sqrt{\hat{\mathbf{v}}^{(t)}}\right\Vert _{1}\\ \leq & \frac{\alpha_{t}}{1-\beta_{1}^{t}}\left(\max_{i}G_{i}\right)\left(2\max_{i}G_{i}\right)\sum_{i=1}^{d}1/c\\ = & \frac{\alpha_{t}}{1-\beta_{1}^{ t}}\left(\max_{i}G_{i}\right)\left(2\max_{i}G_{i}\right)d/c \end{aligned},
\end{equation}
therefore
\begin{equation}
    \begin{aligned} & \left\langle \nabla f\left(\boldsymbol{P}^{(t)}\right),\boldsymbol{\xi}^{(t+1)}-\boldsymbol{\xi}^{(t)}\right\rangle \\ \leq & -\frac{\alpha_{t}}{\left(1-\beta_{1}^{t}\right)\max_{i}\left(G_{i}\right)}\left\Vert \nabla f\left(\boldsymbol{P}^{(t)}\right)\right\Vert _{2}^{2}\\ & +\frac{\alpha_{t}}{1-\beta_{1}^{t}}\left(\max_{i}G_{i}\right)\left(2\max_{i}G_{i}\right)d/c \end{aligned}.
\end{equation}
When $t\geq2$
\begin{equation}
    \begin{aligned} & \left\langle \nabla f\left(\boldsymbol{P}^{(t)}\right),\boldsymbol{\xi}^{(t+1)}-\boldsymbol{\xi}^{(t)}\right\rangle \\ = & \left\langle \nabla f\left(\boldsymbol{P}^{(t)}\right),-\frac{\beta_{1}}{1-\beta_{1}}\mathbf{m}^{(t-1)}\odot\right.\\ & \left.\left(\frac{\alpha_{t}}{1-\beta_{1}^{t}}/\sqrt{\hat{\mathbf{v}}^{(t)}}-\frac{\alpha_{t-1}}{1-\beta_{1}^{t-1}}/\sqrt{\hat{\mathbf{v}}^{(t-1)}}\right)\right\rangle \\ & +\left\langle \nabla f\left(\boldsymbol{P}^{(t)}\right),-\frac{\alpha_{t}}{1-\beta_{1}^{t}}\nabla f\left(\boldsymbol{P}^{(t)}\right)/\sqrt{\hat{\mathbf{v}}^{(t)}}\right\rangle \\ & +\left\langle \nabla f\left(\boldsymbol{P}^{(t)}\right),-\frac{\alpha_{t}}{1-\beta_{1}^{t}}\mathbf{n}_{t}/\sqrt{\hat{\mathbf{v}}^{(t)}}\right\rangle \end{aligned},
\end{equation}
where the first item after the equal sign
\begin{equation}
    \begin{aligned} & \left\langle \nabla f\left(\boldsymbol{P}^{(t)}\right),-\frac{\beta_{1}}{1-\beta_{1}}\mathbf{m}^{(t-1)}\odot\right.\\ & \left.\left(\frac{\alpha_{t}}{1-\beta_{1}^{t}}/\sqrt{\hat{\mathbf{v}}^{(t)}}-\frac{\alpha_{t-1}}{1-\beta_{1}^{t-1}}/\sqrt{\hat{\mathbf{v}}^{(t-1)}}\right)\right\rangle \\ \leq & \frac{\beta_{1}}{1-\beta_{1}}\left\Vert \nabla f\left(\boldsymbol{P}^{(t)}\right)\right\Vert _{\infty}\left\Vert \mathbf{m}^{(t-1)}\right\Vert _{\infty}\cdot\\ & \left\Vert \frac{\alpha_{t}}{1-\beta_{1}^{t}}/\sqrt{\hat{\mathbf{v}}^{(t)}}-\frac{\alpha_{t-1}}{1-\beta_{1}^{t-1}}/\sqrt{\hat{\mathbf{v}}^{(t-1)}}\right\Vert _{1}\\ \leq & \frac{\beta_{1}}{1-\beta_{1}}\left(\max_{i}G_{i}\right)\left(\max_{i}G_{i}\right)\cdot\\ & \sum_{i=1}^{d}\left(\frac{\alpha_{t-1}}{\left(1-\beta_{1}^{t-1}\right)\sqrt{\hat{ v}_{i}^{(t-1)}}}-\frac{\alpha_{t}}{\left(1-\beta_{1}^{t}\right)\sqrt{\hat{v}_{i}^{(t)}}}\right) \end{aligned},
\end{equation}
the second item $\left\langle \nabla f\left(\boldsymbol{P}^{(t)}\right),-\frac{\alpha_{t}}{1-\beta_{1}^{t}}\nabla f\left(\boldsymbol{P}^{(t)}\right)/\sqrt{\hat{\mathbf{v}}^{(t)}}\right\rangle$ is handled similarly with $t=1$.
The third item is calculated as:
\begin{equation}
    \begin{aligned} & \left\langle \nabla f\left(\boldsymbol{P}^{(t)}\right),-\frac{\alpha_{t}}{1-\beta_{1}^{t}}\mathbf{n}_{t}/\sqrt{\hat{\mathbf{v}}^{(t)}}\right\rangle \\ = & \left\langle \nabla f\left(\boldsymbol{P}^{(t)}\right),-\mathbf{n}_{t}\odot\left(\frac{\alpha_{t}}{1-\beta_{1}^{t}}/\sqrt{\hat{\mathbf{v}}^{(t)}}-\frac{\alpha_{t-1}}{1-\beta_{1}^{t-1}}/\sqrt{\hat{\mathbf{v}}^{(t-1)}}\right)\right\rangle \\ & +\left\langle \nabla f\left(\boldsymbol{P}^{(t)}\right),-\frac{\alpha_{t-1}}{1-\beta_{1}^{t-1}}\mathbf{n}_{t}/\sqrt{\hat{\mathbf{v}}^{(t-1)}}\right\rangle \end{aligned},
\end{equation}
where the first term after the equal sign is scaled down as:
\begin{equation}
    \begin{aligned} 
    & \left\langle \nabla f\left(\boldsymbol{P}^{(t)}\right),-\mathbf{n}_{t}\odot\left(\frac{\alpha_{t}}{1-\beta_{1}^{t}}/\sqrt{\hat{\mathbf{v}}^{(t)}}-\frac{\alpha_{t-1}}{1-\beta_{1}^{t-1}}/\sqrt{\hat{\mathbf{v}}^{(t-1)}}\right)\right\rangle 
    \\ \leq & \left\Vert \nabla f\left(\boldsymbol{P}^{(t)}\right)\right\Vert _{\infty}\left\Vert \mathbf{n}_{t}\right\Vert _{\infty}\left\Vert \frac{\alpha_{t}}{1-\beta_{1}^{t}}/\sqrt{\hat{\mathbf{v}}^{(t)}}-\frac{\alpha_{t-1}}{1-\beta_{1}^{t-1}}/\sqrt{\hat{\mathbf{v}}^{(t-1)}}\right\Vert _{1}    \\\leq&\left(\max_{i}G_{i}\right)\left(2\max_{i}G_{i}\right)\cdot\sum_{i=1}^{d}\left(\frac{\alpha_{t-1}}{\left(1-\beta_{1}^{t-1}\right)\sqrt{\hat{v}_{i}^{(t-1)}}}\right.
    \\&\left.-\frac{\alpha_{t}}{\left(1-\beta_{1}^{t}\right)\sqrt{\hat{v}_{i}^{( t)}}}\right) \end{aligned}.
\end{equation}
At last
\begin{equation}
    \begin{aligned} 
    & \left\langle \nabla f\left(\boldsymbol{P}^{(t)}\right),\boldsymbol{\xi}^{(t+1)}-\boldsymbol{\xi}^{(t)}\right\rangle 
    \\ \leq & \frac{\beta_{1}}{1-\beta_{1}}\left(\max_{i}G_{i}\right)\left(\max_{i}G_{i}\right)\cdot
    \\ & \sum_{i=1}^{d}\left(\frac{\alpha_{t-1}}{\left(1-\beta_{1}^{t-1}\right)\sqrt{\hat{v}_{i}^{(t-1)}}}-\frac{\alpha_{t}}{\left(1-\beta_{1}^{t}\right)\sqrt{\hat{v}_{i}^{( t)}}}\right)
    \\ & -\frac{\alpha_{t}}{\left(1-\beta_{1}^{t}\right)\max_{i}\left(G_{i}\right)}\left\Vert \nabla f\left(\boldsymbol{P}^{(t)}\right)\right\Vert_{2}^{2}+\left(\max_{i}G_{i}\right)\left(2\max_{i}G_{i}\right)\cdot
    \\&\sum_{i=1}^{d}\left(\frac{\alpha_{t-1}}{\left(1-\beta_{1}^{t-1}\right)\sqrt{\hat{v}_{i}^{(t-1)}}}-\frac{\alpha_{t}}{\left(1-\beta_{1}^{t}\right)\sqrt{\hat{v}_{i}^{(t)}}}\right)
    \\ & +\left\langle \nabla f\left(\boldsymbol{P}^{(t)}\right),-\frac{\alpha_{t-1}}{1-\beta_{1}^{t-1}}\mathbf{n}_{t}/\sqrt{\hat{\mathbf{v}}^{(t-1)}}\right\rangle \end{aligned}.
\end{equation}

Then, we start sorting the above items out.
When $t=1$,
\begin{equation}
    \begin{aligned} & f\left(\boldsymbol{\xi}^{(t+1)}\right)-f\left(\boldsymbol{\xi}^{(t)}\right)\\ \leq & \frac{L}{2}\cdot0+L\frac{\alpha_{t}^{2}}{\left(1-\beta_{1}^{t}\right)^{2}}\sum_{i=1}^{d}G_{i}^{2}/c^{2}\\ & -\frac{\alpha_{t}}{\left(1-\beta_{1}^{t}\right)\max_{i}\left(G_{i}\right)}\left\Vert \nabla f\left(\boldsymbol{P}^{(t)}\right)\right\Vert _{2}^{2}\\ & +\frac{\alpha_{ t}}{1-\beta_{1}^{t}}\left(\max_{i}G_{i}\right)\left(2\max_{i}G_{i}\right)d/c \end{aligned},
\end{equation}
find the expectation for the random distribution of $\mathbf{n}_{1},\mathbf{n}_{2},\ldots,\mathbf{n}_{t}$ on both sides of the inequality sign as the follows:
\begin{equation}
    \begin{aligned} & \mathbb{E}_{t}\left[f\left(\boldsymbol{\xi}^{(t+1)}\right)-f\left(\boldsymbol{\xi}^{(t)}\right)\right]\\ \leq & L\frac{\alpha_{t}^{2}}{\left(1-\beta_{1}^{t}\right)^{2}}\sum_{i=1}^{d}G_{i}^{2}/c^{2}\\ & -\frac{\alpha_{t}}{\left(1-\beta_{1}^{t}\right)\max_{i}\left(G_{i}\right)}\mathbb{E}_{t}\left[\left\Vert \nabla f\left(\boldsymbol{P}^{(t)}\right)\right\Vert _{ 2}^{2}\right]\\ & +\frac{\alpha_{t}}{1-\beta_{1}^{t}}\left(\max_{i}G_{i}\right)\left(2\max_{i}G_{i}\right)d/c \end{aligned}.
\end{equation}
When $t\geq2$,
\begin{equation}
    \begin{aligned} 
        & f\left(\boldsymbol{\xi}^{(t+1)}\right)-f\left(\boldsymbol{\xi}^{(t)}\right)\\ \leq & \frac{L}{2}\frac{\beta_{1}^{2}}{\left(1-\beta_{1}\right)^{2}}\alpha_{t-1}^{2}\sum_{i=1}^{d}G_{i}^{2}/c^{2}
        \\ & +L\cdot2\frac{\beta_{1}^{2}}{\left(1-\beta_{1}\right)^{2}}\left(\max_{i}G_{i}\right)^{2}\frac{\alpha_{1}}{\left(1-\beta_{1}\right)c}\cdot
        \\ & \sum_{i=1}^{d}\left(\frac{\alpha_{t-1}}{\left(1-\beta_{1}^{t-1}\right)\sqrt{\hat{v}_{i}^{(t-1)}}}-\frac{\alpha_{t}}{\left(1-\beta_{1}^{t}\right)\sqrt{\hat{v}_{i}^{(t)}}}\right)
        \\ & +L\cdot2\frac{\alpha_{t}^{2}}{\left(1-\beta_{1}^{t}\right)^{2}}\sum_{i=1}^{d}G_{i}^{2}/c^{2}
        \\ & +\frac{\beta_{1}}{1-\beta_{1}}\left(\max_{i}G_{i}\right)\left(\max_{i}G_{i}\right)\cdot
        \\ & \sum_{i=1}^{d}\left(\frac{\alpha_{t-1}}{\left(1-\beta_{1}^{t-1}\right)\sqrt{\hat{v}_{i}^{(t-1)}}}-\frac{\alpha_{t}}{\left(1-\beta_{1}^{t}\right)\sqrt{\hat{v}_{i}^{(t)}}}\right)
        \\ & -\frac{\alpha_{t}}{\left(1-\beta_{1}^{t}\right)\max_{i}\left(G_{i}\right)}\left\Vert \nabla f\left(\boldsymbol{P}^{(t)}\right)\right\Vert _{2}^{2}
        \\ & +\left(\max_{i}G_{i}\right)\left(2\max_{i}G_{i}\right)\cdot
        \\&\sum_{i=1}^{d}\left(\frac{\alpha_{t-1}}{\left(1-\beta_{1}^{t-1}\right)\sqrt{\hat{v}_{i}^{(t-1)}}}-\frac{\alpha_{t}}{\left(1-\beta_{1}^{t}\right)\sqrt{\hat{v}_{i}^{(t)}}}\right)
        \\ & +\left\langle \nabla f\left(\boldsymbol{P}^{(t)}\right),-\frac{\alpha_{t-1}}{1-\beta_{1}^{t-1}}\mathbf{n}_{t}/\sqrt{\hat{\mathbf{v}}^{(t-1)}}\right\rangle,
    \end{aligned}
\end{equation}
find the expectation for the random distribution of $\mathbf{n}_{1},\mathbf{n}_{2},\ldots,\mathbf{n}_{t}$ on both sides of the inequality sign as the follows:
\begin{equation}
\label{right_part_left_part}
    \begin{aligned} & \mathbb{E}_{t}\left[f\left(\boldsymbol{\xi}^{(t+1)}\right)-f\left(\boldsymbol{\xi}^{(t)}\right)\right]
    \\ \leq & \frac{L}{2}\frac{\beta_{1}^{2}}{\left(1-\beta_{1}\right)^{2}}\alpha_{t-1}^{2}\sum_{i=1}^{d}G_{i}^{2}/c^{2}
    \\ & +L\cdot2\frac{\beta_{1}^{2}}{\left(1-\beta_{1}\right)^{2}}\left(\max_{i}G_{i}\right)^{2}\frac{\alpha_{1}}{\left(1-\beta_{1}\right)c}\cdot
    \\ & \sum_{i=1}^{d}\left(\frac{\alpha_{t-1}}{\left(1-\beta_{1}^{t-1}\right)\sqrt{\hat{v}_{i}^{(t-1)}}}-\frac{\alpha_{t}}{\left(1-\beta_{1}^{t}\right)\sqrt{\hat{v}_{i}^{(t)}}}\right)
    \\ & +L\cdot2\frac{\alpha_{t}^{2}}{\left(1-\beta_{1}^{t}\right)^{2}}\sum_{i=1}^{d}G_{i}^{2}/c^{2}
    \\ & +\frac{\beta_{1}}{1-\beta_{1}}\left(\max_{i}G_{i}\right)\left(\max_{i}G_{i}\right)\cdot
    \\ & \sum_{i=1}^{d}\left(\frac{\alpha_{t-1}}{\left(1-\beta_{1}^{t-1}\right)\sqrt{\hat{v}_{i}^{(t-1)}}}-\frac{\alpha_{t}}{\left(1-\beta_{1}^{t}\right)\sqrt{\hat{v}_{i}^{(t)}}}\right)
    \\ & -\frac{\alpha_{t}}{\left(1-\beta_{1}^{t}\right)\max_{i}\left(G_{i}\right)}\mathbb{E}_{t}\left[\left\Vert \nabla f\left(\boldsymbol{P}^{(t)}\right)\right\Vert _{2}^{2}\right]    \\&+\left(\max_{i}G_{i}\right)\left(2\max_{i}G_{i}\right)\cdot\sum_{i=1}^{d}\left(\frac{\alpha_{t-1}}{\left(1-\beta_{1}^{t-1}\right)\sqrt{\hat{v}_{i}^{(t-1)}}}\right.
    \\&\left.-\frac{\alpha_{t}}{\left(1-\beta_{1}^{t}\right)\sqrt{\hat{v}_{i}^{(t)}}}\right)
    \\&+\mathbb{E}_{t}\left[\left\langle \nabla f\left(\boldsymbol{P}^{(t)}\right),-\frac{\alpha_{t-1}}{1-\beta_{1}^{t-1}}\mathbf{n}_{t}/\sqrt{\hat{\mathbf{v}}^{(t-1)}}\right\rangle \right] \end{aligned}.
\end{equation}
Since the values of $\boldsymbol{P}^{(t)}$ and $\hat{\mathbf{v}}^{(t-1)}$ have nothing to do with $\mathbf{g}_{t}$, they are statistically independent from $\mathbf{n}_{t}$, so
\begin{equation}
    \begin{aligned} & \mathbb{E}_{t}\left[\left\langle \nabla f\left(\boldsymbol{P}^{(t)}\right),-\frac{\alpha_{t-1}}{1-\beta_{1}^{t-1}}\mathbf{n}_{t}/\sqrt{\hat{\mathbf{v}}^{(t-1)}}\right\rangle \right]\\ = & \mathbb{E}_{t}\left[\left\langle -\frac{\alpha_{t-1}}{1-\beta_{1}^{t-1}}\nabla f\left(\boldsymbol{P}^{(t)}\right)/\sqrt{\hat{\mathbf{v}}^{(t-1)}},\mathbf{n}_{t}\right\rangle \right]\\ = & \left\langle -\frac{\alpha_{t-1}}{1-\beta_{1}^{t-1}}\mathbb{E}_{t}\left[\nabla f\left(\boldsymbol{P}^{(t)}\right)/\sqrt{\hat{\mathbf{v}}^{(t-1)}}\right],\cancelto{\mathbf{0}}{\mathbb{E}_{t}\left[\mathbf{n}_{t}\right]}\right\rangle =0 \end{aligned}.
\end{equation}
Sum $t=1,2,\ldots, T$ on both sides of the inequality sign at the same time.
For the left part of Ineq. \ref{right_part_left_part}, which can be reduced to allow the unequal sign to continue to hold:
\begin{equation}
    \begin{aligned} 
        & \sum_{t=1}^{T}\mathbb{E}_{t}\left[f\left(\boldsymbol{\xi}^{(t+1)}\right)-f\left(\boldsymbol{\xi}^{(t)}\right)\right]\\ = & \sum_{t=1}^{T}\mathbb{E}_{t}\left[f\left(\boldsymbol{\xi}^{(t+1)}\right)\right]-\mathbb{E}_{t}\left[f\left(\boldsymbol{\xi}^{(t)}\right)\right]\\ = & \sum_{t=1}^{T}\mathbb{E}_{t}\left[f\left(\boldsymbol{\xi}^{(t+1)}\right)\right]-\mathbb{E}_{t-1}\left[f\left(\boldsymbol{\xi}^{(t)}\right)\right]\\ = & \mathbb{E}_{T}\left[f\left(\boldsymbol{\xi}^{(T+1)}\right)\right]-\mathbb{E}_{0}\left[f\left(\boldsymbol{\xi}^{(1)}\right)\right] 
    \end{aligned},
\end{equation}
Due to $f\left(\boldsymbol{\xi}^{(T+1)}\right)\geq\min_{\boldsymbol{P}}f\left(\boldsymbol{P}\right)=f\left(\boldsymbol{P}^{\star}\right), \boldsymbol{\xi}^{(1)}=\boldsymbol{P}^{(1)}$ and neither is random, leading to
\begin{equation}
    \begin{aligned}
        &\sum_{t=1}^{T}\mathbb{E}_{t}\left[f\left(\boldsymbol{\xi}^{(t+1)}\right)-f\left(\boldsymbol{\xi}^{(t)}\right)\right]
        \\ \geq & \mathbb{E}_{T}\left[f\left(\boldsymbol{P}^{\star}\right)\right]-\mathbb{E}_{0}\left[f\left(\boldsymbol{P}^{(1)}\right)\right]
        \\ = & f\left(\boldsymbol{P}^{\star}\right)-f\left(\boldsymbol{P}^{(1)}\right) \end{aligned}
\end{equation}
The right part of the unequal sign can be enlarged to keep the unequal sign holding.
Firstly, a series of substitutions are made to simplify the symbols as follows:
\begin{equation}
    \frac{L}{2}\frac{\beta_{1}^{2}}{\left(1-\beta_{1}\right)^{2}}\alpha_{t-1}^{2}\sum_{i=1}^{d}G_{i}^{2}/c^{2}\triangleq C_{1}\alpha_{t-1}^{2},
\end{equation}
i.e., where $C_1 \triangleq \frac{L}{2}\frac{\beta_{1}^{2}}{\left(1-\beta_{1}\right)^{2}}\sum_{i=1}^{d}G_{i}^{2}/c^{2}$.
\begin{equation}
    \begin{aligned} 
    & L\cdot2\frac{\beta_{1}^{2}}{\left(1-\beta_{1}\right)^{2}}\left(\max_{i}G_{i}\right)^{2}\frac{\alpha_{1}}{\left(1-\beta_{1}\right)c}\cdot
    \\ & \sum_{i=1}^{d}\left(\frac{\alpha_{t-1}}{\left(1-\beta_{1}^{t-1}\right)\sqrt{\hat{v}_{i}^{(t-1)}}}-\frac{\alpha_{t}}{\left(1-\beta_{1}^{t}\right)\sqrt{\hat{v}_{i}^{(t)}}}\right)
    \\ \triangleq & C_{2}\sum_{i=1}^{d}\left(\frac{\alpha_{t-1}}{\left(1-\beta_{1}^{t-1}\right)\sqrt{\hat{v}_{i}^{(t-1)}}}-\frac{\alpha_{t}}{\left(1-\beta_{1}^{t}\right)\sqrt{\hat{v}_{i}^{(t)}}}\right) \end{aligned},
\end{equation}
i.e., where $C_2 \triangleq L\cdot2\frac{\beta_{1}^{2}}{\left(1-\beta_{1}\right)^{2}}\left(\max_{i}G_{i}\right)^{2}\frac{\alpha_{1}}{\left(1-\beta_{1}\right)c}$.
\begin{equation}
    \begin{aligned}L\cdot2\frac{\alpha_{t}^{2}}{\left(1-\beta_{1}^{t}\right)^{2}}\sum_{i=1}^{d}G_{i}^{2}/c^{2}\leq & L\cdot2\frac{\alpha_{t}^{2}}{\left(1-\beta_{1}\right)^{2}}\sum_{i=1}^{d}G_{i}^{2}/c^{2}
    \\ \triangleq & C_{3}\alpha_{t}^{2} \end{aligned},
\end{equation}
i.e., where $C_3 \triangleq L\cdot2\frac{1}{\left(1-\beta_{1}\right)^{2}}\sum_{i=1}^{d}G_{i}^{2}/c^{2}$.
\begin{equation}
    \begin{aligned} & \frac{\beta_{1}}{1-\beta_{1}}\left(\max_{i}G_{i}\right)\left(\max_{i}G_{i}\right)\cdot
    \\ & \sum_{i=1}^{d}\left(\frac{\alpha_{t-1}}{\left(1-\beta_{1}^{t-1}\right)\sqrt{\hat{v}_{i}^{(t-1)}}}-\frac{\alpha_{t}}{\left(1-\beta_{1}^{t}\right)\sqrt{\hat{v}_{i}^{(t)}}}\right)\\ \triangleq & C_{4}\sum_{i=1}^{d}\left(\frac{\alpha_{t-1}}{\left(1-\beta_{1}^{t-1}\right)\sqrt{\hat{v}_{i}^{(t-1)}}}-\frac{\alpha_{t}}{\left(1-\beta_{1}^{t}\right)\sqrt{\hat{v}_{i}^{(t)}}}\right) \end{aligned},
\end{equation}
i.e., where $C_4 \triangleq \frac{\beta_{1}}{1-\beta_{1}}\left(\max_{i}G_{i}\right)\left(\max_{i}G_{i}\right)$.
\begin{equation}
    \begin{aligned}&-\frac{\alpha_{t}}{\left(1-\beta_{1}^{t}\right)\max_{i}\left(G_{i}\right)}\mathbb{E}_{t}\left[\left\Vert \nabla f\left(\boldsymbol{P}^{(t)}\right)\right\Vert _{2}^{2}\right]
    \\\leq & -\frac{\alpha_{t}}{\max_{i}\left(G_{i}\right)}\mathbb{E}_{t}\left[\left\Vert \nabla f\left(\boldsymbol{P}^{(t)}\right)\right\Vert _{2}^{2}\right]\\ \triangleq & -C^{\prime}\alpha_{t}\mathbb{E}_{t}\left[\left\Vert \nabla f\left(\boldsymbol{P}^{(t)}\right)\right\Vert _{2}^{2}\right] \end{aligned},
\end{equation}
i.e., where $C^{\prime} \triangleq \frac{1}{\max_{i}\left(G_{i}\right)}$.
\begin{equation}
    \begin{aligned} & \left(\max_{i}G_{i}\right)\left(2\max_{i}G_{i}\right)\cdot\\ & \sum_{i=1}^{d}\left(\frac{\alpha_{t-1}}{\left(1-\beta_{1}^{t-1}\right)\sqrt{\hat{v}_{i}^{(t-1)}}}-\frac{\alpha_{t}}{\left(1-\beta_{1}^{t}\right)\sqrt{\hat{v}_{i}^{(t)}}}\right)\\ \triangleq & C_{5}\sum_{i=1}^{d}\left(\frac{\alpha_{t-1}}{\left(1-\beta_{1}^{t-1}\right)\sqrt{\hat{v}_{i}^{(t-1)}}}-\frac{\alpha_{t}}{\left(1-\beta_{1}^{t}\right)\sqrt{\hat{v}_{i}^{(t)}}}\right) \end{aligned},
\end{equation}
i.e., where $C_{5} \triangleq \left(\max_{i}G_{i}\right)\left(2\max_{i}G_{i}\right)$.

When $t=1$,
\begin{equation}
    L\frac{\alpha_{t}^{2}}{\left(1-\beta_{1}^{t}\right)^{2}}\sum_{i=1}^{d}G_{i}^{2}/c^{2}\leq L\cdot2\frac{\alpha_{t}^{2}}{\left(1-\beta_{1}^{t}\right)^{2}}\sum_{i=1}^{d}G_{i}^{2}/c^{2}=C_{3}\alpha_{t}^{2},
\end{equation}
\begin{equation}
    \begin{aligned}
        &-\frac{\alpha_{t}}{\left(1-\beta_{1}^{t}\right)\max_{i}\left(G_{i}\right)}\mathbb{E}_{t}\left[\left\Vert \nabla f\left(\boldsymbol{P}^{(t)}\right)\right\Vert _{2}^{2}\right]
        \\ \leq & -C^{\prime}\alpha_{t}\mathbb{E}_{t}\left[\left\Vert \nabla f\left(\boldsymbol{P}^{(t)}\right)\right\Vert _{2}^{2}\right],
    \end{aligned}
\end{equation}
\begin{equation}
    \frac{\alpha_{t}}{1-\beta_{1}^{t}}\left(\max_{i}G_{i}\right)\left(2\max_{i}G_{i}\right)d/c\triangleq C_{6}.
\end{equation}
After these substitutions, Ineq. \ref{right_part_left_part} can be written as:
\begin{equation}
    \begin{aligned} 
    & \mathbb{E}_{T}\left[f\left(\boldsymbol{\xi}^{(T+1)}\right)\right]-\mathbb{E}_{0}\left[f\left(\boldsymbol{\xi}^{(1)}\right)\right]
    \\\leq&\sum_{t=2}^{T}C_{1}\alpha_{t-1}^{2}+\sum_{t=1}^{T}C_{3}\alpha_{t}^{2}-\sum_{t=1}^{T}C^{\prime}\alpha_{t}\mathbb{E}_{t}\left[\left\Vert \nabla f\left(\boldsymbol{P}^{(t)}\right)\right\Vert _{2}^{2}\right]
    \\ & +\sum_{t=2}^{T}\left(C_{2}+C_{4}+C_{5}\right)\sum_{i=1}^{d}\left(\frac{\alpha_{t-1}}{\left(1-\beta_{1}^{t-1}\right)\sqrt{\hat{v}_{i}^{(t-1)}}}\right.
    \\&\left.-\frac{\alpha_{t}}{\left(1-\beta_{1}^{t}\right)\sqrt{\hat{v}_{i}^{(t)}}}\right)+C_{6}
    \\=&\sum_{t=2}^{T}C_{1}\alpha_{t-1}^{2}+\sum_{t=1}^{T}C_{3}\alpha_{t}^{2}-\sum_{t=1}^{T}C^{\prime}\alpha_{t}\mathbb{E}_{t}\left[\left\Vert \nabla f\left(\boldsymbol{P}^{(t)}\right)\right\Vert _{2}^{2}\right]+C_{6}
    \\ & +\sum_{i=1}^{d}\left(C_{2}+C_{4}+C_{5}\right)\sum_{t=2}^{T}\left(\frac{\alpha_{t-1}}{\left(1-\beta_{1}^{t-1}\right)\sqrt{\hat{v}_{i}^{(t-1)}}} \right.
    \\& \left. -\frac{\alpha_{t}}{\left(1-\beta_{1}^{t}\right)\sqrt{\hat{v}_{i}^{(t)}}}\right)
    \\=&\sum_{t=2}^{T}C_{1}\alpha_{t-1}^{2}+\sum_{t=1}^{T}C_{3}\alpha_{t}^{2}-\sum_{t=1}^{T}C^{\prime}\alpha_{t}\mathbb{E}_{ t}\left[\left\Vert \nabla f\left(\boldsymbol{P}^{(t)}\right)\right\Vert _{2}^{2}\right]+C_{6}\\ & +\sum_{i=1}^{d}\left(C_{2}+C_{4}+C_{5}\right)\left(\frac{\alpha_{1}}{\left(1-\beta_{1}\right)\sqrt{\hat{v}_{i}^{(1)}}}-\frac{\alpha_{T}}{\left(1-\beta_{1}^{T}\right)\sqrt{\hat{v}_{i}^{(T)}}}\right)
    \\\leq&\left(C_{1}+C_{3}\right)\sum_{t=1}^{T}\alpha_{t}^{2}- C^{\prime}\sum_{t=1}^{T}\alpha_{t}\mathbb{E}_{t}\left[\left\Vert \nabla f\left(\boldsymbol{P}^{(t)}\right)\right\Vert _{2}^{2}\right]+C_{6}\\ & +\sum_{i=1}^{d}\left(C_{2}+C_{4}+C_{5}\right)\frac{\alpha_{1}}{\left(1-\beta_{1}\right)\sqrt{\hat{v}_{i}^{(1)}}}
    \\\leq&\left(C_{1}+C_{3}\right)\sum_{t=1}^{T}\alpha_{t}^{2}-C^{\prime}\sum_{t=1}^{T}\alpha_{t}\mathbb{E}_{t}\left[\left\Vert \nabla f\left(\boldsymbol{P}^{(t)}\right)\right\Vert _{2}^{2}\right]
    \\&+C_{6}+\left(C_{2}+C_{4}+C_{5}\right)\frac{\alpha_{1}d}{\left(1-\beta_{1}\right)c} \end{aligned}
\end{equation}
Combine the results of the scaling on both sides of the unequal sign:
\begin{equation}
    \begin{aligned} 
    & f\left(\boldsymbol{P}^{\star}\right)-f\left(\boldsymbol{P}^{(1)}\right)
    \\\leq&\left(C_{1}+C_{3}\right)\sum_{t=1}^{T}\alpha_{t}^{2}-C^{\prime}\sum_{t=1}^{T}\alpha_{t}\mathbb{E}_{t}\left[\left\Vert \nabla f\left(\boldsymbol{P}^{(t)}\right)\right\Vert _{2}^{2}\right]\\ & +C_{6}+\left(C_{2}+C_{4}+C_{5}\right)\frac{\alpha_{1}d}{\left(1-\beta_{1}\right) c}
    \\ \Longrightarrow & C^{\prime}\sum_{t=1}^{T}\alpha_{t}\mathbb{E}_{t}\left[\left\Vert \nabla f\left(\boldsymbol{P}^{(t)}\right)\right\Vert _{2}^{2}\right]    \\\leq&\left(C_{1}+C_{3}\right)\sum_{t=1}^{T}\alpha_{t}^{2}+f\left(\boldsymbol{P}^{(1)}\right)-f\left(\boldsymbol{P}^{\star}\right)\\ & +C_{6}+\left(C_{2}+C_{4}+C_{5}\right)\frac{\alpha_{1}d}{\left(1-\beta_{1}\right) c} \end{aligned},
\end{equation}
since $\mathbb{E}_{t}\left[\left\Vert \nabla f\left(\boldsymbol{P}^{(t)}\right)\right\Vert _{2}^{2}\right]=\mathbb{E}_{t-1}\left[\left\Vert \nabla f\left(\boldsymbol{P}^{(t)}\right)\right\Vert _{2}^{2}\right]$,
\begin{equation}
    \begin{aligned}
    &C^{\prime}\sum_{t=1}^{T}\alpha_{t}\mathbb{E}_{t}\left[\left\Vert \nabla f\left(\boldsymbol{P}^{(t)}\right)\right\Vert _{2}^{2}\right]
    \\= & C^{\prime}\sum_{t=1}^{T}\alpha_{t}\mathbb{E}_{t-1}\left[\left\Vert \nabla f\left(\boldsymbol{P}^{(t)}\right)\right\Vert _{2}^{2}\right]\\ \geq & C^{\prime}\sum_{t=1}^{T}\alpha_{t}\min_{t=1,2,\ldots,T}\mathbb{E}_{t-1}\left[\left\Vert \nabla f\left(\boldsymbol{P}^{(t)}\right)\right\Vert _{2}^{2}\right]\\ = & C^{\prime}\min_{t=1,2,\ldots,T}\mathbb{E}_{t-1}\left[\left\Vert \nabla f\left(\boldsymbol{P}^{(t)}\right)\right\Vert _{2}^{2}\right]\sum_{t=1}^{T}\alpha_{t}\\ = & C^{\prime}\cdot E\left(T\right)\cdot\sum_{t=1}^{T}\alpha_{t} \end{aligned}.
\end{equation}
Re-order $C_{1}+C_{3}\triangleq C^{\prime\prime}$ and $\underset{\geq0}{\underbrace{f\left(\boldsymbol{P}^{(1)}\right)-f\left(\boldsymbol{P}^{\star}\right)}}+C_{6}+\left(C_{2}+C_{4}+C_{5}\right)\frac{\alpha_{1}d}{\left(1-\beta_{1}\right)c}\triangleq C^{\prime\prime\prime}$, so
\begin{equation}
    \begin{aligned} & C^{\prime}\cdot E\left(T\right)\cdot\sum_{t=1}^{T}\alpha_{t}\leq C^{\prime\prime}\sum_{t=1}^{T}\alpha_{t}^{2}+C^{\prime\prime\prime}\\ \Longrightarrow & E\left(T\right)\leq\frac{C^{\prime\prime}\sum_{t=1}^{T}\alpha_{t}^{2}+C^{\prime\prime\prime}}{C^{\prime}\sum_{t=1}^{T}\alpha_{t}} \end{aligned}
\end{equation}

\clearpage

\section{Experiments}

\subsection{Mathematical Description of Experimental Metrics}
\label{appendix_mathematical_description_of_metrics}

For digital attacks, we use mean Average Precision (mAP) as the evaluation metric to measure the performance of object detection models in the COCO \cite{lin2014microsoft} dataset.
To calculate mAP, we use the following formula:
\begin{equation}
  \text{mAP} = \frac{1}{N} \sum_{i=1}^{N} \text{AP}_i
\end{equation}
where $N$ is the number of classes and $\text{AP}_i$ is the Average Precision for class $i$.
The Average Precision for a class is calculated as follows:
\begin{equation}
  \text{AP}_i = \sum_{r=1}^{R} \text{Precision}(r) \cdot (\text{Recall}(r)-\text{Recall}(r-1)),
\end{equation}
where $R$ is the number of recall levels. 
$\text{Precision}(r)$ and $\text{Recall}(r)$ are the precision and recall values at recall level $r$.
To calculate the precision and recall values, we use the following formulas:
\begin{equation}
  \text{Precision}(r) = \frac{\text{TP}(r)}{\text{TP}(r) + \text{FP}(r)}
\end{equation}
and 
\begin{equation}
  \text{Recall}(r) = \frac{\text{TP}(r)}{\text{TP}(r) + \text{FN}(r)},
\end{equation}
where $\text{TP}(r)$, $\text{FP}(r)$, and $\text{FN}(r)$ are the number of true positives, false positives, and false negatives at recall level $r$.
For physical attacks, we conduct experiments in real-world scenarios with only one targeted object and only focus on if there are any targeted objects are detected, so we calculate the detection rate (DR) the same as the recall value as follows:
\begin{equation}
  \text{DR} = \frac{\text{TP}(r)}{\text{TP}(r) + \text{FN}(r)}.
\end{equation}

Additionally, we evaluate the attack performance with the attack success rate (ASR) by calculating the drop ratio of the detection performance (evaluated by mAP or DR) as follows:
\begin{equation}
  \text{ASR} = 1 - \frac{\text{DP}_{\text{attack}}}{\text{DP}_{\text{clean}}},
\end{equation}
where DP$_{\text{attack}}$ and DP$_{\text{clean}}$ are the detection performance (DP) on the adversarial examples and clean examples, respectively.

\subsection{Experimental Results of Digital Attacks}
\label{appendix_experimental_results_digital_attacks}

The data presented in Table \ref{table_digital_attack2} illustrates the outcomes of digital background attacks executed against a selection of object detectors on the validation set of the COCO dataset, using the metric of mAP0.5:0.95. 
This table highlights the performance of white-box attacks (bolded entries) alongside black-box attacks, with the intensity of the cell color indicating the severity of the attack's impact on detection performance: red signifies a greater negative effect.

A key finding from the table is that the background attacks significantly degrade the performance of various SOTA detectors, such as YOLOv3, YOLOv5, Mask R-CNN, FreeAnchor, FSAF, etc. 
The introduction of adversarial perturbations, even in the background, causes noticeable reductions in the detectors' accuracy. 
Clean images and those with random noise are used as benchmarks to compare against the effects of background attacks.

The experimental results also reveal that the detectors are more susceptible to degradation when the perturbations are optimized specifically for the background, rather than when they are random or focused on the foreground objects. 
This suggests that background features play a critical role in the detectors' decision-making process, a role that was previously undervalued.

\begin{table*}[!htbp]
  \centering
  \tiny
  \setlength{\tabcolsep}{1.224mm}
  \begin{tabular*}{\hsize}{cccccccccccccccccccc}
  \hline
  & \cellcolor[HTML]{EBF1DE}\rotatebox{75}{SSD}   & \cellcolor[HTML]{EBF1DE}\rotatebox{75}{Faster R-CNN} & \cellcolor[HTML]{EBF1DE}\rotatebox{75}{Swin Transformer} & \cellcolor[HTML]{EBF1DE}\rotatebox{75}{YOLOv3} & \cellcolor[HTML]{EBF1DE}\rotatebox{75}{YOLOv5n} & \cellcolor[HTML]{EBF1DE}\rotatebox{75}{YOLOv5s} & \cellcolor[HTML]{EBF1DE}\rotatebox{75}{YOLOv5m} & \cellcolor[HTML]{EBF1DE}\rotatebox{75}{YOLOv5l} & \cellcolor[HTML]{EBF1DE}\rotatebox{75}{YOLOv5x} & \cellcolor[HTML]{EBF1DE}\rotatebox{75}{Cascade R-CNN} & \cellcolor[HTML]{EBF1DE}\rotatebox{75}{RetinaNet} & \cellcolor[HTML]{EBF1DE}\rotatebox{75}{Mask R-CNN}& \cellcolor[HTML]{EBF1DE}\rotatebox{75}{FreeAnchor} & \cellcolor[HTML]{EBF1DE}\rotatebox{75}{FSAF}  & \cellcolor[HTML]{EBF1DE}\rotatebox{75}{RepPoints} & \cellcolor[HTML]{EBF1DE}\rotatebox{75}{TOOD}  & \cellcolor[HTML]{EBF1DE}\rotatebox{75}{ATSS}  & \cellcolor[HTML]{EBF1DE}\rotatebox{75}{FoveaBox} & \cellcolor[HTML]{EBF1DE}\rotatebox{75}{VarifocalNet} \\ \hline
  \cellcolor[HTML]{FDE9D9}Clean        & \cellcolor[HTML]{FE93A6}0.213 & \cellcolor[HTML]{FED3DB}0.384 & \cellcolor[HTML]{FEEFF2}0.460 & \cellcolor[HTML]{FEF8F9}0.482 & \cellcolor[HTML]{FEACBB}0.280 & \cellcolor[HTML]{FECFD8}0.374 & \cellcolor[HTML]{FEEDF0}0.454 & \cellcolor[HTML]{FEFBFB}0.490 & \cellcolor[HTML]{FCFEFF}0.507 & \cellcolor[HTML]{FEDDE3}0.410 & \cellcolor[HTML]{FECBD5}0.364 & \cellcolor[HTML]{FED2DA}0.381 & \cellcolor[HTML]{FED4DC}0.387 & \cellcolor[HTML]{FECFD8}0.374 & \cellcolor[HTML]{FECED6}0.370 & \cellcolor[HTML]{FEEAEE}0.445 & \cellcolor[HTML]{FED7DE}0.394 & \cellcolor[HTML]{FECED7}0.372 & \cellcolor[HTML]{FEDFE5}0.416 \\
  \cellcolor[HTML]{FDE9D9}Random Noise & \cellcolor[HTML]{FE849A}0.173 & \cellcolor[HTML]{FEBECA}0.329 & \cellcolor[HTML]{FEE1E6}0.422 & \cellcolor[HTML]{FEECF0}0.452 & \cellcolor[HTML]{FEB8C5}0.312 & \cellcolor[HTML]{FEC8D2}0.354 & \cellcolor[HTML]{FEE5EA}0.433 & \cellcolor[HTML]{FEF3F5}0.470 & \cellcolor[HTML]{FCFFFF}0.506 & \cellcolor[HTML]{FEC9D3}0.358 & \cellcolor[HTML]{FEBCC8}0.322 & \cellcolor[HTML]{FEC0CC}0.335 & \cellcolor[HTML]{FEC3CE}0.343 & \cellcolor[HTML]{FEC2CD}0.339 & \cellcolor[HTML]{FEBDC9}0.327 & \cellcolor[HTML]{FEDEE4}0.414 & \cellcolor[HTML]{FECAD4}0.361 & \cellcolor[HTML]{FEC0CB}0.333 & \cellcolor[HTML]{FED3DB}0.384 \\
  \cellcolor[HTML]{FDE9D9}SSD              & \cellcolor[HTML]{FE8097}\textbf{0.164} & \cellcolor[HTML]{FEB3C1}0.299          & \cellcolor[HTML]{FED8DF}0.398          & \cellcolor[HTML]{FEE3E8}0.427          & \cellcolor[HTML]{FEAAB9}0.274          & \cellcolor[HTML]{FEB8C5}0.312          & \cellcolor[HTML]{FECED7}0.371          & \cellcolor[HTML]{FEE8EC}0.440          & \cellcolor[HTML]{FEECF0}0.452          & \cellcolor[HTML]{FEBDC9}0.325          & \cellcolor[HTML]{FEB1BF}0.294          & \cellcolor[HTML]{FEB6C3}0.308          & \cellcolor[HTML]{FEB7C4}0.311          & \cellcolor[HTML]{FEB6C3}0.307          & \cellcolor[HTML]{FEB1BF}0.295          & \cellcolor[HTML]{FED0D8}0.376          & \cellcolor[HTML]{FEBAC6}0.317          & \cellcolor[HTML]{FEB4C1}0.301          & \cellcolor[HTML]{FEC4CE}0.344          \\
  \cellcolor[HTML]{FDE9D9}Faster R-CNN     & \cellcolor[HTML]{FE7E95}0.158          & \cellcolor[HTML]{FEA3B4}\textbf{0.257} & \cellcolor[HTML]{FED3DA}0.383          & \cellcolor[HTML]{FEDFE5}0.417          & \cellcolor[HTML]{FEAABA}0.276          & \cellcolor[HTML]{FEB6C3}0.307          & \cellcolor[HTML]{FEC3CE}0.341          & \cellcolor[HTML]{FEE3E8}0.427          & \cellcolor[HTML]{FEE8EC}0.439          & \cellcolor[HTML]{FEB1BF}0.294          & \cellcolor[HTML]{FEA2B3}0.254          & \cellcolor[HTML]{FEA7B7}0.267          & \cellcolor[HTML]{FEA6B6}0.264          & \cellcolor[HTML]{FEA6B6}0.264          & \cellcolor[HTML]{FEA1B2}0.252          & \cellcolor[HTML]{FEC0CB}0.334          & \cellcolor[HTML]{FEA9B9}0.273          & \cellcolor[HTML]{FEA5B5}0.261          & \cellcolor[HTML]{FEB2C0}0.297          \\
  \cellcolor[HTML]{FDE9D9}Swin Transformer & \cellcolor[HTML]{FE8198}0.166          & \cellcolor[HTML]{FEB7C4}0.310          & \cellcolor[HTML]{FED9E0}\textbf{0.401} & \cellcolor[HTML]{FEE4E9}0.429          & \cellcolor[HTML]{FEB0BE}0.292          & \cellcolor[HTML]{FEBCC8}0.322          & \cellcolor[HTML]{FED6DE}0.393          & \cellcolor[HTML]{FEE7EB}0.437          & \cellcolor[HTML]{FEECEF}0.450          & \cellcolor[HTML]{FEC2CD}0.338          & \cellcolor[HTML]{FEB4C2}0.302          & \cellcolor[HTML]{FEBAC6}0.317          & \cellcolor[HTML]{FEBBC7}0.321          & \cellcolor[HTML]{FEB9C6}0.316          & \cellcolor[HTML]{FEB5C2}0.305          & \cellcolor[HTML]{FED4DC}0.387          & \cellcolor[HTML]{FEBECA}0.329          & \cellcolor[HTML]{FEB8C5}0.313          & \cellcolor[HTML]{FEC9D2}0.357          \\
  \cellcolor[HTML]{FDE9D9}YOLOv3           & \cellcolor[HTML]{FE8298}0.168          & \cellcolor[HTML]{FEB8C5}0.312          & \cellcolor[HTML]{FEDAE0}0.402          & \cellcolor[HTML]{FE9FB0}\textbf{0.245} & \cellcolor[HTML]{FEB0BE}0.292          & \cellcolor[HTML]{FEBAC6}0.317          & \cellcolor[HTML]{FECDD6}0.369          & \cellcolor[HTML]{FEDAE1}0.403          & \cellcolor[HTML]{FEE4E9}0.429          & \cellcolor[HTML]{FEC2CD}0.338          & \cellcolor[HTML]{FEB6C3}0.307          & \cellcolor[HTML]{FEBAC6}0.317          & \cellcolor[HTML]{FEBCC8}0.323          & \cellcolor[HTML]{FEBAC7}0.319          & \cellcolor[HTML]{FEB7C4}0.309          & \cellcolor[HTML]{FED5DD}0.390          & \cellcolor[HTML]{FEC0CB}0.334          & \cellcolor[HTML]{FEB9C6}0.315          & \cellcolor[HTML]{FECAD3}0.360          \\
  \cellcolor[HTML]{FDE9D9}YOLOv5n          & \cellcolor[HTML]{FE8097}0.163          & \cellcolor[HTML]{FEB1BF}0.294          & \cellcolor[HTML]{FED9E0}0.401          & \cellcolor[HTML]{FEE7EB}0.437          & \cellcolor[HTML]{FE738C}\textbf{0.128} & \cellcolor[HTML]{FEB8C5}0.313          & \cellcolor[HTML]{FED1D9}0.379          & \cellcolor[HTML]{FEEEF1}0.457          & \cellcolor[HTML]{FEF5F7}0.476          & \cellcolor[HTML]{FEBAC6}0.318          & \cellcolor[HTML]{FEAEBD}0.287          & \cellcolor[HTML]{FEB4C1}0.301          & \cellcolor[HTML]{FEB5C2}0.304          & \cellcolor[HTML]{FEB4C1}0.301          & \cellcolor[HTML]{FEB0BE}0.290          & \cellcolor[HTML]{FECCD5}0.366          & \cellcolor[HTML]{FEB7C4}0.309          & \cellcolor[HTML]{FEB1BF}0.294          & \cellcolor[HTML]{FEC0CB}0.334          \\
  \cellcolor[HTML]{FDE9D9}YOLOv5s          & \cellcolor[HTML]{FE8097}0.163          & \cellcolor[HTML]{FEB5C2}0.305          & \cellcolor[HTML]{FEDCE2}0.408          & \cellcolor[HTML]{FEE3E8}0.428          & \cellcolor[HTML]{FEADBC}0.284          & \cellcolor[HTML]{FE7E95}\textbf{0.157} & \cellcolor[HTML]{FEC9D2}0.357          & \cellcolor[HTML]{FEDFE4}0.415          & \cellcolor[HTML]{FEE6EA}0.435          & \cellcolor[HTML]{FEBFCA}0.330          & \cellcolor[HTML]{FEB2C0}0.297          & \cellcolor[HTML]{FEB8C5}0.312          & \cellcolor[HTML]{FEB9C6}0.315          & \cellcolor[HTML]{FEB8C5}0.312          & \cellcolor[HTML]{FEB4C2}0.302          & \cellcolor[HTML]{FED1D9}0.378          & \cellcolor[HTML]{FEBCC8}0.322          & \cellcolor[HTML]{FEB6C3}0.306          & \cellcolor[HTML]{FEC6D0}0.350          \\
  \cellcolor[HTML]{FDE9D9}YOLOv5m          & \cellcolor[HTML]{FE8198}0.167          & \cellcolor[HTML]{FEB7C4}0.310          & \cellcolor[HTML]{FEDCE2}0.409          & \cellcolor[HTML]{FEE3E8}0.428          & \cellcolor[HTML]{FEB0BE}0.292          & \cellcolor[HTML]{FEB6C3}0.308          & \cellcolor[HTML]{FE889E}\textbf{0.186} & \cellcolor[HTML]{FED5DD}0.390          & \cellcolor[HTML]{FEDCE2}0.408          & \cellcolor[HTML]{FEC1CC}0.336          & \cellcolor[HTML]{FEB5C2}0.304          & \cellcolor[HTML]{FEBAC6}0.317          & \cellcolor[HTML]{FEBBC7}0.320          & \cellcolor[HTML]{FEBAC6}0.317          & \cellcolor[HTML]{FEB6C3}0.308          & \cellcolor[HTML]{FED4DB}0.386          & \cellcolor[HTML]{FEBFCA}0.331          & \cellcolor[HTML]{FEB8C5}0.312          & \cellcolor[HTML]{FEC8D2}0.356          \\
  \cellcolor[HTML]{FDE9D9}YOLOv5l          & \cellcolor[HTML]{FE8298}0.168          & \cellcolor[HTML]{FEB6C3}0.308          & \cellcolor[HTML]{FEDDE3}0.410          & \cellcolor[HTML]{FEE4E9}0.429          & \cellcolor[HTML]{FEB0BE}0.291          & \cellcolor[HTML]{FEB3C1}0.300          & \cellcolor[HTML]{FEC8D2}0.354          & \cellcolor[HTML]{FE889D}\textbf{0.185} & \cellcolor[HTML]{FED9E0}0.400          & \cellcolor[HTML]{FEC1CC}0.336          & \cellcolor[HTML]{FEB3C1}0.300          & \cellcolor[HTML]{FEBAC6}0.317          & \cellcolor[HTML]{FEBBC7}0.320          & \cellcolor[HTML]{FEB9C6}0.316          & \cellcolor[HTML]{FEB6C3}0.306          & \cellcolor[HTML]{FED3DB}0.385          & \cellcolor[HTML]{FEBFCA}0.330          & \cellcolor[HTML]{FEB8C5}0.312          & \cellcolor[HTML]{FEC8D2}0.356          \\
  \cellcolor[HTML]{FDE9D9}YOLOv5x          & \cellcolor[HTML]{FE8198}0.167          & \cellcolor[HTML]{FEB7C4}0.311          & \cellcolor[HTML]{FEDDE3}0.410          & \cellcolor[HTML]{FEE3E8}0.426          & \cellcolor[HTML]{FEB1BF}0.294          & \cellcolor[HTML]{FEB8C5}0.312          & \cellcolor[HTML]{FECBD4}0.362          & \cellcolor[HTML]{FED2DA}0.382          & \cellcolor[HTML]{FE90A4}\textbf{0.206} & \cellcolor[HTML]{FEC2CD}0.338          & \cellcolor[HTML]{FEB5C2}0.305          & \cellcolor[HTML]{FEBAC7}0.319          & \cellcolor[HTML]{FEBCC8}0.323          & \cellcolor[HTML]{FEBBC7}0.320          & \cellcolor[HTML]{FEB7C4}0.310          & \cellcolor[HTML]{FED5DD}0.390          & \cellcolor[HTML]{FEC0CB}0.334          & \cellcolor[HTML]{FEB9C5}0.314          & \cellcolor[HTML]{FECAD4}0.361          \\
  \cellcolor[HTML]{FDE9D9}Cascade R-CNN    & \cellcolor[HTML]{FE7E95}0.159          & \cellcolor[HTML]{FE9FB0}0.246          & \cellcolor[HTML]{FECFD8}0.374          & \cellcolor[HTML]{FEDDE3}0.411          & \cellcolor[HTML]{FEAABA}0.276          & \cellcolor[HTML]{FEB5C2}0.305          & \cellcolor[HTML]{FEC5D0}0.348          & \cellcolor[HTML]{FEE0E6}0.420          & \cellcolor[HTML]{FEDFE5}0.416          & \cellcolor[HTML]{FEA8B8}\textbf{0.271} & \cellcolor[HTML]{FE9EAF}0.243          & \cellcolor[HTML]{FEA2B2}0.253          & \cellcolor[HTML]{FE9FB0}0.245          & \cellcolor[HTML]{FEA1B2}0.251          & \cellcolor[HTML]{FE9AAC}0.232          & \cellcolor[HTML]{FEBBC7}0.321          & \cellcolor[HTML]{FEA2B3}0.255          & \cellcolor[HTML]{FE9FB1}0.247          & \cellcolor[HTML]{FEAAB9}0.275          \\
  \cellcolor[HTML]{FDE9D9}RetinaNet        & \cellcolor[HTML]{FE7E95}0.159          & \cellcolor[HTML]{FEA4B4}0.259          & \cellcolor[HTML]{FED2DA}0.382          & \cellcolor[HTML]{FEE0E6}0.420          & \cellcolor[HTML]{FEA9B9}0.273          & \cellcolor[HTML]{FEB7C4}0.309          & \cellcolor[HTML]{FEC5D0}0.348          & \cellcolor[HTML]{FEE6EA}0.434          & \cellcolor[HTML]{FEE7EB}0.438          & \cellcolor[HTML]{FEB2C0}0.297          & \cellcolor[HTML]{FEA2B2}\textbf{0.253} & \cellcolor[HTML]{FEA7B7}0.268          & \cellcolor[HTML]{FEA6B6}0.265          & \cellcolor[HTML]{FEA7B6}0.266          & \cellcolor[HTML]{FEA2B2}0.253          & \cellcolor[HTML]{FEBFCA}0.330          & \cellcolor[HTML]{FEA8B8}0.270          & \cellcolor[HTML]{FEA5B6}0.263          & \cellcolor[HTML]{FEB3C0}0.298          \\
  \cellcolor[HTML]{FDE9D9}Mask R-CNN       & \cellcolor[HTML]{FE7E95}0.158          & \cellcolor[HTML]{FEA4B4}0.259          & \cellcolor[HTML]{FED1D9}0.379          & \cellcolor[HTML]{FEDEE4}0.414          & \cellcolor[HTML]{FEA9B8}0.272          & \cellcolor[HTML]{FEB6C3}0.306          & \cellcolor[HTML]{FEC4CE}0.344          & \cellcolor[HTML]{FEE2E7}0.425          & \cellcolor[HTML]{FEE5E9}0.431          & \cellcolor[HTML]{FEB0BE}0.292          & \cellcolor[HTML]{FEA3B3}0.256          & \cellcolor[HTML]{FEA4B4}\textbf{0.259} & \cellcolor[HTML]{FEA6B6}0.264          & \cellcolor[HTML]{FEA4B5}0.260          & \cellcolor[HTML]{FEA0B1}0.249          & \cellcolor[HTML]{FEC1CC}0.336          & \cellcolor[HTML]{FEA9B9}0.273          & \cellcolor[HTML]{FEA4B4}0.258          & \cellcolor[HTML]{FEB0BE}0.292          \\
  \cellcolor[HTML]{FDE9D9}FreeAnchor       & \cellcolor[HTML]{FE7F96}0.161          & \cellcolor[HTML]{FEA8B7}0.269          & \cellcolor[HTML]{FED3DB}0.384          & \cellcolor[HTML]{FEDFE5}0.416          & \cellcolor[HTML]{FEABBA}0.278          & \cellcolor[HTML]{FEB5C2}0.304          & \cellcolor[HTML]{FEC5CF}0.346          & \cellcolor[HTML]{FEE4E9}0.429          & \cellcolor[HTML]{FEE5EA}0.433          & \cellcolor[HTML]{FEB5C2}0.304          & \cellcolor[HTML]{FEA8B8}0.270          & \cellcolor[HTML]{FEABBA}0.278          & \cellcolor[HTML]{FEA9B8}\textbf{0.272} & \cellcolor[HTML]{FEAAB9}0.274          & \cellcolor[HTML]{FEA6B6}0.264          & \cellcolor[HTML]{FEC2CD}0.340          & \cellcolor[HTML]{FEACBB}0.281          & \cellcolor[HTML]{FEA9B9}0.273          & \cellcolor[HTML]{FEB8C5}0.312          \\
  \cellcolor[HTML]{FDE9D9}FSAF             & \cellcolor[HTML]{FE7E95}0.159          & \cellcolor[HTML]{FEA4B5}0.260          & \cellcolor[HTML]{FED1D9}0.379          & \cellcolor[HTML]{FEE0E6}0.419          & \cellcolor[HTML]{FEAAB9}0.274          & \cellcolor[HTML]{FEB5C2}0.304          & \cellcolor[HTML]{FEC3CE}0.343          & \cellcolor[HTML]{FEE1E6}0.422          & \cellcolor[HTML]{FEE4E9}0.430          & \cellcolor[HTML]{FEB0BE}0.291          & \cellcolor[HTML]{FEA2B3}0.254          & \cellcolor[HTML]{FEA6B6}0.265          & \cellcolor[HTML]{FEA6B6}0.264          & \cellcolor[HTML]{FEA4B4}\textbf{0.259} & \cellcolor[HTML]{FEA2B2}0.253          & \cellcolor[HTML]{FEBFCA}0.330          & \cellcolor[HTML]{FEA9B8}0.272          & \cellcolor[HTML]{FEA4B4}0.258          & \cellcolor[HTML]{FEB2C0}0.297          \\
  \cellcolor[HTML]{FDE9D9}RepPoints        & \cellcolor[HTML]{FE7F96}0.160          & \cellcolor[HTML]{FEA9B8}0.272          & \cellcolor[HTML]{FED4DC}0.387          & \cellcolor[HTML]{FEE1E6}0.421          & \cellcolor[HTML]{FEABBA}0.277          & \cellcolor[HTML]{FEB7C4}0.310          & \cellcolor[HTML]{FEC7D1}0.352          & \cellcolor[HTML]{FEE6EB}0.436          & \cellcolor[HTML]{FEEBEE}0.448          & \cellcolor[HTML]{FEB5C2}0.304          & \cellcolor[HTML]{FEA8B7}0.269          & \cellcolor[HTML]{FEADBB}0.282          & \cellcolor[HTML]{FEABBA}0.278          & \cellcolor[HTML]{FEAABA}0.276          & \cellcolor[HTML]{FEA4B5}\textbf{0.260} & \cellcolor[HTML]{FEC3CE}0.342          & \cellcolor[HTML]{FEACBB}0.280          & \cellcolor[HTML]{FEABBA}0.277          & \cellcolor[HTML]{FEB8C5}0.313          \\
  \cellcolor[HTML]{FDE9D9}TOOD             & \cellcolor[HTML]{FE7F96}0.160          & \cellcolor[HTML]{FEAFBD}0.288          & \cellcolor[HTML]{FED7DE}0.396          & \cellcolor[HTML]{FEE3E8}0.426          & \cellcolor[HTML]{FEADBC}0.283          & \cellcolor[HTML]{FEB9C5}0.314          & \cellcolor[HTML]{FECDD6}0.368          & \cellcolor[HTML]{FEEAEE}0.446          & \cellcolor[HTML]{FEF1F3}0.463          & \cellcolor[HTML]{FEB9C6}0.316          & \cellcolor[HTML]{FEADBB}0.282          & \cellcolor[HTML]{FEB2C0}0.296          & \cellcolor[HTML]{FEB2C0}0.296          & \cellcolor[HTML]{FEB0BE}0.291          & \cellcolor[HTML]{FEACBB}0.280          & \cellcolor[HTML]{FEC9D2}\textbf{0.357} & \cellcolor[HTML]{FEB3C1}0.299          & \cellcolor[HTML]{FEAFBE}0.289          & \cellcolor[HTML]{FEBECA}0.328          \\
  \cellcolor[HTML]{FDE9D9}ATSS             & \cellcolor[HTML]{FE7E95}0.159          & \cellcolor[HTML]{FEACBB}0.280          & \cellcolor[HTML]{FED6DD}0.391          & \cellcolor[HTML]{FEE0E5}0.418          & \cellcolor[HTML]{FEABBA}0.278          & \cellcolor[HTML]{FEB6C3}0.308          & \cellcolor[HTML]{FEC9D3}0.359          & \cellcolor[HTML]{FEE6EA}0.435          & \cellcolor[HTML]{FEECEF}0.451          & \cellcolor[HTML]{FEB6C3}0.308          & \cellcolor[HTML]{FEAAB9}0.275          & \cellcolor[HTML]{FEAEBC}0.285          & \cellcolor[HTML]{FEADBC}0.284          & \cellcolor[HTML]{FEABBA}0.279          & \cellcolor[HTML]{FEA8B8}0.270          & \cellcolor[HTML]{FEC5CF}0.346          & \cellcolor[HTML]{FEAEBC}\textbf{0.285} & \cellcolor[HTML]{FEABBA}0.277          & \cellcolor[HTML]{FEB9C6}0.316          \\
  \cellcolor[HTML]{FDE9D9}FoveaBox         & \cellcolor[HTML]{FE7F96}0.160          & \cellcolor[HTML]{FEAAB9}0.275          & \cellcolor[HTML]{FED4DC}0.388          & \cellcolor[HTML]{FEE0E6}0.419          & \cellcolor[HTML]{FEABBA}0.279          & \cellcolor[HTML]{FEB5C2}0.304          & \cellcolor[HTML]{FEC5CF}0.346          & \cellcolor[HTML]{FEE4E9}0.429          & \cellcolor[HTML]{FEE5EA}0.432          & \cellcolor[HTML]{FEB6C3}0.306          & \cellcolor[HTML]{FEA8B8}0.271          & \cellcolor[HTML]{FEADBC}0.283          & \cellcolor[HTML]{FEABBA}0.278          & \cellcolor[HTML]{FEABBA}0.277          & \cellcolor[HTML]{FEA7B7}0.267          & \cellcolor[HTML]{FEC2CD}0.340          & \cellcolor[HTML]{FEACBB}0.281          & \cellcolor[HTML]{FEA7B7}\textbf{0.268} & \cellcolor[HTML]{FEB8C5}0.313          \\
  \cellcolor[HTML]{FDE9D9}VarifocalNet     & \cellcolor[HTML]{FE7E95}0.157          & \cellcolor[HTML]{FEA6B6}0.265          & \cellcolor[HTML]{FED4DB}0.386          & \cellcolor[HTML]{FEE2E7}0.424          & \cellcolor[HTML]{FEABBA}0.278          & \cellcolor[HTML]{FEB7C4}0.311          & \cellcolor[HTML]{FEC3CE}0.343          & \cellcolor[HTML]{FEE6EA}0.434          & \cellcolor[HTML]{FEEBEE}0.447          & \cellcolor[HTML]{FEB2C0}0.296          & \cellcolor[HTML]{FEA4B4}0.258          & \cellcolor[HTML]{FEA8B8}0.270          & \cellcolor[HTML]{FEA8B7}0.269          & \cellcolor[HTML]{FEA7B6}0.266          & \cellcolor[HTML]{FEA2B3}0.254          & \cellcolor[HTML]{FEBFCA}0.330          & \cellcolor[HTML]{FEA8B8}0.271          & \cellcolor[HTML]{FEA6B6}0.265          & \cellcolor[HTML]{FEAFBD}\textbf{0.288}
   \\  
  \hline
  \end{tabular*}
  \caption{
  Experimental results of digital background attack on the validation set of COCO in terms of mAP0.5:0.95, where white-box attacks are highlighted in bold and the rest are black-box attacks. 
  The \textbf{redder} the cell, the \textbf{worse} the \textbf{detection performance}.
  Clean and Random Noise mean experiments on clean images and images with random noise, respectively.
The 19 detectors of the first row and the first column are for detection and perturbation optimization, respectively.
  }
  \label{table_digital_attack2}
  \end{table*}


\subsection{Attack Comparison of Object Detection}
\label{appendix_attack_comparison_other_methods}

\begin{figure*}[!htbp]
  \centering
  \includegraphics[width=0.99\linewidth]{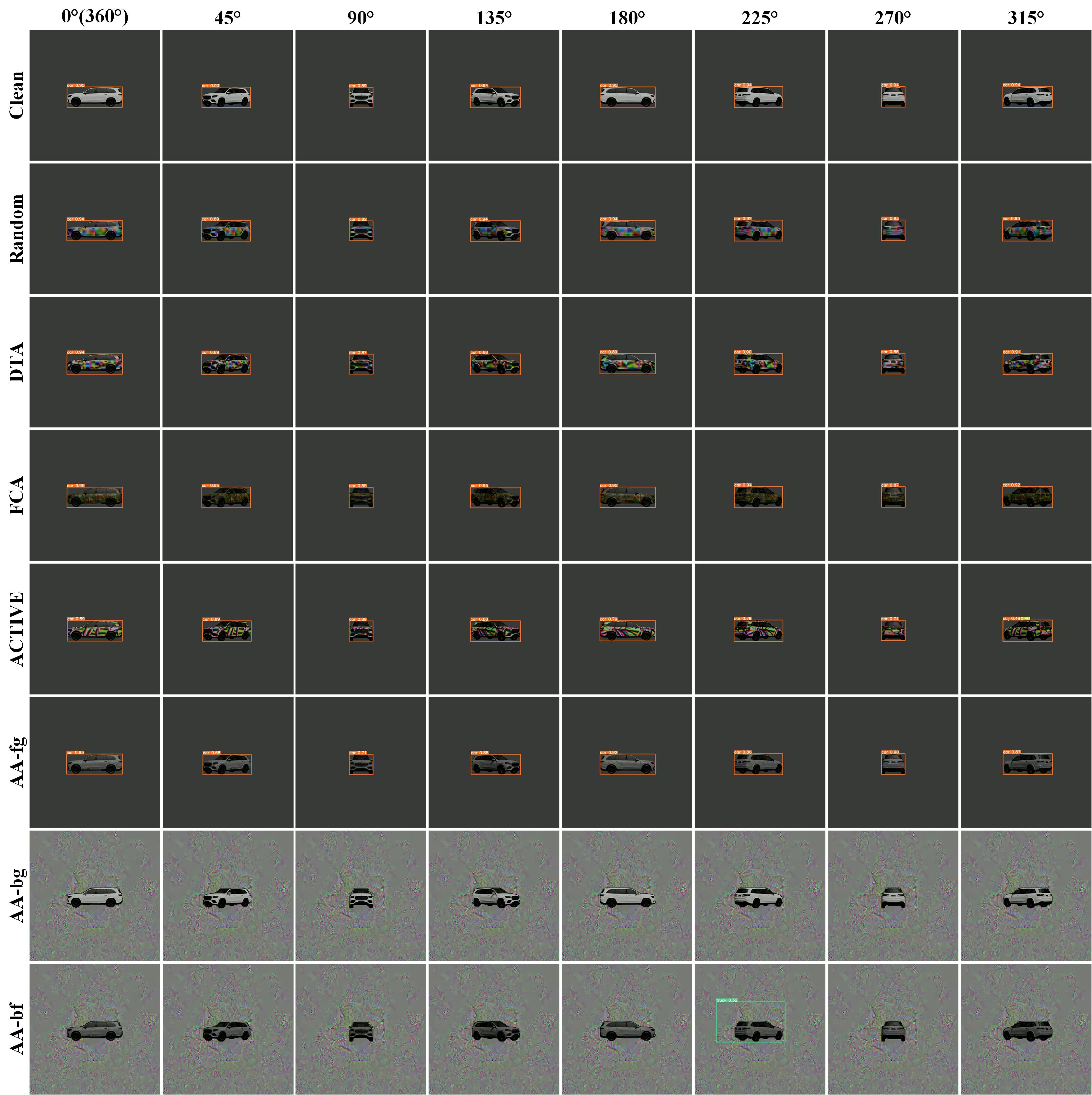} 
  \caption{Qualitative attack comparison of car detection in physically-based simulation and the victim model is YOLOv3. Please zoom in for better visualization.}
  \label{fig_attack_comparison_yolov3_car}
\end{figure*}

\begin{figure*}[!htbp]
  \centering
  \includegraphics[width=0.99\linewidth]{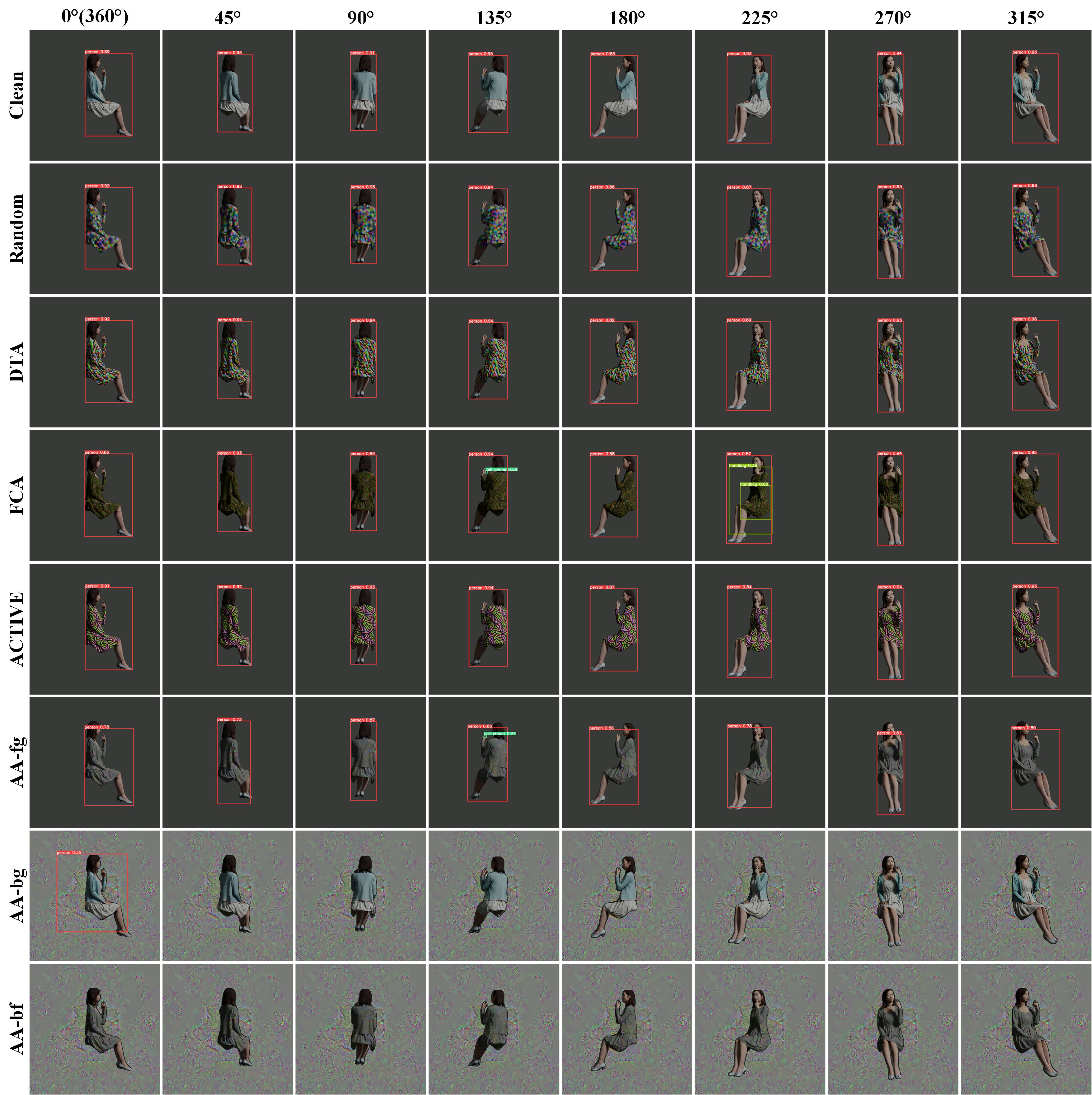} 
  \caption{Qualitative attack comparison of person detection in physically-based simulation and the victim model is YOLOv3. Please zoom in for better visualization.}
  \label{fig_attack_comparison_yolov3_person}
\end{figure*}

\begin{figure}[!htbp]
  \centering
  \begin{subfigure}{0.99\linewidth}
    \includegraphics[width=1\linewidth]{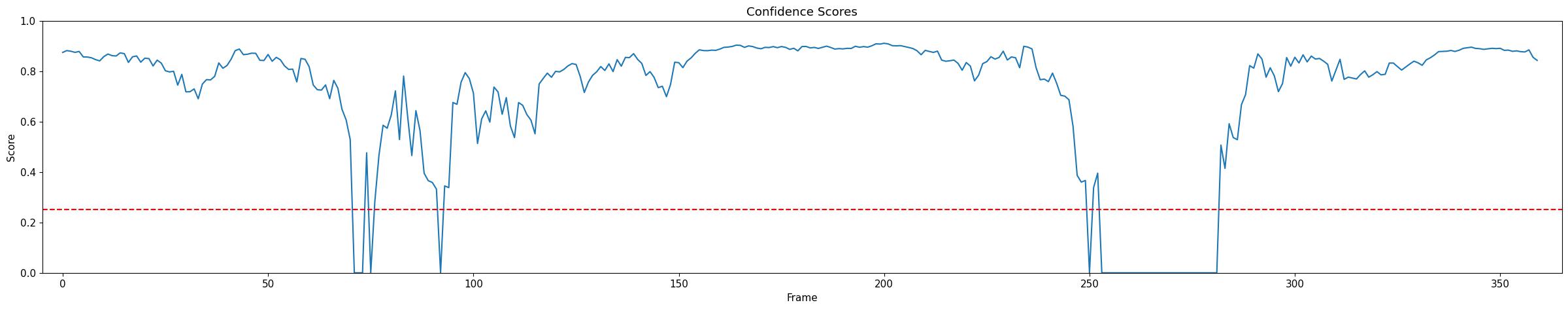}
    \caption{Clean}
  \end{subfigure}
  \begin{subfigure}{0.99\linewidth}
    \includegraphics[width=1\linewidth]{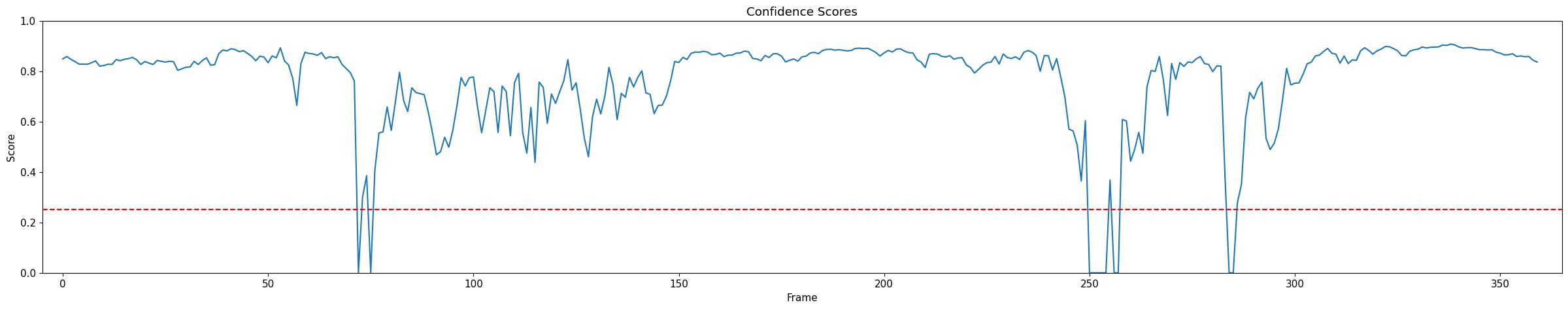}
    \caption{Random}
  \end{subfigure}
  \begin{subfigure}{0.99\linewidth}
    \includegraphics[width=1\linewidth]{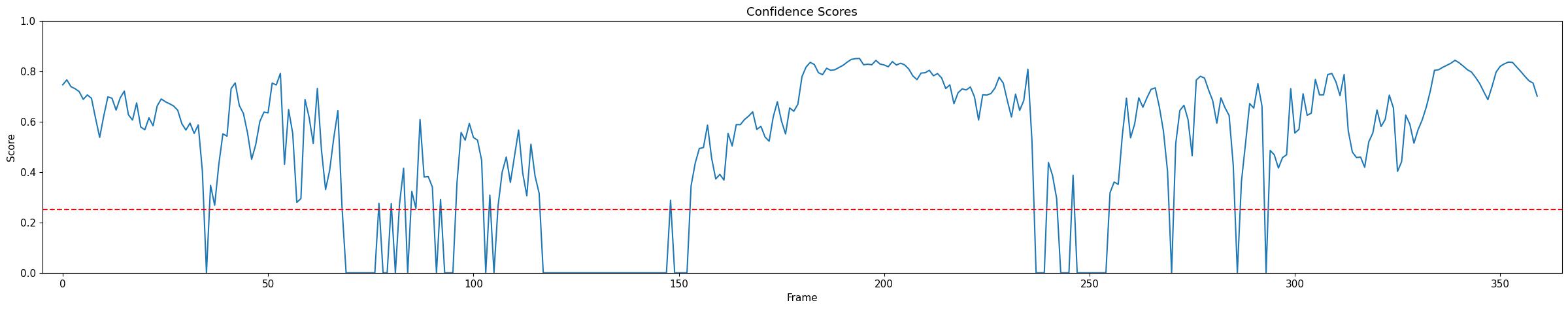}
    \caption{DTA}
  \end{subfigure}
  \begin{subfigure}{0.99\linewidth}
    \includegraphics[width=1\linewidth]{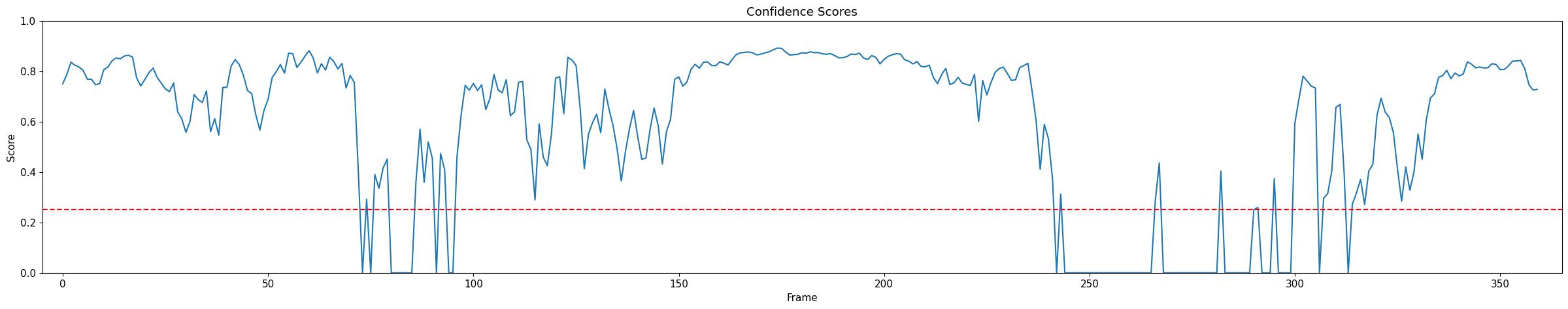}
    \caption{FCA}
  \end{subfigure}
  \begin{subfigure}{0.99\linewidth}
      \includegraphics[width=1\linewidth]{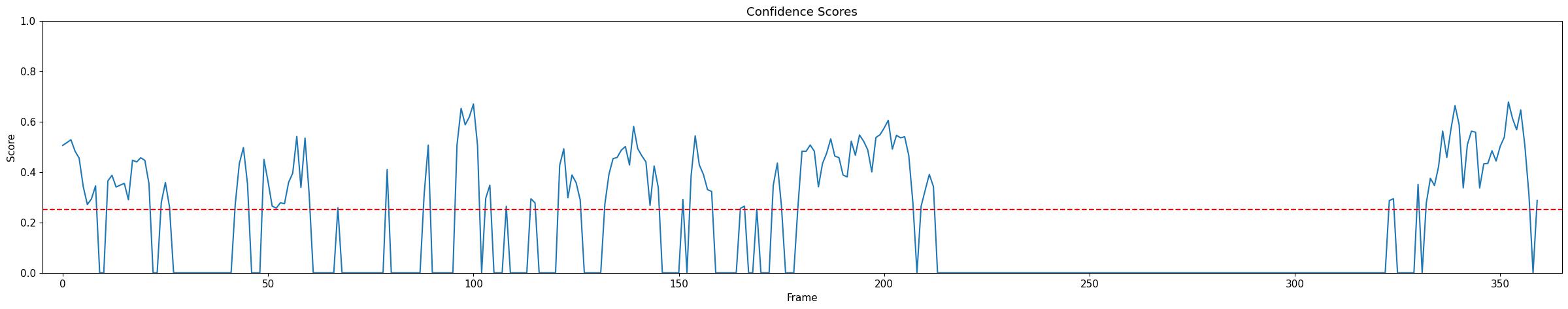}
      \caption{ACTIVE}
    \end{subfigure}
    \begin{subfigure}{0.99\linewidth}
      \includegraphics[width=1\linewidth]{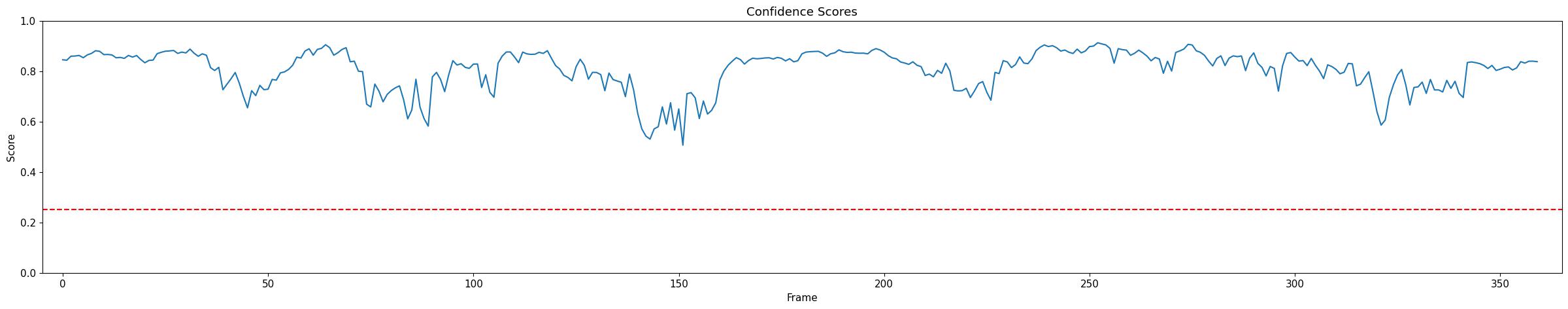}
      \caption{AA-fg}
    \end{subfigure}
    \begin{subfigure}{0.99\linewidth}
      \includegraphics[width=1\linewidth]{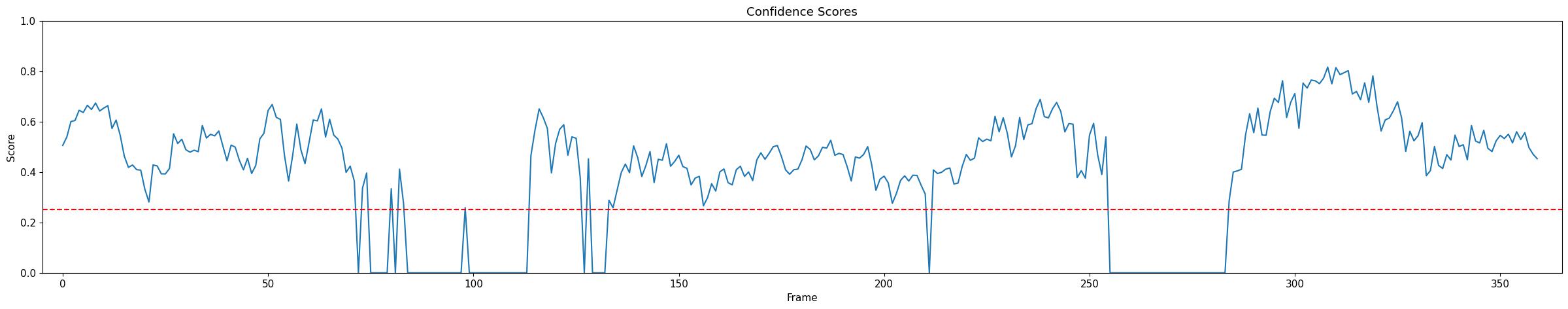}
      \caption{AA-bg}
    \end{subfigure}
    \begin{subfigure}{0.99\linewidth}
      \includegraphics[width=1\linewidth]{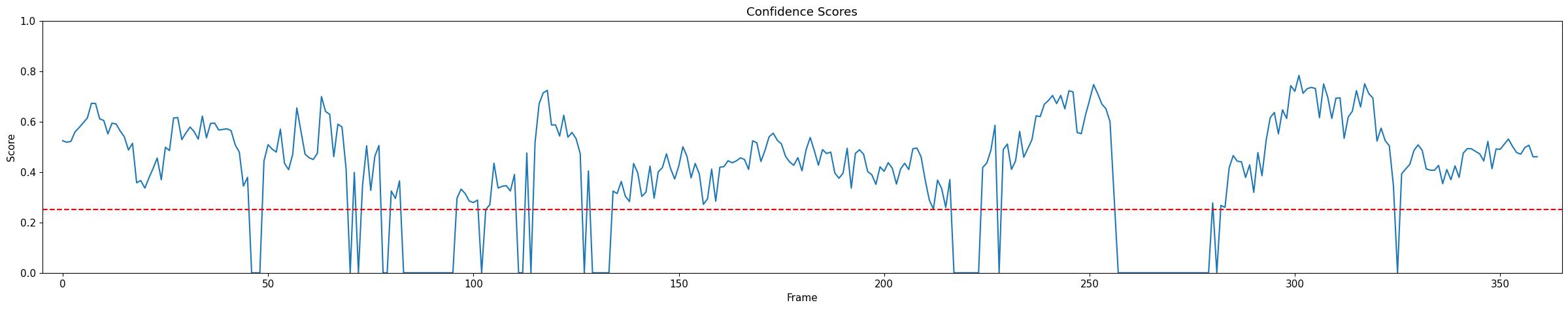}
      \caption{AA-bf}
    \end{subfigure}
  \caption{The graph illustrates a comparison of confidence scores (depicted by the blue line) for a car within a physically-based simulation, utilizing YOLOv3 as the victim model. A confidence threshold, represented by the red dashed line, is established at 0.25. This implies that any confidence score below 0.25 is set as 0 and interpreted as a failure to detect anything. Please zoom in for a better view.}
  \label{fig_scores_line_yolov5s_car}
\end{figure}

\begin{figure}[!htbp]
  \centering
  \begin{subfigure}{0.99\linewidth}
    \includegraphics[width=1\linewidth]{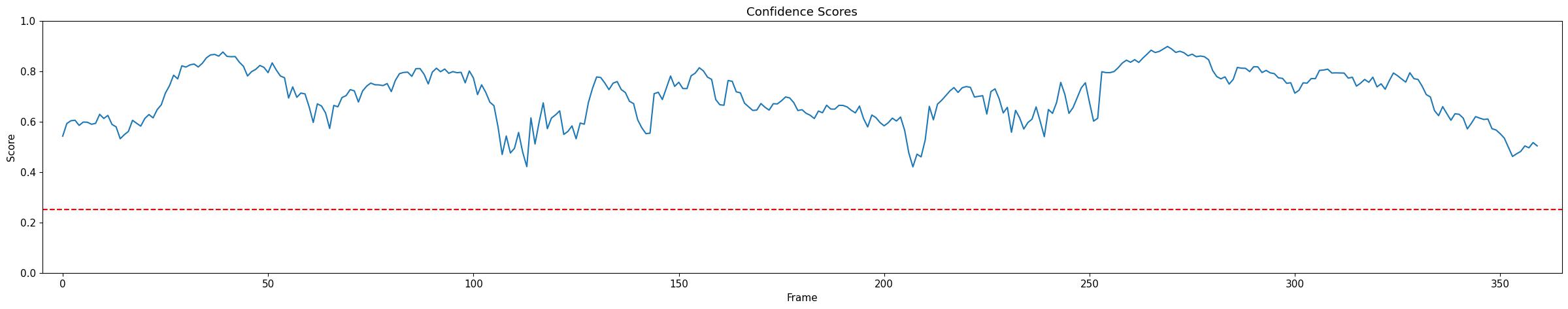}
    \caption{Clean}
  \end{subfigure}
  \begin{subfigure}{0.99\linewidth}
    \includegraphics[width=1\linewidth]{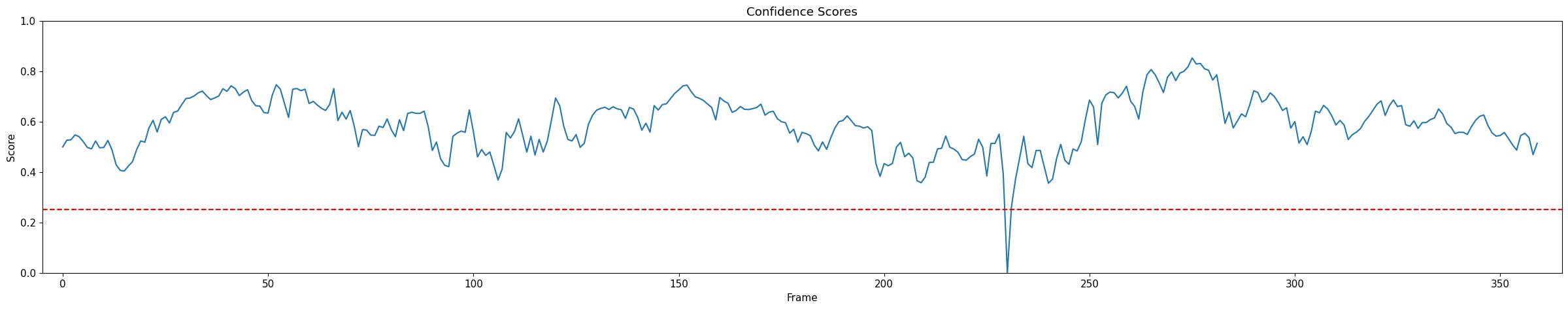}
    \caption{Random}
  \end{subfigure}
  \begin{subfigure}{0.99\linewidth}
    \includegraphics[width=1\linewidth]{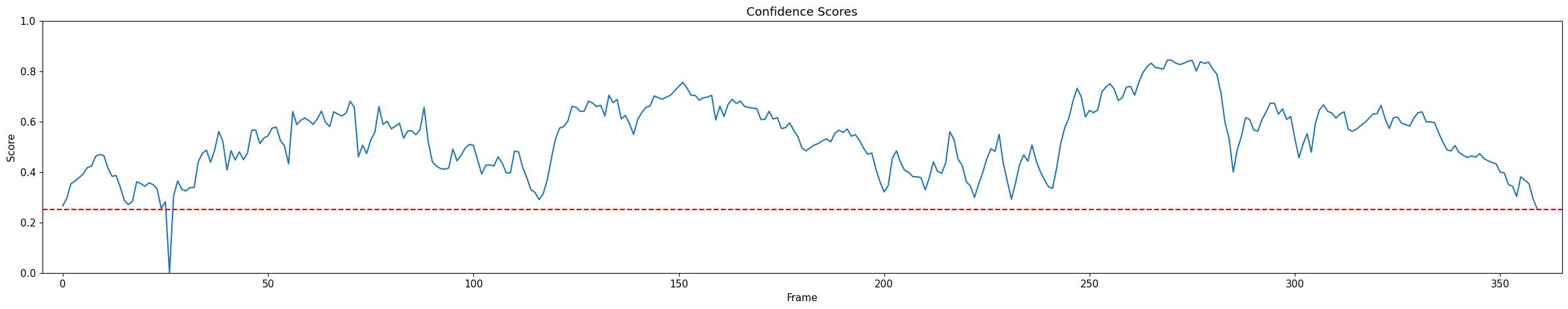}
    \caption{DTA}
  \end{subfigure}
  \begin{subfigure}{0.99\linewidth}
    \includegraphics[width=1\linewidth]{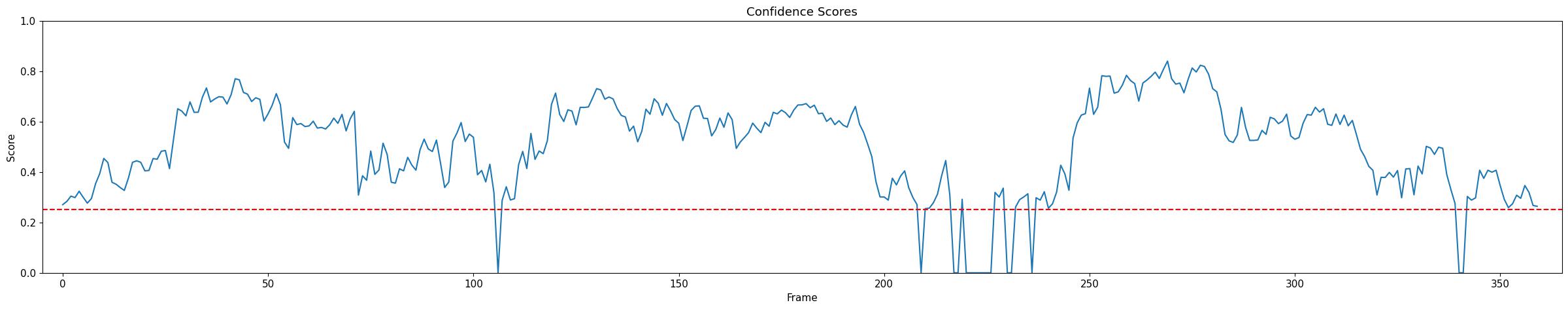}
    \caption{FCA}
  \end{subfigure}
  \begin{subfigure}{0.99\linewidth}
      \includegraphics[width=1\linewidth]{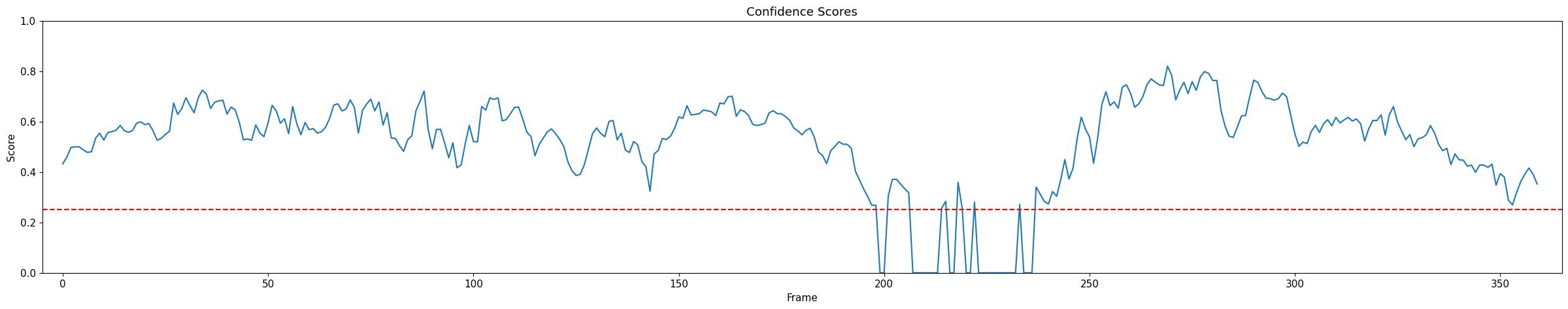}
      \caption{ACTIVE}
    \end{subfigure}
    \begin{subfigure}{0.99\linewidth}
      \includegraphics[width=1\linewidth]{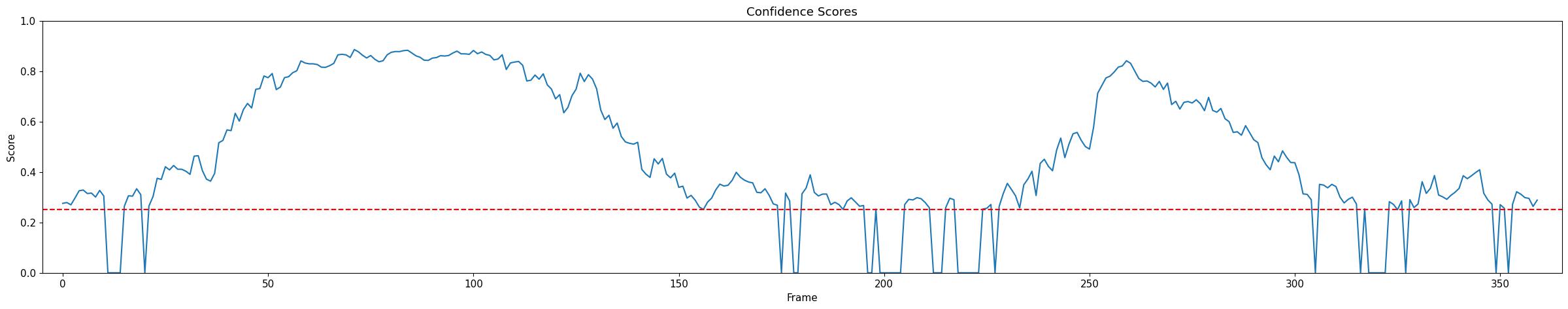}
      \caption{AA-fg}
    \end{subfigure}
    \begin{subfigure}{0.99\linewidth}
      \includegraphics[width=1\linewidth]{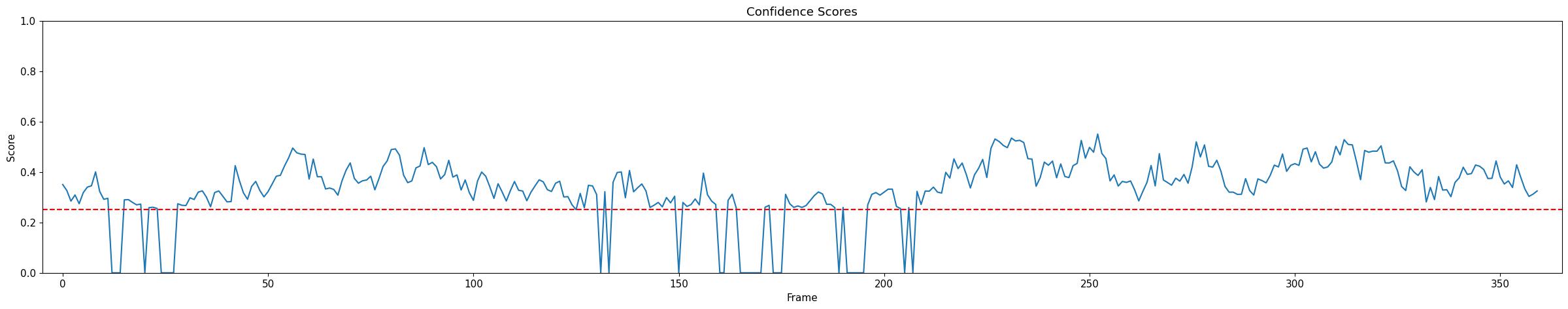}
      \caption{AA-bg}
    \end{subfigure}
    \begin{subfigure}{0.99\linewidth}
      \includegraphics[width=1\linewidth]{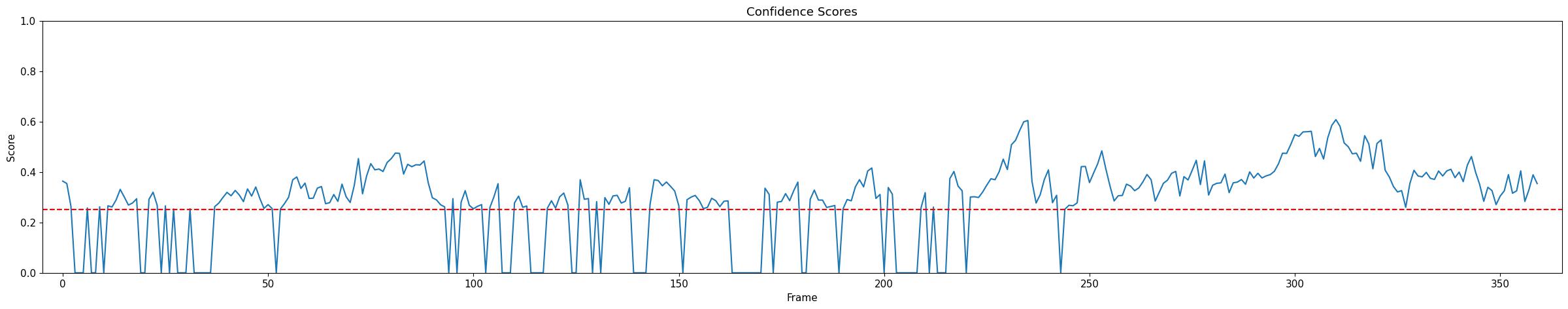}
      \caption{AA-bf}
    \end{subfigure}
  \caption{The graph illustrates a comparison of confidence scores (depicted by the blue line) for a person within a physically-based simulation, utilizing YOLOv3 as the victim model. A confidence threshold, represented by the red dashed line, is established at 0.25. This implies that any confidence score below 0.25 is set as 0 and interpreted as a failure to detect anything. Please zoom in for a better view.}
  \label{fig_scores_line_yolov5s_person}
\end{figure}

Qualitative comparisons shown in Fig. \ref{fig_attack_comparison_yolov3_car}, \ref{fig_attack_comparison_yolov3_person}, \ref{fig_scores_line_yolov5s_car}, \ref{fig_scores_line_yolov5s_person}, \ref{fig_scores_line_yolov3_car}, and \ref{fig_scores_line_yolov3_person} further corroborate those quantitative experimental results, demonstrating that background attacks can successfully conceal objects in physically-based simulations, such as cars and people, when using the YOLOv3 model as the victim. 
These figures provide visual confirmation of the effectiveness of the background attack strategy, showing that it can hide objects from detection by manipulating the scene's background.

Overall, the data and figures indicate that background features are essential components for object detectors and that their manipulation can lead to significant drops in detection performance. 
This underscores the need for more robust detector designs that can withstand adversarial attacks targeting the background, highlighting a critical area for further research and development in the field of computer vision.

\begin{figure*}[!htbp]
  \centering
  \includegraphics[width=0.99\linewidth]{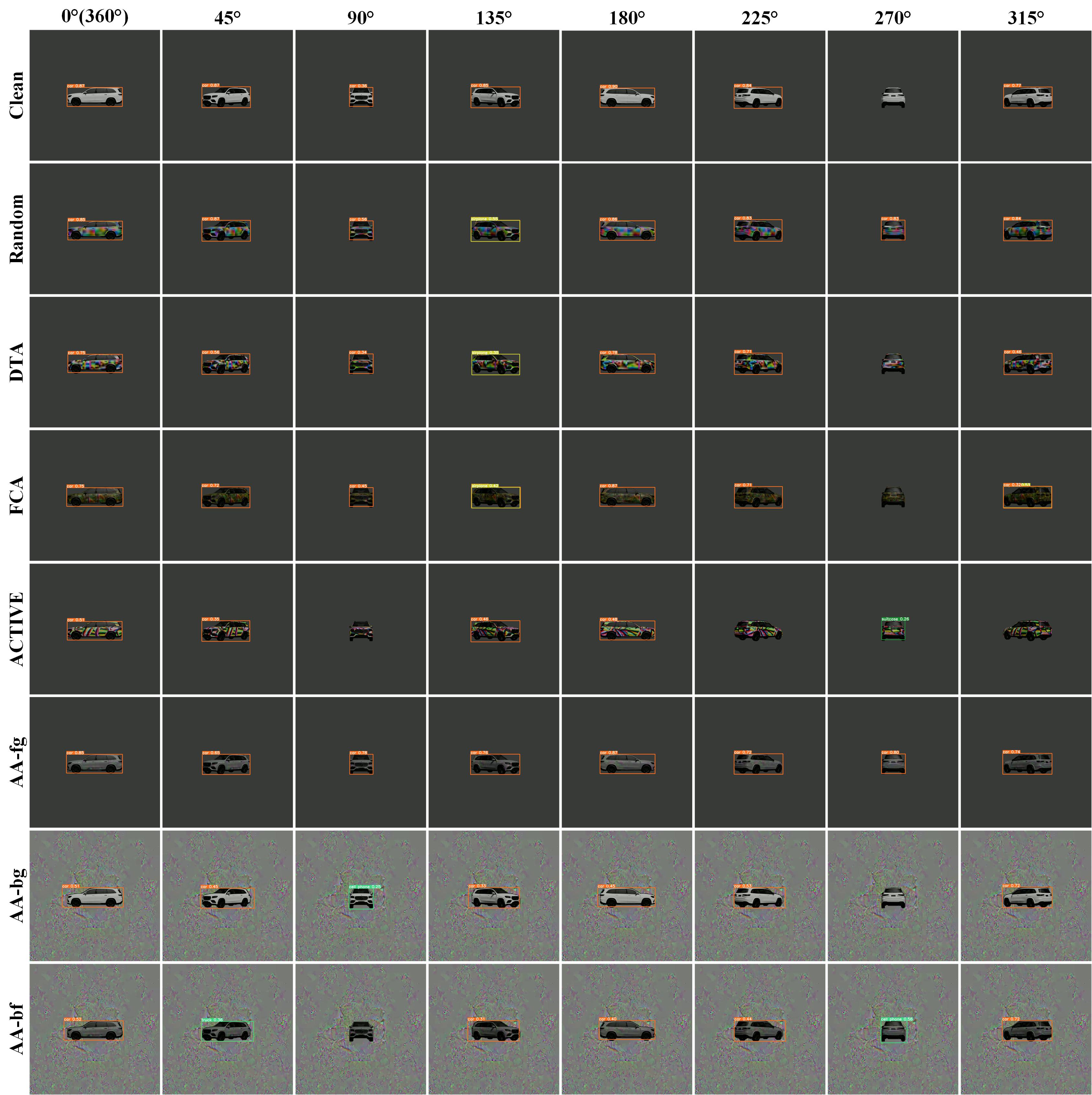} 
  \caption{Qualitative attack comparison of car detection in physically-based simulation and the victim model is YOLOv5s. Please zoom in for better visualization.}
  \label{fig_attack_comparison_yolov5s_car}
\end{figure*}

\begin{figure*}[!htbp]
  \centering
  \includegraphics[width=0.99\linewidth]{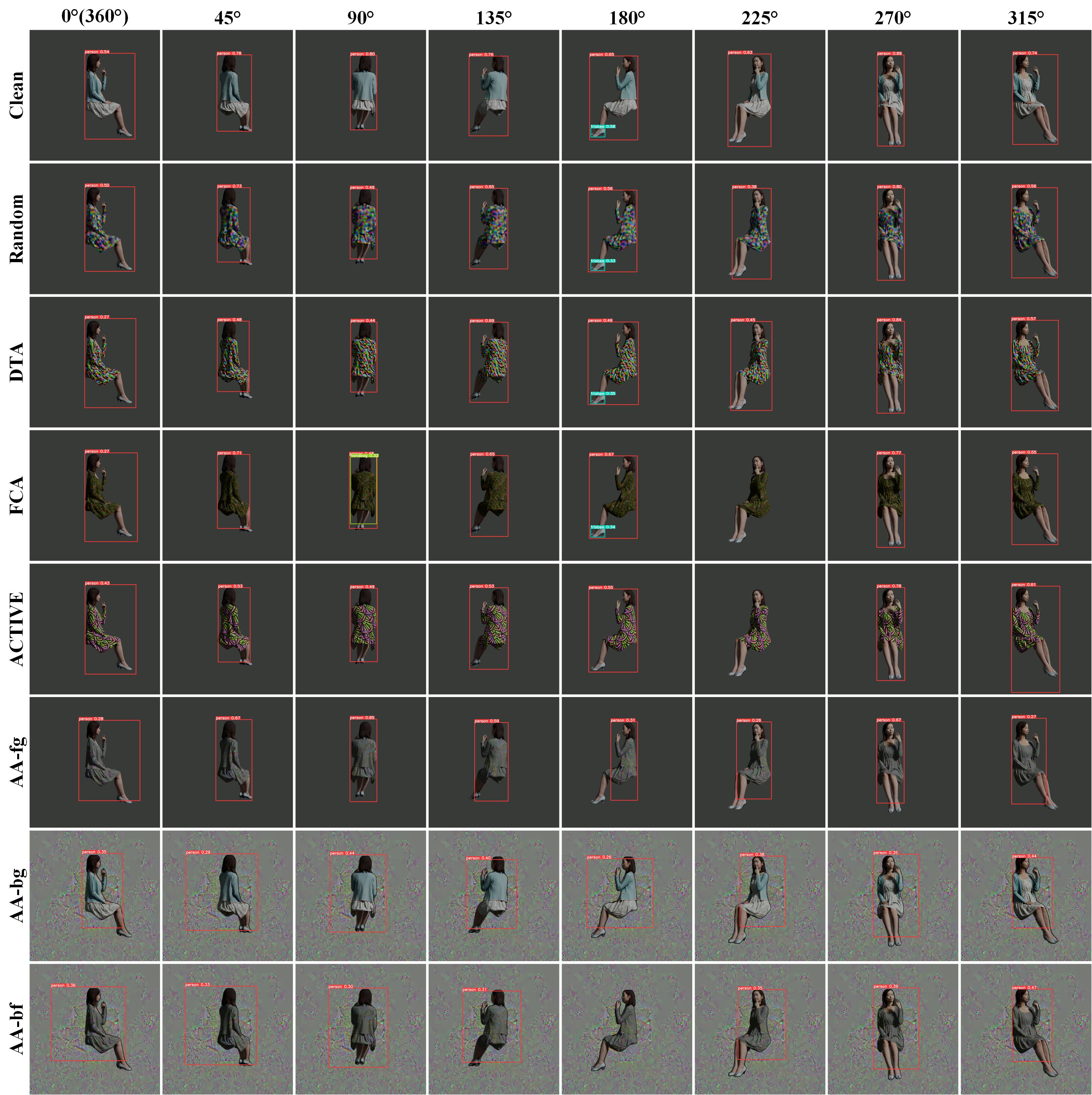} 
  \caption{Qualitative attack comparison of person detection in physically-based simulation and the victim model is YOLOv5s. Please zoom in for better visualization.}
  \label{fig_attack_comparison_yolov5s_person}
\end{figure*}


\subsection{Transfer Attack against Image Classification}
\label{appendix_transfer_attack_other_classification_models}

\begin{figure*}[!htbp]
  \centering
  \includegraphics[width=0.99\linewidth]{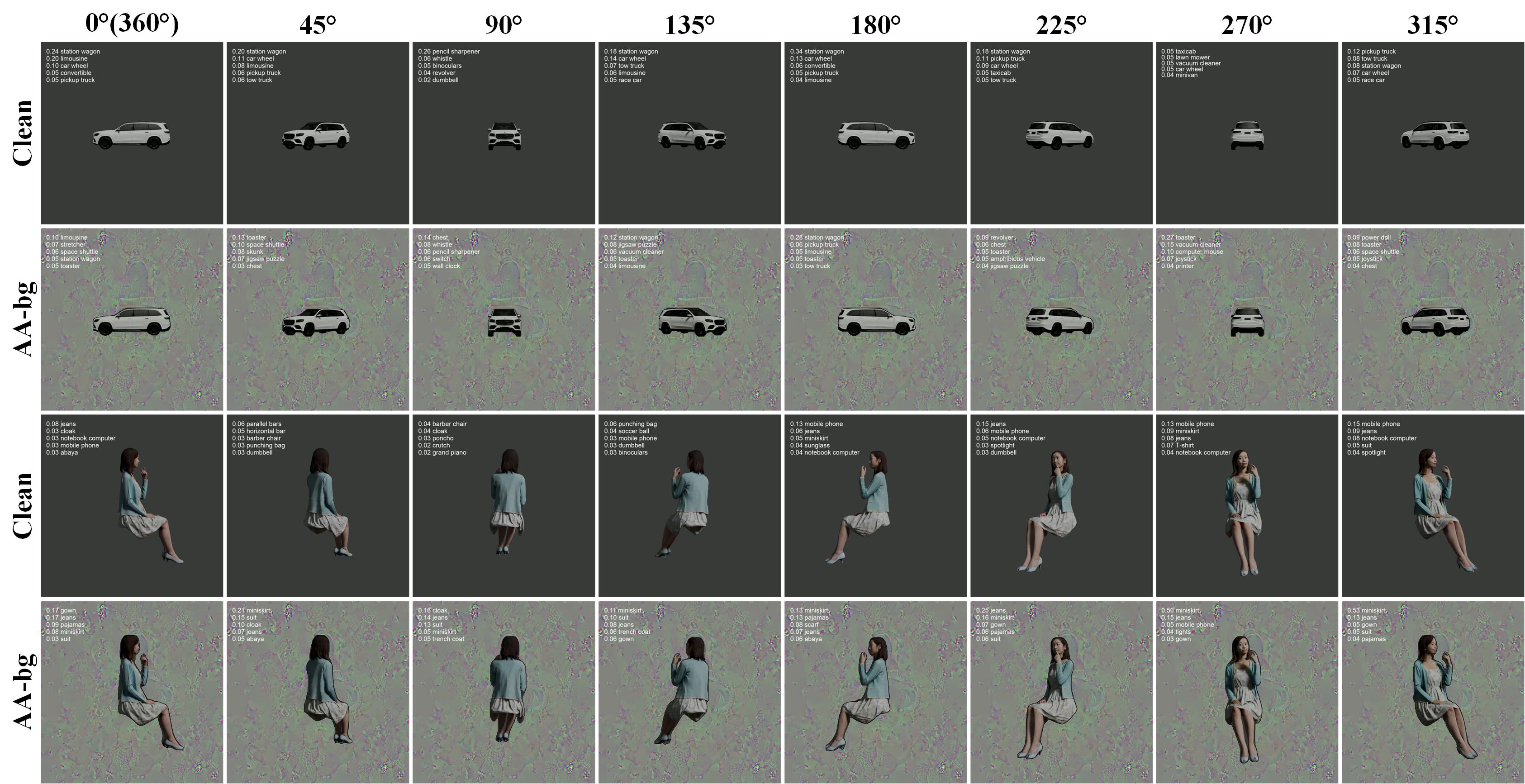} 
  \caption{Trasfer attack against image classification model in physically-based simulation and the victim model is ResNet50. Please zoom in for better visualization.}
  \label{fig_attack_resnet50}
\end{figure*}

\begin{figure*}[!htbp]
  \centering
  \includegraphics[width=0.99\linewidth]{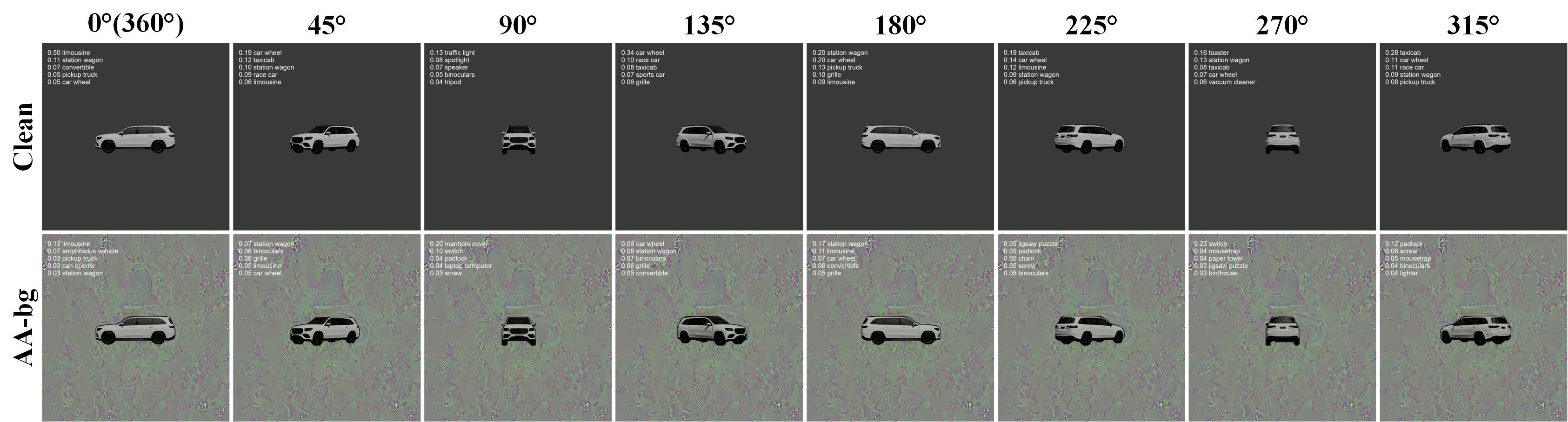} 
  \caption{Trasfer attack against image classification model in physically-based simulation and the victim model is ResNet101. Please zoom in for better visualization.}
  \label{fig_attack_resnet101}
\end{figure*}

\begin{figure*}[!htbp]
  \centering
  \includegraphics[width=0.99\linewidth]{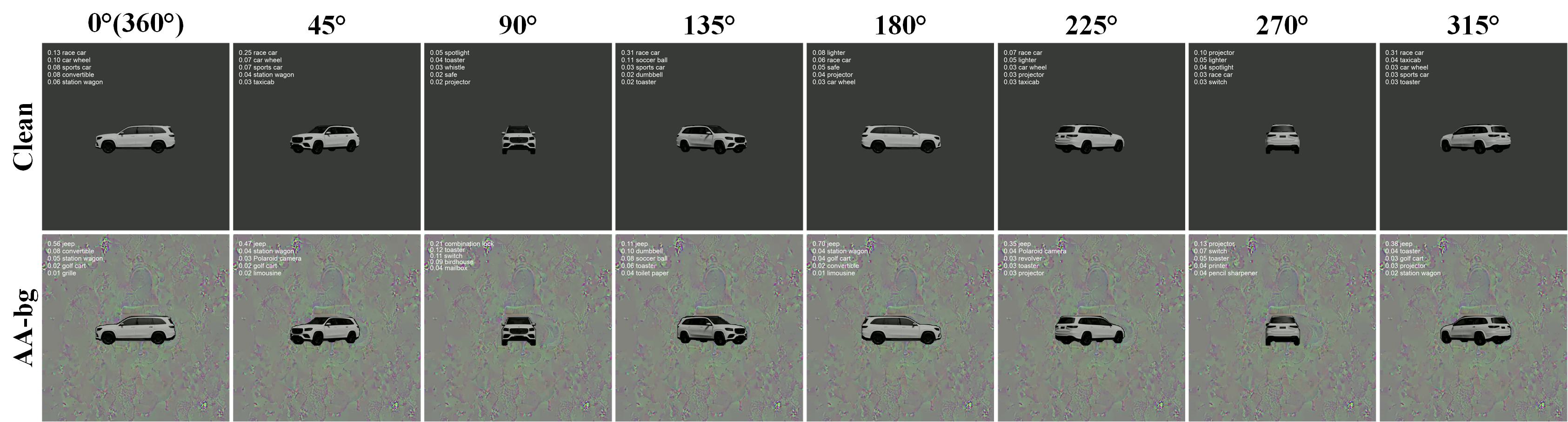} 
  \caption{Trasfer attack against image classification model in physically-based simulation and the victim model is YOLOv5s-cls. Please zoom in for better visualization.}
  \label{fig_attack_yolov5s-cls}
\end{figure*}

\begin{figure*}[!htbp]
  \centering
  \includegraphics[width=0.99\linewidth]{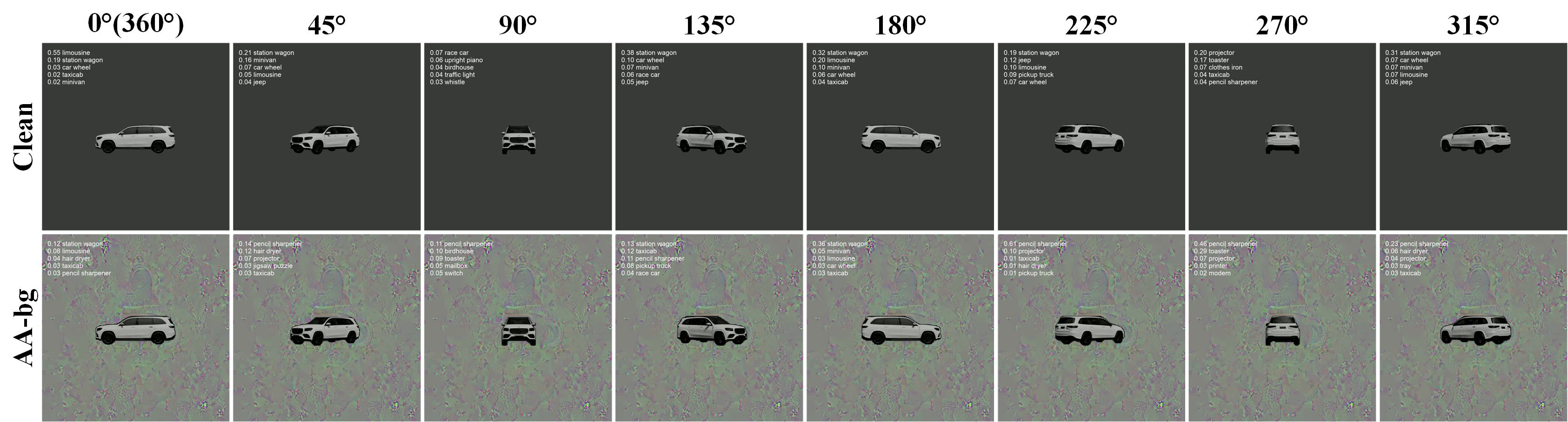} 
  \caption{Trasfer attack against image classification model in physically-based simulation and the victim model is YOLOv5l-cls. Please zoom in for better visualization.}
  \label{fig_attack_yolov5l-cls}
\end{figure*}

\begin{figure*}[!htbp]
  \centering
  \includegraphics[width=0.99\linewidth]{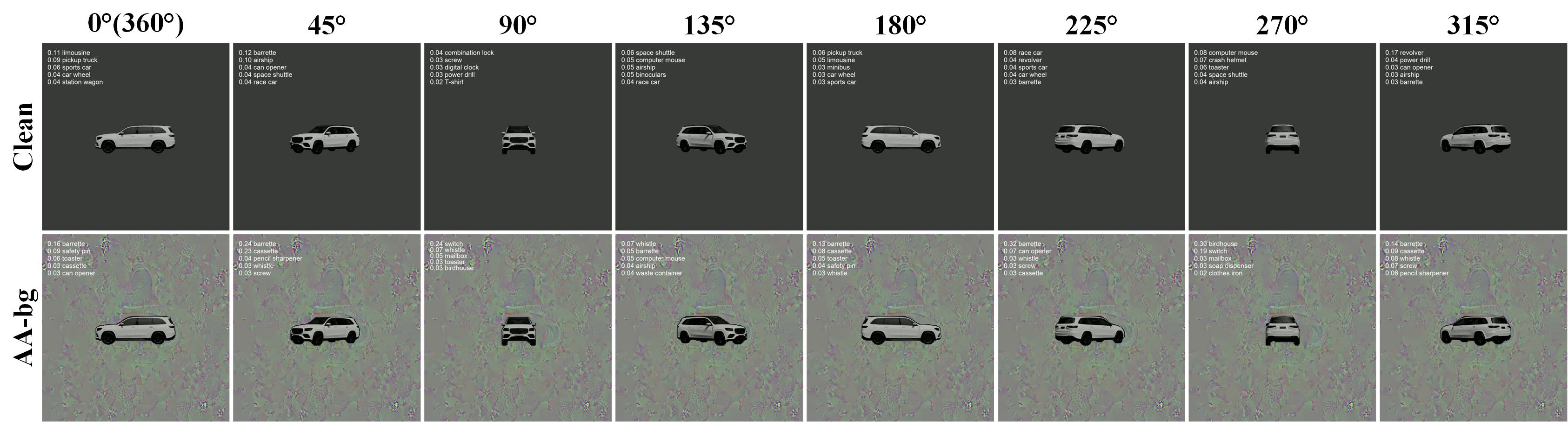} 
  \caption{Trasfer attack against image classification model in physically-based simulation and the victim model is EfficientNet-b1. Please zoom in for better visualization.}
  \label{fig_attack_efficientnet_b1}
\end{figure*}

\begin{figure*}[!htbp]
  \centering
  \includegraphics[width=0.99\linewidth]{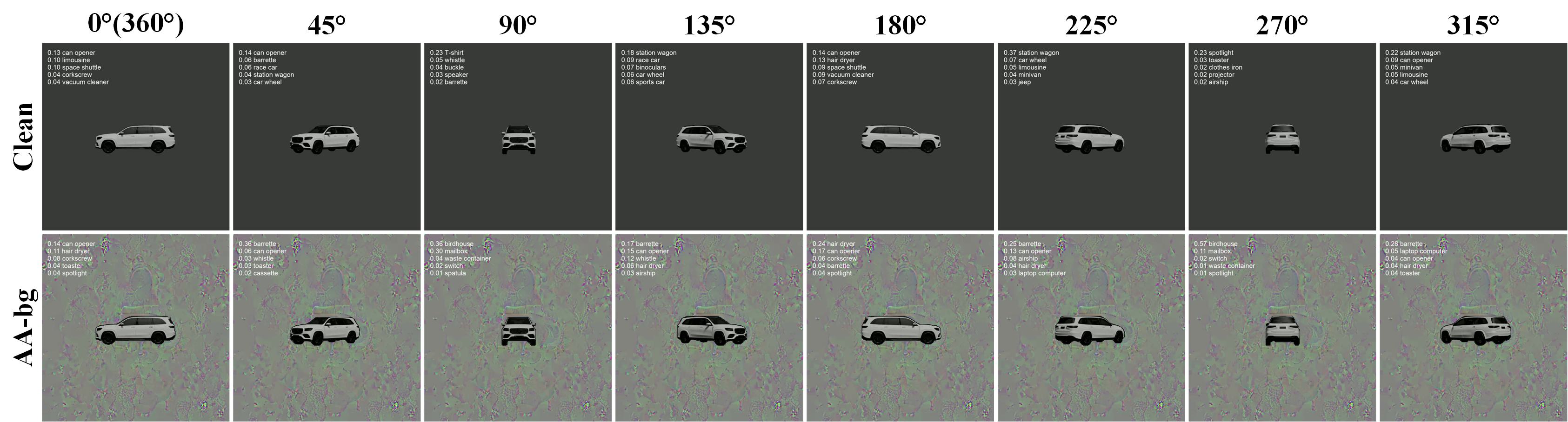} 
  \caption{Trasfer attack against image classification model in physically-based simulation and the victim model is EfficientNet-b3. Please zoom in for better visualization.}
  \label{fig_attack_efficientnet_b3}
\end{figure*}

The transfer attack against image classification models is illustrated through Fig. \ref{fig_attack_resnet50}, \ref{fig_attack_resnet101}, \ref{fig_attack_yolov5s-cls}, \ref{fig_attack_yolov5l-cls}, \ref{fig_attack_efficientnet_b1}, and \ref{fig_attack_efficientnet_b3}, demonstrating the effectiveness of adversarial perturbations across various image classification models in a physically-based simulation setting. 
These figures show how the attack can generalize to different models, including popular architectures like ResNet50, ResNet101, YOLOv5s-cls, YOLOv5l-cls, EfficientNet-b1, and EfficientNet-b3.

In Fig. \ref{fig_attack_resnet50}, the transfer attack against the ResNet50 model is depicted, showcasing the perturbations' ability to cause misclassification. 
Similarly, Fig. \ref{fig_attack_resnet101} displays the impact on ResNet101, revealing comparable results. 
Fig. \ref{fig_attack_yolov5s-cls} and \ref{fig_attack_yolov5l-cls} focus on YOLOv5s-cls and YOLOv5l-cls models, respectively, again confirming the attack's success in misleading these classifiers.

\begin{figure*}[!htbp]
  \centering
  \includegraphics[width=0.99\linewidth]{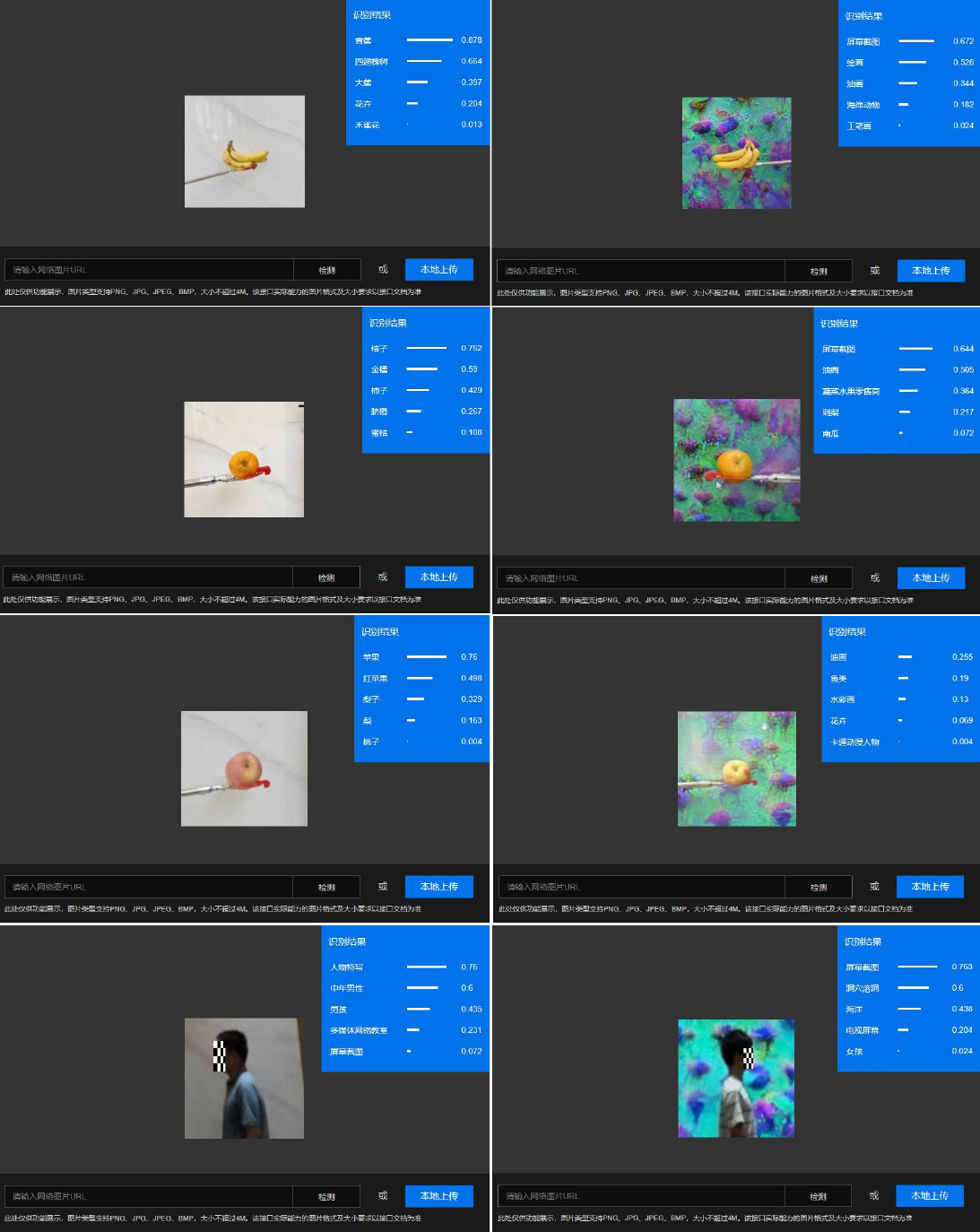} 
  \caption{Physical attacks generalize to image classification (Baidu AI).
  }
  \label{fig_image_classification_physical_attack}
  \end{figure*}

Fig. \ref{fig_attack_efficientnet_b1} and \ref{fig_attack_efficientnet_b3} extend the examination to EfficientNet variants, with b1 and b3 configurations, illustrating that the attack can also affect these efficient architectures. 
Finally, Fig. \ref{fig_image_classification_physical_attack} provides evidence that physical attacks can generalize to black-box image classification models deployed in real scenario applications, like the Baidu AI platform.

All of these figures emphasize the transferability of adversarial perturbations, which can be precomputed and then applied to different models without needing to be reoptimized for each classifier. This characteristic of adversarial attacks poses a significant security concern for image classification systems, as it suggests that a single set of perturbations could potentially compromise multiple models in various settings. The visualizations encourage a closer look at the robustness of image classification models against adversarial attacks, particularly in physically realistic environments. Zooming in on these figures would allow for a more detailed analysis of the perturbations and their effects on the classification outcomes.

\subsection{Transfer Attack against Image Segmentation}
\label{appendix_transfer_attack_other_segmentation_models}

\begin{figure*}[!htbp]
  \centering
  \includegraphics[width=0.99\linewidth]{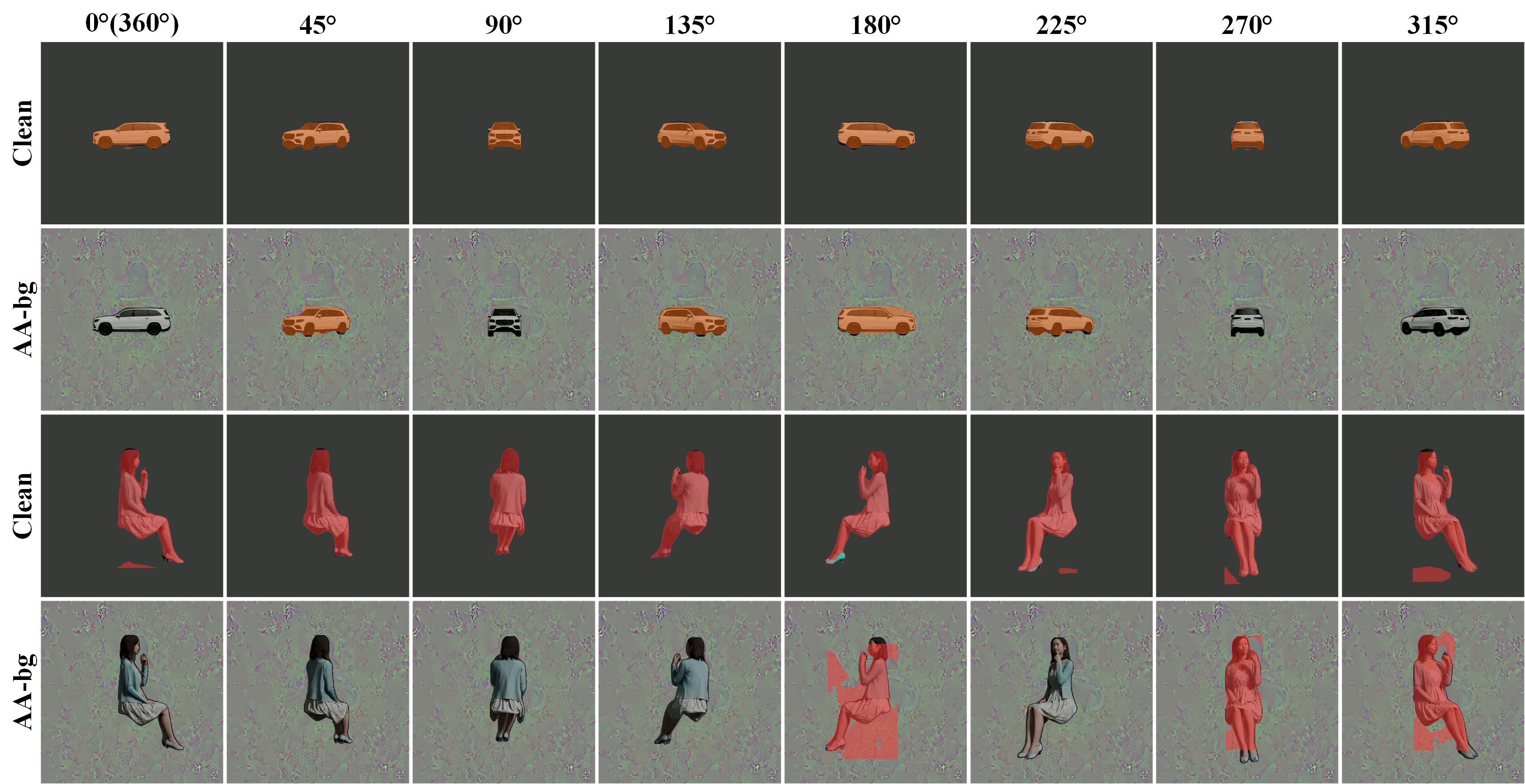} 
  \caption{Trasfer attack against image segmentation model in physically-based simulation and the victim model is YOLOv5s-seg. Please zoom in for better visualization.}
  \label{fig_attack_yolov5s-seg}
\end{figure*}

\begin{figure*}[!htbp]
  \centering
  \includegraphics[width=0.99\linewidth]{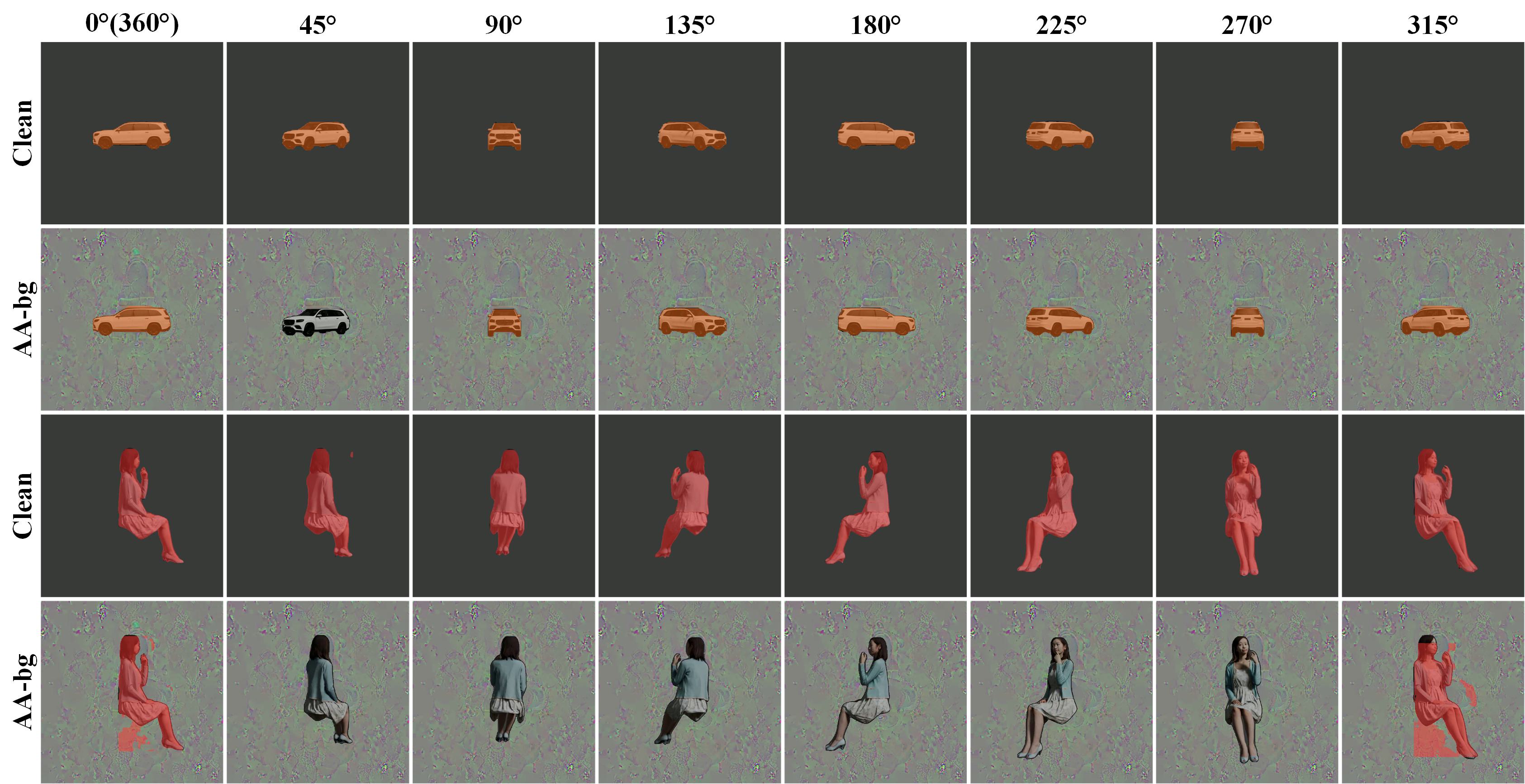} 
  \caption{Trasfer attack against image segmentation model in physically-based simulation and the victim model is YOLOv5l-seg. Please zoom in for better visualization.}
  \label{fig_attack_yolov5l-seg}
\end{figure*}

The transfer attack against image segmentation models is demonstrated through Fig. \ref{fig_attack_yolov5s-seg} and \ref{fig_attack_yolov5l-seg}. 
These figures illustrate the effect of adversarial perturbations on the performance of image segmentation models when applied in a physically-based simulation environment. 
The victim models targeted in this scenario are YOLOv5s-seg and YOLOv5l-seg, respectively.

In Fig. \ref{fig_attack_yolov5s-seg}, the YOLOv5s-seg model is challenged by adversarial perturbations that have been optimized to disrupt its ability to accurately segment objects in the scene. 
The perturbations are designed to blend into the background, but potent enough to significantly alter the model's output. 
As a result, the segmentation masks generated by the model show errors, with incorrect labeling and boundaries of objects in the scene.

Similarly, Fig. \ref{fig_attack_yolov5l-seg} presents the impact of the same attack strategy on the YOLOv5l-seg model. 
Here, too, the adversarial perturbations lead to a noticeable degradation in segmentation quality, demonstrating the transferability of the attack across different models. 
The perturbations effectively mislead the model, causing it to segment objects inaccurately and produce unreliable results.

These figures highlight the vulnerability of image segmentation models to adversarial attacks, even when the attacks are targeted at the background of the image. 
The fact that the attacks are successful across different models and in a physically-based simulation setting suggests that these perturbations can be highly adaptable and pose a significant threat to the reliability of image segmentation systems in real-world applications.

To fully comprehend the effectiveness of these attacks, it is recommended to closely inspect the figures and observe the differences between the clean and adversarially perturbed images. 
This visual analysis reveals the subtle yet powerful influence of the perturbations on the model's performance, indicating the need for robust defense mechanisms to protect against such attacks.

\begin{figure*}[t]
  \centering
  \includegraphics[width=0.99\linewidth]{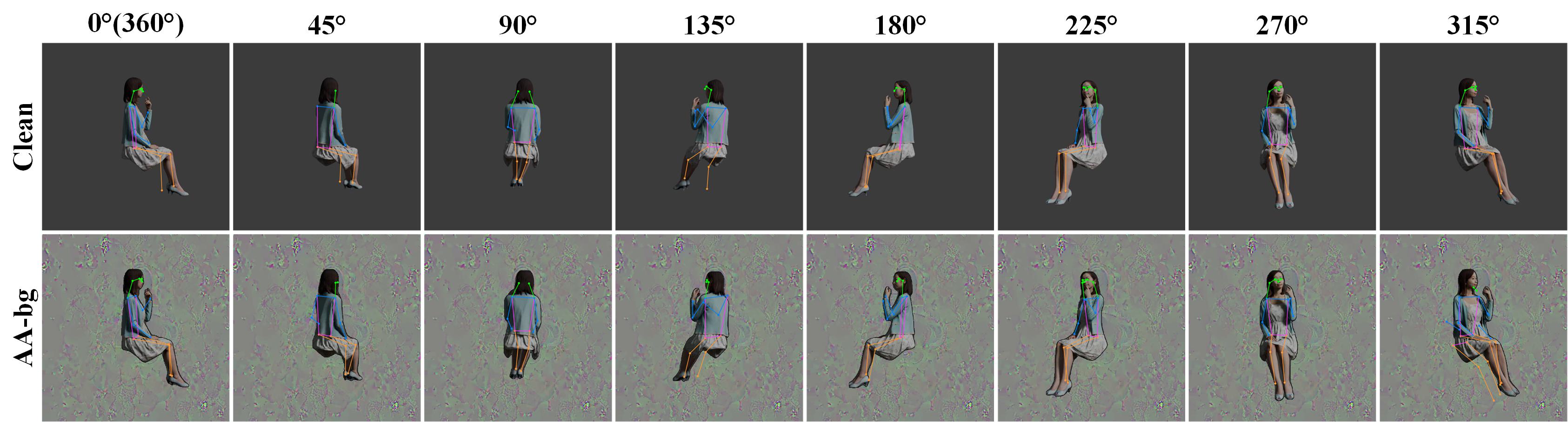} 
  \caption{Trasfer attack against pose estimation model in physically-based simulation and the victim model is YOLOv8n-pose. Please zoom in for better visualization.}
  \label{fig_attack_yolov8n-pose}
\end{figure*}

\subsection{Transfer Attack against Pose Estimation}
\label{appendix_transfer_attack_other_pose_estimation_models}

Fig. \ref{fig_attack_yolov8n-pose} vividly illustrates a transfer attack against the pose estimation model, YOLOv8n-pose, in a physically-based simulation. 
The graphic showcases an image before and after the application of background adversarial perturbations, which reveals the model's compromised performance, evident in the misidentification of joint positions post-perturbation. 
This demonstrates the attack's effectiveness in generalizing across different models, as the same perturbation can disrupt various pose estimation architectures without requiring customization.
The transfer attack underscores the vulnerability of pose estimation models to adversarial manipulation, even in seemingly benign background alterations. 
It highlights the necessity for enhanced robustness and security measures to safeguard against such attacks in real-world applications where precise pose estimation is crucial.

\end{document}